\documentclass[10pt]{article}
\ifdefined\PaperAnon\usepackage[preprint]{tmlr}\else\usepackage[preprint]{tmlr}\fi

\usepackage[utf8]{inputenc}
\usepackage[T1]{fontenc}
\usepackage[english]{babel}
\usepackage{amsmath, amsfonts, amssymb, mathtools, amsthm}
\usepackage{bbm}
\usepackage{xparse}
\usepackage{verbatim}
\usepackage{booktabs}
\usepackage{longtable}
\usepackage{multirow}
\usepackage{float}
\usepackage{afterpage}
\usepackage{needspace}
\providecommand{\FloatBarrier}{\clearpage}
\usepackage{threeparttable}
\usepackage{dcolumn}
\newcolumntype{d}{D{.}{.}{2.3}}
\usepackage{pifont}

\usepackage{adjustbox}
\usepackage{microtype}

\newcommand{\coderepo}{\url{https://github.com/d-rochau/OrbitSSM}}
\usepackage{hyperref}
\usepackage[nameinlink,noabbrev]{cleveref}
\usepackage{breakurl}
\usepackage{xcolor}
\ifdefined\PaperFinal

\fi
\usepackage[normalem]{ulem}

\usepackage{tikz}
\usepackage{subcaption}
 
\theoremstyle{definition}
\newtheorem{theorem}{Theorem}
\newtheorem{lemma}[theorem]{Lemma}
\newtheorem{definition}[theorem]{Definition}
\newtheorem{corollary}[theorem]{Corollary}
\newtheorem{proposition}[theorem]{Proposition}
\newtheorem{remark}[theorem]{Remark}

\AddToHook{env/theorem/before}{\Needspace{4\baselineskip}}
\AddToHook{env/lemma/before}{\Needspace{4\baselineskip}}
\AddToHook{env/definition/before}{\Needspace{4\baselineskip}}
\AddToHook{env/corollary/before}{\Needspace{4\baselineskip}}
\AddToHook{env/proposition/before}{\Needspace{4\baselineskip}}
\AddToHook{env/remark/before}{\Needspace{4\baselineskip}}
 
\newcommand{\R}{\mathbb{R}}
\newcommand{\C}{\mathbb{C}}

\newcommand{\bigO}{\mathcal{O}}

\usetikzlibrary{decorations.pathreplacing}
\usetikzlibrary{shapes.geometric, arrows, fit, automata, positioning, shapes.multipart}

\newcommand{\rprod}{\mathop{\overleftarrow{\textstyle\prod}}\limits}

\title{Diagonal Equivalence in State Space Models:\\Characterization, Retrieval Rank, and Selective Extensions}

\author{Dennis Rochau \\
Institute of Mathematics, Technische Universit\"at Berlin \\
\texttt{rochau@math.tu-berlin.de}}

\begin{document}

\maketitle

\begingroup
\renewcommand{\thefootnote}{}
\footnotetext{LLM-based tools were used for manuscript revision, implementation
assistance and consistency checks. The author reviewed the final content and
takes responsibility for it.}
\endgroup

\begin{abstract}
{We classify real and complex state-space models (SSMs) with linear transitions,
distinguishing fixed-basis diagonal equivalence from input--output equivalence
and tracking coefficient-conditioning windows under coordinate changes.
We characterize diagonal equivalence through commuting transition families
whose members are diagonalizable over the chosen field. With time-shared
token-local matrix maps, we identify this class's state-mediated kernels with
{unnormalized scalar-mode linear attention with multiplicative masks.}
Mamba~3's multiple-input, multiple-output (MIMO) construction lifts one value
per local interface; SSD-MIMO routes independent values for local retrieval rank above one. HybridMamba combines Mamba~1's per-coordinate decay,
Mamba~2-style heads and Mamba~3's complex rotations and two-endpoint writes,
mixing real and rotating state blocks with different decay laws. They recover Mamba~1--3
single-input, single-output (SISO) configurations as exact special cases.
On five large H100 bfloat16 workloads, our fused GPU kernels improve
geometric-mean training time over official Mamba~1/3
in matched SISO recoveries, {with Mamba~2 near parity at SSM-core scope.}
Memory tradeoffs depend on the baseline and metric: saved-for-backward
versus peak allocation.
In small-model multi-token recall experiments, both designs outperform
Mamba~1--3 on the hardest tier; higher routing ranks perform best in gated
or phase-diverse configurations. In single-seed language models with about $500$M parameters,
{both improve on Mamba~1/2 in test loss at $2$B tokens,} but the routing gains do not translate
into lower language-model loss. Our best configurations approach Mamba~3 SISO
through $10$B tokens {while remaining behind it and Mamba~3 MIMO in language-model loss.}}
\end{abstract}

\par\begingroup
\section{Introduction}
\label{sec:introduction}

{Selective state-space models (SSMs) process sequences through a
fixed-size recurrent state, with transition, write and read maps conditioned
on a bounded window of inputs. They offer an alternative to the quadratic
token interactions of dense self-attention~\citep{vaswani2017attention,dao2022flashattention,dao2023flashattention2}.
The state update remains affine in the state.} {Which state-space models
are diagonal-equivalent, meaning that one fixed basis diagonalizes their
transitions for every input and step? When can a model outside this class
nevertheless admit an input--output-equivalent diagonal realization, and
what coefficient-conditioning windows does that require? How can selective
models within the class be extended?}

{For the fixed-basis question, the criterion is precise: over $\R$ or $\C$,
the entire realized transition family admits one diagonalizing basis exactly
when its members commute pairwise and each is diagonalizable
(Theorem~\ref{thm:simultaneous_diagonalization}). A transition need not be
supplied as a diagonal matrix; what matters is one basis that works for every
input and step. Noncommuting pairs therefore rule out diagonal equivalence
for nondegenerate DeltaNet, DeltaProduct and RWKV-7 configurations, even when
individual transitions are diagonalizable~\citep{yang2024deltanet,yang2025gateddelta,siems2025deltaproduct,peng2025rwkv7}.
A defective transition obstructs diagonalization over both fields; over $\R$,
nonreal eigenvalues provide an additional obstruction.}

{{By this criterion, many widely used SSMs and gated linear
recurrences, selective or not, are diagonal-equivalent
(Table~\ref{tab:specializations}). The structured state space sequence model
(S4) used a time-invariant SSM evaluated by convolution~\citep{gu2022efficiently}
and belongs to this class when its learned transition is diagonalizable.
Diagonal state spaces (DSS), diagonal S4 (S4D) and simplified state space
layers (S5) developed diagonal and scan-based formulations
\citep{gupta2022diagonal,gu2022s4d,smith2023simplified}.
Mamba introduced input-dependent selection~\citep{gu2024mamba}, Mamba~2
scalar-transition state space duality (SSD)~\citep{dao2024transformers}, and
Mamba~3 complex dynamics with a revised write rule~\citep{lahoti2026mamba3}.
Further examples include the linear recurrent unit (LRU)~\citep{orvieto2023},
retention network (RetNet)~\citep{sun2023retnet}, gated linear attention
(GLA)~\citep{yang2024gla}, matrix long short-term memory
(mLSTM)~\citep{beck2024xlstm}, real-gated LRU (RG-LRU)~\citep{de2024griffin},
hierarchically gated recurrent network with state expansion
(HGRN2)~\citep{qin2024hgrn2} and adaptive unitary state space model
(AUSSM)~\citep{karuvally2025aussm}.
In a fixed diagonalizing basis, their recurrence cores therefore differ mainly
in how they parametrize per-coordinate decay, phase, writes and reads.}}

\paragraph{Why the fixed basis matters.}
We study recurrences of the form
\begin{align}
  h_t &= A_t h_{t-1}+B_tu_t,&
  y_t &= C_t h_t+D_tu_t,\label{eq:intro_discrete}
\end{align}
At one local interface in real state coordinates,
$u_t\in\R^R$, $h_t\in\R^N$ and $y_t\in\R^O$, with
$A_t\in\R^{N\times N}$, $B_t\in\R^{N\times R}$,
$C_t\in\R^{O\times N}$ and $D_t\in\R^{O\times R}$.
Thus $R$ counts supplied input features and $O$ counts local outputs;
{$R=O=1$ is the scalar, single-input, single-output (SISO) interface;
$R,O>1$ gives a multiple-input, multiple-output (MIMO) interface, without
requiring $R=O$.} The coefficient maps may depend on a wider conditioning
input than $u_t$.

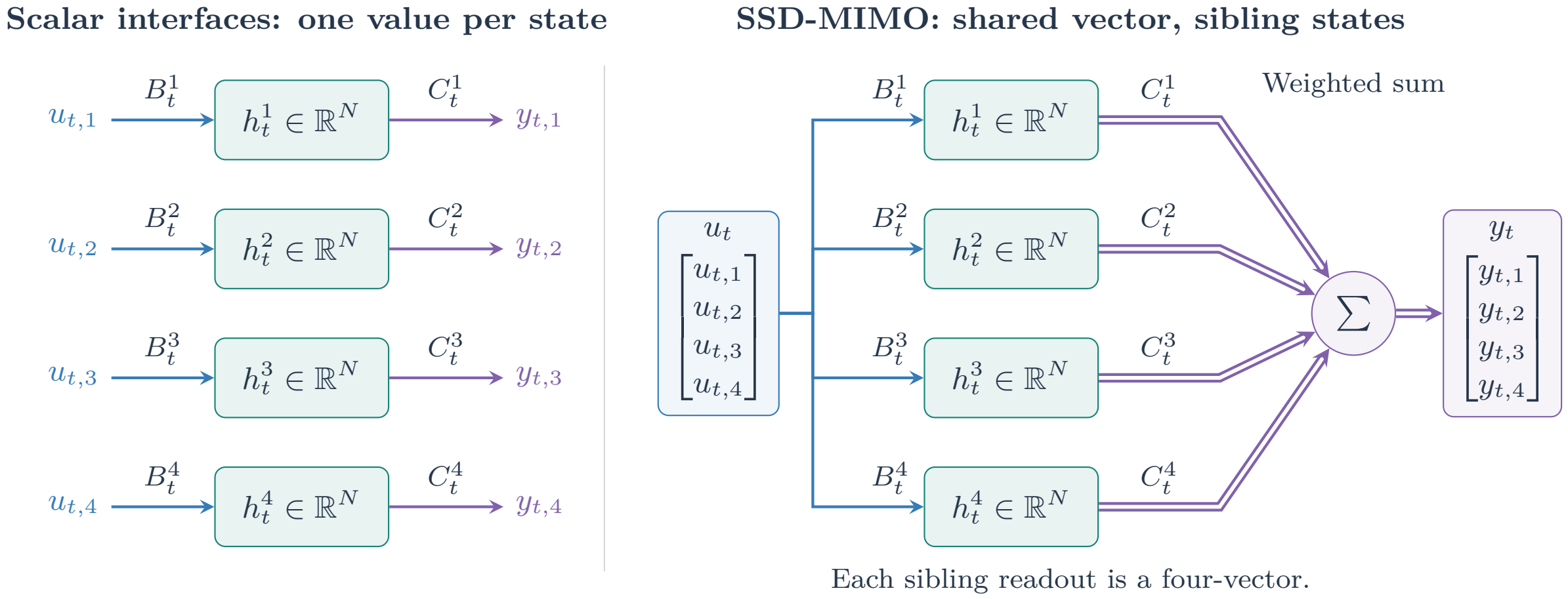
\begin{figure}[t]
\centering
\adjustbox{max width=\linewidth}{\definecolor{routeink}{HTML}{26364A}
\definecolor{routeblue}{HTML}{367BB5}
\definecolor{routeteal}{HTML}{178478}
\definecolor{routeviolet}{HTML}{8161AB}
\begin{tikzpicture}[font=\small,text=routeink,>=stealth,
  state/.style={draw=routeteal,fill=routeteal!10,rounded corners=3pt,
    minimum width=1.75cm,minimum height=.8cm,align=center},
  flow/.style={->,line width=.8pt,draw=routeblue},
  read/.style={->,line width=.8pt,draw=routeviolet},
  vectorread/.style={read,double=white,double distance=1.1pt}]
\node[font=\small\bfseries] at (2.6,5.7) {Scalar interfaces: one value per state};
\node[font=\small\bfseries] at (10.3,5.7) {SSD-MIMO: shared vector, sibling states};
\draw[routeink!20] (5.6,.15) -- (5.6,5.25);
\foreach \r/\yy in {1/4.7,2/3.4,3/2.1,4/.8} {
  \node[text=routeblue] (u\r) at (.25,\yy) {$u_{t,\r}$};
  \node[state] (h\r) at (2.55,\yy) {$h_t^{\r}\in\R^N$};
  \node[text=routeviolet] (y\r) at (4.95,\yy) {$y_{t,\r}$};
  \draw[flow] (u\r) -- node[above,font=\footnotesize] {$B_t^{\r}$} (h\r);
  \draw[read] (h\r) -- node[above,font=\footnotesize] {$C_t^{\r}$} (y\r);
}
\node[draw=routeblue,fill=routeblue!7,rounded corners=3pt,
  minimum width=1.15cm,minimum height=1.6cm,align=center] (u) at (6.75,2.75)
  {$u_t$\\[2pt]$\begin{bmatrix}u_{t,1}\\u_{t,2}\\u_{t,3}\\u_{t,4}\end{bmatrix}$};
\node[draw=routeviolet,fill=routeviolet!8,circle,minimum size=.7cm] (sum) at (13.15,2.75) {$\sum$};
\node[draw=routeviolet,fill=routeviolet!7,rounded corners=3pt,
  minimum width=1.15cm,minimum height=1.6cm,align=center] (out) at (14.65,2.75)
  {$y_t$\\[2pt]$\begin{bmatrix}y_{t,1}\\y_{t,2}\\y_{t,3}\\y_{t,4}\end{bmatrix}$};
\foreach \r/\yy in {1/4.7,2/3.4,3/2.1,4/.8} {
  \node[state] (v\r) at (9.7,\yy) {$h_t^{\r}\in\R^N$};
  \draw[flow] (u.east) -- (7.7,2.75) |- node[pos=.85,above,font=\footnotesize] {$B_t^{\r}$} (v\r.west);
  \draw[vectorread] (v\r.east) -- node[above,font=\footnotesize] {$C_t^{\r}$}
    (11.8,\yy) -- (sum);
}
\draw[vectorread] (sum) -- (out);
\node[font=\footnotesize,align=center] at (13.15,5.05) {Weighted sum};
\node[font=\footnotesize] at (10.3,.05) {Each sibling readout is a four-vector.};
\end{tikzpicture}}
\caption{Local value routing, illustrated at equal total state width $4N$.
\textbf{Left:} four scalar interfaces, with $B_t^r\in\R^{N\times1}$ and
$C_t^r\in\R^{1\times N}$.
\textbf{Right:} one SSD-MIMO group at $R=O=4$ supplies the same vector
to four sibling states, with $B_t^r\in\R^{N\times4}$ and
$C_t^r\in\R^{4\times N}$, then combines their vector readouts as
$y_t=\sum_r\zeta_{r,t}C_t^rh_t^r$, with scalar sibling weights $\zeta_{r,t}$.
Both yield four output coordinates; parameter sharing and computation
need not match; coefficient generation and feedthrough are omitted.}
\label{fig:intro_interfaces}
\end{figure}

Within the characterized class, one fixed change
of coordinates turns the state evolution into independent diagonal
recurrences and transforms the writes and reads accordingly.
{This fixed-basis property is stronger than input--output equivalence:
with invertible transitions, a prefix-dependent change of state coordinates
yields identity transitions and preserves outputs, but may require the full
input prefix in the coefficient maps (Lemma~\ref{lem:gauge_prefix}). A fixed
change of basis preserves the original conditioning window.}
{Affine
composition admits parallel scans for states and gradients even with
non-diagonal transitions; diagonal structure makes the transition
compositions coordinatewise, avoiding dense matrix products.}
Scalar transitions shared across state coordinates also yield a common
history mask and the chunked matrix-multiply form of
SSD~\citep{dao2024transformers}; general diagonal transitions may require separate masks per
coordinate, and coefficient sharing and map costs decide which form is
efficient. {With time-shared token-local maps, the state-mediated
(write--state--read) kernels are exactly those of the masked linear attention
class defined here. The correspondence preserves the complete output,
including the initial-state contribution and direct feedthrough $D_tu_t$
(Theorem~\ref{thm:mla_membership}).}

Our taxonomy records transition families, input windows and local
interfaces, complementing dynamical-system~\citep{sieber2024dsf}, structured linear
controlled differential equation~\citep{slices2025} and test-time regression~\citep{wang2025testtime}
formulations.

\paragraph{Two axes within the class.}
{{Our designs extend different parts of this common structure
(Figure~\ref{fig:intro_interfaces}; Section~\ref{sec:hybridmamba}).
SSD-MIMO routes multiple independent values through MIMO interfaces,
with decay shared across state coordinates within each scan head;
Mamba~3's implemented MIMO construction instead lifts one supplied scalar
per subsystem. HybridMamba (HM) combines per-coordinate decay, head sharing,
complex rotation and the two-endpoint write through SISO interfaces.
Neither design combines full MIMO routing with independently selected
per-coordinate decay. These are distinctions in local interfaces and
transition structure, rather than a whole-architecture expressivity ordering.}}
\paragraph{Evidence and contributions.}
{{Multi-query, multi-key, variable-length associative recall
(MQMKVAR) combines multi-query associative recall (MQAR)'s layout~\citep{arora2023zoologymeasuringimprovingrecall}
with the multi-token keys and variable-length values of Mechanistic Architecture Design (MAD)'s fuzzy in-context
recall~\citep{poli2024mad}, introducing controlled prefix and substring overlap
to test recall under ambiguous keys (Figure~\ref{fig:recall_tasks}).} {We evaluate three difficulty tiers, routing width, gates
and transition mixtures.} Our contributions are:}
\begin{enumerate}
\item {{\emph{A structural framework and taxonomy.} We distinguish fixed-basis
diagonal equivalence from input--output equivalence and give a head-organized
taxonomy of SSMs over $\R$ and $\C$ by transition families,
coefficient-conditioning windows, and local state and interface widths (Section~\ref{sec:system_types}).
{Applying classical simultaneous diagonalization, we
identify the diagonal-equivalent class and its obstructions. We place many
established architectures within this class, including models formulated
with non-diagonal transitions.} We track how moving frames change coefficient windows,
bound local retrieval rank and characterize token-local masked-linear-attention
representations (Sections~\ref{sec:discretization}--\ref{sec:theory}).}}
\item {{\emph{Two designs with validated baseline recovery.}
We introduce SSD-MIMO to route multiple independent values through each local
interface, and HybridMamba to combine per-coordinate decay, head sharing,
complex rotations and two-endpoint writes within one model.
Explicit parameter maps recover Mamba~1, Mamba~2 and Mamba~3 SISO in exact
arithmetic. Copied-weight tests check outputs, input gradients and parameter
gradients against the official implementations, with finite-precision agreement
and its limits reported in Section~\ref{sec:mamba_recovery}.}}
\item {{\emph{Optimized GPU kernels.} Our native kernels support the two
designs' real and complex dynamics, read/write gates, variable initial states
and differentiable state continuation. SSD-MIMO supports SISO and MIMO
interfaces, including unequal input and output widths ($R\ne O$).
For SISO baseline recoveries on H100 with bfloat16 (bf16), we outperform the
official Mamba~1 and Mamba~3 implementations in geometric-mean training speed
at both scan and full-mixer scope across five workloads varying batch size,
sequence length and inner model width.
Our Mamba~2 recovery is near parity with the official implementation at
scan scope but slower at full-mixer scope on average; individual shapes
can reverse these speed comparisons.
Section~\ref{sec:hybridmamba_complexity} reports per-shape times and memory.}}
\item {{\emph{Multi-token recall findings.} On MQMKVAR's multi-token,
overlapping-key tasks, both designs outperform the Mamba~1--3 baselines on
the hardest tier in our small-model setting. Routing two or four independent
values per local interface gives SSD-MIMO its strongest results on the hardest
tier; HybridMamba mixtures lead on the intermediate tier. These gains occur in configurations with write gates or differing sibling
phases; their causal roles are not isolated. Section~\ref{sec:exp_mqmkvar} defines the tiers and reports the shared
learning-rate sweep and baseline budget differences.}}
\item {{\emph{Language-modelling findings.} In approximately $500$M-parameter
models trained on FineWeb-Edu~\citep{penedo2024fineweb}, both SSD-MIMO and
HybridMamba improve on Mamba~1 and
Mamba~2 in test loss at $2$B tokens under the same training recipe.
HybridMamba with $75\%$ complex dynamic rows stays near Mamba~3 SISO through
$10$B tokens, slightly behind on loss; the tested SSD-MIMO configurations
also remain close, while Mamba~3 MIMO has the lowest loss. The recall gains
from wider routing described above do not translate into lower language-model
loss in these experiments. These comparisons use one training seed per
configuration; downstream outcomes and the role of
routing are discussed in Section~\ref{sec:exp_fineweb}.}}
\end{enumerate}

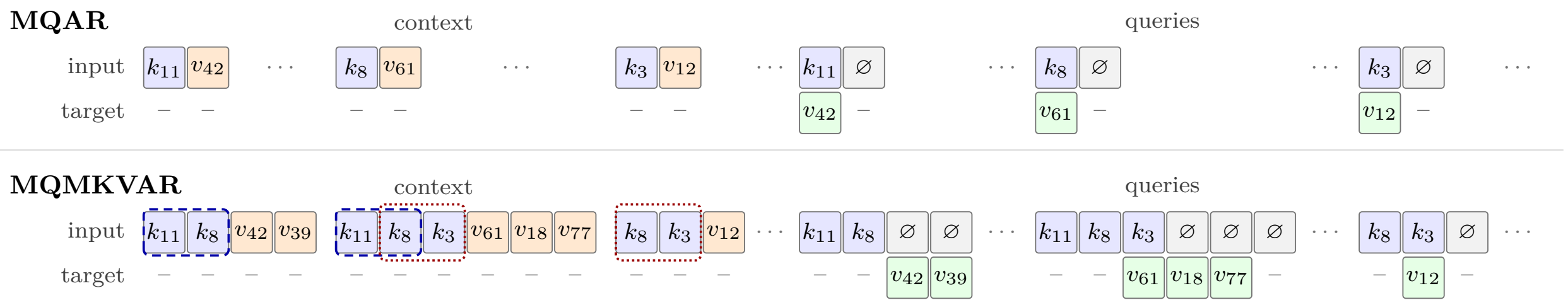
\begin{figure}[t]
\centering
\adjustbox{max width=\linewidth}{\begin{tikzpicture}[
  x=1cm,y=1cm,
  tok/.style={draw=black!55,rounded corners=1pt,minimum width=5mm,
    minimum height=5mm,inner sep=0.5pt,font=\footnotesize},
  key/.style={tok,fill=blue!10},
  val/.style={tok,fill=orange!18},
  pad/.style={tok,fill=black!5},
  target/.style={tok,fill=green!10},
  lab/.style={font=\footnotesize,text=black!75},
  gap/.style={font=\footnotesize,text=black!55},
  title/.style={font=\small\bfseries,anchor=west},
  pre/.style={densely dashed,thick,blue!65!black,rounded corners=2pt},
  suf/.style={densely dotted,thick,red!60!black,rounded corners=2pt}]
\node[title] at (-2.0,0.55) {MQAR};
\node[lab,anchor=east] at (-0.35,0) {input};
\node[lab,anchor=east] at (-0.35,-0.55) {target};
\node[key] at (0,0) {$k_{11}$};
\node[lab] at (0,-0.55) {--};
\node[val] at (0.53,0) {$v_{42}$};
\node[lab] at (0.53,-0.55) {--};
\node[key] at (2.332,0) {$k_{8}$};
\node[lab] at (2.332,-0.55) {--};
\node[val] at (2.862,0) {$v_{61}$};
\node[lab] at (2.862,-0.55) {--};
\node[key] at (5.724,0) {$k_{3}$};
\node[lab] at (5.724,-0.55) {--};
\node[val] at (6.254,0) {$v_{12}$};
\node[lab] at (6.254,-0.55) {--};
\node[key] at (7.95,0) {$k_{11}$};
\node[target] at (7.95,-0.55) {$v_{42}$};
\node[pad] at (8.48,0) {$\varnothing$};
\node[lab] at (8.48,-0.55) {--};
\node[key] at (10.812,0) {$k_{8}$};
\node[target] at (10.812,-0.55) {$v_{61}$};
\node[pad] at (11.342,0) {$\varnothing$};
\node[lab] at (11.342,-0.55) {--};
\node[key] at (14.734,0) {$k_{3}$};
\node[target] at (14.734,-0.55) {$v_{12}$};
\node[pad] at (15.264,0) {$\varnothing$};
\node[lab] at (15.264,-0.55) {--};
\node[gap] at (1.431,0) {$\cdots$};
\node[gap] at (4.293,0) {$\cdots$};
\node[gap] at (7.367,0) {$\cdots$};
\node[gap] at (10.176,0) {$\cdots$};
\node[gap] at (14.098,0) {$\cdots$};
\node[gap] at (16.43,0) {$\cdots$};
\node[lab] at (3.267,0.55) {context};
\node[lab] at (12.1,0.55) {queries};
\draw[black!15] (-2.0,-1.0) -- (16.96,-1.0);
\node[title] at (-2.0,-1.45) {MQMKVAR};
\node[lab,anchor=east] at (-0.35,-2) {input};
\node[lab,anchor=east] at (-0.35,-2.55) {target};
\node[key] at (0.0,-2) {$k_{11}$};
\node[lab] at (0.0,-2.55) {--};
\node[key] at (0.53,-2) {$k_{8}$};
\node[lab] at (0.53,-2.55) {--};
\node[val] at (1.06,-2) {$v_{42}$};
\node[lab] at (1.06,-2.55) {--};
\node[val] at (1.59,-2) {$v_{39}$};
\node[lab] at (1.59,-2.55) {--};
\node[key] at (2.332,-2) {$k_{11}$};
\node[lab] at (2.332,-2.55) {--};
\node[key] at (2.862,-2) {$k_{8}$};
\node[lab] at (2.862,-2.55) {--};
\node[key] at (3.392,-2) {$k_{3}$};
\node[lab] at (3.392,-2.55) {--};
\node[val] at (3.922,-2) {$v_{61}$};
\node[lab] at (3.922,-2.55) {--};
\node[val] at (4.452,-2) {$v_{18}$};
\node[lab] at (4.452,-2.55) {--};
\node[val] at (4.982,-2) {$v_{77}$};
\node[lab] at (4.982,-2.55) {--};
\node[key] at (5.724,-2) {$k_{8}$};
\node[lab] at (5.724,-2.55) {--};
\node[key] at (6.254,-2) {$k_{3}$};
\node[lab] at (6.254,-2.55) {--};
\node[val] at (6.784,-2) {$v_{12}$};
\node[lab] at (6.784,-2.55) {--};
\node[gap] at (7.367,-2) {$\cdots$};
\node[key] at (7.95,-2) {$k_{11}$};
\node[lab] at (7.95,-2.55) {--};
\node[key] at (8.48,-2) {$k_{8}$};
\node[lab] at (8.48,-2.55) {--};
\node[pad] at (9.01,-2) {$\varnothing$};
\node[target] at (9.01,-2.55) {$v_{42}$};
\node[pad] at (9.54,-2) {$\varnothing$};
\node[target] at (9.54,-2.55) {$v_{39}$};
\node[key] at (10.812,-2) {$k_{11}$};
\node[lab] at (10.812,-2.55) {--};
\node[key] at (11.342,-2) {$k_{8}$};
\node[lab] at (11.342,-2.55) {--};
\node[key] at (11.872,-2) {$k_{3}$};
\node[target] at (11.872,-2.55) {$v_{61}$};
\node[pad] at (12.402,-2) {$\varnothing$};
\node[target] at (12.402,-2.55) {$v_{18}$};
\node[pad] at (12.932,-2) {$\varnothing$};
\node[target] at (12.932,-2.55) {$v_{77}$};
\node[pad] at (13.462,-2) {$\varnothing$};
\node[lab] at (13.462,-2.55) {--};
\node[key] at (14.734,-2) {$k_{8}$};
\node[lab] at (14.734,-2.55) {--};
\node[key] at (15.264,-2) {$k_{3}$};
\node[target] at (15.264,-2.55) {$v_{12}$};
\node[pad] at (15.794,-2) {$\varnothing$};
\node[lab] at (15.794,-2.55) {--};
\node[gap] at (10.176,-2) {$\cdots$};
\node[gap] at (14.098,-2) {$\cdots$};
\node[gap] at (16.43,-2) {$\cdots$};
\node[lab] at (3.267,-1.45) {context};
\node[lab] at (12.1,-1.45) {queries};
\draw[pre] (-0.25,-2.28) rectangle (0.78,-1.72);
\draw[pre] (2.082,-2.28) rectangle (3.112,-1.72);
\draw[suf] (2.612,-2.34) rectangle (3.642,-1.66);
\draw[suf] (5.474,-2.34) rectangle (6.504,-1.66);
\end{tikzpicture}}
\caption{{MQMKVAR recall combines MQAR~\citep{arora2023zoologymeasuringimprovingrecall}
with MAD's fuzzy recall~\citep{poli2024mad}. Dashed outlines mark
$k_{11}k_8$, a proper prefix of $k_{11}k_8k_3$; dotted outlines mark
the longer key's suffix $k_8k_3$. Queries match keys from their first token.
By convention, proper-prefix queries start their targets at the first padding
token $\varnothing$, which confirms key completion; other answers start at their
last key token.
Targets align with the input positions predicting them; ``--'' and all
context positions carry no loss. Earlier answers never enter the query
input. ``$\cdots$'' elides context entries, queries and variable-length,
loss-masked padding gaps between queries. Results:
Section~\ref{sec:exp_mqmkvar}; protocol: Appendix~\ref{app:mqmkvar_protocol}.}}
\label{fig:recall_tasks}
\end{figure}

\paragraph{Relation to prior analysis.}
The fixed-basis criterion is classical simultaneous diagonalization
\citep[Sec.~1.3]{horn2013matrix}, and the local rank bound is the
matrix-product rank inequality; held-input discretization and affine scans
use variation of constants~\citep[Ch.~3]{teschl2012ode} and prefix
composition~\citep[Sec.~1.4]{blelloch1990prefix}.
S5 diagonalizes a multidimensional-input/output system
\citep[Sec.~3]{smith2023simplified}, and \citet[Sec.~3]{sieber2024dsf}
compare recurrent architectures in one representation.

Recurrent/attention correspondences include the recurrent evaluation of linear attention
\citep{katharopoulos2020transformers}, SSD's structured and scalar-mask
forms \mbox{\citep[Secs.~3--5]{dao2024transformers}} {and its extension to diagonal SSMs with a
necessary and sufficient condition for a 1-semiseparable masked-attention dual~\citep{hu2025structured}}, implicit attention
with surrounding gates and convolutions \citep[Sec.~3]{zimerman2024unified},
GateLoop's cumulative transitions \citep[Sec.~3.3]{katsch2023gateloop} and
Mamba~3's complex-to-rotary equivalences \citep[Props.~2--4]{lahoti2026mamba3}.
Order restrictions from commuting dynamics are known for diagonal
controlled differential equations \citep[Thm.~4.3]{cirone2024theoretical},
a simultaneously diagonalizable, zero-forcing automaton encoding
\citep[Prop.~1]{terzic2025expressiveness} and finite-precision group
tracking \citep[Sec.~4]{shakerinava2026diagonal}.

{Building on these results, we make explicit the whole realized
transition family, state field, conditioning window and supplied-value
interface. We distinguish a fixed basis from a prefix-dependent frame,
and a kernel factorization from one with token-local generating maps.
Our interior-swap test excludes shared token-local scalar-mode
representations of a given observable kernel.}

{\paragraph{Implementation.}
{We implement SSD-MIMO and HybridMamba in a PyTorch library with fused
Triton/Gluon GPU kernels~\citep{tillet2019triton}.
\ifdefined\PaperAnon The code is provided as an anonymized ZIP archive in the supplementary material.\else The code is publicly available at \coderepo.\fi}
The SSD-MIMO scan extends the SSD decomposition used by Mamba~2/3 to routed
contractions, rotation and the two-endpoint write; one backward kernel is
modelled on Mamba~3's. The per-coordinate scan extends Mamba~1's selective scan
to rotation and mixed decay laws. Tuned for NVIDIA Hopper and Blackwell, the
kernels are timed against Mamba on an H100 (Section~\ref{sec:hm_costs}). All scans
support differentiable initial/final states in whole-sequence, continued and
step evaluation, including training through states carried across segments.}
\par\endgroup

\needspace{6\baselineskip}
\section{State-space models: exact discretization, fixed basis, and evaluation}
\label{sec:discretization}

\par\begingroup
The coefficients below may be noncommuting matrices; Section~\ref{sec:continuous}
defines their transfer operator and sequence kernel before the fixed-basis,
real/complex, SSD and scan analyses.
\par\endgroup

\subsection{Continuous-time formulation, zero-order hold (ZOH), and discretization}
\label{sec:continuous}

We abbreviate ordinary differential equation as ODE and rotary positional
embedding as RoPE.

\par\begingroup
Throughout this section, $\mathbb F\in\{\R,\C\}$ is the \emph{state field}, while $d, S, O \in \mathbb{N}$ denote the external-input dimension, the \emph{combined} state dimension, and the \emph{combined} output dimension of one dynamical system.  Inputs and outputs are real even when the state is complex, and we write $\operatorname{Re}_{\R}:=\operatorname{id}$ and $\operatorname{Re}_{\C}:=\operatorname{Re}$ componentwise.  Section~\ref{sec:system_types} partitions the state into $HP$ subsystems of width $N$, giving $S=HPN$, and uses $O$ for each subsystem's read width (Definition~\ref{def:mhssm}); Sections~\ref{sec:theory_complex}--\ref{sec:scan_grad} concern one general system: $S$ is its real state width, $R$ its write-feature width and $O$ its output width; Section~\ref{sec:theory_complex} realifies $K$ complex state coordinates into $S=2K$ real coordinates.

To relate the complex and real cases, we use the \emph{realification}
$\chi_m\colon\C^m\to\R^{2m}$, also used in
Section~\ref{sec:theory_complex}, defined by
$\chi_m(z):=(\operatorname{Re}z_1,\operatorname{Im}z_1,\ldots,
\operatorname{Re}z_m,\operatorname{Im}z_m)^{\!\top}$.
It is a real-linear bijection preserving the Euclidean norm, with
$\C^m$ viewed as a real vector space. For a complex scalar $a+ib$, with
$a,b\in\R$, define its real matrix representation by
\[
  \mathfrak R(a+ib):=\begin{pmatrix}a&-b\\ b&a\end{pmatrix}.
\]
For a matrix $M=(M_{jk})\in\C^{m\times n}$, define
$\mathfrak R(M):=[\mathfrak R(M_{jk})]_{j,k}\in\R^{2m\times2n}$:
each scalar entry is replaced by its $2\times2$ block. Then
\begin{equation}
  \chi_m(Mz)=\mathfrak R(M)\,\chi_n(z),
  \qquad M\in\C^{m\times n},\ z\in\C^n.
  \label{eq:realification}
\end{equation}
This entrywise block map preserves compatible products,
$\mathfrak R(MN)=\mathfrak R(M)\mathfrak R(N)$; for square $M$, the exponential
power series therefore gives $\mathfrak R(e^M)=e^{\mathfrak R(M)}$.

\begin{definition}\label{def:cts_ssm}
Fix an \emph{initial time} $\tau_{-1}\in\R$ and an initial state
$h_{-1}\in\mathbb F^S$. An \emph{admissible input} is a measurable
signal $x:[\tau_{-1},\infty)\to\R^d$ that is bounded on every compact
time interval. A \textit{continuous-time SSM over $\mathbb F$} has
continuous coefficient maps $A:\R^d\to\mathbb F^{S\times S}$ and
$b:\R^d\to\mathbb F^S$, and a Borel-measurable readout
$f:\R^d\times\mathbb F^S\to\R^O$. For each admissible input, its state and output satisfy
\begin{equation}
 \begin{aligned}
  h'(\tau)&=A(x(\tau))\,h(\tau)+b(x(\tau)),
  &h(\tau_{-1})&=h_{-1},\\
  y(\tau)&=f(x(\tau),h(\tau))\in\R^O.
 \end{aligned}
 \label{eq:ivp}
\end{equation}
Here $h:[\tau_{-1},\infty)\to\mathbb F^S$ is the hidden-state trajectory,
$b$ the forcing, and $y:[\tau_{-1},\infty)\to\R^O$ the
output signal. Both $A$ and $b$ may depend nonlinearly on the input.
The state differential equation is affine in $h$, with linear homogeneous part. Solutions are understood
in the Carath\'eodory sense: $h$ is locally absolutely continuous and
its differential equation holds almost everywhere.
\end{definition}

The locally integrable coefficients and state-Lipschitz bound give a
unique Carath\'eodory state trajectory and measurable output
(Remark~\ref{rem:cts_wellposed} in Appendix~\ref{app:zoh_proof}, following
\citealp[pp.~41--42]{teschl2012ode}); for $\mathbb F=\C$, the same argument
applies to $v=\chi_S(h)$ through
\[
 v'(\tau)=\mathfrak R(A(x(\tau)))\,v(\tau)+\chi_S(b(x(\tau))),
 \qquad v(\tau_{-1})=\chi_S(h_{-1}).
\]
\par\endgroup

\par\begingroup

\begin{definition}\label{def:zoh}
Let $L\in\mathbb N\cup\{\infty\}$,
write $\mathbb T_L:=\{0,\ldots,L-1\}$ for finite $L$ and
$\mathbb T_\infty:=\mathbb N_0$, and take tokens
$(x_t)_{t\in\mathbb T_L}$ in $\R^d$. Choose strictly increasing finite
times $(\tau_t)_{t\in\{-1\}\cup\mathbb T_L}$ and set
$\Delta_t:=\tau_t-\tau_{t-1}>0$ for $t\in\mathbb T_L$.
When $L=\infty$, assume $\tau_t\to\infty$, equivalently
$\sum_{t\ge0}\Delta_t=\infty$. The \emph{zero-order hold} of the
token sequence on this grid is the input
\begin{equation}\label{eq:zoh_input}
  x(\tau) \;=\; x_t
  \qquad\text{for } \tau \in [\tau_{t-1}, \tau_t),\; t\in\mathbb T_L .
\end{equation}
For finite $L$, set $x(\tau)=x_{L-1}$ for $\tau\ge\tau_{L-1}$.
\end{definition}

A continuous selective step $\Delta:\R^d\to(0,\infty)$ gives
$\tau_t=\tau_{t-1}+\Delta(x_t)$; Theorem~\ref{thm:zoh_exact} applies on each
resulting grid. For $L=\infty$, if all tokens lie in a compact set $K$, then
$\min_{x\in K}\Delta(x)>0$, so $\tau_t\to\infty$ as $t\to\infty$.

Each compact time interval meets only finitely many holds, so the held
input is admissible; its coefficients are piecewise constant, and the state
is continuously differentiable inside each hold interval and continuous
across boundaries. Token $x_t$ is held on $[\tau_{t-1},\tau_t)$ and maps
$h_{t-1}$ to the post-update state
\begin{equation}\label{eq:discrete_sampling}
  h_t \;:=\; h(\tau_t), \qquad y_t \;:=\; f(x_t, h_t),
  \qquad t\in\mathbb T_L,
\end{equation}
so it is indexed by the end of its interval; the alternative convention
holds $x_t$ on $[\tau_t,\tau_{t+1})$ and produces $h_{t+1}$. Changing
input values only at the hold boundaries, a countable null set, leaves the
state trajectory unchanged but may change the output there. The discrete
readout pairs $x_t$ with the state it produces; it need not equal $y(\tau_t)$, since at a nonterminal boundary the
right-continuous hold already takes the next token's value.
On each hold interval, the state follows a constant-coefficient linear
ODE with coefficients $A(x_t),b(x_t)$;
subscripted symbols denote the \emph{discrete} step maps defined next.

For non-commutative matrix products, we write
\begin{equation}\label{eq:reverse_product}
  \rprod_{j=i}^{k} A_j \;:=\; A_k\, A_{k-1}\, \cdots\, A_i
  \qquad (i \leq k),
\end{equation}
with the convention that the product over an empty index range ($i > k$) equals the identity matrix.

\par\endgroup

\par\begingroup
\begin{theorem}\label{thm:zoh_exact}
Assume the continuous-time model of Definition~\ref{def:cts_ssm} and the
zero-order-hold input of Definition~\ref{def:zoh}.
There is a unique locally absolutely continuous state trajectory
$h:[\tau_{-1},\infty)\to\mathbb F^S$.
Define $h_t,y_t$ by~\eqref{eq:discrete_sampling}. For $Z \in \mathbb{F}^{S \times S}$ let
\begin{equation}\label{eq:phi_one}
  \varphi(Z) \;:=\; \sum_{j=0}^{\infty} \frac{Z^{\,j}}{(j+1)!}
  \;=\; \int_{0}^{1} e^{rZ}\, dr .
\end{equation}
The series converges for every square matrix $Z$, and $\varphi(Z) = Z^{-1}\bigl(e^{Z} - I\bigr)$ whenever $Z$ is invertible.  Define the discrete coefficients
\begin{equation}\label{eq:zoh_matrices}
  A_t \;:=\; e^{\Delta_t A(x_t)}, \qquad
  b_t \;:=\; \Delta_t\, \varphi\bigl(\Delta_t A(x_t)\bigr)\, b(x_t)
  \;=\; \biggl(\int_{0}^{\Delta_t} e^{r A(x_t)}\, dr\biggr) b(x_t)
  \;\in\; \mathbb{F}^S.
\end{equation}

Then at every finite index $t\in\mathbb T_L$ the sampled state and discrete readout satisfy
\begin{equation}
  h_t \;=\; A_t\, h_{t-1} \;+\; b_t, \qquad y_t \;=\; f(x_t, h_t). \label{eq:recurrence}
\end{equation}
Define the \emph{transfer operator}
\begin{equation}\label{eq:transfer_operator}
  \Phi(t,s)\;:=\;\rprod_{j=s+1}^{t}A_j,
  \qquad -1\le s\le t,\quad t\in\{-1\}\cup\mathbb T_L,
  \qquad \Phi(t,t)=I,
\end{equation}
which propagates the homogeneous state from step $s$ to step $t$.
The closed form is
\begin{equation}\label{eq:telescoped}
 h_t=\Phi(t,-1)h_{-1}+\sum_{s=0}^{t}\Phi(t,s)b_s
    =\biggl(\rprod_{j=0}^{t}A_j\biggr)h_{-1}
      +\sum_{s=0}^{t}\biggl(\rprod_{j=s+1}^{t}A_j\biggr)b_s.
\end{equation}
The recurrence and closed form are exact under the stated hold and grid
assumptions for every positive step size. For $L=\infty$ these are finite-prefix identities;
no convergence of states or infinite matrix products is asserted.
\end{theorem}

Appendix~\ref{app:zoh_proof} proves the result by variation of parameters
and telescoping.
The matrices in~\eqref{eq:zoh_matrices} are the standard ZOH matrices presented
by Mamba~1~\citep[Equation~(4)]{gu2024mamba}; its scan implementation combines the exact
exponential transition with the forcing approximation identified
after~\eqref{eq:exp_euler}.

\par\endgroup

\par\begingroup
\paragraph{The matrix-form specialization.}
Many architectures use forcing and readout maps of the \emph{matrix form}
\begin{equation}
  b(x) \;=\; B(x)\,x, \qquad
  f(x,h) \;=\; \operatorname{Re}_{\mathbb F}\!\big(C(x)h\big)+D(x)\,x,
  \label{eq:matmul_b}
\end{equation}
with $B:\R^d\to\mathbb F^{S\times d}$,
$C:\R^d\to\mathbb F^{O\times S}$ and $D:\R^d\to\R^{O\times d}$,
subject to Definition~\ref{def:cts_ssm}'s regularity assumptions.
A linear preprocessing map $g$ in $B(x)g(x)$ can be absorbed into $B$;
nonlinear $g$ need not admit a continuous factorization
$B(x)g(x)=\tilde B(x)x$ and is covered by the general forcing $b$.
With $A_t=e^{\Delta_t A(x_t)}$, set
\begin{equation}\label{eq:matmul_discretised}
  B_t \;:=\; \Delta_t\, \varphi\bigl(\Delta_t A(x_t)\bigr)\, B(x_t), \qquad C_t \;:=\; C(x_t), \qquad D_t \;:=\; D(x_t),
\end{equation}
so the discretized forcing is $b_t=B_t x_t$.
When $A(x_t)$ is invertible, the equivalent expression is
\[
 B_t=A(x_t)^{-1}\bigl(e^{\Delta_t A(x_t)}-I\bigr)B(x_t)
     =A(x_t)^{-1}(A_t-I)B(x_t).
\]
Define the \emph{sequence kernel}
\begin{equation}\label{eq:sequence_kernel_definition}
  W(t,s):=C_t\,\Phi(t,s)\,B_s\in\mathbb F^{O\times d},
  \qquad 0\le s\le t,\quad t\in\mathbb T_L.
\end{equation}
It combines the write at step $s$, the transfer to step $t$, and the read
at step $t$. Theorem~\ref{thm:zoh_exact} gives the recurrence and closed form
\begin{align}
  h_t &\;=\; A_t\, h_{t-1} \;+\; B_t\, x_t, &
  y_t &\;=\; \operatorname{Re}_{\mathbb F}(C_t h_t) \;+\; D_t\, x_t, \label{eq:abcd_recurrence_state}\\
  h_t &\;=\; \Phi(t,-1)\,h_{-1}
  \;+\; \sum_{s=0}^{t}\Phi(t,s)\,B_s\,x_s\,. \label{eq:abcd_telescoped_state}
\end{align}
Lowercase $b(\cdot), b_t$ denote the general forcing and uppercase $B(\cdot), B_t$ the matrix-form specialization.

Substituting the closed-form state into the readout gives the
\emph{data-dependent causal kernel representation}
\begin{equation}\label{eq:sequence_kernel}
  y_t \;=\; \operatorname{Re}_{\mathbb F}\!\left(
  C_t\,\Phi(t,-1)\,h_{-1}
  \;+\; \sum_{s=0}^{t} W(t,s)\, x_s\right) \;+\; D_t x_t.
\end{equation}
If $A_t,B_t,C_t,D_t$ are independent of the input sequence, the
input--output map is linear for $h_{-1}=0$ and affine for any fixed
$h_{-1}$; input-dependent coefficients can make it nonlinear.

\paragraph{Quadrature discretizations.}\phantomsection\label{sec:disc_alternatives}
Mamba~1--3 use the exponential transition $A_t=e^{\Delta_t A(x_t)}$
and approximate the forcing integral. To obtain these updates, hold
$A(x_t)$ fixed on the interval and choose a continuous forcing
$q_t:[\tau_{t-1},\tau_t]\to\mathbb F^S$ with endpoint values
$q_t^+=q_t(\tau_t)=b(x_t)$ for $t\in\mathbb T_L$ and
$q_t^-=q_t(\tau_{t-1})=b(x_{t-1})$ for $t\ge1$. We initialize the earlier forcing
sample by $q_0^-=q_0(\tau_{-1})=0$.
Variation of parameters gives (Appendix~\ref{app:zoh_proof})
\begin{equation}\label{eq:exp_euler_starting}
  h_t \;=\; A_t\, h_{t-1} \;+\; \int_{\tau_{t-1}}^{\tau_t} e^{(\tau_t - \tau) A(x_t)}\, q_t(\tau)\, d\tau .
\end{equation}
Approximating this integral by $\Delta_t$ times a weighted average of
the integrand's endpoint values gives the discrete update
\begin{equation}\label{eq:exp_trapezoidal_recurrence}
  h_t \;=\; A_t\, h_{t-1} \;+\; \Delta_t\Big[(1-\lambda_t)\, A_t\, q_t^{-} \;+\; \lambda_t\, q_t^{+}\Big],
  \qquad \lambda_t \in [0,1].
\end{equation}
Here $A_t$ transports the earlier sample; $q_0^-=0$ leaves only the
current sample at $t=0$. Following \citet{lahoti2026mamba3}, the weighted
family is exponential-trapezoidal, with the classical trapezoidal rule at $\lambda_t=\tfrac12$;
$\lambda_t=1$ is exponential-Euler, used by Mamba~1/2 with write
$\Delta_tB(x_t)x_t$. The latter also follows by retaining the leading
term of the exact ZOH forcing coefficient:
\begin{equation}\label{eq:exp_euler}
  \Delta_t\,\varphi\bigl(\Delta_t A(x_t)\bigr)
  =\Delta_t I+\tfrac12\Delta_t^2 A(x_t)+\bigO(\Delta_t^3)
  \approx\Delta_t I.
\end{equation}
For bounded $A(x_t)$ this gives $B_t\approx\Delta_t B(x_t)$.
Mamba~3 selects $\lambda_t=\operatorname{sigm}(\mathrm{proj}(x_t))$ per
head.\footnote{We use Mamba~3's terminology. In the exponential-integrator literature, exponential Euler instead uses the $\varphi_1$ coefficient in~\eqref{eq:zoh_matrices}, and exponential trapezoidal includes a $\varphi_2$ correction~\citep{hochbruck2010exponential}.}
For the local error orders relative to the chosen continuous forcing,
the required smoothness and endpoint-weight conditions, and their relation
to the learned gate, see \citet[Section~3.1, Remark~3 and Appendices~A.2--A.3]{lahoti2026mamba3}.

\paragraph{Other schemes.}
S4's bilinear (Tustin) transform~\citep{gu2022efficiently,gu2022s4d} approximates the transition as well as the forcing, and therefore changes the transfer operator. We restrict attention to the exponential schemes above.
\par\endgroup

\par\begingroup
\subsection{Fixed-basis diagonal equivalence}
\label{sec:simultaneous_diagonalization}

From here on we study discrete \emph{state recurrences} without requiring a
continuous-time origin. Their maps may depend on the step and the input
prefix.

A transition with off-diagonal entries in its supplied coordinates may still be diagonal in another fixed basis.  The restriction to a \emph{fixed} basis carries the whole content, since an input-dependent moving frame transforms every realization with invertible transitions into one with identity transitions (Lemma~\ref{lem:gauge_prefix}).

\begin{definition}
\label{def:field_ssm}
Fix $\mathbb F\in\{\R,\C\}$, dimensions $d,S,O\in\mathbb N$, a
horizon $L\in\mathbb N\cup\{\infty\}$ and an input-window bound
$\ell\in\mathbb N\cup\{\infty\}$. For $t\in\mathbb T_L$, write
\[
 x_{t;\ell}:=x_{\max\{0,t-\ell+1\}:t}\quad(\ell<\infty),
 \qquad x_{t;\infty}:=x_{0:t},\qquad
 m_{t,\ell}:=\min\{\ell,t+1\}.
\]
A \emph{discrete SSM realization $\Sigma$ over $\mathbb F$} has step maps
\[
 A_t^{\diamond}:(\R^d)^{m_{t,\ell}}\to\mathbb F^{S\times S},\qquad
 b_t^{\diamond}:(\R^d)^{m_{t,\ell}}\to\mathbb F^S,\qquad
 f_t^{\diamond}:(\R^d)^{m_{t,\ell}}\times\mathbb F^S\to\R^O,
\]
defined for every input $x\in(\R^d)^L$ and run from any initial state
$h_{-1}\in\mathbb F^S$ through
\begin{equation}
  h_t=A_t^{\diamond}(x_{t;\ell})\,h_{t-1}+b_t^{\diamond}(x_{t;\ell}),
  \qquad
  y_t=f_t^{\diamond}(x_{t;\ell},h_t).
  \label{eq:field_ssm_general}
\end{equation}
Write $\mathcal S^{\mathbb F}(\ell,S)$ for these realizations; the field
superscript is explicit, and an omitted $S$ means its fixed contextual value
or, if none is specified, the union over allowed dimensions. The window
bound applies to all three maps for every state argument of the readout;
maps factoring through $x_{t;\ell}$ are identified with their induced
window maps, so the classes are nested in $\ell$. Since $x_{t;L}=x_{0:t}$,
$\ell=L$ is unrestricted causal dependence, including $\ell=L=\infty$;
larger bounds add nothing. An $L=\infty$ realization is
\emph{streaming}; a finite one admits, but does not specify, a streaming
continuation (Appendix~\ref{app:theory_horizon}).

The subscript $t$ indexes the step and $\diamond$ marks a step map.
Fix an arbitrary $x\in(\R^d)^L$; even for an infinite sequence, each step
reads only its finite window $x_{t;\ell}$. Abbreviate the evaluated maps by
\[
 A_t:=A_t^{\diamond}(x_{t;\ell}),\qquad b_t:=b_t^{\diamond}(x_{t;\ell}).
\]
These are relative to the fixed input; otherwise we display the
diamond-marked maps and their arguments. Thus $h_t=A_th_{t-1}+b_t$ and
$y_t=f_t^{\diamond}(x_{t;\ell},h_t)$, with input dependence explicit. States
may be complex, but inputs and outputs remain real. System labels distinguish
the maps $A_t^\diamond[\Sigma],b_t^\diamond[\Sigma],f_t^\diamond[\Sigma]$
from evaluated $A_t[\Sigma],b_t[\Sigma]$; omit $[\Sigma]$ when clear.

The \emph{matrix form} writes the input directly, with $b_t=B_tx_t$:
\begin{equation}
 h_t=A_th_{t-1}+B_tx_t,\qquad
 y_t=\operatorname{Re}_{\mathbb F}(C_th_t)+D_tx_t.
 \label{eq:state_field_matmul}
\end{equation}
Here $M_t:=M_t^{\diamond}(x_{t;\ell})$ for $M\in\{B,C,D\}$, with
$B_t\in\mathbb F^{S\times d}$, $C_t\in\mathbb F^{O\times S}$ and
$D_t\in\R^{O\times d}$. Their input dependence is suppressed only
in the evaluated notation.
The \emph{matrix form with input features} fixes $R\in\mathbb N$ and a causal map
$g_t^{\diamond}:(\R^d)^{m_{t,\ell}}\to\R^R$. Write
$u_t:=g_t^{\diamond}(x_{t;\ell})$ and $b_t=B_tu_t$, giving
\[
 h_t=A_th_{t-1}+B_tu_t,\qquad
 y_t=\operatorname{Re}_{\mathbb F}(C_th_t)+D_tu_t,
 \qquad B_t\in\mathbb F^{S\times R},\quad D_t\in\R^{O\times R}.
\]
The coefficient maps still read the conditioning input $x_{t;\ell}$; they
need not factor through $u_t$. The feature construction is included in
$b_t^{\diamond}$ and $f_t^{\diamond}$; $u_t$ is an optional abbreviation.
Unless stated otherwise, the matrix form below means this feature form;
the direct-input case is $u_t=x_t$ and $R=d$.

\end{definition}

With finite $\ell<L$, a realization runs in memory independent
of $t$: the state plus a buffer of the last $\ell-1$ inputs. {At full-prefix dependence $\ell=L$, every realization with invertible
transitions admits an output-equivalent diagonal realization, even when
fixed-basis diagonal equivalence fails}
(Corollary~\ref{cor:membership_realisation}).

The case $\ell=1$ is \emph{token-local}: $x_{t;1}=x_t$.
In matrix form, $B_t^\diamond,C_t^\diamond,D_t^\diamond,g_t^\diamond$ obey
the same window bound. Explicit step dependence remains allowed.
\emph{Time sharing} instead uses one map each for $A,b,f$ at every step,
after fixed boundary padding of a finite window; this may specify a forcing
sample rather than a padding token. At window one, write
\begin{equation}
 A_t^\diamond(x_t)=A(x_t),\qquad
 b_t^\diamond(x_t)=b(x_t),\qquad
 f_t^\diamond(x_t,h)=f(x_t,h).
 \label{eq:token_local_class}
\end{equation}
Here $A,b,f$ denote discrete updates, distinct from the continuous-time
maps of Section~\ref{sec:continuous}; indexed notation remains valid.
\emph{Selective} maps may depend on the conditioning input, while
input-independent maps are included. Class membership does not require time
sharing.
ZOH updates~\eqref{eq:zoh_matrices} are token-local. The
endpoint-weighted rule~\eqref{eq:exp_trapezoidal_recurrence} keeps
token-local transitions, but its write reads $x_{t-1}$ through $q_t^-$ and
has window two (one if $\lambda_t\equiv1$).

\paragraph{Linear time-invariant systems.}
\phantomsection\label{par:lti}
A matrix-form realization with fixed $A,B,C,D$, independent of time and
input, is \emph{linear time-invariant (LTI)} in its driving features $u$;
with zero initial state its input--output map is linear and time-invariant.
For $u_t=x_t$, this is LTI in the system input itself. Nonlinear feature
generation need not preserve that property. Time sharing alone allows
selective or nonlinear maps, while a singleton realized transition family
fixes only $A$; neither condition alone makes the realization LTI.

For $x\in(\R^d)^L$, extend the transfer and kernel notation of
Section~\ref{sec:continuous} to the causal maps above. For
$t\in\mathbb T_L$ and $-1\le s\le t$, define the transfer
\begin{equation}
 \Phi[x](t,s):=\rprod_{j=s+1}^{t}A_j^{\diamond}(x_{j;\ell}),
 \qquad \Phi[x](t,t)=I.
 \label{eq:field_transfer}
\end{equation}
In the matrix form, define the sequence kernel
\begin{equation}
 W[x](t,s):=C_t^{\diamond}(x_{t;\ell})\,\Phi[x](t,s)\,
 B_s^{\diamond}(x_{s;\ell})\in\mathbb F^{O\times R},\qquad 0\le s\le t.
 \label{eq:field_sequence_kernel}
\end{equation}
Unless stated otherwise, evaluated coefficients, transfers and kernels
refer to a fixed arbitrary input $x\in(\R^d)^L$, with input arguments
displayed only when useful. We write $\Phi(t,s)=\rprod\nolimits_{j=s+1}^{t}A_j$
and $W(t,s)=C_t\Phi(t,s)B_s$. Finite induction gives the state and
matrix-form output:
\begin{equation}
 \begin{aligned}
  h_t&=\Phi(t,-1)\,h_{-1}+\sum_{s=0}^{t}\Phi(t,s)\,b_s,\\
  y_t&=\operatorname{Re}_{\mathbb F}\!\left(
       C_t\Phi(t,-1)h_{-1}+\sum_{s=0}^{t}W(t,s)u_s\right)+D_tu_t.
 \end{aligned}
  \label{eq:field_closed_form}
\end{equation}
For an LTI realization with zero initial state, $\Phi(t,s)=A^{t-s}$, so
\eqref{eq:field_closed_form} takes the convolutional form
\begin{equation}
 y=K*u+Du,\qquad (K*u)_t=\sum_{s=0}^{t}K_{t-s}u_s,\qquad
 K_\tau=\operatorname{Re}_{\mathbb F}\!\left(CA^{\tau}B\right).
 \label{eq:lti_convolution}
\end{equation}

The \emph{realized transition family} of a realization $\Sigma$ is
\begin{equation}
  \mathcal A_\Sigma
  :=\bigl\{A_t^{\diamond}[\Sigma](x_{t;\ell}):
  x\in(\R^d)^L,\ t\in\mathbb T_L\bigr\}
  \;\subseteq\;\mathbb F^{S\times S}.
  \label{eq:realised_transition_set}
\end{equation}

It discards order and multiplicity and need not be finite.

To discuss a real realization $\Sigma$ over $\C$, we extend its state
space from $\R^S$ to $\C^S$, retain its real $A_t,b_t$, and set
$f_t^{\diamond,\C}(x_{t;\ell},z):=f_t^{\diamond}(x_{t;\ell},\operatorname{Re}z)$.
This defines its \emph{complexification} $\Sigma^{\C}$; the realized
transition family is unchanged, now read over $\C$. Runs from real
initial states remain in the invariant subspace $\R^S\subset\C^S$ and
have the original states and outputs.

We write $\operatorname{Diag}_S(\mathbb F)\subset\mathbb F^{S\times S}$ for the diagonal matrices, $\operatorname{GL}_S(\mathbb F)$ for the invertible ones, $\operatorname{spec}(A)\subset\C$ for the spectrum, and $[A,A']:=AA'-A'A$ for the commutator. A family $\mathcal M\subseteq\mathbb F^{S\times S}$ is \emph{abelian} if $[A,A']=0$ for all $A,A'\in\mathcal M$, and \emph{non-abelian} otherwise. A square matrix is \emph{defective} if some eigenvalue has geometric multiplicity strictly below its algebraic multiplicity; Lemma~\ref{lem:transition_boundary}(i) states how diagonalizability depends on defectiveness and, for real matrices, on the field.

\par\endgroup

\par\begingroup
\begin{definition}
\label{def:state_similarity}
For a realization $\Sigma$ and $Q\in\operatorname{GL}_S(\mathbb F)$,
define $\Sigma^Q$ on the same input, state and output spaces and horizon
by the maps
\begin{equation}
 \begin{aligned}
 A_t^{\diamond}[\Sigma^Q](x_{t;\ell})
   &=Q^{-1}A_t^{\diamond}[\Sigma](x_{t;\ell})Q,\\
 b_t^{\diamond}[\Sigma^Q](x_{t;\ell})
   &=Q^{-1}b_t^{\diamond}[\Sigma](x_{t;\ell}),\\
 f_t^{\diamond}[\Sigma^Q](x_{t;\ell},z)
   &=f_t^{\diamond}[\Sigma](x_{t;\ell},Qz),
 \end{aligned}
 \label{eq:similarity_action}
\end{equation}
for every $x\in(\R^d)^L$, $t\in\mathbb T_L$ and $z\in\mathbb F^S$.
Pair each initial state $h_{-1}$ of $\Sigma$ with
$\hat h_{-1}=Q^{-1}h_{-1}$ for $\Sigma^Q$.
Two realizations are \emph{similar over $\mathbb F$}, written
$\Sigma\simeq_{\mathbb F}\Sigma'$, if $\Sigma'=\Sigma^Q$ for some
$Q\in\operatorname{GL}_S(\mathbb F)$; these transformations form a group
action, so similarity is an equivalence relation.
We call $Q$ a \emph{frame}. It is fixed, independent of time and input;
time-dependent frames are treated in Lemma~\ref{lem:gauge_prefix}.

For realizations $\Sigma,\Sigma'$ on the same input/output spaces and
horizon, write $y_t[\Sigma;x,h]$ for the output at step $t$ from initial
state $h$. They are \emph{output-equivalent} if an input-independent
bijection $\mathcal J$ between their initial-state spaces satisfies
\[
 y_t[\Sigma;x,h]=y_t[\Sigma';x,\mathcal J(h)]
 \qquad\text{for every }x,h\text{ and }t\in\mathbb T_L.
\]
When a correspondence $\mathcal J$ is specified, output equivalence refers to
that map; fixed similarity uses $\mathcal J(h)=Q^{-1}h$. Output equivalence
alone imposes no fixed relation between later states.
\end{definition}

\begin{definition}
\label{def:diagonal_equivalent}
A realization is \emph{diagonal over $\mathbb F$} if every realized
transition lies in $\operatorname{Diag}_S(\mathbb F)$, and
\emph{diagonal-equivalent over $\mathbb F$} if it is similar over
$\mathbb F$ to a diagonal realization. For window bound $\ell$ and
state dimension $S$, define
\begin{equation}
 \mathcal D^{\mathbb F}(\ell,S)
 :=\{\Sigma\in\mathcal S^{\mathbb F}(\ell,S):
       \Sigma^Q\text{ is diagonal for some }Q\in\operatorname{GL}_S(\mathbb F)\}.
 \label{eq:diagonal_token_local_class}
\end{equation}
Such a $Q$ is a \emph{witness} for membership, and $\Sigma^Q$ is a
\emph{diagonal representative}. We omit $S$ under the convention of
Definition~\ref{def:field_ssm}; without a specified dimension, this means
the union over $S\in\mathbb N$. The case $\ell=L$ allows unrestricted
causal input dependence.
For a real realization $\Sigma$, the shorthand
$\Sigma\in\mathcal D^{\C}(\ell,S)$ means
$\Sigma^{\C}\in\mathcal D^{\C}(\ell,S)$ under the
complexification defined above, which preserves the window bound and
uses $S$ complex coordinates.
\end{definition}

For fixed field, state dimension and horizon, the window classes satisfy
\[
 \begin{aligned}
 \mathcal C^{\mathbb F}(\ell_1,S)
 &\subseteq\mathcal C^{\mathbb F}(\ell_2,S)
 \quad(\ell_1\le\ell_2),\qquad \mathcal C\in\{\mathcal S,\mathcal D\},\\
 \mathcal D^{\mathbb F}(\ell,S)
 &=\mathcal S^{\mathbb F}(\ell,S)
 \cap\mathcal D^{\mathbb F}(L,S).
 \end{aligned}
\]

\begin{lemma}
\label{lem:similarity_invariance}
Let $\Sigma$ be a general causal realization and
$Q\in\operatorname{GL}_S(\mathbb F)$. For every input $x\in(\R^d)^L$
and initial state $h_{-1}\in\mathbb F^S$, let $(h_t,y_t)$ be its run
and $(\hat h_t,\hat y_t)$ the run of $\Sigma^Q$ from
$\hat h_{-1}=Q^{-1}h_{-1}$. Then, for every $t\in\mathbb T_L$,
\[
 \hat h_t=Q^{-1}h_t,\qquad \hat y_t=y_t.
\]
In the matrix form of Definition~\ref{def:field_ssm}, use the induced
coefficients $\hat B_t=Q^{-1}B_t$, $\hat C_t=C_tQ$ and $\hat D_t=D_t$.
Their kernels $\hat W[x]$ and $W[x]$ agree: suppressing the fixed input,
\begin{equation}
 \hat W(t,s)
 =\bigl(C_tQ\bigr)\bigl(Q^{-1}\Phi(t,s)Q\bigr)\bigl(Q^{-1}B_s\bigr)
 =W(t,s),\qquad 0\le s\le t.
 \label{eq:similarity_kernel_invariance}
\end{equation}
Moreover, for every window bound $\ell$,
\begin{equation}
 \Sigma\in\mathcal C^{\mathbb F}(\ell,S)
 \iff\Sigma^Q\in\mathcal C^{\mathbb F}(\ell,S),
 \qquad \mathcal C\in\{\mathcal S,\mathcal D\}.
 \label{eq:similarity_class_invariance}
\end{equation}
Time sharing is also preserved in both directions. In particular, a
diagonal representative retains the window bound, state dimension and
any time sharing of the original realization.
\end{lemma}

\begin{proof}
Induction from $\hat h_{-1}=Q^{-1}h_{-1}$ gives
$\hat h_t=(Q^{-1}A_tQ)\hat h_{t-1}+Q^{-1}b_t=Q^{-1}h_t$.
The general readout then satisfies
\[
 \hat y_t=f_t^{\diamond}[\Sigma^Q](x_{t;\ell},\hat h_t)
 =f_t^{\diamond}[\Sigma](x_{t;\ell},Q\hat h_t)
 =f_t^{\diamond}[\Sigma](x_{t;\ell},h_t)=y_t.
\]
For the matrix kernel, the transitions telescope to
$\hat\Phi(t,s)=Q^{-1}\Phi(t,s)Q$, giving
\eqref{eq:similarity_kernel_invariance}.
The fixed maps $Q^{-1}A_t^\diamond Q$, $Q^{-1}b_t^\diamond$ and
$f_t^\diamond(x_{t;\ell},Qh)$ introduce no new input arguments or step
dependence; the inverse frame gives the converse for window dependence
and time sharing. Composition of fixed frames preserves the existence
of a diagonalizing frame, giving the $\mathcal D^{\mathbb F}$ equivalence.
\end{proof}
Output equivalence alone need not preserve either class membership or the
input-window bound (Corollary~\ref{cor:realification_window_classes}).

\begin{theorem}
\label{thm:simultaneous_diagonalization}
For $\Sigma\in\mathcal S^{\mathbb F}(\ell,S)$,
the following conditions are equivalent:
\begin{equation}
 \begin{aligned}
 \Sigma\in\mathcal D^{\mathbb F}(\ell,S)
 &\Longleftrightarrow
 \mathcal A_\Sigma\text{ is simultaneously diagonalizable over }\mathbb F\\
 &\Longleftrightarrow
 [A,A']=0\text{ and }A\text{ is }\mathbb F\text{-diagonalizable}
 \quad\forall A,A'\in\mathcal A_\Sigma.
 \end{aligned}
 \label{eq:simultaneous_algebraic_test}
\end{equation}
Under these conditions, there is a single $Q\in\operatorname{GL}_S(\mathbb F)$
with $Q^{-1}AQ\in\operatorname{Diag}_S(\mathbb F)$ for all
$A\in\mathcal A_\Sigma$. Every such $Q$ gives the diagonal representative
$\Sigma^Q$, with $\hat h_t=Q^{-1}h_t$ and the same outputs for
corresponding initial states. For a fixed arbitrary input $x\in(\R^d)^L$, write
\[
 \Lambda_t:=A_t[\Sigma^Q]=Q^{-1}A_t[\Sigma]Q
 \in\operatorname{Diag}_S(\mathbb F).
\]
In the matrix form of Definition~\ref{def:field_ssm}, its recurrence is
\begin{align}
  \hat h_t&=\Lambda_t\hat h_{t-1}+Q^{-1}B_tu_t,
  &y_t&=\operatorname{Re}_{\mathbb F}
       \!\left(C_tQ\,\hat h_t\right)+D_tu_t,
  \label{eq:diagonal_equivalent_recurrence}
\end{align}
with the feedthrough $D_t$ untouched. Within the declared window class, membership is decided by transitions
alone; forcing and readout are otherwise unrestricted.

\end{theorem}

\begin{proof}
For the first equivalence, Definition~\ref{def:diagonal_equivalent} requires
one $Q$ for which $\Sigma^Q$ has diagonal transitions on every input and
step. By~\eqref{eq:similarity_action}, this is exactly
$Q^{-1}AQ\in\operatorname{Diag}_S(\mathbb F)$ for every
$A\in\mathcal A_\Sigma$.

For the second equivalence, a common diagonalizing basis makes every
member diagonalizable and every pair commute. Conversely, suppose the
members commute and are individually diagonalizable over $\mathbb F$.
Choose $A_1,\ldots,A_m\in\mathcal A_\Sigma$ spanning
$\operatorname{span}_{\mathbb F}\mathcal A_\Sigma$; this is possible with
$m\le S^2$, even if the family is infinite. The finite commuting family
$A_1,\ldots,A_m$ has a common diagonalizing basis
$Q$~\citep[Thm.~1.3.21 and Cor.~1.3.30]{horn2013matrix}. Every other member is a linear
combination of these matrices, so
\[
 A=\sum_{k=1}^{m}c_kA_k
 \quad\Longrightarrow\quad
 Q^{-1}AQ=\sum_{k=1}^{m}c_k\bigl(Q^{-1}A_kQ\bigr)
 \in\operatorname{Diag}_S(\mathbb F).
\]
Thus the same $Q$ diagonalizes the entire realized family. Substituting
this $Q$ into~\eqref{eq:similarity_action} gives the stated diagonal
representative and~\eqref{eq:diagonal_equivalent_recurrence}.
\end{proof}

Commutativity is field-independent; diagonalizability is checked over
$\mathbb F$. Appendix~\ref{app:theory_horizon} gives the finite-horizon
consequences.

For an SSM obtained by ZOH discretization as in Section~\ref{sec:continuous},
a common eigenbasis of its continuous-time generators is inherited by
its discrete transitions: if $A(x)=Q\Lambda(x)Q^{-1}$ with diagonal
$\Lambda(x)$ and fixed $Q$, then
\[
 A_t=e^{\Delta_t A(x_t)}=Qe^{\Delta_t\Lambda(x_t)}Q^{-1}.
\]
Thus the resulting discrete realization belongs to
$\mathcal D^{\mathbb F}(1,S)$, with the same witness $Q$.
It is also time-shared when its step-size and coefficient laws are shared;
a nonuniform prescribed time grid need not have this property.

\begin{remark}
\label{rem:transfer_bound}
Let $\Sigma\in\mathcal D^{\mathbb F}(\ell,S)$ with witness $Q$, so that
$A_t=Q\Lambda_tQ^{-1}$ with diagonal $\Lambda_t$. Then
\[
 \|\Phi(t,s)\|_2\le\kappa_2(Q)\,\max_{n}\prod_{r=s+1}^{t}\bigl|(\Lambda_r)_{nn}\bigr|,
 \qquad \kappa_2(Q)=\|Q\|_2\,\|Q^{-1}\|_2 .
\]
\end{remark}
Eigenvalue moduli at most one keep the transfer, and with it the gradients through the
carried state, bounded up to the fixed factor $\kappa_2(Q)$. A bounded state under bounded writes
$b_t$ needs more: moduli at most some $\rho<1$ give
$\|h_t\|_2\le\kappa_2(Q)\bigl(\|h_{-1}\|_2+\sup_s\|b_s\|_2/(1-\rho)\bigr)$, whereas with moduli equal
to one, constant forcing can grow the state linearly. Practical parameterizations keep moduli at most one: decays
$e^{\Delta a}$ with $a<0$ and gates in $(0,1)$ give moduli below one, and
rotations give modulus one. For non-abelian families, moduli alone do not
bound products of non-normal factors; delta-rule transitions
$I-\beta kk^{\top}$ with unit-norm keys and $\beta\in[0,2]$ are symmetric
with spectrum in $[-1,1]$, hence non-expansive.

\paragraph{Fixed and moving frames.}
Allowing the frame to depend on the prefix changes the question: it can
move temporal dependence from the transition into the write and read maps.

\begin{lemma}
\label{lem:gauge_prefix}
Let $\Sigma\in\mathcal S^{\mathbb F}(\ell,S)$ have
horizon $L$. For each input $x$, let
$Q_t[x]\in\operatorname{GL}_S(\mathbb F)$ depend only on $x_{0:t}$,
with $Q_{-1}$ independent of the input. Fix an arbitrary
$x\in(\R^d)^L$ and use the evaluated coefficients $A_t,b_t$ and, in the
matrix form, $B_t,C_t,D_t$; write $Q_t:=Q_t[x]$.
For $t\in\mathbb T_L$, set $\tilde h_t:=Q_t^{-1}h_t$ and define the
transformed readout by
\[
 \tilde f_t^{\diamond}(x_{t;L},z)
 :=f_t^{\diamond}(x_{t;\ell},Q_t[x]z).
\]
Then the transformed recurrence and output are
\begin{equation}
  \tilde h_t=\big(Q_t^{-1}A_tQ_{t-1}\big)\tilde h_{t-1}+Q_t^{-1}b_t,
  \qquad
  \tilde y_t=\tilde f_t^{\diamond}(x_{t;L},\tilde h_t)=y_t,
  \qquad
  \tilde h_{-1}=Q_{-1}^{-1}h_{-1}.
  \label{eq:frame_change}
\end{equation}
The transformed realization lies in $\mathcal S^{\mathbb F}(L,S)$
on the same horizon $L$. Here $f_t^{\diamond}$ and $y_t$ belong to $\Sigma$, while
$\tilde f_t^{\diamond}$ and $\tilde y_t$ belong to the transformed
realization. Its
transfer is $\tilde\Phi(t,s)=Q_t^{-1}\Phi(t,s)Q_s$; in the matrix form,
\[
 W(t,s)=(C_tQ_t)\,\tilde\Phi(t,s)\,(Q_s^{-1}B_s).
\]

Assume now that every realized transition is invertible, and set
$\Pi_t:=\Phi(t,-1)$, with $\Pi_{-1}:=I$.
\begin{enumerate}
\item[\textnormal{(i)}] \emph{Prefix-frame construction.}
The choice $Q_t:=\Pi_t$ gives $\Phi(t,s)=\Pi_t\Pi_s^{-1}$ and
turns~\eqref{eq:frame_change} into a
\emph{diagonal realization with identity transitions}:
\[
 \tilde h_t=\tilde h_{t-1}+\Pi_t^{-1}b_t
 =h_{-1}+\sum_{s=0}^{t}\Pi_s^{-1}b_s,
 \qquad y_t=f_t^{\diamond}(x_{t;\ell},\Pi_t\tilde h_t).
\]
In the matrix form, $W(t,s)=(C_t\Pi_t)(\Pi_s^{-1}B_s)$.
This realization is in $\mathcal D^{\mathbb F}(L,S)$;
the prefix products now enter the write and read maps.
\item[\textnormal{(ii)}] \emph{Characterization of diagonalizing frames.}
Any frame producing diagonal transitions
$\Lambda_t:=Q_t^{-1}A_tQ_{t-1}$ must satisfy
\begin{equation}
 Q_t=\Pi_t\,Q_{-1}\Big(\rprod_{j=0}^{t}\Lambda_j\Big)^{-1},
 \qquad \Lambda_t\in\operatorname{Diag}_S(\mathbb F).
 \label{eq:diag_frame}
\end{equation}
Conversely, any fixed $Q_{-1}\in\operatorname{GL}_S(\mathbb F)$ and
causal invertible diagonal maps $\Lambda_t$ define a diagonalizing frame
by~\eqref{eq:diag_frame}. The prefix construction chooses
$Q_{-1}=I$, $\Lambda_t=I$.
\end{enumerate}
Independently of invertibility, a constant diagonalizing frame exists
exactly when $\mathcal A_\Sigma$ is simultaneously diagonalizable over
$\mathbb F$ (Theorem~\ref{thm:simultaneous_diagonalization}).
\end{lemma}

\begin{proof}
Substitute $h_t=Q_t\tilde h_t$ into $h_t=A_th_{t-1}+b_t$ to obtain
\eqref{eq:frame_change}. Its readout is unchanged since
$Q_t\tilde h_t=h_t$. Multiplying the transformed transitions cancels
adjacent frame factors, giving $\tilde\Phi(t,s)=Q_t^{-1}\Phi(t,s)Q_s$
and the matrix-kernel identity.

For (i), $\Pi_t=A_t\Pi_{t-1}$ gives
$\Pi_t^{-1}A_t\Pi_{t-1}=I$; summing the resulting recurrence gives the
stated closed form. For (ii), writing $A_t=Q_t\Lambda_tQ_{t-1}^{-1}$
and telescoping gives
$\Pi_t=Q_t\bigl(\rprod\nolimits_{j=0}^{t}\Lambda_j\bigr)Q_{-1}^{-1}$.
Each $\Lambda_j$ is invertible because $A_j$ and the frames are invertible,
so rearranging gives~\eqref{eq:diag_frame}. Conversely, if
$P_t:=\rprod_{j=0}^t\Lambda_j$, $P_{-1}:=I$, and $Q_t=\Pi_tQ_{-1}P_t^{-1}$, then
$Q_t^{-1}A_tQ_{t-1}=P_tP_{t-1}^{-1}=\Lambda_t$; the frame is causal
and invertible. Finally, setting $Q_t\equiv Q$
reduces diagonalization to $Q^{-1}AQ\in\operatorname{Diag}_S(\mathbb F)$
for every realized transition $A$.
\end{proof}

\begin{corollary}
\label{cor:membership_realisation}
Every $\Sigma\in\mathcal S^{\mathbb F}(\ell,S)$ with
invertible transitions has the output-equivalent identity-transition
realization $\widetilde\Sigma\in\mathcal D^{\mathbb F}(L,S)$
of Lemma~\ref{lem:gauge_prefix}(i), on the same horizon $L$.
The construction need not preserve a smaller window bound or time sharing.
If $\Sigma\in\mathcal D^{\mathbb F}(L,S)$ already, both
realizations lie in that full-window diagonal class; this does not make
them fixed-frame similar.
\end{corollary}

Fixed similarity preserves window bounds; a prefix frame guarantees
only full-prefix dependence. Thus for unrestricted causal maps ($\ell=L$) with invertible transitions,
noncommutativity excludes
fixed diagonal equivalence but alone gives no input--output expressivity
separation. Its identity-transition scan still requires forming and applying
$\Pi_t,\Pi_t^{-1}$ at controlled cost and conditioning. See
Appendix~\ref{app:gauge_examples} for per-coordinate potentials and the
accumulator form.

\paragraph{Leaving the class.}
Negating~\eqref{eq:simultaneous_algebraic_test} leaves exactly two exits from
the class. Noncommutativity and, over $\R$, nonreal spectrum survive
sufficiently small perturbations; defectiveness need not.

\begin{lemma}
\label{lem:transition_boundary}
For a realization $\Sigma$ over $\mathbb F$ with realized transition
family $\mathcal A_\Sigma$:
\begin{enumerate}
\item[\textnormal{(i)}] \emph{Individual obstruction --- $\mathbb F$-dependent.}  If some
$A\in\mathcal A_\Sigma$ is not diagonalizable over $\mathbb F$, then
$\Sigma\notin\mathcal D^{\mathbb F}(L)$, even when $\mathcal A_\Sigma$ is
abelian.  For a real matrix,
	\begin{align*}
	  A\text{ is diagonalizable over }\C
	  &\iff A\text{ is nondefective},\\
	  A\text{ is diagonalizable over }\R
	  &\iff A\text{ is nondefective and }
	  \operatorname{spec}(A)\subseteq\R.
	\end{align*}
Thus defectiveness obstructs both fields, whereas nonreal spectrum alone
obstructs only real diagonalization.  These are individual obstructions;
whole-family membership additionally requires pairwise commutativity.
\item[\textnormal{(ii)}] \emph{Pairwise obstruction --- field-independent and robust.}  If
$\mathcal A_\Sigma$ is non-abelian, that is,
$[A,A']\ne0$ for some $A,A'\in\mathcal A_\Sigma$, then
$\Sigma\notin\mathcal D^{\mathbb F}(L)$ for both fields.
Moreover, the pair $(A,A')$ has a neighborhood in
$(\mathbb F^{S\times S})^2$ containing no commuting --- hence no
simultaneously diagonalizable --- pair: noncommutativity survives every
sufficiently small perturbation of the realized transitions.
\end{enumerate}
\end{lemma}

\begin{proof}
The exclusions follow by the contrapositive of
Theorem~\ref{thm:simultaneous_diagonalization}. The field criteria in part~(i) are standard
\citep[Thm.~1.3.29 and p.~77]{horn2013matrix}.
The robustness claim in part~(ii) holds because
the commutator map $(A,A')\mapsto[A,A']$ is continuous
and nonzero at $(A,A')$.
\end{proof}

\begin{remark}
\label{rem:constant_transition_boundary}
For a realization with constant transition $A$ and readouts continuous in
the state, diagonal realizations can approximate its outputs on each
fixed finite input run, from each prescribed initial state. Over $\C$
this holds for every $A$; a real realization is first complexified.
Approximation of a real $A$ by real-diagonalizable transitions is possible
exactly when $A$ has real spectrum; this spectral condition concerns
transition approximation, not arbitrary output approximation. The approximants retain the state
width over the chosen field, input-window bound and any time sharing.
This is finite-run output convergence, not exact fixed similarity,
uniform infinite-horizon approximation, or convergence of diagonal
coordinates or parameters. The construction is in
Appendix~\ref{app:diagonal_approximation}.
\end{remark}
\par\endgroup

\par\begingroup
\subsection{Complex-diagonal SSMs and real diagonal evaluation}
\label{sec:theory_complex}

\begin{definition}
\label{def:complex_ssm}
Specialize Definition~\ref{def:field_ssm} to a complex realization with
$K\in\mathbb N$ state coordinates, window bound $\ell$, and diagonal
transitions. In its matrix
form with input features $u_t=g_t^{\diamond}(x_{t;\ell})\in\R^R$, it has
initial state $z_{-1}\in\C^K$, output $y_t\in\R^O$, and recurrence
\begin{equation}
 \begin{aligned}
 z_t&=\Lambda_t z_{t-1}+B_t^{\C}u_t,
 &\Lambda_t&=\operatorname{diag}_{k=1}^K
       \!\big(\rho_t^{(k)}e^{i\theta_t^{(k)}}\big),\\
 y_t&=\operatorname{Re}(C_t^{\C}z_t)+D_tu_t,
 &\rho_t^{(k)}&\ge0,\qquad\theta_t^{(k)}\in\R,
 \end{aligned}
 \label{eq:complex_ssm}
\end{equation}
where $B_t^{\C}\in\C^{K\times R}$, $C_t^{\C}\in\C^{O\times K}$ and
$D_t\in\R^{O\times R}$. The superscript $\C$ identifies the complex
representation. The polar form allows every complex
diagonal entry.
\end{definition}

For any complex realization $\Sigma$ with $K$ state coordinates, its
\emph{realification} $\Sigma_{\R}$ has state $\chi_K(h_t)$, initial state
$\chi_K(h_{-1})$, and maps $\mathfrak R\circ A_t^\diamond$,
$\chi_K\circ b_t^\diamond$, and
$f_t^\diamond(x_{t;\ell},\chi_K^{-1}(v))$. By~\eqref{eq:realification},
it preserves outputs, the window bound and time sharing.

Using the maps \hyperref[eq:realification]{$\chi_K$ and $\mathfrak R$}
of Section~\ref{sec:continuous}, define the real write/read blocks and
rotation matrix by
\begin{equation}
 \begin{gathered}
 B_t^{(k)}=\begin{pmatrix}\operatorname{Re}(B_t^{\C})_{k,:}\\
                         \operatorname{Im}(B_t^{\C})_{k,:}\end{pmatrix},
 \quad
 C_t^{(k)}=\begin{pmatrix}\operatorname{Re}(C_t^{\C})_{:,k}&
                         -\operatorname{Im}(C_t^{\C})_{:,k}\end{pmatrix},\\
 \mathcal R(\theta)=\mathfrak R(e^{i\theta})
 =\begin{pmatrix}\cos\theta&-\sin\theta\\
                  \sin\theta&\cos\theta\end{pmatrix}.
 \end{gathered}
 \label{eq:complex_real_maps}
\end{equation}
For all $\theta,\theta'\in\R$,
\[
 \mathcal R(\theta)\mathcal R(\theta')=\mathcal R(\theta+\theta'),
 \qquad
 \mathcal R(\theta)^{-1}=\mathcal R(\theta)^\top=\mathcal R(-\theta).
\]
Here $B_t^{(k)}\in\R^{2\times R}$ and $C_t^{(k)}\in\R^{O\times2}$.

For $\Sigma\in\mathcal D^{\C}(\ell,K)$ and a fixed diagonalizing frame,
Theorem~\ref{thm:rotation_fold} gives two real forms with $2K$
coordinates: a recurrence on rotation--scaling pairs, and a real diagonal
recurrence with the rotations folded into the write and read maps, which
Mamba~3 calls the \emph{RoPE trick}~\citep{lahoti2026mamba3}.

\begin{theorem}
\label{thm:rotation_fold}
Let $\Sigma\in\mathcal D^{\C}(\ell,K)$ be a complex
realization with state $w_t$, output $y_t$, evaluated coefficients
$A_t,b_t$ and readout $f_t^\diamond$. Fix a diagonalizing frame $Q$.
Its diagonal representative $\Sigma_D:=\Sigma^Q$ has state
$z_t=Q^{-1}w_t$, the same outputs $y_t$, and transition
$\Lambda_t=Q^{-1}A_tQ=\operatorname{diag}_k
(\rho_t^{(k)}e^{i\theta_t^{(k)}})$, in the polar notation of
Definition~\ref{def:complex_ssm}. Write $\Sigma_{\R}$ and
$(\Sigma_D)_{\R}$ for their realifications. The following real realizations have
$S=2K$ state coordinates and the same outputs for every input and
corresponding initial states. Their state correspondences are fixed
in (i) and generally moving in (ii).

\textnormal{(i)} \emph{Realification.}
Set $h_t=\chi_K(z_t)=\chi_K(Q^{-1}w_t)$, including at $t=-1$.
The realification $(\Sigma_D)_{\R}$, called its
\emph{rotational real representation}, has maps
\[
 \begin{gathered}
 \hat A_t=\mathfrak R(\Lambda_t)
   =\bigoplus_{k=1}^K\rho_t^{(k)}\mathcal R(\theta_t^{(k)}),
 \qquad \hat b_t=\chi_K(Q^{-1}b_t),\\
 \hat f_t^\diamond(x_{t;\ell},h)
   =f_t^\diamond(x_{t;\ell},Q\chi_K^{-1}(h)),
 \end{gathered}
\]
and recurrence
\[
 h_t=\hat A_th_{t-1}+\hat b_t,\qquad
 \hat y_t=\hat f_t^\diamond(x_{t;\ell},h_t)=y_t.
\]
It is fixed-similar over $\R$ to $\Sigma_{\R}$, through
$\mathfrak R(Q)$.

When $\Sigma$ is in matrix form, $\Sigma_D$ has the form
\eqref{eq:complex_ssm}. Form $B_t^{(k)},C_t^{(k)}$ from its coefficients
using~\eqref{eq:complex_real_maps}, retaining the original features and
feedthrough (Lemma~\ref{lem:similarity_invariance}). The pair coordinates
$h_t^{(k)}=\chi_1(z_t^{(k)})$ then satisfy
\begin{align}
 h_t^{(k)}&=\rho_t^{(k)}\mathcal R(\theta_t^{(k)})h_{t-1}^{(k)}
                  +B_t^{(k)}u_t,\label{eq:rotssm_state}\\
 y_t&=\sum_{k=1}^K C_t^{(k)}h_t^{(k)}+D_tu_t.
 \label{eq:paired_read}
\end{align}
For an already-diagonal $\Sigma$, one may choose $Q=I$.

\textnormal{(ii)} \emph{Real diagonal recurrence.}
Define
\begin{equation}
 \Theta_t^{(k)}=\sum_{s=0}^t\theta_s^{(k)},\qquad
 \Theta_{-1}^{(k)}=0,\qquad
 Q_t=\bigoplus_{k=1}^K\mathcal R(\Theta_t^{(k)}).
 \label{eq:accum_angle}
\end{equation}
The folded realization $\widetilde\Sigma$ has state
$\tilde h_t=Q_t^{-1}h_t$, initial state
$\tilde h_{-1}=h_{-1}=\chi_K(Q^{-1}w_{-1})$ and maps
\[
 \begin{gathered}
 \tilde A_t=\bigoplus_{k=1}^K\rho_t^{(k)}I_2,
 \qquad \tilde b_t=Q_t^{-1}\hat b_t,\\
 \tilde f_t^\diamond(x_{t;L},h)
   =f_t^\diamond(x_{t;\ell},Q\chi_K^{-1}(Q_th)).
 \end{gathered}
\]
Thus
\[
 \tilde h_t=\tilde A_t\tilde h_{t-1}+\tilde b_t,\qquad
 \tilde y_t=\tilde f_t^\diamond(x_{t;L},\tilde h_t)=y_t.
\]
The frame $Q_t$ may depend on the input prefix, while $Q$ and $\chi_K$
are fixed. Given $Q_t$, the complex states are recovered by
\[
 z_t=\chi_K^{-1}(Q_t\tilde h_t),\qquad w_t=Qz_t.
\]

In matrix form, rotate the write and read blocks of (i) as
\begin{equation}
 \tilde B_t^{(k)}=\mathcal R(-\Theta_t^{(k)})B_t^{(k)},\qquad
 \tilde C_t^{(k)}=C_t^{(k)}\mathcal R(\Theta_t^{(k)}).
 \label{eq:folded_proj}
\end{equation}
With $\tilde h_t^{(k)}=\mathcal R(-\Theta_t^{(k)})h_t^{(k)}$, the
recurrence becomes
\begin{equation}
 \begin{aligned}
 \tilde h_t^{(k)}&=\rho_t^{(k)}\tilde h_{t-1}^{(k)}
                        +\tilde B_t^{(k)}u_t,
 &\tilde h_{-1}^{(k)}&=h_{-1}^{(k)},\\
 y_t&=\sum_{k=1}^K\tilde C_t^{(k)}\tilde h_t^{(k)}+D_tu_t.
 \end{aligned}
 \label{eq:folded_scan}
\end{equation}
Each pair shares one real nonnegative multiplier; no rotation is
performed inside the recurrence.
\end{theorem}

\begin{proof}
For general $\Sigma$, apply Lemma~\ref{lem:similarity_invariance} with
$Q$ and then realify using~\eqref{eq:realification}. Preservation of
products and inverses, together with
$\chi_K(Q^{-1}w)=\mathfrak R(Q)^{-1}\chi_K(w)$, gives the maps and
state/output correspondence in (i). The matrix readout specialization follows from
$\operatorname{Re}(cz)=(\operatorname{Re}c,-\operatorname{Im}c)\chi_1(z)$.
Part (ii) applies the moving-frame identity~\eqref{eq:frame_change} of
Lemma~\ref{lem:gauge_prefix} with this $Q_t$. Explicitly,
$\Theta_t^{(k)}=\Theta_{t-1}^{(k)}+\theta_t^{(k)}$ gives
\[
 \mathcal R(-\Theta_t^{(k)})\,
 \rho_t^{(k)}\mathcal R(\theta_t^{(k)})\,
 \mathcal R(\Theta_{t-1}^{(k)})=\rho_t^{(k)}I_2.
\]
Substitution gives the folded write and read; $\Theta_{-1}^{(k)}=0$
preserves the initial state. No inverse of a transition is needed.
\end{proof}

\paragraph{Relative rotations and RoPE.}
In matrix form, $(\Sigma_D)_{\R}$ and $\widetilde\Sigma$ from
Theorem~\ref{thm:rotation_fold}(i)--(ii) have the same sequence kernel on the features.
Writing $\Phi_{\C}(t,s)=\prod_{r=s+1}^t\Lambda_r$ for the complex
diagonal transfer,
\[
 \begin{aligned}
 W_{\mathrm{real}}(t,s)
 &=\operatorname{Re}\!\big(C_t^{\C}\Phi_{\C}(t,s)B_s^{\C}\big)\\
 &=\sum_{k=1}^K\Big(\prod_{r=s+1}^t\rho_r^{(k)}\Big)
 \underbrace{C_t^{(k)}\mathcal R(\Theta_t^{(k)}-\Theta_s^{(k)})B_s^{(k)}}_
 {\displaystyle\tilde C_t^{(k)}\tilde B_s^{(k)}},\qquad 0\le s\le t.
 \end{aligned}
\]
The read at $t$ and write at $s$ interact through
$\Theta_t^{(k)}-\Theta_s^{(k)}=\sum_{r=s+1}^t\theta_r^{(k)}$.
Standard RoPE uses prescribed, input-independent frequencies, so its
relative angles depend only on $t-s$~\citep{su2021roformer}; here the
increments may depend on the input, making the relative rotation
content-dependent, which learned fixed frequencies would not.

Mamba~3 ties the modulus across pairs while retaining per-pair angles
\citep{lahoti2026mamba3}; Theorem~\ref{thm:rotation_fold} does not require
that tying.

Appendix~\ref{app:transition_boundary_proofs} gives the fixed real
block characterization of complex diagonal equivalence, with scalar
coordinates and rotation--scaling pairs at the original real dimension.

\begin{corollary}
\label{cor:realification_window_classes}
Let $\Sigma\in\mathcal D^{\C}(\ell,K)$ be a complex
realization of horizon $L$, and let $\Sigma_D=\Sigma^Q$ be a diagonal
representative. Write $\Sigma_{\R}$ and $(\Sigma_D)_{\R}$ for their
realifications, and $\widetilde\Sigma$ for the rotation fold in
Theorem~\ref{thm:rotation_fold}(ii). Then these realizations,
related by fixed similarity ($Q$, $\mathfrak R(Q)$), realification
($\R$), complexification ($\C$) and the fold, lie in the classes shown:
\[
\begin{tikzpicture}[baseline=(c.base),every node/.style={inner sep=3pt},
  >=stealth, lab/.style={font=\scriptsize}]
 \node (a) at (0,0) {$\Sigma\in\mathcal D^{\C}(\ell,K)$};
 \node (b) at (5.2,0) {$\Sigma_D\in\mathcal D^{\C}(\ell,K)$};
 \node (f) at (10.0,0) {$\widetilde\Sigma\in\mathcal D^{\R}(L,2K)$};
 \node (c) at (0,-1.5) {$\Sigma_{\R}\in\mathcal S^{\R}(\ell,2K)$};
 \node (d) at (5.2,-1.5) {$(\Sigma_D)_{\R}\in\mathcal S^{\R}(\ell,2K)$};
 \node (e) at (0,-3.0) {$(\Sigma_{\R})^{\C}\in\mathcal D^{\C}(\ell,2K)$};
 \draw[<->] (a) -- node[lab,above]{$Q$} (b);
 \draw[->] (b) -- node[lab,above]{fold} (f);
 \draw[->] (a) -- node[lab,left]{$\R$} (c);
 \draw[->] (b) -- node[lab,right]{$\R$} (d);
 \draw[<->] (c) -- node[lab,above]{$\mathfrak R(Q)$} (d);
 \draw[->] (c) -- node[lab,left]{$\C$} (e);
\end{tikzpicture}
\]%
All have identical outputs under the initial-state correspondences of
the theorem; $(\Sigma_{\R})^{\C}$ reproduces these outputs from the
corresponding real initial states. If some realized transition of $\Sigma$
has a nonreal eigenvalue, neither realification belongs to
$\mathcal D^{\R}(\ell',2K)$ for any window bound $\ell'$.
The folded realization need not
retain the original window bound or time sharing.
\end{corollary}

\begin{proof}
\emph{Square.} Theorem~\ref{thm:rotation_fold} defines $\Sigma_D=\Sigma^Q$ and
gives the fixed real similarity $\mathfrak R(Q)$ between the realifications;
both similarities are invertible. By Lemma~\ref{lem:similarity_invariance},
$\Sigma_D$ keeps the window bound and time sharing of $\Sigma$ and has diagonal
transitions $\Lambda_t$, so $\Sigma_D\in\mathcal D^{\C}(\ell,K)$. Realification
composes every map with the fixed $\chi_K$ and $\mathfrak R$, which preserve
products by~\eqref{eq:realification}; it therefore preserves outputs, the window
bound and time sharing, so $\Sigma_{\R},(\Sigma_D)_{\R}\in\mathcal S^{\R}(\ell,2K)$.
\par\smallskip\noindent\emph{Complexification and fold.} By
Theorem~\ref{thm:rotation_fold}(i), $\Sigma_{\R}$ and the rotational real
representation $(\Sigma_D)_{\R}$ are fixed-similar through $\mathfrak R(Q)$, and
so are their complexifications. Its transitions $\bigoplus_k\rho_t^{(k)}\mathcal R(\theta_t^{(k)})$ have the
form~\eqref{eq:real_block_characterization} with $Q=I$ and $J=0$, so the
complexification of $(\Sigma_D)_{\R}$ is complex diagonal-equivalent by
Corollary~\ref{cor:abelian_closure}, hence so is
$(\Sigma_{\R})^{\C}$; complexification keeps the maps and the window bound, so
$(\Sigma_{\R})^{\C}\in\mathcal D^{\C}(\ell,2K)$.
Theorem~\ref{thm:rotation_fold}(ii) gives $\widetilde\Sigma$ with real diagonal
transitions $\bigoplus_k\rho_t^{(k)}I_2$ and causal maps, so
$\widetilde\Sigma\in\mathcal D^{\R}(L,2K)$. Its window bound and time sharing can
be lost: take $d=K=1$, $A_t=e^{ix_t}$, general forcing $b_t=1$, and $f_t^\diamond(x_t,z)=\operatorname{Re}z$.
Its folded forcing $(\cos\Theta_t,-\sin\Theta_t)^\top$, with
$\Theta_t=\sum_{s=0}^t x_s$, depends on every token in the prefix.
Choosing instead constant increments $\theta_t=\pi/2$ makes that
forcing position-dependent, showing that time sharing can also be lost.
\par\smallskip\noindent\emph{Exclusion.} If $A_t$ has a nonreal eigenvalue
$\lambda$, then $\mathfrak R(A_t)$ has the nonreal eigenvalues
$\lambda,\bar\lambda$, so neither realification is real diagonalizable, and
Lemma~\ref{lem:transition_boundary}(i) excludes $\mathcal D^{\R}(\ell',2K)$ for
every $\ell'$.
\end{proof}

Output equivalence alone thus preserves neither fixed diagonal equivalence
over a specified field nor the window bound: in the proof's example
($\ell=1$), both realifications keep window one, whereas for $L>1$ the
fold depends on the entire prefix.

\par\endgroup

\par\begingroup
\subsection{State space duality for scalar transitions}
\label{sec:ssd}

State space duality (SSD) evaluates a scalar-transition recurrence
by masked matrix products~\citep{dao2024transformers}. The scalar may
vary with the token but is shared across state coordinates, giving one
temporal mask and allowing chunkwise general matrix multiplication (GEMM).

Specialize the real matrix form with input features of
Definition~\ref{def:field_ssm} to $A_t=\alpha_t I_S$, $\alpha_t\in\R$,
on a finite restriction of length $L\in\mathbb N$. For a streaming
realization, this may be any finite prefix. Use the evaluated coefficients
and features $u_t=g_t^{\diamond}(x_{t;\ell})\in\R^R$ for this restriction, with
$B_t\in\R^{S\times R}$, $C_t\in\R^{O\times S}$ and
$h_{-1}\in\R^S$. All maps may use $x_{t;\ell}$ for any window
bound $\ell$, including the unrestricted case $\ell=L$.
We omit the feedthrough $D_tu_t$ to simplify the formulas; it is a
separate additive term with straightforward evaluation and differentiation.

\begin{proposition}\label{prop:ssd}
Let $\Gamma_{t,s}:=\prod_{j=s+1}^{t}\alpha_j$ for $-1\le s\le t<L$,
so $\Phi(t,s)=\Gamma_{t,s}I_S$, and set $\Gamma_{t,s}=0$ for $s>t$.
The \emph{history mask} $\Gamma=(\Gamma_{t,s})_{t,s=0}^{L-1}\in\R^{L\times L}$
has unit diagonal. Then
\begin{equation}\label{eq:ssd_recurrence}
 \begin{aligned}
 h_t&=\Gamma_{t,-1}h_{-1}+\sum_{s=0}^{t}\Gamma_{t,s}B_su_s,\\
 y_t&=C_t\Gamma_{t,-1}h_{-1}
       +\sum_{s=0}^{t}\Gamma_{t,s}(C_tB_s)u_s.
 \end{aligned}
\end{equation}
For write-feature coordinate $r$ and output coordinate $o$, define
$\mathsf B^r,\mathsf C^o\in\R^{L\times S}$ and $U^r\in\R^L$ by
\[
 (\mathsf B^r)_{t,n}=(B_t)_{n,r},\qquad
 (\mathsf C^o)_{t,n}=(C_t)_{o,n},\qquad
 (U^r)_t=(u_t)_r.
\]
Let $Y^o\in\R^L$ collect output coordinate $o$, and define
\[
 (Y_{\mathrm{init}}^o)_t=\Gamma_{t,-1}(C_th_{-1})_o.
\]
Contracting the read and write matrices first gives form (a):
\begin{equation}\label{eq:ssd_matrix}
 Y^o=Y_{\mathrm{init}}^o+
 \sum_{r=1}^{R}
   \Big(\Gamma\odot\big(\mathsf C^o(\mathsf B^r)^{\!\top}\big)\Big)U^r.
\end{equation}
Alternatively, absorb the features into the writes. Define
$V,\mathsf H\in\R^{L\times S}$ and $\gamma\in\R^L$ by
$V_{t,:}=(B_tu_t)^\top$, $\mathsf H_{t,:}=h_t^\top$ and
$\gamma_t=\Gamma_{t,-1}$. Form (b) computes the states first:
\[
 \mathsf H=\gamma h_{-1}^\top+\Gamma V,
 \qquad y_t=C_t(\mathsf H_{t,:})^\top.
\]
Both associations give the same outputs.
\end{proposition}

\begin{proof}
For $A_j=\alpha_jI_S$, the transfer~\eqref{eq:field_transfer} is
$\Phi(t,s)=\Gamma_{t,s}I_S$. Substituting $b_s=B_su_s$ and omitting
feedthrough in~\eqref{eq:field_closed_form} proves~\eqref{eq:ssd_recurrence}.
For form (a), contracting the state coordinate gives
\[
 \bigl[\mathsf C^o(\mathsf B^r)^{\!\top}\bigr]_{t,s}
 =\sum_{n=1}^{S}(C_t)_{o,n}(B_s)_{n,r}=(C_tB_s)_{o,r}.
\]
Using $\Gamma_{t,s}=0$ for $s>t$, the proposed matrix output therefore has
coordinate
\[
 \begin{aligned}
 \left[Y_{\mathrm{init}}^o+\sum_{r=1}^{R}
   \bigl(\Gamma\odot\mathsf C^o(\mathsf B^r)^{\!\top}\bigr)U^r\right]_t
 &=\Gamma_{t,-1}(C_th_{-1})_o
   +\sum_{r=1}^{R}\sum_{s=0}^{t}\Gamma_{t,s}(C_tB_s)_{o,r}(u_s)_r\\
 &=\Gamma_{t,-1}(C_th_{-1})_o
   +\sum_{s=0}^{t}\Gamma_{t,s}(C_tB_su_s)_o=(y_t)_o.
 \end{aligned}
\]
For form (b), aggregating the writes first gives, row by row,
\[
 \begin{aligned}
 [\gamma h_{-1}^{\top}+\Gamma V]_{t,:}
 &=\Gamma_{t,-1}h_{-1}^{\top}
   +\sum_{s=0}^{t}\Gamma_{t,s}(B_su_s)^{\top}=h_t^{\top},\\
 C_t(\mathsf H_{t,:})^{\top}
 &=\Gamma_{t,-1}C_th_{-1}
   +\sum_{s=0}^{t}\Gamma_{t,s}C_tB_su_s=y_t.
 \end{aligned}
\]
Thus both contraction orders include the same transported initial state
and give the same output.
\end{proof}

Form (a) has $OR$ read--write contractions sharing one mask; form (b)
aggregates the writes first and computes one state sequence shared by all
$O$ readouts. At $R=O=1$ this is Mamba~2's scalar-write identity.
Here $\mathsf C^o$ contains the queries, $\mathsf B^r$ the keys and $U^r$
the values. The unnormalized masked-attention form is
\begin{equation}
 Y=[\Gamma\odot(\mathsf Q\mathsf K^\top)]\mathsf V,
 \qquad \mathsf Q,\mathsf K\in\R^{L\times d_k},\quad
 \mathsf V\in\R^{L\times d_v}.
 \label{eq:masked_attention_matrix}
\end{equation}
Equation~\eqref{eq:ssd_matrix} sums this expression with
$\mathsf Q=\mathsf C^o$, $\mathsf K=\mathsf B^r$, $\mathsf V=U^r$,
plus $Y_{\mathrm{init}}^o$. Its mask is the scalar-transition product;
masked linear attention membership additionally requires token-local generating maps
(Section~\ref{sec:theory_mla}). Normalization is a separate readout
\citep{katharopoulos2020transformers,dao2024transformers}.

Subsystems sharing $\alpha_t$ reuse one mask; coordinate-dependent decay
requires up to $S$ masks. A shared modulus lets a complex-diagonal system
use one mask after Theorem~\ref{thm:rotation_fold}, with folded endpoints
whose accumulated angles are scalar prefix sums and may depend on the
full prefix. Continuation carries the folded state and these angles.
Applying the identities within chunks and carrying the boundary state
between them avoids materializing the full $L\times L$ mask
(Section~\ref{sec:hybridmamba_complexity}).
For positive scalar decays, this chunked evaluation applies the rescaling
of Corollary~\ref{cor:gauge_schedule} with chunk-local cumulative
log-decays, avoiding the potentially overflowing global factors
$e^{\pm\mu_t}$.

\par\endgroup

\par\begingroup
\subsection{Parallel scan and gradients}
\label{sec:scan_grad}

Work on a finite restriction of length $L\in\mathbb N$. For a streaming
realization, this may be any finite prefix; the loss includes only
its steps $0\le t<L$. The maps retain their declared window bound
$\ell$; for a finite restriction, any larger bound reduces to $L$.

An associative scan of $(p_0,\ldots,p_{L-1})$ under $\star$ returns
the inclusive prefixes $p_0\star\cdots\star p_t$ in
$\bigO(\log L)$ combine depth on $L$ processors
\citep{kogge1973parallel,blelloch1990prefix,merrill2016singlepass}.
The forward construction carries $h_t$ (Lemma~\ref{lem:scan_op}); reverse
scanning gives the sensitivities below. Scalar addition and multiplication
give the prefix sums $\sum_{s=0}^t a_s$ and products $\prod_{s=0}^t a_s$.

\begin{lemma}\label{lem:ssm_gradients}
Consider $h_t = A_t h_{t-1} + b_t$ for $0\le t<L$, with
$A_t\in\R^{S\times S}$, $b_t,h_t\in\R^S$ and initial state $h_{-1}$.
Treat $(A_t,b_t)_{t=0}^{L-1}$ and $h_{-1}$ as independent inputs to the
recurrence. Let $\mathcal L=\sum_t\mathcal L_t(y_t)$ depend on the states only
through $y_t=f_t^{\diamond}(x_{t;\ell},h_t)$, with $\mathcal L_t$ differentiable
and $f_t^{\diamond}$ differentiable in its state argument.
Derivatives through the causal coefficient maps and the readout's input
dependence are chained separately.  Gradients with respect to vectors are row vectors; write $\varepsilon_t := \frac{\partial\mathcal L_t}{\partial y_t}\frac{\partial f_t^{\diamond}}{\partial h}(x_{t;\ell},h_t)\in\R^{1\times S}$ for the \emph{direct} sensitivity of the loss to $h_t$ through its own readout, and $\mathrm d\mathcal L/\mathrm d(\cdot)$ for total derivatives along the recurrence. For the real matrix readout
$y_t=C_th_t+D_tu_t$, this specializes to
$\varepsilon_t=(\partial\mathcal L_t/\partial y_t)C_t$. Then:
\begin{enumerate}
\item[\textnormal{(i)}] \emph{Forcing.}
The state and forcing sensitivities agree,
$\mathrm d\mathcal L/\mathrm d h_t=\mathrm d\mathcal L/\mathrm d b_t$, with
\begin{equation}\label{eq:gradient_b}
\frac{\mathrm d\mathcal{L}}{\mathrm d b_t}
\;=\; \sum_{k=t}^{L-1} \varepsilon_k\, \Phi(k,t),
\qquad\text{equivalently}\qquad
\frac{\mathrm d\mathcal{L}}{\mathrm d b_{t-1}}
\;=\; \varepsilon_{t-1} \;+\; \frac{\mathrm d\mathcal{L}}{\mathrm d b_t}\, A_t,
\quad
\frac{\mathrm d\mathcal{L}}{\mathrm d b_{L-1}} \;=\; \varepsilon_{L-1}.
\end{equation}
For $1\le t<L$, this is a time-reversed affine recurrence,
right-multiplied by $A_t$ where the forward recurrence left-multiplies.
\item[\textnormal{(ii)}] \emph{Transition.}
\begin{equation}\label{eq:gradient_A}
\frac{\mathrm d\mathcal{L}}{\mathrm d A_t}
\;=\; \Big(\frac{\mathrm d\mathcal{L}}{\mathrm d b_t}\Big)^{\!\top} h_{t-1}^{\!\top}
\;\in\;\R^{S\times S},
\qquad
\Big(\frac{\mathrm d\mathcal{L}}{\mathrm d A_t}\Big)_{ij}
\;=\; \Big(\frac{\mathrm d\mathcal{L}}{\mathrm d b_t}\Big)_{\!i}(h_{t-1})_j ,
\end{equation}
of which only the diagonal $\frac{\mathrm d\mathcal{L}}{\mathrm d b_t}\odot h_{t-1}^{\!\top}$ survives when $A_t$ is constrained to be diagonal. For the scalar constraint $A_t=\alpha_t I_S$,
\[
 \frac{\mathrm d\mathcal L}{\mathrm d\alpha_t}
 =\frac{\mathrm d\mathcal L}{\mathrm d b_t}h_{t-1}.
\]
\item[\textnormal{(iii)}] \emph{Initial state.}
\begin{equation}\label{eq:gradient_hinit}
\frac{\mathrm d\mathcal{L}}{\mathrm d h_{-1}} \;=\; \frac{\mathrm d\mathcal{L}}{\mathrm d b_0}\,A_0.
\end{equation}
For the stated loss, the terminal sensitivity is
$\mathrm d\mathcal L/\mathrm d h_{L-1}=\varepsilon_{L-1}$.
\end{enumerate}
\end{lemma}

If the final state is passed to a later chunk, the
sensitivity arriving from that chunk is added to the terminal condition
$\varepsilon_{L-1}$.

\begin{lemma}\label{lem:scan_op}
Define a \emph{forward} operator $\star$ and a \emph{reverse} operator $\star_{\mathrm{R}}$ on pairs $(A, v) \in \mathbb{R}^{S \times S} \times \mathbb{R}^S$ by
\begin{equation}\label{eq:scan_op}
(A_1, v_1) \;\star\; (A_2, v_2) \;:=\; \bigl(A_2\, A_1,\; A_2\, v_1 + v_2\bigr),
\end{equation}
\begin{equation}\label{eq:reverse_scan_op}
(A_1, v_1) \;\star_{\mathrm{R}}\; (A_2, v_2) \;:=\; \bigl(A_1 A_2,\; A_2^{\!\top} v_1 + v_2\bigr).
\end{equation}
Both are associative for arbitrary $A_t \in \mathbb{R}^{S \times S}$, and they coincide on pairs whose matrix components are diagonal. Under these operators:
\begin{enumerate}
\item[\textnormal{(i)}] \emph{Forward states.} With $p_0 := (A_0,\, A_0 h_{-1} + b_0)$ and $p_t := (A_t, b_t)$ for $t \geq 1$, the left-to-right scan recovers the states of~\eqref{eq:field_ssm_general},
\begin{equation}\label{eq:scan_output}
p_0 \star p_1 \star \cdots \star p_t \;=\; \bigl(\Phi(t,-1),\; h_t\bigr);
\end{equation}
\item[\textnormal{(ii)}] \emph{Reverse gradients.} With $\bar p_{L-1} := (I,\, \varepsilon_{L-1}^{\!\top})$ and $\bar p_t := (A_{t+1},\, \varepsilon_t^{\!\top})$ for $t < L-1$, the right-to-left scan recovers the forcing gradients of~\eqref{eq:gradient_b},
\begin{equation}\label{eq:reverse_scan_output}
\bar p_{L-1} \star_{\mathrm{R}} \bar p_{L-2} \star_{\mathrm{R}} \cdots \star_{\mathrm{R}} \bar p_t \;=\; \Bigl(\Phi(L-1,t),\; \bigl(\tfrac{\mathrm d\mathcal{L}}{\mathrm d b_t}\bigr)^{\!\top}\Bigr).
\end{equation}
\end{enumerate}
\end{lemma}

With the forward states, the reverse-scan values $\mathrm d\mathcal L/\mathrm d b_t$
give the transition and initial-state gradients
(Lemma~\ref{lem:ssm_gradients}); parameter and input gradients follow by
chaining through the coefficient maps, including the readout's direct
derivatives. Proofs are in Appendix~\ref{app:scan_grad_details}. The scan
constructions use only the associativity of affine composition; the
transition matrices need neither commute nor be invertible. A combine
costs $\bigO(S^3)$ for dense transitions and $\bigO(S)$ for diagonal or
scalar ones, excluding the generation of the causal coefficients and
readouts, whose cost and parallelism depend on their specified maps.
For complex states, apply the lemmas after realification.
\par\endgroup

\section{A unified taxonomy for head-organized SSMs}
\label{sec:system_types}

\par\begingroup
Architectures partition the $S$-dimensional state of
Section~\ref{sec:discretization} into one or several subsystems. A declared
shape records state and interface widths, and a separate annotation records
the transition class (Figure~\ref{fig:class_overview}); the construction
specifies the maps and parameter sharing. Table~\ref{tab:specializations}
thus compares existing architectures, models outside the diagonal class and
our designs (Section~\ref{sec:hybridmamba}) by transition class and by
interface widths separately.
\par\endgroup

\begin{figure}[t]
\centering
\adjustbox{max width=\linewidth}{\definecolor{classink}{HTML}{26364A}
\definecolor{classblue}{HTML}{0072B2}
\definecolor{classviolet}{HTML}{80649E}
\definecolor{classteal}{HTML}{009E73}
\definecolor{classrust}{HTML}{D55E00}
\begin{tikzpicture}[yscale=.93,font=\small,text=classink,align=center,>=stealth,
  boundary/.style={fill=white,inner sep=3pt}]
\node[font=\small\bfseries] at (7.5,7.18)
  {Transition classes and a common window restriction $\ell<\hat\ell$};
\filldraw[fill=classblue!4,draw=classblue!85,line width=1pt]
  (7.5,3.25) ellipse [x radius=7.4cm,y radius=3.3cm];
\filldraw[fill=classviolet!7,draw=classviolet,dashed,line width=1pt]
  (6.3,3.05) ellipse [x radius=5.3cm,y radius=2.65cm];
\filldraw[fill=classteal!9,draw=classteal,line width=1.5pt]
  (6.3,2.45) ellipse [x radius=4.65cm,y radius=1.4cm];
\begin{scope}
\clip (7.5,3.25) ellipse [x radius=7.4cm,y radius=3.3cm];
\clip (-.1,-.2) rectangle (6.3,6.7);
\fill[classink,opacity=.06] (-.1,-.2) rectangle (6.3,6.7);
\foreach \x in {-7,-6.5,...,6.5}
  \draw[classink!24,line width=.3pt] (\x,-.2) -- ++(7,7);
\end{scope}
\begin{scope}
\clip (7.5,3.25) ellipse [x radius=7.4cm,y radius=3.3cm];
\draw[classink!85,densely dashed,line width=1pt] (6.3,-.1) -- (6.3,6.7);
\end{scope}
\node[boundary,text=classblue!80!black] at (8.6,6.52)
  {$\mathcal S^{\mathbb F}(\hat\ell,S)$};
\node[boundary,font=\footnotesize] at (3.45,5.99)
  {$\mathcal S^{\mathbb F}(\ell,S)$};
\node[boundary,text=classviolet!80!black] at (8.7,5.42)
  {Commuting families};
\node[font=\scriptsize,text=classviolet!80!black] at (8.25,4.48)
  {Commuting; some member\\not diagonalizable over $\mathbb F$\\
   (defective; over $\mathbb R$ also\\nonreal eigenvalues)};
\node[boundary,font=\footnotesize,text=classteal!60!black] at (8.0,1.15)
  {$\mathcal D^{\mathbb F}(\hat\ell,S)$: diagonal-equivalent\\one fixed basis};
\node[font=\footnotesize,text=classteal!60!black,fill=white,fill opacity=.85,text opacity=1,inner sep=3pt]
  at (4.0,2.45) {$\mathcal D^{\mathbb F}(\ell,S)$};
\node[font=\footnotesize,text=classrust!85!black] at (13.0,3.15)
  {Noncommuting\\$[A,A']\ne0$\\for some $A,A'$};
\node[font=\footnotesize] at (7.5,-.65)
  {Shaded portions: window $\ell$.\qquad Whole ellipses (both sides): window $\hat\ell$.};
\end{tikzpicture}}
\caption{{Transition classes at common field $\mathbb F$, state dimension
$S\ge2$ (for $S=1$ they coincide) and horizon $L$. Whole ellipses use window bound $\hat\ell$;
the shaded portions impose $1\le\ell<\hat\ell\le L$ on every class, so
$\mathcal D^{\mathbb F}(\ell,S)=\mathcal D^{\mathbb F}(\hat\ell,S)
\cap\mathcal S^{\mathbb F}(\ell,S)$, and likewise for commuting families.
Diagonal equivalence requires, and is guaranteed by, pairwise commutativity
and individual diagonalizability over $\mathbb F$ of the entire realized
transition family (Theorem~\ref{thm:simultaneous_diagonalization}).}}
\label{fig:class_overview}
\end{figure}

\needspace{8\baselineskip}
\subsection{Head-organized realizations}
\label{sec:def_general}

\par\begingroup
\begin{definition}\label{def:mhssm}
Fix $\mathbb F\in\{\R,\C\}$, $d,H,P,R,N,O\in\mathbb N$,
a horizon $L\in\mathbb N\cup\{\infty\}$ and a window bound
$\ell\in\mathbb N\cup\{\infty\}$ (default $1$ in this section, including the matrix form).
A \emph{head-organized SSM} is a realization
$\Sigma\in\mathcal S^{\mathbb F}(\ell,HPN)$ in the sense
of Definition~\ref{def:field_ssm}, equipped with a declared partition into
$H$ heads of $P$ subsystems, each with state width $N$ over $\mathbb F$,
declared write-feature width $R$ and real output width $O$. We set
\[
 \operatorname{shape}(\Sigma):=(H,P;R,N,O).
\]
Write $j=(\eta,p)$ for head $\eta\in\{1,\ldots,H\}$ and subsystem
$p\in\{1,\ldots,P\}$. The shared \emph{system input}
$x=(x_t)_{t\in\mathbb T_L}$, $x_t\in\R^d$, conditions the component maps;
subsystem $j$ has state $h_t^j\in\mathbb F^N$ and output $y_t^j\in\R^O$.
For a fixed input $x$, its causal maps follow Definition~\ref{def:field_ssm},
with evaluations
\[
 A_t^j:=A_t^{\diamond,j}(x_{t;\ell})\in\mathbb F^{N\times N},\qquad
 b_t^j:=b_t^{\diamond,j}(x_{t;\ell})\in\mathbb F^N.
\]
From the specified initial state $h_{-1}^j$, they give
\begin{equation}
 h_t^j=A_t^j h_{t-1}^j+b_t^j,\qquad
 y_t^j=f_t^{\diamond,j}(x_{t;\ell},h_t^j).
 \label{eq:mhssm_state}
\end{equation}
In the \emph{matrix form}, write
$M_t^j:=M_t^{\diamond,j}(x_{t;\ell})$ for $M\in\{B,C,D\}$.
Set $u_t^j:=g_t^{\diamond,j}(x_{t;\ell})\in\R^R$ and use
\begin{equation}
 b_t^{\diamond,j}(x_{t;\ell})=B_t^j u_t^j,\qquad
 f_t^{\diamond,j}(x_{t;\ell},h)=\operatorname{Re}_{\mathbb F}(C_t^j h)+D_t^j u_t^j.
 \label{eq:mhssm_features}
\end{equation}
Here $B_t^j\in\mathbb F^{N\times R}$,
$C_t^j\in\mathbb F^{O\times N}$ and $D_t^j\in\R^{O\times R}$.
All component maps use the declared window; coefficients need not factor
through $u_t^j$.
For finite-window time-shared maps, we write $A^j(x_{t;\ell}),b^j(x_{t;\ell}),
f^j(x_{t;\ell},h)$ and similarly $B^j,C^j,D^j,g^j$, with the fixed
boundary padding of Section~\ref{sec:discretization}.
Otherwise the step index and diamond remain.
Stacking the subsystem vectors $(v^j)_j$ in lexicographic order of
$j=(\eta,p)$ (head first) gives
\[
 h_t=(h_t^j)_j\in\mathbb F^{HPN},\qquad
 u_t=(u_t^j)_j\in\R^{HPR},\qquad
 y_t^{\mathrm{pre}}=(y_t^j)_j\in\R^{HPO}.
\]
An optional stateless aggregation map
$\psi_t^\diamond:(\R^d)^{m_{t,\ell}}\times\R^{HPO}\to\R^{O_{\mathrm{out}}}$,
with $O_{\mathrm{out}}\in\mathbb N$ specified separately, gives
$y_t=\psi_t^\diamond(x_{t;\ell},y_t^{\mathrm{pre}})$; by default it is the
identity on local outputs, with $O_{\mathrm{out}}=HPO$.
The combined realization has $S=HPN$ and maps
\[
 \begin{gathered}
 A_t[\Sigma]=\bigoplus_j A_t^j,\qquad
 b_t[\Sigma]=(b_t^j)_j,\\
 f_t^{\diamond}[\Sigma](x_{t;\ell},h)
   =\psi_t^\diamond\!\left(x_{t;\ell},
       \bigl(f_t^{\diamond,j}(x_{t;\ell},h^j)\bigr)_j\right).
 \end{gathered}
\]
All local maps and $\psi_t^\diamond$ satisfy the declared window bound
and any time-sharing requirement.
\end{definition}

Shape describes the declared presentation, though the realization may
admit others; time is indexed below and subsystems above. Widths $(R,O)$
give single-input, single-output (SISO, $R=O=1$), multiple-input, single-output
(MISO, $R>1,O=1$), single-input, multiple-output (SIMO, $R=1,O>1$), or
multiple-input, multiple-output (MIMO, $R,O>1$). The write-feature width $R$ is not an invariant of $b_t^j$:
{$u_t^j=x_t$ with $R=d$ gives the direct-input matrix form when
the forcing admits $b_t^j=B_t^jx_t$; general forcing need not have this form.}
Meanwhile,
Table~\ref{tab:specializations} uses the smallest exhibited $R,O$ under
the reductions below. A convolution inside $g_t^{\diamond,j}$ leaves the
one-step feature width $R$ unchanged; multi-endpoint writes combine such
samples in $b_t^j$. Stacked widths are logical, so shared features need
not be stored repeatedly. Aggregation $\psi_t^\diamond$ may use
input-dependent gates, weighted sums and output projections; if nonlinear,
it need not preserve matrix-form readout. Section~\ref{sec:discretization}
applies to subsystems and their combination at state/output widths $(N,O)$
and $(S,O_{\mathrm{out}})$; $d$ is shared, while $O_{\mathrm{out}}$ need not
equal $d$. Class membership (e.g.,
$\Sigma\in\mathcal D^{\mathbb F}(\ell,HPN)$), time sharing and transition
structure are separate from shape and justified by the maps and ties.
The field determines coordinate count: realifying a complex presentation
changes $(H,P;R,N,O)$ to $(H,P;R,2N,O)$; complexifying a real one to test
diagonal equivalence does not change its declared real shape
(Section~\ref{sec:simultaneous_diagonalization}).

\paragraph{Declared input boundary.}
The declared input $x$ supplies every branch used by the realization.
All intervening projections, convolutions and gates are included in its
maps, and their temporal dependence is included in $\ell$.

\paragraph{Exhibited interface reductions.}\phantomsection\label{par:interface_reductions}
For a fixed state partition, Table~\ref{tab:specializations} uses the smallest $R,O$ exhibited
by linear interface reductions. Suppressing $j$, take window-dependent
real matrices $V_t\in\R^{R\times R'}$ and $U_t\in\R^{O'\times O}$.
An input factorization
$u_t=V_t\tilde g_t^\diamond(x_{t;\ell})$, with
$\tilde u_t:=\tilde g_t^\diamond(x_{t;\ell})\in\R^{R'}$, and a local output
contraction $z_t=U_t y_t\in\R^{O'}$ give
\[
 \hat B_t=B_tV_t,\qquad \hat C_t=U_tC_t,\qquad
 \hat D_t=U_tD_tV_t.
\]
These factor the supplied interfaces without absorbing values into
coefficient branches, and retain the declared window. The complete
aggregation must admit
$\psi_t^\diamond(x,(y^j)_j)=\tilde\psi_t^\diamond(x,(U_t^jy^j)_j)$.
A nonlinear operation before contraction is allowed only if this
factorization still holds: normalization may depend on components
that $U_t^j$ discards.
These are exhibited factorizations, not a global minimum over arbitrary
feature maps. Nonlinear preprocessing may discard information even at
unchanged width; Theorem~\ref{thm:crosstalk} concerns linear transport of
the supplied values with coefficients fixed. Corollary~\ref{cor:m3_value_rank}
exhibits the Mamba~3 scalar-lift and local-read reductions.

\paragraph{Examples of head grouping.}
Mamba~2 shares transition, step and write/read vectors across a
head's $P$ subsystems with different scalar values; output gates lie in
$\psi$, and feedthrough may be head-shared or per-channel. Local widths
depend on the declared interface and partition, while $HP$ need not equal
$d$ (as in S5). Any $H'P'=HP$ regroups the same subsystems without changing
their maps or sharing.

\paragraph{Transition classes.}
For a fixed transition family, changing writes or readout
aggregation preserves membership in $\mathcal D^{\mathbb F}(\ell)$ as long as
the new maps keep window $\ell$; maps with a larger window move the
realization to that larger class. For block-diagonal transitions,
$[A,A']=\bigoplus_j[A^j,(A')^j]$, and a transition is diagonalizable
exactly when all its blocks are. Theorem~\ref{thm:simultaneous_diagonalization}
therefore applies blockwise. Table~\ref{tab:specializations} uses
the nested classes
$\mathcal D^{\R}(\ell)\subseteq\mathcal D^{\C}(\ell)$ and, for real realizations,
$\mathcal S_{\mathrm{nd}}^{\R}(\ell,S)
:=\mathcal S^{\R}(\ell,S)\setminus\mathcal D^{\C}(\ell,S)$,
using complexification. The latter denotes failure even over
$\C$, with the transition column naming the obstruction; annotations omit
$S=HPN$. Each class refers to the displayed recurrence at the smallest
established window among exhibited presentations with the same input
boundary, state size and interfaces, not a global minimum. For Mamba~3,
this uses the two-endpoint write~\eqref{eq:exp_trapezoidal_recurrence}
with rotations in the transition; Theorem~\ref{thm:rotation_fold} folds
them into potentially prefix-dependent endpoints. The table thus classifies a presentation, not the
computation an implementation runs, and the compared presentations need not be diagonal-equivalent:
Mamba~3's kernels evaluate the fold, an output-equivalent realization in $\mathcal D^{\R}(L)$
(Corollary~\ref{cor:realification_window_classes}). Taking the smallest window matters, because with a
prefix-dependent frame every realization with invertible transitions becomes diagonal (Lemma~\ref{lem:gauge_prefix}).

\par\endgroup

\subsection{Architecture classification}
\label{sec:classification}

\par\begingroup
\paragraph{SSM core.}
The \emph{SSM core} comprises the recurrence, maps producing its
driving features and coefficients from the \emph{core input}
$x=(x_t)$, $x_t\in\R^d$, and the stateless aggregation $\psi$ of
Definition~\ref{def:mhssm}. Convolutions and shifts before the recurrence
belong to the core; processing only subsystem outputs and the core input,
including stacking, combining and output gates, belongs to $\psi$.
Mamba~2/3 compute coefficients and gate from the same input, placing the
gate in $\psi$; Mamba~1 computes coefficients from its convolved value
branch and its gate from a separate branch outside the core
(Figure~\ref{fig:hm_block}). Table~\ref{tab:specializations} uses each
model's stated defaults; {$S$ counts recurrent-state coordinates,
separately from buffers used to generate coefficients.}

\paragraph{Windows.}
Windows include every conditioning branch and are measured on the
core input. A width-$k_{\mathrm{conv}}$ convolution gives window
$k_{\mathrm{conv}}$; shifting those features one token adds one. A
two-endpoint write (Section~\ref{sec:disc_alternatives}) also reuses the
previous features, giving $k_{\mathrm{conv}}+1$ or two without convolution.
Without these operations, maps are token-local (window one), though the
state retains history. If preprocessing feeds only the drive $u_t$, as
with a convolution on the value branch alone, the coefficient maps stay
token-local and Theorem~\ref{thm:mla_membership} takes $u$ as the value
sequence; coefficients computed from convolved features, as in Mamba~1/2,
inherit its window. A two-endpoint write cannot be absorbed this way:
$x_{t-1}$ enters its write coefficients.

The families below differ by commutativity of the entire realized
transition family and individual diagonalizability over $\mathbb F$
(Theorem~\ref{thm:simultaneous_diagonalization}). For diagonal-equivalent
SSMs with time-shared token-local maps, the state-mediated kernel is masked
linear attention over that field (Theorem~\ref{thm:mla_membership}(i));
nonlinear $\psi$, such as mLSTM's division, acts afterward. See
Appendix~\ref{app:classification_detail} for coefficient laws,
constructions and qualifications.

\par\endgroup

\begin{table}[!t]
\centering
\caption{{SSM cores of published architectures with shapes $(H,P;R,N,O)$
(Definition~\ref{def:mhssm}; $N$ counts real state coordinates per subsystem).
Each row shows the presentation with the smallest established window,
usually the unfolded one; the class argument bounds that window, and all
maps are time-shared.
$\mathcal S_{\mathrm{nd}}^{\R}(\ell)=\mathcal S^{\R}(\ell)\setminus\mathcal D^{\C}(\ell)$:
real cores not diagonal-equivalent even over $\C$ for generic weights (Section~\ref{sec:classification_boundary}).
Input-dep.: input-dependent discrete maps among $A,B,C$ (driving features
and $\psi$ not listed; Appendix~\ref{app:classification_detail}).
$k_{\mathrm{conv}}$: default short-convolution width ($1$ if none);
$k_{\mathrm{shift}}$: H3's shift width; $d_{\mathrm{ssm}}$: number of scalar
driving channels; $d_h$: H3's key/value width per head; $n_h$:
DeltaProduct's updates per token. ``Scalar (decay)'': one decay per subsystem.}
{The last group previews our designs (Section~\ref{sec:hybridmamba}).}}
\label{tab:specializations}
\smallskip
\footnotesize
\setlength{\tabcolsep}{4.5pt}
\renewcommand{\arraystretch}{1.18}
\adjustbox{max width=\linewidth}{%
\begin{tabular}{@{}lllcl@{}}
\toprule
{\textbf{Model}} & {\textbf{Declared shape}} & {\textbf{Class}} & {\textbf{Input-dep.}} & {\textbf{Transition structure}} \\
\midrule
{Mamba 1~\mbox{\citep{gu2024mamba}}} & {$(d_{\mathrm{ssm}},1;1,N,1)$} & {$\mathcal D^{\R}(k_{\mathrm{conv}})$\textsuperscript{f}} & {$A,B,C$} & {diagonal} \\
{Mamba 2~\mbox{\citep{dao2024transformers}}} & {$(H,P;1,N,1)$} & {$\mathcal D^{\R}(k_{\mathrm{conv}})$} & {$A,B,C$} & {scalar} \\
{GLA~\mbox{\citep{yang2024gla}}} & {$(H,P;1,N,1)$} & {$\mathcal D^{\R}(1)$} & {$A,B,C$} & {diagonal} \\
{mLSTM~\mbox{\citep{beck2024xlstm}}} & {$(H,P+1;1,N,1)$\textsuperscript{d}} & {$\mathcal D^{\R}(k_{\mathrm{conv}})$} & {$A,B,C$} & {scalar} \\
{RG-LRU~\mbox{\citep{de2024griffin}}} & {$(d_{\mathrm{ssm}},1;1,1,1)$} & {$\mathcal D^{\R}(k_{\mathrm{conv}})$} & {$A,B$} & {scalar} \\
{HGRN2~\mbox{\citep{qin2024hgrn2}}} & {$(H,P;1,N,1)$} & {$\mathcal D^{\R}(1)$} & {$A,B,C$} & {diagonal} \\
\midrule
{S4~\mbox{\citep{gu2022efficiently}}} & {$(d_{\mathrm{ssm}},1;1,N,1)$} & {$\mathcal D^{\C}(1)$\textsuperscript{e}} & {--} & {DPLR-derived} \\
{S4D~\mbox{\citep{gu2022s4d}} / DSS~\mbox{\citep{gupta2022diagonal}}} & {$(d_{\mathrm{ssm}},1;1,N,1)$} & {$\mathcal D^{\C}(1)$} & {--} & {complex diag.} \\
{GSS~\mbox{\citep{mehta2022gss}}, DSS core} & {$(d_{\mathrm{ssm}},1;1,N,1)$} & {$\mathcal D^{\C}(1)$} & {--} & {complex diag.} \\
{H3~\mbox{\citep{fu2023h3}}, diagonal core} & {$(H,d_h^2;1,N,1)$} & {$\mathcal D^{\C}(k_{\mathrm{shift}})$\textsuperscript{g}} & {--} & {complex diag.} \\
{S5~\mbox{\citep{smith2023simplified}}} & {$(1,1;d,N,d)$} & {$\mathcal D^{\C}(1)$} & {--} & {complex diag.} \\
{LRU~\mbox{\citep{orvieto2023}}} & {$(1,1;d,N,d)$} & {$\mathcal D^{\C}(1)$} & {--} & {complex diag.} \\
{RetNet~\mbox{\citep{sun2023retnet}}, unfolded} & {$(H,P;1,N,1)$} & {$\mathcal D^{\C}(1)$} & {$B,C$} & {complex diag., scalar decay} \\
\midrule
{Mamba 3 SISO~\mbox{\citep{lahoti2026mamba3}}} & {$(H,P;1,N,1)$} & {$\mathcal D^{\C}(2)$\textsuperscript{a}} & {$A,B,C$} & {complex diag., scalar decay} \\
{Mamba 3 MIMO~\mbox{\citep{lahoti2026mamba3}}} & {$(H,P;1,N,1)$\textsuperscript{c}} & {$\mathcal D^{\C}(2)$\textsuperscript{a}} & {$A,B,C$} & {complex diag., scalar decay} \\
{AUSSM~\mbox{\citep{karuvally2025aussm}}} & {$(d_{\mathrm{ssm}},1;1,N,1)$} & {$\mathcal D^{\C}(k_{\mathrm{conv}})$} & {$A,B$} & {complex diag., unitary} \\
\midrule
\midrule
{H3~\mbox{\citep{fu2023h3}}, shift core} & {$(H,d_h;1,N,1)$} & {$\mathcal S_{\mathrm{nd}}^{\R}(1)$} & {--} & {nilpotent shift, defective} \\
\midrule
{DeltaNet~\mbox{\citep{schlag2021fastweight,yang2024deltanet}}} & {$(H,P;1,N,1)$} & {$\mathcal S_{\mathrm{nd}}^{\R}(k_{\mathrm{conv}})$} & {$A,B,C$} & {identity $-$ rank one, non-abelian} \\
{Gated DeltaNet~\mbox{\citep{yang2025gateddelta}}} & {$(H,P;1,N,1)$} & {$\mathcal S_{\mathrm{nd}}^{\R}(k_{\mathrm{conv}})$} & {$A,B,C$} & {scaled identity $-$ rank one, non-abelian} \\
{DeltaProduct~\mbox{\citep{siems2025deltaproduct}}} & {$(H,P;n_h,N,1)$} & {$\mathcal S_{\mathrm{nd}}^{\R}(k_{\mathrm{conv}})$} & {$A,B,C$} & {product of $n_h$ such factors, non-abelian} \\
{RWKV-7~\mbox{\citep{peng2025rwkv7}}} & {$(H,P;1,N,1)$} & {$\mathcal S_{\mathrm{nd}}^{\R}(2)$} & {$A,B,C$} & {diagonal $-$ rank one, non-abelian} \\
\midrule
{\textit{HybridMamba} (\S\ref{sec:hm_m1})} & {$(H,P;1,N,1)$} & {$\mathcal D^{\C}(\ell)$\textsuperscript{b}} & {$A,B,C$} & {complex diag., per-row decay} \\
{\textit{SSD-MIMO} (\S\ref{sec:hm_mhr})} & {$(H,GR;R,N,O)$} & {$\mathcal D^{\C}(\ell)$\textsuperscript{b}} & {$A,B,C$} & {complex diag., scalar decay} \\
\bottomrule
\end{tabular}}
\smallskip
\begin{minipage}{\linewidth}\footnotesize
\noindent{\textsuperscript{a} Convolution-free Mamba~3 before the phase fold.}
\par\noindent{\textsuperscript{b} $\ell=k_{\mathrm{conv}}+\iota$, with $k_{\mathrm{conv}}=1$ without a
convolution and $\iota=1$ if two-endpoint writes or $\mu$ rows (Section~\ref{sec:hm_m1}) are used
(either or both), $\iota=0$ otherwise. Both default configurations use
$k_{\mathrm{conv}}=4$ and two-endpoint writes, so $\ell=5$.}
\par\noindent{\textsuperscript{c} Input rank one by Corollary~\ref{cor:m3_value_rank}
(data-independent scalar lift), with the local read contraction absorbed ($O=1$).
With the optional normalization across subsystems before contraction (off by
default), $\psi$ needs all $R_{\mathrm{M3}}$ rank outputs: $(H,P;1,N,R_{\mathrm{M3}})$;
the class is unchanged.}
\par\noindent{\textsuperscript{d} Normalizer as an extra unit-value subsystem;
division and output gate in the aggregation $\psi$.}
\par\noindent{\textsuperscript{e} Diagonalizable at HiPPO-LegS initialization (distinct
eigenvalues $-(n+1)$) and generically after training, but not guaranteed;
bilinear discretization keeps the eigenvectors.}
\par\noindent{\textsuperscript{f} Real default; the complex variant~\citep[Section~3.6]{gu2024mamba} belongs to $\mathcal D^{\C}(k_{\mathrm{conv}})$ (Appendix~\ref{app:class_real}).}
\par\noindent\textsuperscript{g} Window from the shift filter on the drive only; $A,B,C$ are constant.
\end{minipage}
\end{table}

\par\begingroup
\paragraph{Constant transitions.}
A singleton transition family commutes, leaving only
diagonalizability over $\mathbb F$. The structured state space sequence
model (S4) uses a diagonal plus low rank (DPLR) parameterization and
qualifies when its learned transition is diagonalizable (Table~\ref{tab:specializations},
footnote~e). Diagonal S4 (S4D), diagonal state spaces (DSS), gated state
spaces (GSS), simplified state space layers (S5), the linear recurrent
unit (LRU) and the retention network (RetNet) use fixed complex bases;
DSS's softmax variant fixes the horizon, and RetNet is shown unfolded
before rotations move into queries and keys. Linear time-invariant (LTI)
cores with token-local drives have window one and, at zero initial state,
the causal convolution~\eqref{eq:lti_convolution}, which S4 evaluates via
its kernel $K$~\citep{gu2022efficiently}. H3's diagonal core instead has
window $k_{\mathrm{shift}}$ (Appendix~\ref{app:class_constant}).

\paragraph{Input-dependent real diagonal transitions.}
Mamba~1/2, gated linear attention (GLA), matrix long short-term
memory (mLSTM), real-gated LRU (RG-LRU) and the hierarchically gated
recurrent network with state expansion (HGRN2) use real diagonal,
including scalar, transitions. Token-dependent coefficients rescale
coordinates without changing direction, so both conditions hold in this
basis. Appendix~\ref{app:class_real} constructs mLSTM's extra normalizer
subsystem (footnote~d) and exact stabilizer rescaling. Although usually
described as a linear recurrent neural network (RNN) or gated linear attention~\citep{beck2025tfla},
mLSTM is thus a real-diagonal SSM here, unlike LSTM and sLSTM, whose
gates read the previous hidden state.

\paragraph{Input-dependent rotations.}
Mamba~3, the adaptive unitary state space model (AUSSM), and our
multi-head HybridMamba and multiple-input, multiple-output state space
duality (SSD-MIMO) designs (Section~\ref{sec:hybridmamba}) use
input-dependent rotations on fixed coordinate pairs. Both coordinates
share one decay, except in AUSSM, which has none. Rotations on the same
pairs commute and share a fixed complex diagonalizing basis, so these
cores are complex diagonal-equivalent, in $\mathcal D^{\C}(\ell)$
(Appendix~\ref{app:class_rotations}).

\paragraph{Inside $\mathcal S^{\mathbb F}(\ell)$, outside $\mathcal D^{\mathbb F}(\ell)$.}
\phantomsection\label{sec:classification_boundary}
Hungry Hungry Hippos (H3) has a constant shift that is defective
for $N>1$: a nonzero nilpotent shift has no eigenbasis. DeltaNet, Gated
DeltaNet and DeltaProduct apply $I-\beta_tk_tk_t^{\!\top}$ along each key;
non-parallel, non-orthogonal keys yield noncommuting updates, so no common
eigenbasis serves all inputs. Receptance weighted key value (RWKV-7) has
analogous diagonal-minus-rank-one transitions. Appendix~\ref{app:class_nondiagonal}
gives attained noncommuting pairs. Their $\mathcal S_{\mathrm{nd}}^{\R}$
entries hold for generic parameters; degenerate ones, such as a key map
with one-dimensional image, can give diagonal-equivalent realizations.

\paragraph{Outside the SSM class $\mathcal S^{\mathbb F}(\ell)$.}
Nonlinear hidden-state feedback makes gated recurrent units (GRUs),
long short-term memory (LSTM) and scalar-memory LSTM (sLSTM) non-affine
in their specified recurrent state.
Full softmax attention keeps a growing key/value history;
Hyena~\citep{poli2023hyena} specifies its long-convolution filter as a
learned function of lag over the horizon. These specified architectures
do not supply an SSM core of fixed width $S$ independent of the horizon.
A long convolution is an SSM only if its kernel can be generated by a
finite state, $K_\tau=CA^\tau B$ with state width independent of the horizon;
S4's and H3's kernels are generated this way, whereas a freely learned
kernel such as Hyena's generally is not.
These are architecture classifications, not input--output separations:
unrestricted causal maps can compute from history even when the
readout ignores the state (Appendix~\ref{app:class_outside}).
\par\endgroup

\section{Retrieval rank and the attention correspondence}
\label{sec:theory}

\subsection{Setup: the retrieval operator}
\label{sec:theory_setup}

\par\begingroup

Fix one subsystem of shape $(H,P;R,N,O)$ with a matrix-form readout,
drop its index, and prescribe its initial state $h_{-1}$ independently of
the current input sequence. Work in real coordinates, realifying
a complex subsystem when needed, so $N$ counts real state coordinates.
The evaluated maps may have any input-window bound $\ell$. Use
$\Phi(t,s)=\rprod_{q=s+1}^{t}A_q$ and the closed form~\eqref{eq:field_closed_form}.
For a write $b_s=B_su_s$, with $B_s\in\R^{N\times R}$ and
$C_t\in\R^{O\times N}$, its state-mediated retrieval operator is the kernel of~\eqref{eq:field_sequence_kernel}:
\begin{equation}
  W(t,s) \;=\; C_t\, \Phi(t,s)\, B_s \;\in\; \R^{O \times R}.
  \label{eq:retrieval_operator}
\end{equation}
With the evaluated coefficients held fixed, $W(t,s)$ maps the supplied
value at $s$ to its contribution to the state-mediated read at $t$.
The initial-state contribution is separate: it changes the output but
not $W(t,s)$ or the retrieval-rank bounds below.

For the exhibited interface reductions of Section~\ref{par:interface_reductions},
write $u_s=V_s\tilde u_s$, $\tilde u_s\in\R^{R'}$, and
$z_t=U_ty_t\in\R^{O'}$. The effective endpoints and operator are
\[
 \hat B_s=B_sV_s,\qquad \hat C_t=U_tC_t,\qquad
 W_{\mathrm{eff}}(t,s)=\hat C_t\Phi(t,s)\hat B_s\in\R^{O'\times R'}.
\]
For an already reduced interface, take $V_s=I$, $U_t=I$, $R'=R$ and $O'=O$.
We study the rank of $W_{\mathrm{eff}}$ as a linear map on the supplied
value space $\R^{R'}$, holding the evaluated coefficients and interface
maps fixed.

\begin{theorem}
\label{thm:crosstalk}
Let $\hat B\in\R^{N\times R'}$, $\hat C\in\R^{O'\times N}$ and
$\Phi\in\R^{N\times N}$ be the effective write, effective read and transfer
of one subsystem at a fixed pair $(t,s)$. Set
$W_{\mathrm{eff}}=\hat C\Phi\hat B$.
For every $\Phi$, including singular transfers,
\[
 \operatorname{rank}W_{\mathrm{eff}}
 \le\min\!\big(\operatorname{rank}\hat C,\operatorname{rank}\hat B\big)
 \le\min(O',\operatorname{rank}\hat B).
\]
For fixed $\hat B$ and nonsingular $\Phi$, there exists an unrestricted
effective read $\hat C\in\R^{O'\times N}$ attaining
$r_{\mathrm{val}}:=\min(O',\operatorname{rank}\hat B)$, the \emph{value-routing
rank ceiling}. Constrained reads $\hat C=UC$ retain the bound but need
their own attainment witness. In particular, $O'=1$ or
$\operatorname{rank}\hat B\le1$ implies rank at most one. No diagonality,
commutativity or decay assumption is required.
\end{theorem}

\begin{proof}
The rank-product inequality~\citep[Sec.~0.4.5(c)]{horn2013matrix} gives
the first inequality; $\operatorname{rank}\hat C\le O'$ gives the second.
To show that the ceiling is attainable, suppose $\Phi$ is nonsingular.
The image of $\Phi\hat B$ then has dimension $\operatorname{rank}\hat B$.
Set $k=\min(O',\operatorname{rank}\hat B)$ and choose $k$ orthonormal
vectors in this image. Take their transposes as the first $k$ rows of
$\hat C$, with all remaining rows zero. Then $\hat C\Phi\hat B$ has
rank $k$, attaining the ceiling.
\end{proof}
\citet{ghodsi2025heads} counts the attention heads needed to represent a linear SSM by the
dimension of its lag operators; Theorem~\ref{thm:crosstalk} instead bounds the rank of one local
interface.

\paragraph{Linear aggregation across subsystems.}
Let $J=\{j_1,\ldots,j_m\}$ be subsystem indices $j=(\eta,p)$ sharing
$\tilde u_s\in\R^{R'}$, with $u_s^j=V_s^j\tilde u_s$.
For an aggregation of Section~\ref{sec:def_general} of the form
$\psi_t^\diamond(x_{t;\ell},(y^j)_j)=\sum_{j\in J}M_t^j y^j$,
where $M_t^j\in\R^{O_{\mathrm{out}}\times O}$ uses only the declared
input window, set $\hat B_s^j=B_s^jV_s^j$ and
$L_t^j=M_t^jC_t^j\Phi^j(t,s)\in\R^{O_{\mathrm{out}}\times N}$ and
$\hat B_s^{\mathrm{stack}}=[(\hat B_s^{j_1})^\top,\ldots,
(\hat B_s^{j_m})^\top]^\top\in\R^{mN\times R'}$.
Then $W_{\mathrm{agg}}(t,s)=\sum_{j\in J}L_t^j\hat B_s^j$
factors through $\hat B_s^{\mathrm{stack}}$, giving
\begin{equation}
 \operatorname{rank}W_{\mathrm{agg}}(t,s)
 \le\min\!\big(O_{\mathrm{out}},\operatorname{rank}\hat B_s^{\mathrm{stack}}\big)
 \le\min(O_{\mathrm{out}},R').
  \label{eq:aggregation_rank}
\end{equation}
The shared input and linear aggregation are additional assumptions;
Corollary~\ref{cor:static_grouping} gives the block factorization and proof.

\paragraph{Mamba~3's scalar-lift MIMO interface.}
The following corollary applies Theorem~\ref{thm:crosstalk} to its scalar
value lift and identifies when the local readout also reduces to width one.
\begin{corollary}
\label{cor:m3_value_rank}
Suppose $u_s=v\tilde u_s$ with $\tilde u_s\in\R$ and fixed
$v\in\R^R$, as in Mamba~3 MIMO. The effective write and feedthrough
are $\hat B_s=B_sv$ and $\hat D_s=D_sv$, exhibiting $R'=1$.
For any transition sequence, the single-write retrieval operator
\[
 W_{\mathrm{eff}}(t,s)=C_t\Phi(t,s)B_sv\in\R^{O\times1}
\]
has rank at most one. The same per-sample bound holds for the
two-endpoint write~\eqref{eq:exp_trapezoidal_recurrence}.
If the complete aggregation factors through a local contraction
$U_ty_t$ with $U_t\in\R^{1\times O}$, the effective read and feedthrough
are $U_tC_t$ and $U_tD_tv$, exhibiting $O'=1$ as well.
\end{corollary}
\begin{proof}
The scalar lift gives $\operatorname{rank}(B_sv)\le1$, so
Theorem~\ref{thm:crosstalk} applies without any decay or diagonality
assumption. For a two-endpoint write, both injections of the sample
$\tilde u_s$ are columns multiplying that same scalar; after transport,
their sum is still an $O\times1$ operator. The output contraction
preserves this bound, and substitution gives the reduced interfaces.
\end{proof}

\par\endgroup

\subsection{Masked linear attention: definition and exact membership}
\label{sec:theory_mla}

\par\begingroup
We compare state-mediated kernels with unnormalized masked linear
attention, using time-shared token-local features and multiplicative
scalar masks. \citet{hu2025structured} extend state space duality to real diagonal SSMs, analysed with scalar
inputs, and characterize through semiseparable structure when an SSM has a 1-semiseparable
masked-attention dual. Theorem~\ref{thm:mla_membership} states the class equality for realizations
with one fixed diagonalizing basis over $\R$ or $\C$, with vector writes and the initial state, and
adds an exclusion test for non-abelian kernels. A prefix-dependent Gram factorization alone is insufficient:
every invertible transition family admits one (Lemma~\ref{lem:gauge_prefix});
token locality is therefore an essential restriction
(Remark~\ref{rem:mla_positioning}).

Use the matrix interfaces of Sections~\ref{sec:simultaneous_diagonalization}
and~\ref{sec:classification} at layer scope: state width $S$, conditioning
input $x_t\in\R^d$, supplied features $u_t=g(x_t)\in\R^R$, and output
width $O_{\mathrm{out}}$. Here $g$ is the value map, commonly $g(x)=W_Vx$. The matrix $U$ below
stacks these values; $B$ (or the mode covectors $k^{(n)}$) writes them
into memory. For a prescribed initial state $h_{-1}\in\mathbb F^S$
independent of the current input sequence,
\begin{equation}
 \begin{gathered}
 y_t=y_t^{\mathrm{init}}+
 \operatorname{Re}_{\mathbb F}\!\Big(\sum_{s=0}^{t}W[x](t,s)u_s\Big)+D(x_t)u_t,\\
 y_t^{\mathrm{init}}=
 \operatorname{Re}_{\mathbb F}\!\big(C(x_t)\Phi(t,-1)h_{-1}\big),\\
 W[x](t,s)=C(x_t)\Phi(t,s)B(x_s),\qquad
 \Phi(t,s)=\rprod_{r=s+1}^{t}A(x_r),
 \end{gathered}
 \label{eq:mla_transfer}
\end{equation}
where $B(x)\in\mathbb F^{S\times R}$ and
$C(x)\in\mathbb F^{O_{\mathrm{out}}\times S}$.
A nonlinear aggregation $\psi$ is outside this kernel comparison.
Locality is relative to the declared input: after temporal preprocessing,
$x_t$ must include \emph{all} coefficient branches. A representation in
these processed inputs need not be token-local in the original sequence;
swap witnesses must lie in the domain being compared.

The following form covers linear attention~\citep{katharopoulos2020transformers},
decayed retention~\citep{sun2023retnet}, gated linear attention~\citep{yang2024gla},
rotary position masks~\citep{su2021roformer} and the structured masked attention
of state space duality~\citep[Section~4]{dao2024transformers}; in attention
notation the values are $v_t=g(x_t)=u_t$.

\begin{definition}
\label{def:linear_attention}
Fix an input sequence $x=(x_t)_{t\in\mathbb T_L}$, $x_t\in\R^d$, of horizon
$L\in\mathbb N\cup\{\infty\}$, with values $u_t=g(x_t)\in\R^R$.
For each mode $n\in\{1,\ldots,S\}$, choose time-shared token-local maps
\[
 \delta^{(n)}:\R^d\to\mathbb F,\qquad
 q^{(n)}:\R^d\to\mathbb F^{O_{\mathrm{out}}},\qquad
 k^{(n)}:\R^d\to\mathbb F^R,
\]
the mode's scalar transition, query (output direction) and key (write
direction), and write $\delta^{(n)}_t=\delta^{(n)}(x_t)$,
$q^{(n)}_t=q^{(n)}(x_t)$, $k^{(n)}_t=k^{(n)}(x_t)$.
Given an initial memory $z_{-1}\in\mathbb F^S$, independent of the current
input sequence, and a feedthrough $D$, the \emph{masked linear attention}
evaluates scalar mode memories $z^{(n)}_t\in\mathbb F$ and outputs
$y_t\in\R^{O_{\mathrm{out}}}$ by
\begin{equation}
 z^{(n)}_t=\delta^{(n)}_t z^{(n)}_{t-1}+(k^{(n)}_t)^*u_t,\qquad
 y_t=\operatorname{Re}_{\mathbb F}\sum_{n=1}^{S}q^{(n)}_t z^{(n)}_t+D(x_t)u_t,
 \label{eq:mla_recurrence}
\end{equation}
where $*$ denotes conjugate transpose. Its \emph{kernel} is
\begin{equation}
 W[x](t,s)=\sum_{n=1}^{S}\gamma^{(n)}(t,s)\,q^{(n)}_t(k^{(n)}_s)^*,
 \qquad \gamma^{(n)}(t,s)=\prod_{r=s+1}^{t}\delta^{(n)}_r,
 \label{eq:mla_kernel}
\end{equation}
for $s\le t$, with empty products equal to one; a kernel of this form is a
\emph{masked linear attention kernel} with $S$ modes. It is causal by
construction.
\end{definition}
Each mode is one scalar memory: linear attention with $d_k$ key features and $d_v$
value channels has $d_kd_v$ modes, one per entry of its state matrix (see below).

Unrolling~\eqref{eq:mla_recurrence} gives
\begin{equation}
 z^{(n)}_t=\gamma^{(n)}(t,-1)\,z^{(n)}_{-1}
        +\sum_{s=0}^{t}\gamma^{(n)}(t,s)\,(k^{(n)}_s)^*u_s,
 \label{eq:mla_operators}
\end{equation}
the transported initial memory plus the kernel action, so that
$y_t=\operatorname{Re}_{\mathbb F}\bigl(\sum_n q^{(n)}_t\gamma^{(n)}(t,-1)z^{(n)}_{-1}
+\sum_{s\le t}W[x](t,s)u_s\bigr)+D(x_t)u_t$; $z_{-1}=0$ recovers the
usual attention sum over the supplied tokens.
The mask $\gamma^{(n)}$ accumulates the mode's transitions; it is not a
value map. Constant $\delta^{(n)}=1$ gives a plain causal mask,
$0<\delta^{(n)}<1$ decay, $\delta^{(n)}_t\in(0,1)$ input-dependent forgetting and
$\delta^{(n)}=e^{i\omega}$ a rotary mask; amplification and sign changes are also
allowed. A \emph{mask-tie class} is a maximal set of modes sharing one transition
map $\delta^{(n)}$, hence one value $\delta^{(n)}_t$ at every step and one mask
$\gamma^{(n)}$; the shared value may still vary with $t$. Its dimension is its
mode count. Modes describe the
combined kernel; no architectural head count is prescribed, and a
mask-tie class need not coincide with a head.

\begin{theorem}
\label{thm:mla_membership}
Let $\Sigma\in\mathcal S^{\mathbb F}(1,S)$ have time-shared token-local matrix maps
and kernel~\eqref{eq:mla_transfer}.
\begin{enumerate}
\item[\textnormal{(i)}] If $\Sigma\in\mathcal D^{\mathbb F}(1,S)$ and $Q$ is a fixed
diagonalizing basis, the maps
\begin{equation}
 \delta^{(n)}(x)=\big(Q^{-1}A(x)Q\big)_{nn},\qquad
 q^{(n)}(x)=C(x)Qe_n,\qquad k^{(n)}(x)^*=e_n^{\!\top}Q^{-1}B(x),
 \qquad n=1,\ldots,S,
 \label{eq:mla_features}
\end{equation}
with mode memory $z_{-1}=Q^{-1}h_{-1}$ and the same value map and feedthrough,
define a masked linear attention with $S$ modes and kernel $W$ that reproduces
the complete output of $\Sigma$, including $y_t^{\mathrm{init}}$.
Conversely, for every masked linear attention with $S$ modes, the
realization with $A(x)=\operatorname{diag}_n\delta^{(n)}(x)$, $C(x)e_n=q^{(n)}(x)$,
$e_n^{\!\top}B(x)=k^{(n)}(x)^*$, initial state $h_{-1}=z_{-1}$ and the same value
map and feedthrough lies in $\mathcal D^{\mathbb F}(1,S)$ and has the same kernel
and output; its maps are~\eqref{eq:mla_features} with $Q=I$.
Hence, for the matrix interfaces above,
\begin{equation}
 \{W:\Sigma\in\mathcal D^{\mathbb F}(1),\ \text{maps shared across time}\}
 =\{\text{masked linear attention kernels over }\mathbb F\}.
 \label{eq:mla_class_equality}
\end{equation}
\item[\textnormal{(ii)}] Suppose an input $x$ and strictly interior times
$s<r<r'<t$ witness
\begin{equation}
 W[x^{(r\leftrightarrow r')}](t,s)\ne W[x](t,s),
 \label{eq:mla_witness}
\end{equation}
where $x^{(r\leftrightarrow r')}$ exchanges $x_r,x_{r'}$.
Then $W$ is neither a masked linear attention kernel over $\R$ or $\C$ with finitely many
modes nor a pointwise limit of such kernels.
\end{enumerate}
\end{theorem}

Appendix~\ref{app:theory_sec4_proofs} proves the theorem. Part~(ii)
tests one kernel entry with fixed source and read tokens, not the complete
output. Shared scalar masks, including relative-position phases and lag
dependence such as $\gamma(t,s)=c^{t-s}$, are invariant under interior swaps.
For a non-abelian realization, an exposed change
$C(x_t)(\Phi'-\Phi)B(x_s)\ne0$, where
$\Phi'=\Phi[x^{(r\leftrightarrow r')}](t,s)$, at $s<r<r'<t$ for a swap
in the stated domain, excludes its kernel; degenerate endpoints may hide the
change, so this test is sufficient, not necessary.

Part~(i) and its converse also hold for step-indexed token-local maps
$A_t(x_t),B_t(x_t),C_t(x_t)$ and
$\delta^{(n)}(t,\cdot),q^{(n)}(t,\cdot),k^{(n)}(t,\cdot)$ when one fixed basis
diagonalizes the realized family. Part~(ii) needs time sharing: a single
diagonal mode with $\delta_t(x)=e^{tx}$ and unit endpoints detects interior
order. The scan--attention identity itself does not require time sharing
\citep{dao2024transformers,ali2024hidden,zimerman2024unified,katsch2023gateloop}.

\paragraph{Matrix form and the attention correspondence.}
On a sequence of length $L$, let the mask $\Gamma_c\in\mathbb F^{L\times L}$ of a
mask-tie class $c$ have entries $\gamma_c(t,s)$ for $s\le t$ and zero otherwise,
stack $u_t^\top$ as
the rows of $U$, and collect the mode coordinates over time in the vectors
$\mathbf q^{(n)}_o,\mathbf k^{(n)}_r\in\mathbb F^L$ with entries
$(q^{(n)}_t)_o$ and $\overline{(k^{(n)}_t)_r}$.
\begin{corollary}
\label{cor:mla_matrix}
Take a masked linear attention with $S$ modes and mode memory $z_{-1}$,
without feedthrough. Group its modes into mask-tie classes $c$ with
masks $\Gamma_c$, and let $\mathsf C^o_c$ and $\mathsf B^r_c$ have the columns
$\mathbf q^{(n)}_o$ and $\mathbf k^{(n)}_r$ of the modes in class $c$. Then every
output coordinate $o$ satisfies
\begin{equation}
 Y_{:,o}=\sum_c\Bigl(Y^{\mathrm{init}}_{c,:,o}+\operatorname{Re}_{\mathbb F}\sum_{r=1}^{R}
  \bigl[\Gamma_c\odot\bigl(\mathsf C^o_c(\mathsf B^r_c)^\top\bigr)\bigr]U_{:,r}\Bigr),
 \label{eq:mla_matrix}
\end{equation}
where $Y^{\mathrm{init}}_{c,t,o}=\operatorname{Re}_{\mathbb F}\!\left(
\gamma_c(t,-1)\sum_{n\in c}(q^{(n)}_t)_o z^{(n)}_{-1}\right)$.
Each class term is the matrix
form~\eqref{eq:ssd_matrix}. Through the modes of
Theorem~\ref{thm:mla_membership}(i) with $z_{-1}=Q^{-1}h_{-1}$, this also gives the
output of every $\Sigma\in\mathcal D^{\mathbb F}(1,S)$
{with time-shared token-local matrix maps and no feedthrough;
adding $D(x_t)u_t$ restores the general matrix-form output.}
\end{corollary}
\begin{proof}
By the converse in Theorem~\ref{thm:mla_membership}(i), the attention is the
diagonal realization with transitions $\operatorname{diag}_n\delta^{(n)}_t$. The modes
of class $c$ form a subsystem with scalar transition $\delta_{c,t}I$, read and write maps given by
$\mathsf C^o_c$ and $\mathsf B^r_c$, and initial state
$(z^{(n)}_{-1})_{n\in c}$. Proposition~\ref{prop:ssd}, whose argument holds verbatim
over $\mathbb F$, gives its output as the class term, and the output is the sum over
classes.
\end{proof}
The standard head~\eqref{eq:masked_attention_matrix} has
$R=O_{\mathrm{out}}=d_v$, $S=d_kd_v$, zero initial memory,
$u_t=\mathsf V_{t,:}^\top$, and one mask class $\Gamma$. Its mode $(a,b)$,
$a\le d_k$, $b\le d_v$, is entry $(a,b)$ of the usual
$d_k\times d_v$ memory: set $q^{(a,b)}_t=\mathsf Q_{t,a}e_b$ and
$k^{(a,b)}_t=\mathsf K_{t,a}e_b$. Then
$\mathsf C^o(\mathsf B^r)^\top=\mathsf Q\mathsf K^\top$ for $o=r$ and is zero
otherwise, so~\eqref{eq:mla_matrix} gives
$Y=[\Gamma\odot(\mathsf Q\mathsf K^\top)]\mathsf V$.
Input-dependent masks per key coordinate, $\delta^{(a,b)}_t=\alpha_{t,a}$,
give gated linear attention~\citep{yang2024gla}.
Modes permit independent read and write directions, but their masks are
causal scalar-transition products, not arbitrary lower-triangular masks.
Normalized linear attention~\citep{katharopoulos2020transformers} divides
its masked-linear-attention numerator by a key-only memory; adding that
memory as a subsystem makes a diagonal SSM with nonlinear aggregation, as
for mLSTM (Section~\ref{sec:classification}); only the division lies outside
this kernel comparison. Softmax attention instead
keeps every past key and value, a shift-register state that grows with the
context and is not diagonalizable. Multiple heads use block-supported modes;
Appendix~\ref{app:mla_heads} details their concatenation and fixed output
projections.
\par\endgroup

\par\begingroup

\paragraph{Example: rotary features.}\phantomsection\label{ex:rope}
For an attention query $\mathsf q_t\in\R^2$ and key
$\mathsf k_s\in\R^2$, RoPE~\citep{su2021roformer} rotates each pair by its
absolute position at frequency $\omega$, giving
\begin{equation}
 \bigl(\mathcal R(t\omega)\mathsf q_t\bigr)^{\!\top}\bigl(\mathcal R(s\omega)\mathsf k_s\bigr)
 =\mathsf q_t^{\top}\mathcal R\bigl((s-t)\omega\bigr)\mathsf k_s .
 \label{eq:rope_score}
\end{equation}
One complex mode realizes this score. For time-shared token-local
$\mathsf q(x)=W_Qx$ and $\mathsf k(x)=W_Kx\in\R^2$, set
$\mathsf q^{\C}=\mathsf q[1]+i\mathsf q[2]$,
$\mathsf k^{\C}=\mathsf k[1]+i\mathsf k[2]$, and
\begin{equation}
 \delta^{(1)}=e^{-i\omega},\qquad
 q^{(1)}(x)=\overline{\mathsf q^{\C}(x)},\qquad
 k^{(1)}(x)=\overline{\mathsf k^{\C}(x)}.
 \label{eq:rope_route}
\end{equation}
Since $\overline{\mathsf q^{\C}_t}\mathsf k^{\C}_s
=\mathsf q_t^{\top}\mathsf k_s
+i(\mathsf q_t[1]\mathsf k_s[2]-\mathsf q_t[2]\mathsf k_s[1])$,
the real part of the kernel~\eqref{eq:mla_kernel} is
\begin{equation}
 \begin{aligned}
 \operatorname{Re}W(t,s)&=\operatorname{Re}\bigl(e^{-i\omega(t-s)}\,\overline{\mathsf q^{\C}_t}\,\mathsf k^{\C}_s\bigr)\\
 &=\mathsf q_t^{\top}\mathsf k_s\cos\bigl(\omega(t-s)\bigr)
  +(\mathsf q_t[1]\mathsf k_s[2]-\mathsf q_t[2]\mathsf k_s[1])\sin\bigl(\omega(t-s)\bigr)
 =\mathsf q_t^{\top}\mathcal R\bigl((s-t)\omega\bigr)\mathsf k_s ,
 \end{aligned}
 \label{eq:rope_ssm}
\end{equation}
This equals~\eqref{eq:rope_score}; relative position enters only
through $\gamma^{(1)}(t,s)=e^{-i\omega(t-s)}$, with time-shared maps.
The converse of Theorem~\ref{thm:mla_membership}(i) gives a complex diagonal
SSM with transition $e^{-i\omega}$, read $\overline{\mathsf q^{\C}}$, and
write $\mathsf k^{\C}$. Its rotational real representation
(Theorem~\ref{thm:rotation_fold}(i))
has $A=\mathcal R(-\omega)$, $B(x)=\mathsf k(x)$, and
$C(x)=\mathsf q(x)^{\top}$: Mamba~3's rotation law at constant angle and
unit decay. A width-$2K$ head uses $K$ pairs with frequencies $\omega_k$,
giving $K$ complex modes per value coordinate and transition
$\bigoplus_k\mathcal R(-\omega_k)$. Decay moduli $\rho^{(k)}$ may vary by
pair under Theorem~\ref{thm:rotation_fold}; Mamba~3 shares them across pairs.
An input-dependent angle $\theta(x_r)$ in place of $-\omega$ gives
$\delta^{(1)}_r=e^{i\theta(x_r)}$, the rotation law~\eqref{eq:hm_rotation_law}
of our designs and Mamba~3~\citep{lahoti2026mamba3}. Folded RoPE rotates
$\mathsf q_t,\mathsf k_s$ by absolute positions and has step-indexed maps;
the extension of part~(i), not part~(ii), applies.

\paragraph{Applications of the classification table.}
Table~\ref{tab:specializations}'s cores in
$\mathcal D^{\mathbb F}(1)$ with time-shared maps have masked linear
attention kernels over $\mathbb F$; RetNet's rotational core needs complex
modes. Coefficient-branch convolutions and two-endpoint writes read earlier
tokens, breaking token locality on original block inputs. A drive-only
convolution can instead be absorbed into the values
(Section~\ref{sec:classification}). Failing these hypotheses neither
excludes another masked linear attention representation nor ranks
expressivity. Mamba~3's per-sample rank bound is
Corollary~\ref{cor:m3_value_rank}. Folded rotations can make endpoints
prefix-dependent, so a real folded scan alone does not establish real masked
linear attention; Section~\ref{sec:hm_kernels} gives the conditions for our
designs.

\par\endgroup

\section{Extending dynamics and routing within the diagonal-equivalent class}
\label{sec:hybridmamba}

\par\begingroup
We develop two complementary extensions within the
$\mathcal D^{\C}(\ell)$ class. SSD-MIMO provides multidimensional local
writes and reads with scalar per-scan-head decay, folding rotation into the
endpoint maps when present.
The multi-head HybridMamba (HM) extends Mamba~1's per-coordinate decay: it combines per-coordinate and per-head laws, with real or
complex dynamics, in one row-partitioned state. Appendix~\ref{app:implementation} gives the map constructions, state-continuation interface and Mamba
specializations; chunk algorithms are in Appendix~\ref{app:kernel_details}.
Both designs run on custom fused Triton and Gluon GPU kernels
(Section~\ref{sec:hm_kernels}), with PyTorch reference scans for validation.

\subsection{Mixer, SSM core and scan algorithms}
\label{sec:hm_kernels}

\paragraph{Mixer.}
Both designs use the Mamba-style mixer in Figure~\ref{fig:hm_block}.
Input $z_t\in\R^{d_{\mathrm{model}}}$ projects to values $v_t=W_vz_t\in\R^d$
and gate $z_t^{\mathrm g}=W_gz_t\in\R^d$; optional short convolution and a sigmoid linear unit (SiLU)
produce the recurrence drive $u_t\in\R^d$. {The core boundary
(Section~\ref{sec:classification}) depends on the variant: in (a), it includes
the value projection, so $x_t=z_t$ and coefficients use $s_t=x_t$
(default, Mamba~2/3); in (b), it starts after that projection, so $x_t=v_t$
and coefficients use $s_t=u_t$ (Mamba~1).}
In case (a), all projections may be packed into one. The core output
$y_t$ is multiplied by $\operatorname{SiLU}(z_t^{\mathrm g})$ and projected
to $z'_t$. Mamba~2's preset also convolves B/C; Mamba~3's bypasses convolution
and drive SiLU. Section~\ref{sec:mamba_recovery}'s presets recover each block exactly.

\begin{figure}[t]
\centering
\adjustbox{max width=\linewidth}{%
\definecolor{mixink}{HTML}{26364A}
\definecolor{mixblue}{HTML}{367BB5}
\definecolor{mixteal}{HTML}{178478}
\definecolor{mixgold}{HTML}{C48A25}
\begin{tikzpicture}[font=\footnotesize, >=stealth, text=mixink,
 line/.style={->,draw=mixink,line width=.7pt},
 coeffline/.style={->,draw=mixteal!85!black,line width=.7pt},
 core/.style={draw=mixink!75,dashed,rounded corners=4pt,line width=.8pt},
 box/.style={rounded corners=2pt,minimum height=8mm,
             inner sep=3pt,align=center,line width=.7pt},
 projection/.style={box,draw=mixblue,fill=mixblue!10},
 coefficient/.style={box,draw=mixteal,fill=mixteal!8,
                     minimum height=4.5mm,minimum width=7mm},
 activation/.style={box,draw=mixgold,fill=mixgold!12}]
\foreach \panel/\shift/\source in {a/0/z,b/-4.15/u}{%
\begin{scope}[yshift=\shift cm]
  \ifnum\pdfstrcmp{\panel}{a}=0
    \node[anchor=west,font=\small] at (-.4,2.05)
      {\textbf{(a)} $s_t=x_t$ \quad Default; Mamba 2/3 lineage; core input: $x_t=z_t$};
    \draw[core] (.2,-1.75) rectangle (16.1,1.8);
    \node[anchor=south east,fill=white,inner sep=2pt]
      at (15.9,-1.65) {SSM core (including $\psi$)};
  \else
    \node[anchor=west,font=\small] at (-.4,2.05)
      {\textbf{(b)} $s_t=u_t$ \quad Mamba 1 lineage; core input: $x_t=v_t$};
    \draw[core] (2.95,-.98) rectangle (10.73,1.8);
    \node[anchor=south west,inner sep=2pt,font=\small]
      at (3.0,-.95) {SSM core};
  \fi
  \node (z) at (-.4,0) {$z_t$};
  \node[projection,minimum width=16mm] (lin) at (1.5,0) {Value/gate\\projections};
  \node[projection,minimum width=15mm] (conv) at (4.0,0)
    {Conv1d\\$+\,\mathrm{SiLU}$\\{\scriptsize(optional)}};
  \coordinate (drive) at (5.5,0);
  \node[box,draw=mixteal,fill=mixteal!14,minimum width=25mm]
    (ssm) at (7,0) {SSM recurrence};
  \node[box,draw=mixteal,fill=mixteal!8,minimum width=7mm]
    (read) at (9.5,0) {Read};
  \node[draw=mixteal,fill=white,circle,minimum size=5mm,inner sep=0pt,line width=.6pt]
    (plus) at (10.45,0) {$+$};
  \node[draw=mixgold,fill=mixgold!12,circle,minimum size=6mm,
        inner sep=0pt,line width=.7pt] (mul) at (11.7,0) {$\odot$};
  \node[box,draw=mixink!65,fill=black!3,densely dotted,
        minimum width=16mm] (norm) at (13.2,0) {RMSNorm\\(optional)};
  \node[projection,minimum width=16mm] (out) at (15.1,0) {Output\\projection};
  \node (y) at (16.65,0) {$z'_t$};
  \node[activation,minimum width=14mm,minimum height=5mm]
    (silu) at (7,-1.2) {SiLU};
  \draw[line] (z) -- (lin);
  \draw[line] (drive) -- (5.5,-.62) -- node[below,pos=.88,font=\scriptsize,inner sep=1.5pt] {$D_tu_t$} (10.45,-.62) -- (plus.south);
  \draw[line] (lin) -- node[above,pos=.35,inner sep=1.5pt] {$v_t$} (conv);
  \draw[line] (conv) -- node[above,pos=.5,inner sep=1.5pt] {$u_t$} (ssm);
  \draw[line] (ssm) -- node[above,pos=.5,inner sep=1.5pt] {$h_t$} (read);
  \draw[line] (read) -- (plus);
  \draw[line] (plus) -- node[above,pos=.5,inner sep=1.5pt] {$y_t$} (mul);
  \draw[line] (mul) -- (norm);
  \draw[line] (norm) -- (out);
  \draw[line] (out) -- (y);
  \draw[line] (lin.south) |- node[below,pos=.75] {$z_t^{\mathrm g}$} (silu.west);
  \draw[line] (silu.east) -| (mul.south);
  \foreach \name/\pos/\symbol in {B/6.1/B_t,delta/6.85/\Delta_t,rate/7.6/A_t,theta/8.35/\theta_t,C/9.5/C_t}{%
    \node[coefficient] (\name) at (\pos,.85) {$\symbol$};
    \draw[coeffline] (\pos,1.3) -- (\name.north);
  }
  \draw[coeffline] (B.south) -- (6.1,.4);
  \draw[coeffline] (delta.south) -- (6.85,.4);
  \draw[coeffline] (rate.south) -- (7.6,.4);
  \draw[coeffline] (theta.south) -- (8.35,.4);
  \draw[coeffline] (C.south) -- (read.north);
  \node[font=\scriptsize] at (7.55,1.55) {Coefficient maps (main examples)};
  \ifnum\pdfstrcmp{\panel}{a}=0
    \draw[draw=mixteal!85!black,line width=.7pt] (.35,0) -- (.35,1.3) -- (9.5,1.3);
    \fill[mixteal!85!black] (.35,0) circle (1.2pt);
  \else
    \draw[draw=mixteal!85!black,line width=.7pt] (drive) -- (5.5,1.3) -- (9.5,1.3);
    \fill[mixteal!85!black] (drive) circle (1.2pt);
  \fi
\end{scope}}
\end{tikzpicture}}
\caption{{Shared HybridMamba/SSD-MIMO mixer. Conv1d is one-dimensional convolution;
RMSNorm is root-mean-square normalization. Dashed boundaries define the core
input $x_t$ separately in each variant: (a) before the value projection,
$x_t=z_t$, with coefficients from $s_t=x_t$ (default, Mamba~2/3 lineage);
(b) after it, $x_t=v_t$, with coefficients from $s_t=u_t$ (Mamba~1 lineage).
Boxes show step $\Delta_t$, decay $A_t$ (per coordinate in Mamba~1/HM,
or head-shared), angle $\theta_t$ (with rotation), write $B_t$ and read $C_t$;
gates, trapezoid weight and potential are omitted. The read adds $D_tu_t$.
B/C normalization is on by default; output root-mean-square (RMS)
normalization is off by default and on in the Mamba~2 preset.}}
\label{fig:hm_block}
\end{figure}
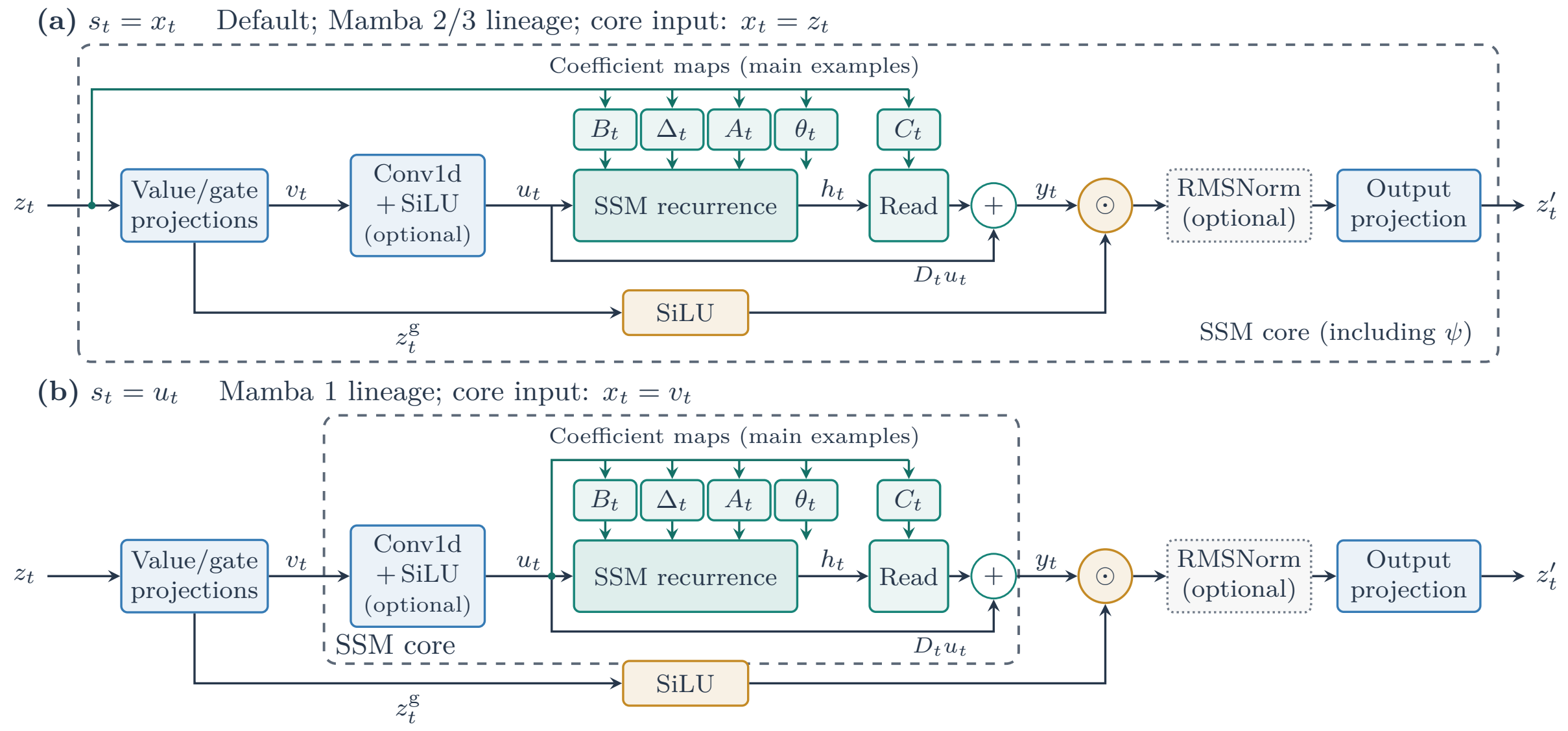

\paragraph{Indexing.}
Both designs split drive width $d=HP$ into $H$ heads $\eta$ of $P$ coordinates,
carrying $dN$ real state coordinates. Subsystem $(\eta,p)$ has $N$ state rows
and drive $u_t[\eta,p]$. SSD-MIMO splits $p=(g,r)$, $P=GR$: each head has
$G$ groups of $R$ \emph{siblings}, each with its own state and shared drive
$u_t[\eta,g,:]\in\R^R$. A \emph{scan head} $(\eta,r)$ shares one scalar
decay (Section~\ref{sec:hm_mhr}); $k$ indexes rotating pairs.
Endpoint maps $\bar B,\bar C$ are base projections, optionally gated and then
shifted by additive biases (Section~\ref{sec:hm_transition_laws}). Coefficient
maps share along $p$ in HM and $g$ in SSD-MIMO; base write/read projections
also share across heads and siblings, which differ in biases, steps and rates.
This sharing reduces coefficient-generation costs; the recurrence still
updates $dN$ state coordinates.

\paragraph{Recurrence.}
The SSM core evaluates the recurrence of Section~\ref{sec:discretization},
\[
 h_t=A_th_{t-1}+b_t,\qquad y_t=C_th_t+D_tu_t,
\]
per subsystem with state $h_t\in\R^N$, input $u_t\in\R^R$ and output $y_t\in\R^O$, where
$A_t\in\R^{N\times N}$ is diagonal or $2\times2$-block diagonal, $B_t\in\R^{N\times R}$, $C_t\in\R^{O\times N}$,
$D_t\in\R^{O\times R}$ ($R=O=1$ in HM), and $b_t=\Delta_tB_tu_t$ under Euler forcing or the
two-endpoint write~\eqref{eq:hm_disc}, which also uses $u_{t-1}$. Section~\ref{sec:hm_transition_laws} gives the class and input window of both designs.

\paragraph{Scan algorithms.}
Both chunked scans carry state across boundaries
(Section~\ref{sec:hybridmamba_complexity}), differing in transition resolution
and write/read shapes. Our fused Triton/Gluon kernels~\citep{tillet2019triton}
accept initial and return final recurrent state, accumulated phase and previous
write; the mixer also carries convolution history. All carries and rotations
are differentiable, supporting training across segments
(Appendix~\ref{app:state_validation}). GPU-specific chunk lengths, tiles and
backward schedules change execution, not the recurrence.

\paragraph{SSD-MIMO scan.}
Each scan head uses $A_t=\alpha_{\eta,r,t}I$, in folded coordinates when
rotating (Section~\ref{sec:hm_mhr}); a chunk is a masked matrix contraction.
Each sibling takes $R$ inputs through $B_t\in\R^{N\times R}$ and returns
$O$ outputs through $C_t\in\R^{O\times N}$. Complex blocks fold rotation
into these maps with phase per pair $k$ (Section~\ref{sec:hm_transition_laws}).
Intra-chunk work grows with $RO$~\eqref{eq:scan_work_bounds}: $R=O=8$
costs about $64$ times the $R=O=1$ term. The scan covers SSD-MIMO, HM's
flat rows and Mamba~2/3 SISO presets. Chunk-level kernels form masked
read/write scores and evaluate $R\times O$ routed contractions, rotation,
the two-endpoint write and carry, with dedicated $R=O=1$ paths. They use
Proposition~\ref{prop:ssd}'s SSD decomposition, as do Mamba~2/3
\citep{dao2024transformers,lahoti2026mamba3}.
Appendices~\ref{app:gemm_details} and~\ref{app:two_endpoint_eval} give the
routed contractions/chunk factorization and two-endpoint evaluation/numerics.

\paragraph{Per-coordinate scan.}
Diagonal transitions may vary across all $N$ state coordinates. Complex rows
use real arithmetic, folding rotation into $B_t,C_t$
(Section~\ref{sec:hm_transition_laws}) or applying it inside the scan for
$P<16$, where folding costs more. A rotating pair shares decay and phase
across its two coordinates: these vary per $k$, with $N/2$ pairs if all rows
rotate. Each log decay must sum at most two terms, each a static
subsystem/coordinate coefficient times a selected head-shared schedule
(Appendix~\ref{app:design_interface}). The scan covers HM's per-coordinate
rows and the Mamba~1 preset; Appendix~\ref{app:kernel_schedules} gives both
scans' schedules.

\subsection{Transition laws, rotation and the write}
\label{sec:hm_transition_laws}

Let $s_t\in\{x_t,u_t\}$ denote the selected coefficient source.
The selected step $\Delta_{\eta,t}>0$ is the softplus of a projection of $s_t$. Each law has a negative
rate $a$; the decay applied per step is $\alpha=\exp(\Delta a)\in(0,1)$, except under HM's potential law (below).
In Table~\ref{tab:hm_laws}, $W_a$ and $b_a$ are rate-projection weights
and biases, $A_{\log}$ is the learned log-rate magnitude, and
$\operatorname{act}$ denotes the selected rate activation.

\begin{table}[!ht]
\centering\small
\caption{Decay laws. Rates $a$ are negative, so the applied decay $\alpha=\exp(\Delta a)$ lies in $(0,1)$ (potential law: see text). Heavy-tail selection is the
default for the SSD dynamic rate and the HM dynamic and per-head dynamic rates.}
\label{tab:hm_laws}
\begin{tabular}{@{}p{.19\linewidth}p{.42\linewidth}p{.32\linewidth}@{}}
\toprule
Family/law & Formula & Resolution/default \\
\midrule
SSD dynamic & $a_t=-\max(\operatorname{ht}(W_as_t+b_a),10^{-4})$ & Default; head/token, optionally sibling; zero-initialized bias \\
SSD static & $a=-\exp(A_{\log})$ & Per $(\eta,r)$; spread rate ladder \\
HM static & $a=-\exp(A_{\log})$ & Default; $(\eta,p,n)$, tied within complex pairs; no rate bias \\
HM dynamic & $a_t=-\operatorname{act}(W_as_t)$ & Per $n$ or head; no bias/floor \\
HM potential & $\log\alpha=\Delta a+c_n^j(\mu_t-\mu_{t-1})$ & Bounded selected potential \\
HM flat dynamic & $a_t=-\max(\operatorname{act}(W_as_t+b_a),10^{-4})$ & Head/token; zero-initialized bias \\
HM flat static & $a=-\exp(A_{\log})$ & Per head \\
\bottomrule
\end{tabular}
\end{table}

Here $\operatorname{ht}(v)=1+v$ for $v\ge0$ and $(1-v)^{-1}$ otherwise.
For HM potential, $a_n^j=-\exp(A_{\log,n}^j)$ and
\begin{equation}
 \mu_t=\tanh(W_\mu s_t),\qquad
 c_n^j=G_\mu\operatorname{sigm}(v_n^j),\qquad
 G_\mu=\tfrac12\log\mu_{\mathrm{cap}},
 \label{eq:hm_bounded_mu}
\end{equation}
where $W_\mu$ is the bias-free potential-projection matrix and
$\operatorname{sigm}$ is the logistic sigmoid,
$\mu_{\mathrm{cap}}>1$ is finite (default $4$), the potential starts at $\mu_{-1}=0$, and $v_n^j$ is the
stored gain logit. Thus $0<c_n^j<G_\mu$. Potential increments telescope:
in exact arithmetic their factor on any span is in
$(1/\mu_{\mathrm{cap}},\mu_{\mathrm{cap}})$, and the total decay product
is in $(0,\mu_{\mathrm{cap}})$
(Corollary~\ref{cor:gauge_cases}).
Here $G_\mu$ is a gain bound, distinct from the SSD channel-group count $G$.

\paragraph{Endpoint maps and gates.}
Both designs form their write and read maps from base projections $\widehat B_t,\widehat C_t$
of $s_t$, RMS-normalized along the state axis by default:
\begin{equation}
 \bar B_t=\phi_B(s_t)\odot\widehat B_t+\beta,\qquad
 \bar C_t=\phi_C(s_t)\odot\widehat C_t+\gamma,\qquad
 \phi_B(s_t)=\phi(W^B_gs_t),\quad \phi_C(s_t)=\phi(W^C_gs_t),
 \label{eq:endpoint_gates}
\end{equation}
where $\phi_B,\phi_C$ are the write and read gates: a gate activation $\phi$, by default
$\phi(v)=1+v$, of gate projections $W^B_g,W^C_g$. SSD-MIMO initializes these
projections to zero, so its gates start at one; HM uses uniform fan-in
initialization when its optional gates are enabled. $\odot$ is the elementwise
product, broadcast along the input or output axis,
and $\beta,\gamma$ are static additive biases. A gate rescales the base map per
state row before the bias is added, and the rotation below then acts on the result. Without gates,
$\phi\equiv1$. Sections~\ref{sec:hm_mhr} and~\ref{sec:hm_m1} state which indices gates and biases
resolve and when gates are on; Appendix~\ref{app:design_maps} gives the per-design shapes.

\paragraph{Paired rotation.}
One rotation law serves both designs on their complex row blocks, with
$K_{\mathrm{rot}}$ rotating pairs; real rows have identity rotation, including the read offset.
In the default interleaved layout, pair $k$ is $(2k,2k+1)$. With the subsystem index
suppressed (Sections~\ref{sec:hm_mhr} and~\ref{sec:hm_m1} resolve it), pair $k$ carries
\begin{equation}
 A_t|_k=\alpha_{k,t}\,\mathcal R(\theta_{k,t}),\qquad
 \theta_{k,t}=-\,\omega_k\tanh(q_{k,t})\,\Delta_t,\qquad
 \Theta_{k,t}=\sum_{s\le t}\theta_{k,s},
 \label{eq:hm_rotation_law}
\end{equation}
the rotational real representation of the complex eigenvalue $\alpha_{k,t}e^{i\theta_{k,t}}$
(Theorem~\ref{thm:rotation_fold}(i)), where $q_{k,t}$ is the projected angle feature and
$\omega_k=\pi\tanh(\varphi_k)$ the frequency,
with a learned $\varphi_k$ initialized at $3.8$, so that $\omega_k$ starts at $0.999\pi$. The phase
changes by $\theta_{k,t}$ per step whatever the decay rate: in $\exp(\Delta(a+i\omega))$ the rate
$a<0$ sets only the modulus, so the step is the clock.
This is a data-dependent, trainable RoPE, related to selective RoPE~\citep{movahedi2025selectiverope}: with a constant angle per step it reduces to RoPE
(Section~\ref{ex:rope}); here the increment depends on the token through $q_{k,t}$ and $\Delta_t$.
As in Mamba~3, the rotation is folded into the endpoint maps
(Theorem~\ref{thm:rotation_fold}(ii)),
\[
 \widetilde B_t=\mathcal R(-\Theta_t)\,\bar B_t,\qquad
 \widetilde C_t=\bar C_t\,\mathcal R(\Theta_t+\nu),\qquad
 \widetilde C_t\widetilde B_s=\bar C_t\,\mathcal R(\Theta_t-\Theta_s+\nu)\,\bar B_s ,
\]
where $\mathcal R(\theta)$ is the $2\times2$ rotation of Section~\ref{sec:theory_complex},
$\bar B,\bar C$ are the endpoint maps after their additive biases, and $\nu$ holds one static
offset per pair. By default both designs apply gate $\to$ bias $\to$ rotation, so the ungated
score depends on the phase difference alone. The offset is carried on the read side: a
write-side offset would enter the score only through $\nu^C-\nu^B$, an exact gauge redundancy.
Zero phase is the real case; rotation is off by default in both designs.

\paragraph{Discretization.}
SSD rows and HM per-coordinate and flat dynamic rows default to the
two-endpoint rule, while HM flat static rows use exponential Euler; the
two-endpoint rule uses a selected per-head weight:
\begin{equation}
 h_t=A_th_{t-1}+\Delta_t\big[\lambda_{\eta,t}v_t+(1-\lambda_{\eta,t})A_tv_{t-1}\big],
 \qquad \lambda_{\eta,t}=\operatorname{sigm}(w_\lambda^\top s_t+b_\lambda),
 \label{eq:hm_disc}
\end{equation}
where $v_t$ is the \emph{undiscretized} write in the state coordinates of $h_t$,
$v_{-1}=0$, and $A_t$ is the full transition --- decay and rotation --- acting on
the transported previous write. Exponential Euler is $\lambda\equiv1$,
$b_t=\Delta_tv_t$, used by the exact Mamba~1 and Mamba~2 specializations. The step
multiplies the write exactly once (no $\Delta^2$), and the read map is never
discretized.

\paragraph{Class and window.}
{Before folding, transitions consist of real diagonal rows and
rotation--scaling pairs. The latter share a fixed complex diagonalizing
basis (Corollary~\ref{cor:abelian_closure}).} Without complex rows, both designs therefore lie in
$\mathcal D^{\R}(\ell)$ (Definition~\ref{def:diagonal_equivalent}). With complex rows, the
implemented recurrence is the realification of a diagonal complex realization, so the designs lie in $\mathcal D^{\C}(\ell)$, as their rows in
Table~\ref{tab:specializations} of Section~\ref{sec:classification} record, and not in
$\mathcal D^{\R}(\ell)$ {if some realized rotation increment lies
outside $\pi\mathbb Z$, giving nonreal eigenvalues. The fold of
Theorem~\ref{thm:rotation_fold}(ii) gives a real diagonal recurrence with
window bound $L$; it need not retain a smaller window because the folded
maps carry accumulated phase}
(Corollary~\ref{cor:realification_window_classes}). The window $\ell$ comes only from the
convolution, the two-endpoint write and the potential, since all other coefficient maps are
token-local: a width-$k_{\mathrm{conv}}$ convolution gives window $k_{\mathrm{conv}}$, and the
two-endpoint write and the potential each reach back one token, so either of them, or both
together, add one. Hence $\ell=k_{\mathrm{conv}}+1=5$ by default, $\ell=2$ without convolution, as
for Mamba~3, and $\ell=1$ with Euler writes, no convolution and no potential rows. In that token-local case, the time-shared state-mediated kernel
is a masked linear attention kernel over the class's field (Theorem~\ref{thm:mla_membership}(i)).
Gates preserve this: they act before the rotation or, after it, are pair-tied and commute with it
(Section~\ref{sec:hm_m1}).
\par\endgroup

\subsection{SSD-MIMO: vector-valued local inputs and outputs}
\label{sec:hm_mhr}
\par\begingroup
With the siblings of Section~\ref{sec:hm_kernels}, group $(\eta,g)$ has the shared drive
$u_{\eta,g,t}=u_t[\eta,g,:]\in\R^R$, and $o$ indexes output coordinates. The design has
$HGR=d$ input channels, $HGRN=dN$ state coordinates and $HGO$ output channels,
hence is width-preserving at the default $O=R$. The implementation supports
$R,O\in\{1,2,4,8\}$ and at most one real and one complex row block, which share
one decay law. Since the scan work grows with $RO$ (Section~\ref{sec:hm_kernels}), $R=O=8$ is
implemented but not recommended; where multi-value routing is wanted, we recommend $R=O=2$, the
most optimized configuration, or $R=O=4$ when more values per step are needed. The declared shape is
$(H,GR;R,N,O)$: each sibling receives all $R$ supplied channels and
writes their linear combination through an $N\times R$ map, permitting
multiple independent value directions within one subsystem. Coefficient
maps may depend on the complete declared input, beyond this local drive.
In folded coordinates $\tilde h_{\eta,g,r,t}=\mathcal R(-\Theta_{\eta,r,t})h_{\eta,g,r,t}$,
the transition is scalar, and the recurrence with the default two-endpoint write~\eqref{eq:hm_disc}
and the group readout are
\begin{equation}
 \begin{aligned}
 \tilde h_{\eta,g,r,t}
 &=\alpha_{\eta,r,t}\tilde h_{\eta,g,r,t-1}
   +\Delta_{\eta,r,t}\big[\lambda_{\eta,t}\widetilde B_{\eta,r,t}u_{\eta,g,t}
   +(1-\lambda_{\eta,t})\,\alpha_{\eta,r,t}\widetilde B_{\eta,r,t-1}u_{\eta,g,t-1}\big],\\
 y_{\eta,g,t}
 &=\sum_{r=1}^{R}\zeta_{\eta,r,t}
   \left(\widetilde C_{\eta,r,t}\tilde h_{\eta,g,r,t}
   +D_{\eta,g,r,:}u_{\eta,g,t}[r]\right)\in\R^O.
 \end{aligned}
 \label{eq:ssdmimo_recurrence}
\end{equation}
with $\tilde h_{\eta,g,r,t}\in\R^N$, $\widetilde B_{\eta,r,t}\in\R^{N\times R}$, $\widetilde C_{\eta,r,t}\in\R^{O\times N}$,
$D_{\eta,g,r,:}\in\R^O$ and scalar sibling weights $\zeta_{\eta,r,t}$.
Folding moves the rotation of the previous write into $\widetilde B_{\eta,r,t-1}$, so only
the decay $\alpha_{\eta,r,t}$ transports it. The skip acts inside the
weighted sibling sum and permits $O\ne R$.
On complex blocks, pair $k$ of scan head $(\eta,r)$ follows~\eqref{eq:hm_rotation_law} with its
decay $\alpha_{\eta,r,t}$ and step $\Delta_{\eta,r,t}$; biases enter per scan head and never along the
group axis $g$. SSD-MIMO can also learn $\omega_k$ per head
and pair or per head, use per-head angle features $q_{\eta,k,t}$, and carries the read offsets per head
and pair, $\nu_\eta$.
The mixer's output gate acts after the sibling sum, on both state and skip
contributions, with one factor per group and output coordinate.

At $R>1$ the default sibling weights are
$\zeta_{\eta,r,t}=\zeta_{\eta,r}+(W_{\mathrm{mix}}s_t)_{\eta,r}$,
with static weights initialized to $1/R$ and the selected projection to zero.
They are shared over $g,o$ and are not softmax-normalized. Static learned
mixing remains an option. At $R=1$ the default has no mixing parameters and
the reduction is the identity, including when $O>1$; the output gate remains.

\paragraph{Variation across siblings.}
If transitions, writes and initial states agree across $r$, all sibling states coincide, and
varying only the read cannot separate them. By default the siblings of a group therefore
differ in their steps $\Delta_{\eta,r,t}$, hence also in their phases on complex blocks, and on real
blocks in write gates ($R>1$) and read gates ($O>1$) of~\eqref{eq:endpoint_gates} that vary per sibling: the write gate
$\phi_B(s_t)[:,r]$ scales every input column of sibling $r$, the read gate resolves output coordinates, both
are shared over the groups $g$, and the biases $\beta^{\eta,r}\in\R^{N\times R}$,
$\gamma^{\eta,r}\in\R^{O\times N}$ resolve the scan head. Complex blocks are ungated by default,
since their siblings already differ in phase through their steps; gates are pair-tied when enabled. The additive biases are initialized to differ across siblings,
and the dynamic rate can optionally vary per sibling. These map gates differ from the mixer output
gate (Appendix~\ref{app:design_maps}).

\paragraph{Local transport rank.}
At fixed coefficient schedules one SSD-MIMO sibling can attain local
transport rank $\min(N,R,O)$ (Theorem~\ref{thm:crosstalk}). Among the selective models of
Table~\ref{tab:specializations}, SSD-MIMO is thus the only one that is MIMO in this rank sense:
S5 and LRU attain local rank above one with time-invariant maps, and Mamba~3, including its
MIMO variant with its data-independent scalar lift, keeps rank one
(Corollary~\ref{cor:m3_value_rank}). The sibling sum is a linear aggregation over subsystems in the sense of
Section~\ref{sec:theory_setup}, with aggregation matrices $M_t^r=\zeta_{\eta,r,t}I_O$ on the shared
drive of width $R'=R$. By~\eqref{eq:aggregation_rank}, the whole group's transfer from
$u_{\eta,g,s}$ to $y_{\eta,g,t}$ therefore has rank at most $\min(O,R)$: summing siblings
does not raise the routing rank beyond the supplied width. Several scalar-input subsystems can
also attain full aggregate rank, so this local interface distinction is
not a whole-model expressivity ordering; comparisons must control state
budgets and parameter sharing. Appendix~\ref{app:ssd_local_rank} gives
the witness and the group aggregation bound.
\par\endgroup

\subsection{HybridMamba: the hybrid of the Mamba~1--3 laws}
\label{sec:hm_m1}
\par\begingroup
HybridMamba generalizes Mamba~1's per-coordinate diagonal decay to multi-head
systems with data-dependent complex states: it shares Mamba~3's two-endpoint write and
rotation law while keeping per-row decay.
Subsystem $j=(\eta,p)$ receives the scalar $u_t^j=u_t[\eta,p]$ and carries $h_t^j\in\R^N$:
\begin{equation}
 h_{t}^j=A_{t}^jh_{t-1}^j+\Delta_{\eta,t}\big[\lambda_{\eta,t}B_{\eta,t}u_{t}^j
   +(1-\lambda_{\eta,t})A_t^jB_{\eta,t-1}u_{t-1}^j\big],
 \qquad y_{t}^j=C_{\eta,t}h_{t}^j+D_\eta u_{t}^j,
 \label{eq:hm_recurrence}
\end{equation}
with $B_{\eta,t}\in\R^{N\times1}$, $C_{\eta,t}\in\R^{1\times N}$ and the default two-endpoint
write~\eqref{eq:hm_disc}, or Euler forcing as in the Mamba~1 preset;
{$A_t^j\in\R^{N\times N}$ is block diagonal, with scalar real rows
and $2\times2$ rotation--scaling blocks on complex pairs.}
Thus the declared local widths are $R=O=1$ and the shape is
$(H,P;1,N,1)$, as recorded in Table~\ref{tab:specializations}.
Heads differ through their biases, steps and rates. Static rates may vary over
$(\eta,p,n)$. Rotation ties each pair's modulus and phase.
Unlike the complex variant of Mamba~1~\citep[Section~3.6]{gu2024mamba},
which drives each mode's decay and phase by the same step $\Delta_t$, so that
each eigenvalue moves along a fixed spiral, HybridMamba selects the phase
through input-dependent angle features~\eqref{eq:hm_rotation_law},
independently of its decay law. Pair $k$ of subsystem $(\eta,p)$ carries
$A_t|_k=\alpha_{\eta,p,k,t}\,\mathcal R(\theta_{\eta,k,t})$ by~\eqref{eq:hm_rotation_law} with the head step
$\Delta_{\eta,t}$; the modulus is tied within the pair but may vary over $(\eta,p)$ and $k$. HM
implements the write and read gates of~\eqref{eq:endpoint_gates}, with head-shared gate features
and per-head biases, before or after the rotation, but leaves them off by default
(Section~\ref{sec:exp_mqmkvar}); its post-rotation gates are pair-tied, so they commute with the
rotation (Section~\ref{sec:hm_transition_laws}). HM's flat rows, which run on the SSD-MIMO scan, rotate as in Mamba~3:
$\omega=\pi$ fixed, $\nu=0$ and a head-shared angle feature; rotation settings are global per HM core,
so a core with complex flat rows also fixes the frequencies and offsets of its per-coordinate rows.
The step is one projection from $s_t$ to the $H$ heads by default; when the heads are many and
narrow, a Mamba~1-style low-rank bottleneck of rank $\lceil d_{\mathrm{model}}/16\rceil$ is cheaper and
used instead, and either can be configured. Mamba~1 is the case $H=d$, $P=1$ with static real rows,
Euler writes and Mamba~1's maps (Section~\ref{sec:mamba_recovery}).

\paragraph{State-row mixtures.}
\phantomsection\label{sec:hm_mixture}
Partitioning the state rows into blocks $\mathcal N_q$ gives independent recurrences, a direct
sum of block subsystems whose readouts add,
\begin{equation}
 \{1,\ldots,N\}=\biguplus_q\mathcal N_q,\qquad
 y_{t}^j=\sum_q C_{\eta,t}[:,\mathcal N_q]\,h_{t}^j[\mathcal N_q]+D_\eta u_{t}^j.
 \label{eq:hm_rowsplit}
\end{equation}
Each block chooses a law and real or paired state, preserving rotation
pairs; readouts add without a convexity constraint. The step, the coefficient source and the
base B/C projections are shared across blocks, each block reading its own rows. One state can
thus mix Mamba~1-style per-coordinate decay, Mamba~2-style per-head decay and Mamba~3-style rotation
with the two-endpoint write; the subcomponents behave like their Mamba counterparts but are not
exact Mamba~1--3 layers, whose maps and steps would differ. Per-coordinate blocks
use the per-coordinate scan; flat blocks use SSD at $R=O=1$.
Appendices~\ref{app:impl_mh_state} and~\ref{app:mamba_mapping} give state
continuation, flat rows and exact specializations.
\par\endgroup

\needspace{6\baselineskip}
\par\begingroup
\subsection{Computational cost}
\label{sec:hybridmamba_complexity}

{For batch size $\mathsf B$, drive width $d$ and state width $N$, let
$W$ denote SSM-core work and $M$ memory saved for backward, measured in
scalar entries. The bounds exclude surrounding projections; $M$ excludes
parameters and temporary workspace, counting inputs and outputs only when
saved. HM's reverse affine recurrence (Lemma~\ref{lem:ssm_gradients}) and
the adjoints of SSD's local and chunk-state products preserve their respective
forward work orders.
The bounds $W$ therefore apply to forward and backward separately, including
state recomputation (Appendix~\ref{app:backward_cost}).
Both routes use bounded tiles of length $T$, with
$m=\lceil L/T\rceil$ growing with sequence length.
Let $s_{\mathrm{HM}},s_{\mathrm{SSD}}\in\{0,1\}$ indicate whether recurrent
boundary states are saved or recomputed.

\begin{subequations}\label{eq:scan_work_bounds}
\paragraph{HybridMamba.}
The coordinate kernel uses Triton/Gluon's hierarchical associative scan:
thread-local prefixes, Kogge--Stone-style lane shuffles and inter-warp
propagation. The shuffle work is $\bigO(w\log w)$ over $w$ lanes, with
fixed warp width $w=32$. For bounded tiles, scanning and carrying across
chunks costs $\bigO(\mathsf BmTdN)=\bigO(\mathsf BLdN)$.
Grouping $d=HP$ coordinates into heads changes coefficient reuse, not the
$dN$ state recurrences. HM's flat rows use the SSD algorithm below.

With shared coefficients, drive/output tensors ($\mathsf BLd$), head clocks
($\mathsf BLH$) and base endpoint, rate and angle features ($\mathsf BLN$)
give $\bigO(\mathsf BL(d+N))$ storage since $H\le d$.
Recurrent checkpoints store a $dN$-coordinate state every $T_{\mathrm c}$
tokens, where $T_{\mathrm c}\mid T$ is also the backward tile length;
the forward scan still spans $T$. Rotating rows temporarily form
$\mathsf BLHN_{\mathrm f}$ folded-map entries, with $N_{\mathrm f}\le N$
coordinates on this route and endpoint biases shared across $P$.
These maps are recomputed, but phase checkpoints every $T_\phi$ tokens
($T_\phi=128$ here) remain saved independently of $s_{\mathrm{HM}}$. Thus,
\begin{equation}
\begin{aligned}
 W_{\mathrm{HM}}&=\bigO(\mathsf BLdN),\\
 M_{\mathrm{HM}}&=\bigO\bigl(\mathsf BL(d+N)
       +s_{\mathrm{HM}}\mathsf BdN\lceil L/T_{\mathrm c}\rceil
       +\mathsf BHN_{\mathrm f}\lceil L/T_\phi\rceil\bigr).
\end{aligned}
\end{equation}

\paragraph{SSD-MIMO.}
Each of $HR=d/G$ scan heads processes its $G$ value streams in tiles of
width $G_t\le G$. For each write/read pair $(r,o)$, chunk endpoint maps
$B_r,C_o\in\R^{T\times N}$ form $C_oB_r^\top$; the decay-weighted score
$K_{o,r}=\Gamma\odot(C_oB_r^\top)$ then multiplies routed inputs
$X_r\in\R^{T\times G_t}$ (Appendix~\ref{app:gemm_details}).
The score is shared across streams, but recomputed for each group tile.
The $RO$ pairs cost $\bigO(RO(T^2N+T^2G_t))$ per tile and chunk;
multiplying by $\mathsf Bm\,HR(G/G_t)=\mathsf Bm d/G_t$ gives
$\bigO(\mathsf BLd\,TRO(1+N/G_t))$.
Chunk-state writes multiply $(N\times TR)(TR\times G_t)$ matrices;
incoming-carry reads multiply $(TO\times N)(N\times G_t)$ matrices.
Together they add $\bigO(\mathsf BLdN(R+O))$, with the same order for their
adjoints. The scalar boundary-state recurrence and its adjoint cost only
$\bigO(\mathsf Bm dN)$, already covered by this term.
For $R=O=1$ and $G_t=G=P$, the score term is Mamba~2's $T(1+N/P)$.

Saved sibling drives/readouts cost $\mathsf BLd(R+O)$; accumulated phases
and rotating SISO's folded maps contribute at most $\mathsf BLHRN$.
The two-endpoint write saves the local $O\times R$ product
$\widetilde C_t\widetilde B_t$ per token and scan head ($\mathsf BLHRRO$).
Shared endpoint maps contribute $\mathsf BLN(R^2+O)$ for $R$
sibling-specific $N\times R$ write maps and one $O\times N$ read map.
Sharing the write map reduces $NR^2$ to $NR$; sibling-dependent write gates
materialize the extra axis, while per-head biases are applied in the kernel.
The general fused route and rotation-free SISO route reconstruct boundary
states by default ($s_{\mathrm{SSD}}=0$); the specialized rotating SISO
backward retains them ($s_{\mathrm{SSD}}=1$). Hence,
\begin{equation}
\begin{aligned}
 W_{\mathrm{SSD}}&=\bigO\bigl(\mathsf BLd[N(R+O)+TRO(1+N/G_t)]\bigr),\\
 M_{\mathrm{SSD}}&=\bigO\bigl(\mathsf BL[d(R+O)+HR(N+RO)+N(R^2+O)]
       +s_{\mathrm{SSD}}\mathsf BdN\lceil L/T\rceil\bigr).
\end{aligned}
\end{equation}
\end{subequations}}
\phantomsection\label{sec:hm_costs}
\begin{table}[t]
\centering\footnotesize
\caption{{{Same-function ours/official ratios ($<1$: faster or smaller) over
Table~\ref{tab:bench_cost}'s five shapes. Panels show forward-and-backward
time, peak allocation and retained-for-backward storage for aligned SSM-core
kernels and eager full mixers; memory accounting is defined in
Table~\ref{tab:bench_cost}. We measure on an H100 PCIe GPU in bf16;
Mamba~1 uses $N=16$, Mamba~2/3 $N=128$. Final rows give equal-weight geometric means of unrounded per-shape
ratios. Table~\ref{tab:app_parity_summary} adds four-Large/six-LM summaries,
forward-only and graph timings; additional HM shapes use the labelled earlier
buffer schedule. Absolute times: Tables~\ref{tab:app_parity_m1},
\ref{tab:app_parity_1b_m2} and~\ref{tab:app_parity_1c_m3}.}}}
\label{tab:bench_parity}
\adjustbox{max width=\linewidth}{\begin{tabular}{@{}lrrrrrr@{}}
\toprule
& \multicolumn{2}{c}{HM / Mamba 1} & \multicolumn{2}{c}{SSD / Mamba 2} & \multicolumn{2}{c}{SSD / Mamba 3 SISO} \\
\cmidrule(lr){2-3}\cmidrule(lr){4-5}\cmidrule(l){6-7}
Shape & Kernel & Mixer & Kernel & Mixer & Kernel & Mixer \\
\midrule
\multicolumn{7}{l}{\emph{Forward-and-backward time ratios}} \\
\addlinespace[2pt]
Balanced & 0.68 & 0.87 & 0.91 & 1.07 & 0.84 & 0.87 \\
Long & 0.79 & 0.90 & 0.87 & 1.05 & 0.77 & 0.76 \\
Wide & 0.72 & 0.94 & 0.95 & 1.02 & 0.83 & 0.89 \\
Batch & 0.72 & 0.82 & 0.85 & 1.00 & 0.89 & 0.86 \\
LM & 0.83 & 0.94 & 1.52 & 1.48 & 1.17 & 0.95 \\
\addlinespace[2pt]
Geomean (5 shapes) & 0.74 & 0.89 & 1.00 & 1.11 & 0.89 & 0.86 \\
\midrule
\multicolumn{7}{l}{\emph{Peak memory ratios}} \\
\addlinespace[2pt]
Balanced & 1.09 & 0.86 & 2.56 & 1.27 & 0.95 & 1.01 \\
Long & 1.10 & 0.86 & 2.53 & 1.26 & 0.95 & 1.01 \\
Wide & 1.08 & 0.86 & 2.62 & 1.28 & 0.95 & 1.01 \\
Batch & 1.09 & 0.86 & 2.10 & 1.23 & 0.97 & 1.03 \\
LM & 1.09 & 0.86 & 2.33 & 1.26 & 0.96 & 1.02 \\
\addlinespace[2pt]
Geomean (5 shapes) & 1.09 & 0.86 & 2.42 & 1.26 & 0.96 & 1.01 \\
\midrule
\multicolumn{7}{l}{\emph{Retained for backward ratios}} \\
\addlinespace[2pt]
Balanced & 1.11 & 1.06 & 0.54 & 1.03 & 0.90 & 1.08 \\
Long & 1.11 & 1.06 & 0.57 & 1.06 & 0.90 & 1.08 \\
Wide & 1.11 & 1.05 & 0.53 & 1.02 & 0.90 & 1.07 \\
Batch & 1.10 & 1.06 & 0.62 & 1.11 & 0.91 & 1.08 \\
LM & 1.11 & 1.06 & 0.57 & 1.06 & 0.90 & 1.08 \\
\addlinespace[2pt]
Geomean (5 shapes) & 1.11 & 1.06 & 0.56 & 1.05 & 0.90 & 1.08 \\
\bottomrule
\end{tabular}
}
\end{table}

\begin{table}[t]
\centering\footnotesize
\caption{{{Absolute forward-and-backward time (ms) at aligned SSM-core scope, with retained-for-backward
storage and peak allocation (MiB) at the LM shape. Shapes
$(\mathsf B,d_{\mathrm{inner}},L)$: Balanced $(16,4096,2048)$, Long $(2,2048,32768)$,
Wide $(4,8192,4096)$, Batch $(128,1024,1024)$, LM $(6,2048,2048)$
(tensor sizes: Table~\ref{tab:bench_shapes}). We measure on an H100 PCIe GPU
in bf16, with $N=128$ and $P=64$ unless indicated ($P=1$ for Mamba~1). Mamba rows use official
kernels. Retained counts unique nonparameter backing storage held for
backward, including saved caller inputs/outputs; aliases count once.
Peak is the maximum allocation during one forward/backward call above resident
parameters, inputs and output cotangent, including output, workspace and
new gradients. The metrics overlap and must not be added; neither measures
full-model memory. Rows generally compute different mappings, so these are
costs rather than recovery ratios. Mamba~3 MIMO's timing metric is explained below.}}}
\label{tab:bench_cost}
\adjustbox{max width=\linewidth}{\begin{tabular}{@{}lrrrrrr@{\hspace{1.2em}}r@{}}
\toprule
& \multicolumn{4}{c}{Kernel time (ms)} & \multicolumn{3}{c}{LM (one shape)} \\
\cmidrule(lr){2-5}\cmidrule(l){6-8}
Model & Balanced & Long & Wide & Batch & Time & Retained & Peak \\
& & & & & (ms) & (MiB) & (MiB) \\
\midrule
Mamba 1 (official) & 119.5 & 121.1 & 114.2 & 136.4 & 20.70 & 116.0 & 178 \\
Mamba 2 (official) & 9.63 & 10.69 & 8.79 & 11.59 & 2.01 & 104.3 & 239 \\
Mamba 3 SISO (official) & 10.95 & 17.52 & 11.75 & 11.03 & 2.49 & 507.9 & 759 \\
Mamba 3 MIMO R=4 (official)\textsuperscript{a} & 65.76 & 148.86 & 73.92 & 65.82 & 14.27 & 203.6 & 1618 \\
\midrule
HM m1exact (P=1) & 70.65 & 83.11 & 72.00 & 70.98 & 15.82 & 201.0 & 284 \\
HM-0, P=64 & 94.98 & 138.9 & 95.44 & 95.51 & 19.30 & 275.5 & 453 \\
HM-75, P=64 & 106.5 & 150.6 & 106.8 & 111.3 & 22.51 & 443.8 & 1043 \\
HM-100, P=64 & 104.9 & 149.0 & 104.1 & 111.2 & 22.09 & 289.8 & 1223 \\
HM-0, P=32 & 94.98 & 141.7 & 95.88 & 95.70 & 19.25 & 284.5 & 464 \\
HM-75, P=32 & 111.9 & 163.3 & 113.8 & 117.7 & 24.15 & 456.4 & 1632 \\
HM-100, P=32 & 111.4 & 163.1 & 112.1 & 120.1 & 23.92 & 299.6 & 2001 \\
\midrule
SSD m2exact (R=O=1) & 8.66 & 8.99 & 8.35 & 9.67 & 3.04 & 59.2 & 555 \\
SSD m3exact (R=O=1) & 9.18 & 13.51 & 9.76 & 9.76 & 3.12 & 459.0 & 728 \\
SSD R1O1 real & 13.60 & 14.14 & 13.32 & 15.09 & 3.37 & 125.3 & 619 \\
SSD R1O1 complex & 12.44 & 16.76 & 13.34 & 13.09 & 3.16 & 514.5 & 776 \\
SSD R2O2 real & 43.48 & 51.57 & 43.40 & 48.55 & 8.86 & 273.1 & 1050 \\
SSD R2O2 complex & 68.83 & 81.93 & 67.42 & 76.68 & 13.98 & 468.9 & 1442 \\
SSD R4O4 real & 182.7 & 202.4 & 180.0 & 200.2 & 35.38 & 639.5 & 2334 \\
SSD R4O4 complex & 328.0 & 361.7 & 324.3 & 355.1 & 63.71 & 1030.3 & 3125 \\
\bottomrule
\end{tabular}
}
\par\smallskip
\noindent\begin{minipage}{\linewidth}\footnotesize
\textsuperscript{a} Official Mamba~3 MIMO at kernel scope reports summed profiler GPU-kernel
time over three calls, excluding the TileLang wrapper's persistent host overhead;
the other timing rows report elapsed CUDA-event time. Its eager mixer timing is
therefore n/a; its CUDA-graph replay is reported separately.
\end{minipage}

\end{table}
{\paragraph{Measured cost.}
We compare the Mamba recoveries with the official modules on matched functions
(Table~\ref{tab:bench_parity}, ratios ours/official), using copied weights and
output checks before timing. Measurements use H100 PCIe, bf16, PyTorch~2.14 and
Triton~3.8. Kernel scope is aligned within each comparison pair; mixer timings
cover the full module. Each side uses its production configuration.
We report three-round medians from the first valid comparison-group attempt,
with per-group canary spread at most $1.15$; whole-run drift is recorded rather
than gated. Appendix~\ref{app:speed_cells} gives the protocol and limitations.
The main tables share four Large shapes with $\mathsf BLd_{\mathrm{inner}}=2^{27}$
and one LM shape, $(\mathsf B,d_{\mathrm{inner}},L)=(6,2048,2048)$.
The appendix additionally reports six-LM summaries; HM measurements beyond
these five shapes retain the earlier backward-buffer schedule and are labelled separately.

Across these five shapes, kernel forward-and-backward time is lower on
geometric mean for the Mamba~1 function ($0.74\times$, $N=16$) and
Mamba~3 SISO ($0.89\times$), with Mamba~2 at near parity ($1.00\times$).
The eager full mixers are also faster than Mamba~1/3 ($0.89/0.86\times$),
but slower than Mamba~2 ($1.11\times$).
Individual shapes can be slower: at the LM shape,
Mamba~2 recovery has kernel/mixer ratios $1.52/1.48$, and Mamba~3 recovery
has $1.17/0.95$.

On geometric mean, kernel memory saved for backward is about $44\%$ lower
than Mamba~2 and $10\%$ lower than Mamba~3, but $11\%$ higher than Mamba~1.
Kernel peak allocation is $4\%$ lower than Mamba~3, $9\%$ higher than
Mamba~1 and $2.42\times$ Mamba~2. Full mixers hold $5$--$8\%$ more memory
for backward; their peak allocation is $14\%$ lower than Mamba~1,
near Mamba~3 and $26\%$ higher than Mamba~2.
Mixer measurements include projections, coefficient preparation and output
processing; these comparisons do not isolate the effect of fusion.
The memory figures describe the measured scopes, not end-to-end training.

HM reduces backward temporary buffers and their overlapping lifetimes,
without reducing memory saved for backward. Relative to the unreduced-buffer
schedule, HM-0/75/100 at $P=64$ reduce geometric-mean kernel peak allocation
by $48$--$51\%$ and mixer peak by $35$--$45\%$ over the five shapes.
Mixer training time decreases by $4.2\%$/$0.7\%$ for HM-0/75 but increases
by $3.8\%$ for HM-100. At $P=32$, rotating HM-75/100 mixers are
$3.9\%$/$7.8\%$ slower on geometric mean, up to $18\%$ on one shape;
the reduced peak therefore comes with a configuration-dependent timing
tradeoff (Table~\ref{tab:hm_workspace_tradeoff}).

Table~\ref{tab:bench_cost} reports absolute scan costs across model
configurations that generally compute different input--output mappings,
rather than matched-function recovery ratios. From R2O2 to R4O4, kernel forward-and-backward time grows
$3.92$--$4.20\times$ for real rows and $4.41$--$4.81\times$ for complex rows
across the five shapes. Model head width is fixed, but kernel head count,
kernel width and execution configurations change with rank. These deployed
costs do not establish a universal $4\times$ scaling bound or isolate the
cause of the complex overhead.
A small baseline-tracking ablation with 32-bit floating-point (fp32) input and output is given in
Appendix~\ref{app:fp32_baseline_tracking}; the main timing and memory results
use bf16.
}

\par\endgroup

\subsection{Recovering the Mamba configurations}
\label{sec:mamba_recovery}
\par\begingroup

Table~\ref{tab:mamba_recovery} summarizes the exact mappings of
Appendix~\ref{app:mamba_mapping}, where finite-precision agreement is
checked separately.
\begin{table}[!htbp]
\centering\small
\caption{Mamba SISO configurations recovered in exact arithmetic after
matching coefficients, projections, normalization and initial state.}
\label{tab:mamba_recovery}
\begin{tabular}{@{}l l p{.57\linewidth}@{}}
\toprule
Target & Design & Matching settings \\
\midrule
Mamba~1 & HybridMamba & Real static per-coordinate decay; exponential Euler;
state-branch coefficients; matched convolution and projections. \\
Mamba~2 & SSD-MIMO & Scalar decay, scalar local interfaces; exponential Euler;
joint drive/write/read convolution and output normalization. \\
Mamba~3 SISO & SSD-MIMO & Scalar local interfaces; matched rate law and
rotation, two-endpoint write, write/read normalization and biases;
convolution-free mixer. \\
\bottomrule
\end{tabular}
\end{table}

\paragraph{Finite-precision agreement.}
With copied weights and identical inputs, all tested outputs, input gradients
and parameter gradients agree with the official modules within twice the
arithmetic-derived tolerance at $\mathsf B=6$, $L=2048$: $60/60$, $75/75$ and
$80/80$ tensor--seed comparisons for Mamba~1, 2 and 3, respectively, in each
of bf16 and fp32. The full matrix uses Blackwell, with an A100 cross-check
(Appendix~\ref{app:numerics_exact}). In bf16 the median ours--official output
errors are $3.5\cdot10^{-3}$, $1.4\cdot10^{-3}$ and $4.3\cdot10^{-3}$.
At $\mathsf B=2$, $L=16384$, Mamba~3's bf16 angle-projection gradient exceeds
twice the tolerance in all five seeds ($75/80$ comparisons pass).
fp32 input/output does not imply matched scan arithmetic: official Mamba~3
uses bf16 scan operands, whereas our default uses TensorFloat-32 (tf32) products.

\paragraph{Gradient accuracy of the rotation.}
The angle-projection gradient is sensitive to long context and strong forgetting
(Appendix~\ref{app:numerics_redteam}). Across nine constructed scenarios,
official Mamba~3's median angle-gradient error against 64-bit floating point (fp64) is $1.4$--$7.4$
times ours in bf16 and $14$--$49$ times ours with fp32 input/output; the latter
comparison uses different operand precisions. These ratios do not imply that
ours always meets tolerance. With tf32 operands, our angle gradient fails at
strong forgetting tokens; full-fp32 products pass the nine scenarios but fail
a separate, stronger boundary probe. The bf16 long-context stress also has
two gradient tensors exceeding tolerance on Blackwell but not in the A100
cross-check. Neither our models nor the official baselines showed training
instability attributable to these effects in our experiments.

Trained independently under the same recipe, from independent initializations,
the matched Mamba~3 SISO configuration and the reference end within $0.0032$ nats
of each other at $1$B tokens (test loss $3.2335$ versus $3.2367$, one seed each);
this is a training-level check, not a parity test; downstream
comparisons and the real-only variant are in
Appendix~\ref{app:lm_final_breakdowns}.

\par\endgroup

\section{Experiments}
\label{sec:experiments}

\par\begingroup
We evaluate complete configurations on recall and language modelling, measuring
recall separately from text loss: Zoology links much of efficient LMs' quality
gap to in-context recall, and Based shows its tradeoff with recurrent state size
\citep{arora2023zoologymeasuringimprovingrecall,arora2024simple}. Our tasks build on
Zoology's MQAR and Based's LM recall evaluation; Section~\ref{sec:hm_costs} reports cost.
\par\endgroup

\subsection{Multi-token associative recall (MQMKVAR)}
\label{sec:exp_mqmkvar}
\par\begingroup

MQMKVAR extends Zoology's MQAR~\citep{arora2023zoologymeasuringimprovingrecall}
and MAD's fuzzy recall~\citep{poli2024mad} with controlled prefix/substring
overlap, variable evaluation key lengths, masked ambiguous continuations and
padded answer slots (Figure~\ref{fig:recall_tasks}; Appendix~\ref{app:mqmkvar_protocol}).
No earlier answer enters the input: each value list must be recalled from the
context state. Easy uses Zoology's MQAR generator; medium/hard use ours with
vocabulary $8192$, $32$ pairs, key/value lengths up to $12$ and sequence length
up to $2048$ (Table~\ref{tab:mq_protocol}). We target confusable multi-token keys,
without comparing difficulty to MAD. Easy saturates at three-seed means of
$99.9$--$100\%$ for all configurations, including attention; medium/hard separate
them, spanning $22.2$--$96.9\%$ and $1.1$--$82.6\%$
(Table~\ref{tab:mqmkvar_main}), the separation MQMKVAR was designed to provide.

Easy/medium share peak learning rate (LR) $10^{-3}$ without per-model tuning
(Appendix~\ref{app:mqmkvar_analysis}). Hard adds $5\times10^{-4}$ and
$2\times10^{-3}$ for every configuration, three seeds per rate; the table reports
each configuration's best eligible rate on this common grid
(Appendix~\ref{app:mq_lr_sweep}).
\paragraph{Parameter budget and configurations.}
Our non-embedding budgets are within $\pm2\%$ of Mamba~1's (M1)
$1{,}220{,}352$ parameters. Mamba~2/3 (M2/M3) retain upstream head width $64$
and expansion $2$, restricting $d_{\text{model}}$ to steps of $32$:
$288$ gives $1.158$M/$1.183$M, below the anchor, so we use $320$,
$16$--$18\%$ above it. Attention follows Zoology's Figure~2
\citep{arora2023zoologymeasuringimprovingrecall} and meets the M1 size floor.
SSD-MIMO R$k$O$k$ denotes $R=O=k$, with real-row gates by default for $k>1$
(Section~\ref{sec:hm_mhr}); -r/-c means all-real/half-complex rows.
R1O1 and HM mixtures are ungated; HM's $A$ and ``dynamic'' denote static
per-coordinate decay and input-dependent rate rows. Appendix~\ref{app:mqmkvar_analysis}
gives gate masks, mixtures and per-seed scores.
\begin{table}[!ht]
\centering\footnotesize
\caption{{Held-out token accuracy (percent), three-seed mean [min--max].
Easy/medium use peak LR $10^{-3}$; hard selects each row's highest mean
among rates with three completed seeds. Selected rates and diverged runs
(which make a rate ineligible) appear in Appendix~\ref{app:mq_lr_sweep}.
Bold marks the best medium/hard means; easy saturates.
Widths are model/inner; r/c denote real/complex.}}
\label{tab:mqmkvar_main}
\adjustbox{max width=\linewidth}{\begin{tabular}{@{}l r c c c c c@{}}
\toprule
Configuration & Non-emb. & $d_{\text{model}}$ / $d_{\text{inner}}$ & Easy & Medium & Hard & Hard LR \\
\midrule
M1 & 1,220,352 & 256 / 512 & 99.9 [99.9--100.0] & 89.3 [88.8--89.7] & 69.6 [67.3--72.8] & $10^{-3}$ \\
M2\textsuperscript{a} & 1,410,300 & 320 / 640 & 100.0 [100.0--100.0] & 91.7 [88.7--97.3] & 42.0 [29.6--66.0] & $5\!\times\!10^{-4}$ \\
M3\textsuperscript{a} & 1,438,952 & 320 / 640 & 99.9 [99.6--100.0] & 76.6 [58.1--87.1] & 27.8 [14.3--48.2] & $5\!\times\!10^{-4}$ \\
Attention & 1,236,368 & 392 / --- & 100.0 [100.0--100.0] & 22.2 [22.2--22.2] & 1.1 [0.4--1.9] & $10^{-3}$ \\
\midrule
R1O1-r & 1,203,486 & 296 / 576 & 100.0 [100.0--100.0] & 89.6 [89.4--89.7] & 67.0 [63.9--71.7] & $5\!\times\!10^{-4}$ \\
R1O1-c & 1,223,070 & 296 / 576 & 99.9 [99.9--100.0] & 89.8 [89.5--90.3] & 69.4 [61.2--79.3] & $5\!\times\!10^{-4}$ \\
R2O2-r & 1,217,308 & 248 / 448 & 100.0 [100.0--100.0] & 91.3 [91.0--91.5] & 82.0 [81.1--83.3] & $10^{-3}$ \\
R2O2-r, gates off & 1,206,492 & 312 / 448 & 100.0 [100.0--100.0] & 89.4 [89.3--89.6] & 60.3 [51.0--76.9] & $5\!\times\!10^{-4}$ \\
R2O2-c & 1,207,944 & 248 / 512 & 100.0 [100.0--100.0] & 91.1 [90.8--91.5] & \textbf{82.6} [82.1--83.5] & $10^{-3}$ \\
R2O2-c, gates off & 1,217,256 & 280 / 512 & 100.0 [100.0--100.0] & 90.4 [89.9--90.8] & 82.2 [81.4--82.6] & $10^{-3}$ \\
R2O2-c, head frequency & 1,195,992 & 248 / 512 & 100.0 [100.0--100.0] & 91.0 [90.5--91.7] & 82.0 [80.9--83.2] & $5\!\times\!10^{-4}$ \\
R4O4-r & 1,204,744 & 176 / 384 & 100.0 [100.0--100.0] & 91.5 [91.1--92.0] & 82.0 [80.9--83.4] & $10^{-3}$ \\
R4O4-c & 1,214,248 & 208 / 384 & 100.0 [100.0--100.0] & 90.5 [89.6--91.6] & 82.2 [81.9--82.7] & $5\!\times\!10^{-4}$ \\
\midrule
HM $A$+dynamic & 1,236,416 & 256 / 512 & 100.0 [100.0--100.0] & 93.5 [91.2--96.0] & 80.6 [79.8--81.4] & $10^{-3}$ \\
HM complex-$A$ mix & 1,237,472 & 256 / 512 & 100.0 [100.0--100.0] & 95.1 [90.9--98.1] & 79.6 [79.1--80.0] & $10^{-3}$ \\
HM complex-dynamic mix & 1,237,472 & 256 / 512 & 100.0 [100.0--100.0] & \textbf{96.9} [96.5--97.2] & 80.7 [79.8--81.5] & $5\!\times\!10^{-4}$ \\
HM all-$A$ real & 1,236,416 & 256 / 512 & 100.0 [100.0--100.0] & 91.2 [90.3--92.9] & 79.2 [78.2--80.1] & $5\!\times\!10^{-4}$ \\
HM all-$A$ complex & 1,200,922 & 256 / 496 & 100.0 [100.0--100.0] & 92.3 [89.5--96.9] & 69.0 [61.4--74.6] & $10^{-3}$ \\
\bottomrule
\end{tabular}
}
\par\smallskip\noindent\begin{minipage}{\linewidth}\footnotesize
\textsuperscript{a}Upstream default head width and expansion at the next supported width above the budget.
\end{minipage}
\end{table}

\begin{figure}[!ht]
\centering
\includegraphics[width=\linewidth]{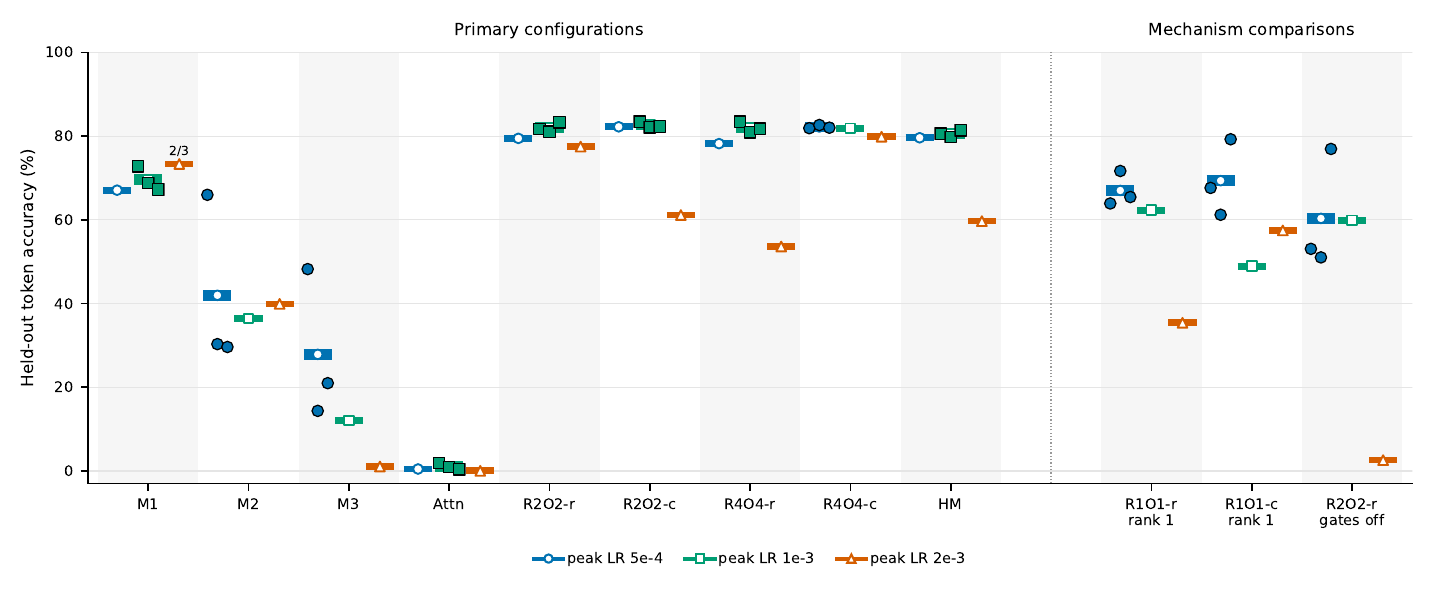}
\caption{Hard-tier held-out token accuracy at each swept peak LR, rates
increasing from left to right within each configuration. Horizontal lines
mark seed means; the selected eligible rate has a thick line and filled
seed markers. Mamba~1's $2/3$ marks one diverged seed at $2\times10^{-3}$.
The mechanism group shows rank-one SSD and real R2O2 with gates off.
HM denotes $A$+dynamic; r/c denote real/complex.}
\label{fig:mqmkvar_final_points}
\end{figure}

\paragraph{Local width and recall.}
Real/complex R2O2 and R4O4 beat all three Mamba baselines on hard at each
configuration's best eligible LR, despite M2/M3's $16$--$18\%$ larger budgets
(Table~\ref{tab:mqmkvar_main}). Real R4O4 matches real R2O2 at width $176$
versus $248$ ($29\%$ narrower); complex R4O4 does not improve on R2O2.
Thus widening beyond two gives no further hard-tier gain here; on medium,
HM's complex mixtures lead.
\paragraph{Gates and sibling diversity.}
At the shared LR, removing gates strongly hurts real-row hard-tier recall but
barely changes the complex configuration (Table~\ref{tab:mq_lr_sweep}).
The asymmetry persists after selecting each configuration’s best eligible LR: real gates-off reaches $60.3$ [$51.0$--$76.9$], versus $82.0$ with gates, whereas complex gates-off reaches $82.2$, versus $82.6$ for the complex default.

For the half-complex default, only real-block gates are removed. Write gates
vary real siblings' write directions; complex siblings' selected steps also
vary their phases, consistent with the stronger real-row gate effect. This
remains a hypothesis: steps and additive biases also differ across siblings
(Section~\ref{sec:hm_mhr}), and budget matching changes model width.
At $2\times10^{-3}$, the highest tested LR, both gates-off configurations fail
on all three seeds (at most $3.8$), versus gated scores of $77.5$ and $61.1$
(Table~\ref{tab:mq_complete}).
\paragraph{Routing rank.}
At each configuration’s best eligible LR, ungated R2O2-c reaches $82.2$ [$81.4$--$82.6$], against $69.4$ [$61.2$--$79.3$] for R1O1-c, with disjoint seed ranges. For real rows, gated R2O2-r reaches $82.0$, but ungated R2O2-r, at $60.3$, falls below R1O1-r ($67.0$) on average.
The ungated rank-one HM mixture also reaches $80.6$ [$79.8$--$81.4$]. These configuration comparisons do not isolate the effect of routing rank (Figure~\ref{fig:mqmkvar_final_points}).

\paragraph{Attention reference.}
Attention solves easy but remains low on medium at the shared LR and on hard
throughout the sweep, which does not exhaust attention tuning. Replacing only
learned absolute positions with RoPE gives low scores on medium/hard
($0.7$/$0.3\%$, three seeds per tier).
\paragraph{HybridMamba.}
At the Mamba~1 budget, the three HM mixtures with dynamic rows exceed all three Mamba baselines on medium and hard. On medium, the complex-dynamic mix has the highest mean of all configurations, $96.9$ [$96.5$--$97.2$]; on hard, the mixtures reach $79.6$--$80.7$, the best within $1.9$ points of the best SSD-MIMO configuration.
The gains build up from Mamba~1’s static design. With the same static per-coordinate decay, HM's maps, two-endpoint write and head organization raise the hard mean from $69.6$ to $79.2$; dynamic rows add $1.4$ points on hard and $2.3$ on medium; complex dynamic rows add a further $3.4$ on medium and are neutral on hard. Complex rows with only static decay hurt on hard ($69.0$). Each step changes several settings at once, so the steps are not isolated mechanisms.

HM uses neither map gates nor potential rows: earlier studies found gates
not worth their width cost at this budget, and potential rows less accurate
and stable than dynamic rows (Appendix~\ref{app:mqmkvar_analysis}).
\par\endgroup

\subsection{Language-model pretraining on FineWeb-Edu}
\label{sec:exp_fineweb}
\par\begingroup

We train approximately $500$M-parameter models on FineWeb-Edu
\citep{penedo2024fineweb}, one seed per arm, one recipe and no per-model tuning.
Table~\ref{tab:lm_final_main} reports $2$B/$10$B-token test loss and downstream
scores; Figure~\ref{fig:lm_loss_curves} shows validation loss.
Recall-6 averages Based/Just Read Twice's SWDE, FDA, SQuAD, TriviaQA, NQ-1024
and DROP tasks, also reported by Mamba~3~\citep[Table~4]{lahoti2026mamba3}.
Our needle-in-a-haystack (NIAH) diagnostic is our own implementation of RULER-style single- and
multi-key needles
\citep{hsieh2024ruler} at $512$--$2048$ tokens; Mamba~3 also uses RULER.
Appendices~\ref{app:fineweb_protocol} and~\ref{app:lm_final_breakdowns} give the
recipe/configurations and per-task/paired comparisons. Paired bootstraps on
shared evaluation items estimate score-difference intervals; significance
means Holm-adjusted $p<0.05$. These intervals capture finite-evaluation-set
uncertainty, not variation across training seeds.
\begin{table}[!t]
\centering\small
\setlength{\tabcolsep}{3pt}
\caption{FineWeb-Edu language modelling, one training seed per model,
$504.5$--$508.0$M parameters, at most the Mamba~3 SISO anchor ($507.95$M)
and within $1\%$ below it. The $2$B checkpoints are on the constant-LR
trajectory; $10$B continues from $2$B with cosine decay to $10^{-5}$.
Loss is test-shard cross-entropy (nats). Zero-shot (ZS) columns show LAMBADA,
HellaSwag, PIQA, ARC-Easy, ARC-Challenge, WinoGrande and OpenBookQA,
followed by their average (ZS); HellaSwag and ARC-Challenge use normalized
accuracy; Recall-6 and NIAH are in percent. Underlining marks the best
value per column and budget, without implying significance.
Attention is a context reference trained with the Mamba recipe.
HM-$x$ is HybridMamba with $x\%$ complex dynamic rows; the remaining rows are
split equally between real static and real dynamic decay (Appendix~\ref{app:lm_final_breakdowns}).}
\label{tab:lm_final_main}
\adjustbox{max width=\linewidth}{\begin{tabular}{@{}lrrrrrrrrrrr@{}}
\toprule
 & & \multicolumn{8}{c}{Zero-shot accuracy (\%)} & \multicolumn{2}{c}{Retrieval (\%)} \\
\cmidrule(lr){3-10}\cmidrule(l){11-12}
Model & Loss & LAMB. & Hella. & PIQA & ARC-E & ARC-C & Wino. & OBQA & Avg. & Recall-6 & NIAH \\
\midrule
\multicolumn{12}{@{}l}{\emph{2B tokens}} \\
Mamba~1 & 3.0979 & 25.5 & 34.3 & 64.6 & 55.2 & 26.3 & 50.3 & \underline{21.6} & 39.7 & 14.8 & 19.4 \\
Mamba~2 & 3.0793 & 24.9 & 35.0 & 63.2 & 54.0 & 26.2 & 50.6 & 20.4 & 39.2 & 15.8 & 35.8 \\
Mamba~3 SISO & 3.0592 & 27.4 & \underline{35.2} & 63.9 & \underline{55.7} & 26.5 & 51.1 & 19.4 & 39.9 & 16.8 & 25.6 \\
Mamba~3 MIMO & \underline{3.0561} & 27.3 & 34.9 & 64.6 & 55.6 & 27.2 & 49.4 & 20.8 & \underline{40.0} & 16.8 & 27.4 \\
Attention & 3.1439 & 25.3 & 32.8 & 62.8 & 52.0 & 25.4 & 51.2 & 18.4 & 38.3 & \underline{25.0} & \underline{56.8} \\
\addlinespace[2pt]
SSD-MIMO R2O2-r & 3.0736 & 23.5 & 34.3 & 63.3 & 54.6 & 25.8 & 49.8 & 20.2 & 38.8 & 18.3 & 39.0 \\
SSD-MIMO R2O2-c & 3.0682 & 25.2 & 34.7 & 64.6 & 54.7 & \underline{28.2} & 50.9 & 20.2 & 39.8 & 17.0 & 36.8 \\
HM-0 & 3.0745 & \underline{28.7} & 35.0 & 64.3 & 53.9 & 26.5 & \underline{51.9} & 19.2 & 39.9 & 16.2 & 22.6 \\
HM-25 & 3.0701 & 28.2 & 34.8 & 64.5 & 55.0 & 25.3 & \underline{51.9} & 19.2 & 39.8 & 17.0 & 16.3 \\
HM-75 & 3.0639 & 26.7 & 35.1 & \underline{65.2} & 54.5 & 24.9 & 51.1 & 20.4 & 39.7 & 16.9 & 33.7 \\
HM-100 & 3.0698 & 28.3 & 34.7 & 63.7 & 55.2 & 25.6 & 51.1 & 18.8 & 39.6 & 16.9 & 18.5 \\
\midrule
\multicolumn{12}{@{}l}{\emph{10B tokens}} \\
Mamba~3 SISO & 2.6991 & 36.7 & 44.8 & 68.0 & 64.3 & 30.2 & 52.2 & \underline{25.6} & 46.0 & 27.3 & 38.8 \\
Mamba~3 MIMO & \underline{2.6933} & 38.0 & \underline{45.2} & 68.7 & 64.4 & \underline{31.9} & \underline{53.7} & 23.4 & \underline{46.5} & 27.9 & \underline{46.8} \\
\addlinespace[2pt]
SSD-MIMO R2O2-r & 2.7184 & 36.1 & 43.9 & 68.1 & \underline{65.0} & 30.6 & 51.5 & 23.4 & 45.5 & 27.3 & 36.5 \\
SSD-MIMO R2O2-c & 2.7101 & 36.4 & 44.5 & 67.8 & 64.1 & 31.4 & 53.3 & \underline{25.6} & 46.2 & \underline{28.0} & 42.8 \\
HM-75 & 2.7057 & \underline{38.8} & 44.7 & \underline{69.3} & 62.7 & 29.4 & 52.6 & 25.0 & 46.1 & 27.6 & 37.8 \\
\bottomrule
\end{tabular}
}
\end{table}

\paragraph{Loss and downstream development.}
Mamba~3 MIMO leads on loss at both budgets and $10$B zero-shot/NIAH means.
At $2$B, our primary SSD-MIMO arms improve on Mamba~1 by $0.024$--$0.030$
nats and match or slightly improve on Mamba~2 ($0.006$--$0.011$ nats),
but trail both Mamba~3 variants. At $10$B, R2O2-c trails Mamba~3 SISO by
$0.011$ nats with NIAH $4.0$ points higher; R2O2-r loses its early Recall-6
lead over SISO. The baseline ordering agrees with Mamba~2 over Mamba~1 and
Mamba~3 over Mamba~2~\citep{dao2024transformers,lahoti2026mamba3}.
\paragraph{Routing rank on language.}
SSD-MIMO's MQMKVAR rank advantage does not transfer to language modelling here.
Rank-one Mamba~3 MIMO (Corollary~\ref{cor:m3_value_rank}) beats R2O2-c at $10$B
on loss ($2.6933$ versus $2.7101$) and NIAH ($+4.1$ points, significant),
with Recall-6 level ($27.9$ versus $28.0$). By our analysis its advantage is not routing
rank: each channel mixes $R$ head-shared projections for its write/read
directions, whereas SISO shares one write/read map per head. This suggests
rank one suffices here and richer per-channel coefficients matter more than
routing more values; differing decay laws and state geometries prevent an
isolated ablation.
\paragraph{HybridMamba.}
HM improves on its Mamba~1 starting point, as on MQMKVAR. At $2$B, HM-75 beats
Mamba~1/2 on loss ($3.0639$ versus $3.0979$/$3.0793$) and Recall-6
($+2.1$/$+1.0$ points); NIAH is $14.2$ above Mamba~1 and $2.1$ below Mamba~2,
all four retrieval-score differences significant. It narrows Mamba~1's loss gap to Mamba~3 SISO from $0.039$
to $0.005$ nats. At $10$B it stays near SISO: $0.007$ nats behind
($2.7057$ versus $2.6991$), with ZS/Recall-6/NIAH differences
$+0.1$/$+0.2$/$-1.0$ points, none significant; its $2$B NIAH lead of $8.1$
points disappears. Per-row decay (Section~\ref{sec:hm_m1}) is neither a clear
advantage nor disadvantage here. Without multi-input routing, HM-75 has our
lowest $10$B loss but trails Mamba~3 MIMO ($0.012$ nats; NIAH $-9.1$ points,
significant). Of four tested complex fractions, excluding Mamba~3's $50\%$,
$75\%$ gives the lowest loss and highest NIAH: at $2$B its NIAH exceeds HM-0/25 by $11.1$/$17.3$ points;
HM-100 loses $15.2$ points and $0.006$ nats (Table~\ref{tab:lm_complex_fraction}).
All four beat Mamba~1/2's $2$B loss; ZS and Recall-6 means differ by at most
$0.3$ and $0.8$ points. Appendix~\ref{app:lm_final_breakdowns} gives full scores,
paired comparisons and the format check.
\paragraph{Attention context.}
At $2$B, attention leads Recall-6 ($25.0$) and NIAH ($56.8$) despite higher
loss ($3.1439$). Against Mamba~3 SISO it gains $8.2$ Recall-6 and $31.2$ NIAH points
and loses $1.6$ ZS points, all significant, without attention-specific tuning.
Its Recall-6 lead is mainly extraction: SWDE/FDA $37.2$/$20.9$, versus
R2O2-r's $16.8$/$3.9$, with the other four tasks closer.
The attention--SSM ordering reverses between medium/hard MQMKVAR and LM
retrieval, but the references differ in size (two layers/one head versus
$29$ layers/$16$ heads), data, objectives and scoring: scale is not isolated.
Replacing learned absolute positions with RoPE does not resolve the low
MQMKVAR scores in the tested configuration.
\begin{figure}[!htbp]
\centering
\includegraphics[width=\linewidth]{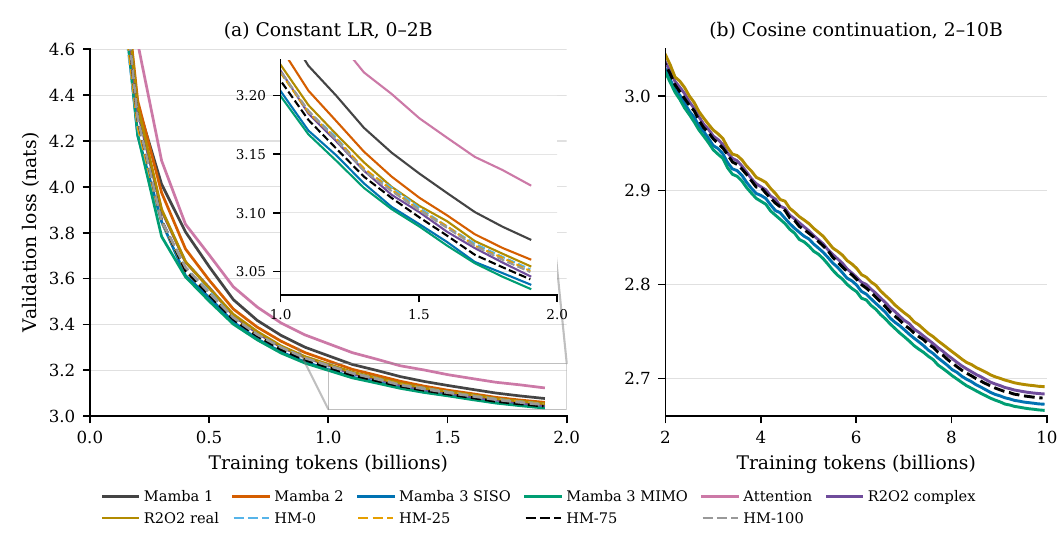}
\caption{Validation-loss curves, one seed per arm, without smoothing:
(a) the eleven main-table configurations at constant LR up to $2$B tokens,
with a $1$--$2$B zoom; (b) the five configurations continued from $2$B to
$10$B with cosine decay. Fixed-window validation losses, distinct
from the test-shard losses of Table~\ref{tab:lm_final_main}.}
\label{fig:lm_loss_curves}
\end{figure}
\par\endgroup

\needspace{8\baselineskip}
\section{Conclusion}
\label{sec:conclusion}
\par\begingroup

Fixed diagonal equivalence holds exactly when the realized transitions commute
pairwise and are individually diagonalizable over the chosen field; their failures are the two exits
(Theorem~\ref{thm:simultaneous_diagonalization}). {Many widely used SSMs satisfy this criterion, including
Mamba~1--3 and S4 when its learned transition is diagonalizable. In a fixed
diagonalizing basis, their cores differ mainly in decay, phase, writes and reads}
(Table~\ref{tab:specializations}). Local retrieval rank is at most
$\min(O,\operatorname{rank}B_s)$ for any transfer, attainable with nonsingular
transfer and unrestricted reads (Theorem~\ref{thm:crosstalk}). With time-shared
token-local matrix maps, diagonal-equivalent state-mediated kernels are exactly masked
linear attention: transitions, reads and writes in the fixed diagonalizing basis
give its decays, queries and keys (Theorem~\ref{thm:mla_membership}(i)).

SSD-MIMO widens local writes and reads, the taxonomy's only selective model
with local rank above one;
HybridMamba combines Mamba~1--3 decay laws in one state, extending Mamba~1 to
multi-head, data-dependent complex states. Both support input-dependent rotation phases and
differentiable carries. At about $1$M parameters, SSD-MIMO's R2O2 configurations
reach about $82\%$ hard MQMKVAR recall, compared with $67$--$69\%$ for R1O1,
while rank-one HM with per-coordinate decay reaches
$80\%$: rank is one way to pack values into a small state. At about $500$M,
scalar-input subsystems already supply rank in aggregate
(Section~\ref{sec:hm_mhr}), possibly explaining why multi-rank routing improves
neither LM loss nor retrieval over rank-one Mamba~3 MIMO at $10$B tokens,
where it leads NIAH and matches Recall-6. {Local rank one appears sufficient in these comparisons;} complex SSD-MIMO stays close
to Mamba~3 SISO, and HM remains competitive with Mamba~3, reaching near parity with SISO
through $10$B tokens but trailing MIMO.

\paragraph{Limitations.}
Local rank permits more independent values per step, not necessarily better
trained models or lower loss. MQMKVAR compares complete configurations differing
in geometry, initialization, steps and biases, so phase diversity remains a
hypothesis; three seeds at each of three common rates measure sensitivity only within
that grid. LM evidence covers one seed per arm, one scale and recipe through
$10$B tokens; small loss gaps are untested. Tasks needing more values per step
may benefit from multi-input routing, but whether MQMKVAR's gain reflects task
or small model scale remains untested.
\par\endgroup

\makeatletter
\if@accepted
\needspace{14\baselineskip}
\subsubsection*{Acknowledgements}
The research leading to these results is funded by the German Federal Ministry for
Economic Affairs and Energy within the project ``NXT GEN AI METHODS -- Generative
Methoden f\"ur Perzeption, Pr\"adiktion und Planung''. The author would like to thank
the consortium for the successful cooperation.

The author thanks the Gauss Centre for
Supercomputing e.V.\ (GCS) for providing computing time on the Supercomputer JUPITER at
the J\"ulich Supercomputing Centre via the John von Neumann Institute for Computing.
JUPITER is supported by the EuroHPC JU and GCS through funding by the European
Commission, the German Federal Ministry of Research, Technology and Space, and the
Ministry of Culture and Science of the State of North Rhine-Westphalia.
\fi
\makeatother

\FloatBarrier
\bibliographystyle{tmlr}
\bibliography{diagonal_equivalence}

\appendix
\FloatBarrier

\section{State-space theory: proofs and examples}
\label{app:sec2_details}

\par\begingroup
\subsection{Proof of Theorem~\ref{thm:zoh_exact}}
\label{app:zoh_proof}
\par\endgroup

\par\begingroup
\begin{remark}
\label{rem:cts_wellposed}
Under Definition~\ref{def:cts_ssm}'s assumptions, the state equation is well posed. Here $\|\cdot\|$
denotes the Euclidean vector norm and its induced matrix operator norm.
On each compact time interval the input range lies in a compact ball, so
continuity of $A,b$ makes $A\circ x,b\circ x$ measurable and bounded,
hence locally integrable. For $F_x(\tau,h):=A(x(\tau))h+b(x(\tau))$, the bound
$\|F_x(\tau,h)-F_x(\tau,\tilde h)\|\le\|A(x(\tau))\|\,\|h-\tilde h\|$
gives global Lipschitz continuity in the state with a locally integrable
Lipschitz bound. Thus a unique Carath\'eodory solution exists on every
finite interval $[\tau_{-1},T]$
\citep[Corollary~2.6 and the subsequent Carath\'eodory remark,
pp.~41--42]{teschl2012ode}. By uniqueness these solutions agree on
overlapping intervals, defining a unique solution on $[\tau_{-1},\infty)$.
For $\mathbb F=\C$, use the realified initial-value problem displayed
after Definition~\ref{def:cts_ssm}, with $v=\chi_S(h)$.
The maps $\chi_S$ and $\chi_S^{-1}$ preserve local absolute continuity and
commute with differentiation almost everywhere. The existence and uniqueness of $v$ therefore give the existence and
uniqueness of $h=\chi_S^{-1}\circ v$.
The Borel-measurable readout then defines the measurable output
$y(\tau)=f(x(\tau),h(\tau))$ uniquely.
\end{remark}

\begin{proof}
The held input is admissible, so Remark~\ref{rem:cts_wellposed} gives
its unique state trajectory. Realification preserves matrix products,
exponentials and integration, transferring the real calculation below
to $\mathbb F=\C$.

\textbf{Single-interval recurrence.} Fix a finite index $t\in\mathbb T_L$ and its interval $[\tau_{t-1},\tau_t]$. Consider first the more general interval problem $h'(\tau) = A(x_t)\, h(\tau) + q_t(\tau)$ with the coefficient held at $A(x_t)$ and an integrable forcing $q_t \colon [\tau_{t-1}, \tau_t] \to \mathbb{F}^S$; the quadrature schemes at the end of Section~\ref{sec:continuous} start from this problem, and the held input~\eqref{eq:zoh_input} is the case $q_t \equiv b(x_t)$. The state equation is understood almost everywhere, as in Definition~\ref{def:cts_ssm}. By the variation-of-parameters formula for linear systems with time-varying forcing~\citep{teschl2012ode}, the unique such solution with initial condition $h(\tau_{t-1}) = h_{t-1}$ is
\begin{equation}\label{eq:proof_voc_general}
  h(\tau) \;=\; e^{(\tau - \tau_{t-1}) A(x_t)}\, h_{t-1} \;+\; \int_{\tau_{t-1}}^{\tau} e^{(\tau - \sigma) A(x_t)}\, q_t(\sigma)\, d\sigma ,
  \qquad \tau \in [\tau_{t-1}, \tau_t].
\end{equation}

At $\tau=\tau_t$, with $h_t=h(\tau_t)$ and $A_t=e^{\Delta_t A(x_t)}$,
this gives the endpoint formula~\eqref{eq:exp_euler_starting}:
\begin{equation*}
 h_t=A_t h_{t-1}+\int_{\tau_{t-1}}^{\tau_t}
       e^{(\tau_t-\sigma)A(x_t)}q_t(\sigma)\,d\sigma.
 \tag{\ref{eq:exp_euler_starting}}
\end{equation*}

Now impose the hold~\eqref{eq:zoh_input}: each token $x_t$ specifies the
input on the whole interval $[\tau_{t-1},\tau_t)$. On the closed interval,
$x(\sigma)=x_t$ except possibly at $\tau_t$, so
$A(x(\sigma))=A(x_t)$ and $q_t(\sigma)=b(x_t)$ almost everywhere.
Changing the signal only at isolated endpoints leaves its state integral
equation unchanged; changing a token changes the input on its entire hold
interval. Thus the constant forcing $b(x_t)$ leaves the integral.
The remaining integral is evaluated by the substitution $r = \tau_t - \sigma$ and termwise integration of the exponential series, which converges uniformly on this finite interval by comparison with the
scalar series $e^{\Delta_t\|A(x_t)\|}$,
\[
  \int_{\tau_{t-1}}^{\tau_t} e^{(\tau_t - \sigma) A(x_t)}\, d\sigma
  \;=\; \int_{0}^{\Delta_t} e^{r A(x_t)}\, dr
  \;=\; \sum_{j \geq 0} \frac{\Delta_t^{\,j+1}}{(j+1)!}\, A(x_t)^{j}
  \;=\; \Delta_t\, \varphi\bigl(\Delta_t A(x_t)\bigr) ,
\]
in the notation of~\eqref{eq:phi_one}. The same comparison shows absolute
convergence in operator norm of the $\varphi$ series for every real or
complex matrix argument. Termwise multiplication gives $Z\varphi(Z)=e^Z-I$, yielding the stated inverse
formula whenever $Z$ is invertible. Evaluating~\eqref{eq:proof_voc_general} at $\tau = \tau_t$ therefore gives
\begin{equation}
  h_t \;=\; e^{\Delta_t A(x_t)}\, h_{t-1}
  \;+\; \biggl(\int_{0}^{\Delta_t} e^{r A(x_t)}\, dr\biggr)\, b(x_t)
  \;=\; A_t\, h_{t-1} \;+\; b_t.
\end{equation}
Here $A_t=e^{\Delta_t A(x_t)}$ and
$b_t=\Delta_t\varphi(\Delta_t A(x_t))b(x_t)$ are the discrete coefficients
of~\eqref{eq:zoh_matrices}, giving~\eqref{eq:recurrence}. The $t$-th LTI subsystem maps $h_{t-1}$ at time $\tau_{t-1}$ exactly to $h_t$ at time $\tau_t$. Whenever $t+1\in\mathbb T_L$, the next subsystem on $[\tau_t, \tau_{t+1}]$ takes $h_t$ as its initial condition. Consequently, the discrete state sequence is the exact trace of the continuous-time trajectory at the partition points; the discrete readout $y_t = f(x_t, h_t)$ of~\eqref{eq:discrete_sampling} is a definition and needs no argument.

\textbf{Closed form.} The base case is $h_0=A_0h_{-1}+b_0$.
At the induction step, multiply the expression for $h_t$ on the left by
$A_{t+1}$ and add $b_{t+1}$. This yields, at every finite index,
\begin{equation}
 h_t=(A_t\cdots A_0)h_{-1}+\sum_{s=0}^{t}(A_t\cdots A_{s+1})b_s
     =\Phi(t,-1)h_{-1}+\sum_{s=0}^{t}\Phi(t,s)b_s,
\end{equation}
where $\Phi(t,s):=A_t\cdots A_{s+1}$ and the empty product is
$\Phi(t,t)=I$, as in~\eqref{eq:transfer_operator}.
For $L=\infty$, this induction proves the identity at every finite index.
To cover continuous time, fix any finite $T\ge\tau_{-1}$. Since
$\tau_t\to\infty$, there is a finite $n$ with $\tau_n>T$; the holds
indexed $0,\ldots,n$ cover $[\tau_{-1},T]$. Their solutions match at the
boundaries by the recurrence and agree with the unique global solution.
As $T$ is arbitrary, this covers all of $[\tau_{-1},\infty)$.
\end{proof}
\par\endgroup

\par\begingroup
\subsection{Consequences of the fixed-basis criterion}
\label{app:theory_horizon}

\paragraph{Streaming membership from finite truncations.}

\begin{corollary}
\label{cor:horizon_independence}
Fix $\ell\in\mathbb N\cup\{\infty\}$ and $S\in\mathbb N$.
Let $\Sigma\in\mathcal S^{\mathbb F}(\ell,S)$ have
horizon $L=\infty$ and, for $L'\in\mathbb N$, let
$\Sigma|_{L'}$ denote the restriction of its causal transition, forcing, and
readout rules to inputs $x\in(\R^d)^{L'}$.  Then
\begin{equation}
  \mathcal A_{\Sigma|_{L'}}\subseteq\mathcal A_{\Sigma|_{L''}}
  \quad(L'\le L''),
  \qquad
  \mathcal A_\Sigma=\bigcup_{L'\in\mathbb N}\mathcal A_{\Sigma|_{L'}},
  \label{eq:horizon_monotone}
\end{equation}
and $\Sigma\in\mathcal D^{\mathbb F}(\ell,S)$ if and only if
$\Sigma|_{L'}\in\mathcal D^{\mathbb F}(\ell,S)$ for every
$L'\in\mathbb N$.
\end{corollary}

\begin{proof}
The step-$t$ coefficients read only the prefix $x_{0:t}$, so the
restrictions are well defined, a transition realized at step $t<L'$ on
$x\in(\R^d)^{L'}$ is realized at the same step on every extension of $x$,
and every transition realized by $\Sigma$ reads a finite prefix, hence lies
in $\mathcal A_{\Sigma|_{L'}}$ for all $L'>t$; this
gives~\eqref{eq:horizon_monotone}. Each restriction retains the
window bound $\ell$ and state dimension $S$.
Criterion~\eqref{eq:simultaneous_algebraic_test} is decided on pairs, and
any pair in the union lies in a single $\mathcal A_{\Sigma|_{L'}}$, so it
holds on $\mathcal A_\Sigma$ iff it holds on every
$\mathcal A_{\Sigma|_{L'}}$;
Theorem~\ref{thm:simultaneous_diagonalization}, applied to $\Sigma$ and to
each $\Sigma|_{L'}$, converts this into the membership claim.
\end{proof}

For time-shared token-local transitions the conclusion is stronger:
$\mathcal A_{\Sigma|_{L'}}=\{A(x_0):x_0\in\R^d\}$ for every $L'\ge1$, so one
pairwise test settles every horizon at once.  It does not extend to a
transition rule with explicit position or prefix dependence, whose realized
family may grow with $L'$: a short truncation can then lie in the class when a
longer one does not, and the corollary is what remains.

\paragraph{Further consequences.}\label{app:criterion_consequences}
A single diagonalizable transition is diagonal-equivalent however dense, and so
is any family $A_t=g_t(A)$, defined as $g_t(A):=Q\,g_t(\Lambda)\,Q^{-1}$ for functions
$g_t\colon\operatorname{spec}(A)\to\mathbb F$ on the eigenvalues of one diagonalizable
$A=Q\Lambda Q^{-1}$, with $g_t(\Lambda)=\operatorname{diag}(g_t(\lambda_n))$, since the
eigenbasis stays fixed; this covers polynomials, $a_tA$ and $e^{\Delta_tA}$.  Membership is decided by
the realized transitions $\mathcal A_\Sigma$ alone, not by forcing or readout:
realizations with the same transitions but different write or read maps get the
same verdict and the same diagonalizing basis, although they need not be similar to
each other.  And similarity preserves neither implementation cost
nor conditioning, the witness $Q$ being free to be dense or ill-conditioned.
The converse to the inherited-eigenbasis statement in
Section~\ref{sec:simultaneous_diagonalization} fails: exponentials $e^{\Delta M_t}$ can share
one eigenbasis while their generators $M_t$ do not, since $M\mapsto e^{\Delta M}$ is not
injective. For example, the generators of rotations about the $x$- and $y$-axis
in $\R^3$, $K_x=e_3e_2^{\!\top}-e_2e_3^{\!\top}$ and $K_y=e_1e_3^{\!\top}-e_3e_1^{\!\top}$, do not
commute, yet $e^{2\pi K_x}=e^{2\pi K_y}=I$. The failure is joint only: for a single
matrix, $e^{\Delta M}$ diagonalizable implies $M$ diagonalizable over $\C$. Otherwise $M$
has a Jordan block $\lambda I+N$ with $N\ne0$, and $e^{\Delta M}$ has the block
$e^{\Delta\lambda}e^{\Delta N}$; since $e^{\Delta N}\ne I$ has $1$ as its only eigenvalue,
it is not diagonalizable.
\par\endgroup

\subsection{Approximation of constant transitions}
\label{app:diagonal_approximation}

\par\begingroup
\begin{proof}[Construction for Remark~\ref{rem:constant_transition_boundary}]
Let $\Sigma\in\mathcal S^{\mathbb F}(\ell,S)$ have constant transition
$A$ and horizon $L$. We construct diagonal approximants over the chosen field.
\begin{enumerate}
\item[\textnormal{(i)}] Over $\C$, choose complex-diagonalizable
$A_n\in\C^{S\times S}$ with $A_n\to A$
\citep[Cor.~5.1]{serre2010matrices}. For a real $\Sigma$, apply this
construction to its complexification $\Sigma^{\C}$.
\item[\textnormal{(ii)}] For a real target $A$, a sequence of real matrices
$A_n\to A$ diagonalizable over $\R$ exists exactly when the target has
real spectrum, $\operatorname{spec}(A)\subseteq\R$. For sufficiency, perturb the diagonal
of a real Jordan form $A=PJP^{-1}$ to obtain triangular $J_n\to J$
with distinct real eigenvalues, and set $A_n=PJ_nP^{-1}$.
Necessity follows from continuity of the spectrum. This condition concerns
transition approximation, not arbitrary output approximation.
\end{enumerate}
In either case, retain the forcing and readout of the chosen realization
($\Sigma^{\C}$ when complexifying), denoted by $b_t^\diamond,f_t^\diamond$.
Define $\Sigma_n$ on the same horizon and state space by
\[
 h_t^{(n)}=A_nh_{t-1}^{(n)}+b_t,\qquad
 y_t^{(n)}=f_t^\diamond(x_{t;\ell},h_t^{(n)}),\qquad h_{-1}^{(n)}=h_{-1}.
\]
Choose $Q_n$ over the corresponding field
with $\Lambda_n=Q_n^{-1}A_nQ_n$ diagonal. The diagonal realization
$\widehat\Sigma_n:=\Sigma_n^{Q_n}$ is explicitly
\[
 \widehat h_t^{(n)}=\Lambda_n\widehat h_{t-1}^{(n)}+Q_n^{-1}b_t,
 \qquad
 \widehat y_t^{(n)}=f_t^\diamond(x_{t;\ell},Q_n\widehat h_t^{(n)}),
 \qquad \widehat h_{-1}^{(n)}=Q_n^{-1}h_{-1}.
\]
It retains the window bound $\ell$, dimension $S$ over the chosen field,
and any time sharing (Lemma~\ref{lem:similarity_invariance}).
For every fixed input $x\in(\R^d)^L$, original initial state
$h_{-1}\in\mathbb F^S$ and $L'\in\mathbb N$ with $L'\le L$, the finite closed form
\eqref{eq:field_closed_form} and $A_n^j\to A^j$ give
\[
 \max_{0\le t<L'}\|Q_n\widehat h_t^{(n)}-h_t\|\longrightarrow0.
\]
If $\Sigma$'s readout is continuous in its state argument at each
target state $h_t$ for this input, then
\[
 \widehat y_t^{(n)}=y_t[\Sigma_n;x,h_{-1}],
 \qquad
 \max_{0\le t<L'}\|\widehat y_t^{(n)}-y_t[\Sigma;x,h_{-1}]\|
 \longrightarrow0.
\]
For real $\Sigma$, complexification preserves its real-initial runs and
readout continuity, so the target outputs are those of $\Sigma$ itself.
Thus even a defective constant transition admits diagonal output
approximants in this sense. This need not give exact fixed similarity
or convergence of the diagonal-coordinate states or parameters.
\end{proof}
\par\endgroup

\par\begingroup
\subsection{Moving frames: potentials and the accumulator form}
\label{app:gauge_examples}

\begin{corollary}
\label{cor:gauge_cases}
Let $\hat\Sigma$ be a realization with diagonal transition
$\hat A_t=\operatorname{diag}(\hat\lambda_{n,t})_{n=1}^{S}$, $\hat\lambda_{n,t}\ne0$, and transfer
$\hat\Phi(t,s)$. Let $\mu_{n,t}\in\R$ ($1\le n\le S$, $t\ge0$, $\mu_{n,-1}:=0$) be a causal
input-dependent sequence (a \emph{potential}), $\mu_t:=(\mu_{n,t})_{n=1}^{S}$, and let
$\Sigma_\mu$ be $\hat\Sigma$ with the transition $A_t=\operatorname{diag}(\lambda_{n,t})$,
$\lambda_{n,t}:=\hat\lambda_{n,t}\,e^{\mu_{n,t}-\mu_{n,t-1}}$. Then:
\begin{enumerate}
\item[\textnormal{(i)}] $\Sigma_\mu$ has transfer
$\Phi_\mu(t,s)=\operatorname{diag}\!\big(e^{\mu_{n,t}-\mu_{n,s}}\big)\,\hat\Phi(t,s)$. If
$\mu_{\min}\le\mu_{n,t}\le\mu_{\max}$ for all $n$ and $t\ge-1$, then
$|\Phi_\mu(t,s)_{nn}|\le e^{\mu_{\max}-\mu_{\min}}\,|\hat\Phi(t,s)_{nn}|$.
\item[\textnormal{(ii)}] Through the frame $Q_t=\operatorname{diag}(e^{\mu_t})$, $\Sigma_\mu$ is
equivalent to the realization $\tilde\Sigma$ with the transitions of $\hat\Sigma$ and the
forcing and readout rescaled per token,
\[
 \tilde b_t=\operatorname{diag}(e^{-\mu_t})\,b_t,\qquad
 \tilde f_t^{\diamond}(x_{t;L},h)=f_t^{\diamond}(x_{t;\ell},Q_th),\qquad
 \tilde C_t=C_tQ_t\quad\text{in the matrix form.}
\]
\end{enumerate}
\end{corollary}

\begin{proof}
(i) The modulation telescopes:
$\prod_{r=s+1}^{t}\hat\lambda_{n,r}e^{\mu_{n,r}-\mu_{n,r-1}}=e^{\mu_{n,t}-\mu_{n,s}}\hat\Phi(t,s)_{nn}$,
and $\mu_{n,t}-\mu_{n,s}\le\mu_{\max}-\mu_{\min}$. (ii) This is Lemma~\ref{lem:gauge_prefix} with the
stated frame: $Q_t^{-1}A_tQ_{t-1}=\operatorname{diag}(\hat\lambda_t)$.
\end{proof}
Individual steps may amplify, $|\lambda_{n,t}|>|\hat\lambda_{n,t}|$, but any interval gains
at most the factor $e^{\mu_{\max}-\mu_{\min}}$ over the base system; outputs of $\Sigma_\mu$ are those
of $\tilde\Sigma$, which has the base decay and rescaled write and read maps. With token-local base maps, a token-local
potential gives $\Sigma_\mu$ window two, through $\mu_{t-1}$, while $\tilde\Sigma$ keeps
token-local maps. For the bounded potential law~\eqref{eq:hm_bounded_mu}, the effective
potential $c_n^j\mu_t$ lies in $(-G_\mu,G_\mu)$, so (i) applies with
$\mu_{\min}=-G_\mu$ and $\mu_{\max}=G_\mu$. Each endpoint difference is
strictly less than $2G_\mu=\log\mu_{\mathrm{cap}}$ in magnitude, so the endpoint factor lies in
$(1/\mu_{\mathrm{cap}},\mu_{\mathrm{cap}})$, and the negative base rate gives total decay in
$(0,\mu_{\mathrm{cap}})$ in exact arithmetic. Composing a frame with free untied write/read
gates does not enlarge the realization class at full-prefix window bound.

\begin{corollary}
\label{cor:gauge_schedule}
Let $\Sigma$ have a real diagonal transition $A_t=\operatorname{diag}(\lambda_{n,t})$ with $\lambda_{n,t}>0$, and write $\mu_{n,t}:=\sum_{j=0}^{t}\log\lambda_{n,j}$ and $\mu_{n,-1}:=0$.  Then Corollary~\ref{cor:gauge_cases}(ii) at base $\hat\lambda\equiv1$ gives $\Sigma$ the input--output map of the decay-free accumulator
\[
  \tilde h_t \;=\; \tilde h_{t-1} \;+\; \operatorname{diag}(e^{-\mu_t})\,b_t,
  \qquad
  y_t \;=\; f_t^{\diamond}\big(x_{t;\ell},\operatorname{diag}(e^{\mu_t})\,\tilde h_t\big),
  \qquad
  \tilde h_{-1}=h_{-1},
\]
whose kernel carries only the differences $e^{\mu_{n,t}-\mu_{n,s}}=\Phi(t,s)_{nn}$.  These invertible diagonal recurrences therefore admit an accumulator
form.
\end{corollary}
Here $\mu_t$ is the cumulative log-decay, so the accumulator's maps depend on the
whole prefix, and large cumulative log-decays make the global rescaling unsuitable in
finite precision. For a complex-diagonal transition the statement applies to the modulus
after the fold of Theorem~\ref{thm:rotation_fold}.
\par\endgroup

\par\begingroup
\subsection{Real block characterization of complex diagonal equivalence}
\label{app:transition_boundary_proofs}

\begin{corollary}
\label{cor:abelian_closure}
\label{cor:rotational_is_dselC}
For a real realization $\Sigma$ with $S$ state coordinates, its
complexification $\Sigma^{\C}$ satisfies
$\Sigma^{\C}\in\mathcal D^{\C}(L)$ if and only if there exist
$J,K\in\mathbb N_0$, $J+2K=S$, and one fixed $Q\in\operatorname{GL}_S(\R)$ such that
for every input and time,
\begin{equation}
 Q^{-1}A_t[\Sigma]Q=
 \operatorname{diag}(a_t^{(1)},\ldots,a_t^{(J)})
 \;\oplus\;\bigoplus_{k=1}^K\rho_t^{(k)}\mathcal R(\theta_t^{(k)}),
 \quad a_t^{(j)}\in\R,\quad\rho_t^{(k)}\ge0.
 \label{eq:real_block_characterization}
\end{equation}
If every realized transition is normal ($A_tA_t^{\top}=A_t^{\top}A_t$), $Q$ can be chosen orthogonal.
In particular, the complexification of every rotational real
representation in Theorem~\ref{thm:rotation_fold}(i) lies in
$\mathcal D^{\C}(L)$.
\end{corollary}

\begin{proof}[Proof of Corollary~\ref{cor:abelian_closure}]
Suppose $\Sigma^{\C}\in\mathcal D^{\C}(L)$. By
Theorem~\ref{thm:simultaneous_diagonalization}, its transition family has
a common complex eigenbasis. Group the basis vectors into joint
eigenspaces according to their eigenvalues for every matrix in the
family. Because every transition $A_t$ of the real realization is real,
$A_tv=\lambda_tv$ implies $A_t\bar v=\bar\lambda_t\bar v$, so conjugation maps each
joint eigenspace to the one with the conjugate eigenvalue profile.

A joint eigenspace whose eigenvalues are all real is invariant under
conjugation and is the complexification of its real part; choose a real
basis there. Each remaining joint eigenspace $E$ is paired with $\bar E\ne E$. Choose one
member $E$ of every pair and enumerate bases of all chosen members as
$v_k=p_k+iq_k$, $k=1,\ldots,K$ ($K=0$ if all eigenvalues are real); then the
$\bar v_k=p_k-iq_k$ span the partners $\bar E$, and $p_k,q_k$ span the real invariant
planes. Splitting $A_tv_k=\rho_t^{(k)}e^{i\theta_t^{(k)}}v_k$ into real and imaginary
parts gives
\[
 A_t\begin{pmatrix}p_k&-q_k\end{pmatrix}
 =\begin{pmatrix}p_k&-q_k\end{pmatrix}\rho_t^{(k)}\mathcal R(\theta_t^{(k)}),
\]
the realification $\mathfrak R(\rho_t^{(k)}e^{i\theta_t^{(k)}})$ of
Section~\ref{sec:continuous}. The same splitting for $\bar v_k$ gives the identical
plane with $(p_k,q_k)$ and $\mathcal R(-\theta_t^{(k)})$, so one member of each pair
suffices. Collecting these real bases gives one
$Q\in\operatorname{GL}_S(\R)$ and the fixed block decomposition
\eqref{eq:real_block_characterization} for the entire family.

Conversely, each scalar block is diagonal and every rotation--scaling
block is diagonalized by the same complex matrix
\[
 U=\frac1{\sqrt2}\begin{pmatrix}1&1\\-i&i\end{pmatrix},\qquad
 U^{-1}\rho\mathcal R(\theta)U
   =\operatorname{diag}(\rho e^{i\theta},\rho e^{-i\theta}).
\]
Combining $Q$ with these fixed block eigenbases simultaneously
diagonalizes every transition, so
$\Sigma^{\C}\in\mathcal D^{\C}(L)$.

If the family is normal, choose a common orthonormal complex eigenbasis
\citep[Thm.~2.5.5]{horn2013matrix}. Distinct joint eigenspaces are
orthogonal. For a unit vector $v=p+iq$, $p,q\in\R^S$, in a nonreal joint eigenspace,
$v$ and $\bar v$ lie in distinct joint eigenspaces and are orthogonal, so
$\bar v^{\,*}v=v^{\!\top}v=\|p\|^2-\|q\|^2+2i\,p^{\!\top}q=0$; hence
$\|p\|=\|q\|=1/\sqrt2$ and $p\perp q$. For orthonormal $v_k,v_j$, $j\ne k$, the same
computation with $v_j^{\,*}v_k=0$ and $\bar v_j^{\,*}v_k=0$ makes $p_j,q_j$ orthogonal to
$p_k,q_k$. Choosing $\sqrt2(p_k,-q_k)$ for each $k$ and real orthonormal bases in the
real joint eigenspaces makes $Q$ orthogonal.
\end{proof}
The block characterization also allows unpaired real scalar blocks;
it need not have $S=2K$.

A completely paired form at the same real dimension is possible exactly
when every real joint eigenvalue profile has even multiplicity: pair
the corresponding equal scalar blocks. Otherwise some scalar blocks
must remain. For example, the constant transition $\operatorname{diag}(1,2)$ defines a realization in
$\mathcal D^{\R}(L)$ but is not similar over $\R$ to a single
rotation--scaling block. This does not affect the exact $K$-complex to
$2K$-real identification in Theorem~\ref{thm:rotation_fold}(i).
\par\endgroup

\par\begingroup
\subsection{Proofs of the gradient and scan lemmas}
\label{app:scan_grad_details}

Every index in this subsection is a discrete step of the recurrence~\eqref{eq:field_ssm_general}.

\begin{proof}[Proof of Lemma~\ref{lem:ssm_gradients}]
Gradients with respect to vectors are row vectors, so Jacobians multiply on the right.  Unrolling, each $h_k$ is a function of $h_{-1}$, $A_{0:k}$ and $b_{0:k}$, and $\mathcal L=\sum_k \mathcal L_k(f_k^{\diamond}(x_{k;\ell},h_k))$, so for any scalar entry $q$ of these inputs
\begin{equation}\label{eq:proof_total_gradient}
\frac{\mathrm d\mathcal L}{\mathrm d q} \;=\; \sum_{k=0}^{L-1} \varepsilon_k\, \frac{\mathrm d h_k}{\mathrm d q} ,
\end{equation}
with $\varepsilon_k$ the direct sensitivity of the lemma (there is no $\varepsilon_{-1}$), whereas the total derivative satisfies $\mathrm d\mathcal L/\mathrm d h_t = \varepsilon_t + (\mathrm d\mathcal L/\mathrm d h_{t+1})\,A_{t+1}$ for $t < L-1$ and $\mathrm d\mathcal L/\mathrm d h_{L-1} = \varepsilon_{L-1}$.
\textit{(i)} $b_t$ enters only at step $t$ with $\mathrm d h_t/\mathrm d b_t = I$, so $\mathrm d h_k/\mathrm d b_t = \Phi(k,t)$ for $k \ge t$ and $0$ before; substituting into~\eqref{eq:proof_total_gradient} gives the closed form, and splitting off the $k = t-1$ term of the same sum at $t-1$ (empty product, contributing $\varepsilon_{t-1}$) and factoring $A_t$ to the right of the remainder gives the recursion.
\textit{(ii)} $A_t$ enters only through $h_t = A_t h_{t-1} + b_t$; since $h_t$ and $b_t$ enter identically from step $t$ on, $\mathrm d\mathcal L/\mathrm d h_t = \mathrm d\mathcal L/\mathrm d b_t$; with $\partial(h_t)_i/\partial(A_t)_{ij} = (h_{t-1})_j$ the chain rule gives $\mathrm d\mathcal L/\mathrm d(A_t)_{ij} = (\mathrm d\mathcal L/\mathrm d b_t)_i\,(h_{t-1})_j$, the matrix $(\mathrm d\mathcal L/\mathrm d b_t)^{\!\top} h_{t-1}^{\!\top}$, of which a diagonal constraint on $A_t$ keeps the entries $i=j$. The scalar constraint $A_t=\alpha_tI$ sums these diagonal entries, giving $(\mathrm d\mathcal L/\mathrm d b_t)h_{t-1}$.
\textit{(iii)} $h_{-1}$ enters only through $h_0 = A_0 h_{-1} + b_0$ with $\mathrm d h_0/\mathrm d h_{-1} = A_0$, so $\mathrm d\mathcal L/\mathrm d h_{-1} = (\mathrm d\mathcal L/\mathrm d b_0)\,A_0$; at the other end $h_{L-1}$ is read only by $\mathcal L_{L-1}$, so $\mathrm d\mathcal L/\mathrm d h_{L-1} = \varepsilon_{L-1}$.
\end{proof}

\begin{proof}[Proof of Lemma~\ref{lem:scan_op}]
\textbf{Associativity.} For $\star$,
\begin{align*}
&\bigl((A_1, v_1) \star (A_2, v_2)\bigr) \star (A_3, v_3)
=\; (A_2 A_1,\, A_2 v_1 + v_2) \star (A_3, v_3) \\
&\;=\; (A_3 A_2 A_1,\, A_3 A_2 v_1 + A_3 v_2 + v_3)
=(A_1, v_1) \star (A_3 A_2,\, A_3 v_2 + v_3) \\
&\;=\;(A_1, v_1) \star \bigl((A_2, v_2) \star (A_3, v_3)\bigr).
\end{align*}
For $\star_{\mathrm{R}}$ the same computation applies with $(A_2 A_3)^{\!\top} = A_3^{\!\top} A_2^{\!\top}$: both bracketings evaluate to $(A_1 A_2 A_3,\; A_3^{\!\top} A_2^{\!\top} v_1 + A_3^{\!\top} v_2 + v_3)$.

\textbf{Forward states.} Induction on $t$. For $t = 0$, the scan output is $p_0 = (A_0,\, A_0 h_{-1} + b_0) = (A_0,\, h_0)$. For $t \geq 1$, assume $p_0 \star \cdots \star p_{t-1} = (\rprod_{j=0}^{t-1} A_j,\, h_{t-1})$. Then
\[
p_0 \star \cdots \star p_t = \bigl(A_t\,\Phi(t-1,-1),\; A_t h_{t-1} + b_t\bigr) = \bigl(\Phi(t,-1),\, h_t\bigr).
\]

\textbf{Reverse gradients.} Transposing the recursion in~\eqref{eq:gradient_b} into columns gives
\[
  \Bigl(\frac{\mathrm d\mathcal L}{\mathrm d b_{t-1}}\Bigr)^{\!\top}
  \;=\; \varepsilon_{t-1}^{\!\top} \;+\; A_t^{\!\top}\Bigl(\frac{\mathrm d\mathcal L}{\mathrm d b_t}\Bigr)^{\!\top},
  \qquad
  \Bigl(\frac{\mathrm d\mathcal L}{\mathrm d b_{L-1}}\Bigr)^{\!\top} \;=\; \varepsilon_{L-1}^{\!\top} .
\]
We show by downward induction on $t$ that the scan equals $\bigl(P_t,\, (\mathrm d\mathcal L/\mathrm d b_t)^{\!\top}\bigr)$ with $P_t := \Phi(L-1,t)$ (empty product $I$ at $t = L-1$). The base case is $\bar p_{L-1} = (I,\, \varepsilon_{L-1}^{\!\top}) = \bigl(P_{L-1},\, (\mathrm d\mathcal L/\mathrm d b_{L-1})^{\!\top}\bigr)$. For the step, appending $\bar p_{t-1} = (A_t,\, \varepsilon_{t-1}^{\!\top})$ gives
\[
  \Bigl(P_t,\, \bigl(\tfrac{\mathrm d\mathcal L}{\mathrm d b_t}\bigr)^{\!\top}\Bigr)
  \;\star_{\mathrm{R}}\; \bigl(A_t,\, \varepsilon_{t-1}^{\!\top}\bigr)
  \;=\; \Bigl(P_t A_t,\; A_t^{\!\top}\bigl(\tfrac{\mathrm d\mathcal L}{\mathrm d b_t}\bigr)^{\!\top} + \varepsilon_{t-1}^{\!\top}\Bigr)
  \;=\; \Bigl(P_{t-1},\, \bigl(\tfrac{\mathrm d\mathcal L}{\mathrm d b_{t-1}}\bigr)^{\!\top}\Bigr),
\]
the second component by the transposed recursion and the first because $P_t A_t = P_{t-1}$. This is~\eqref{eq:reverse_scan_output}.
\end{proof}
\par\endgroup

\section{Retrieval and architecture classification details}
\label{app:theory_scope}
\par\begingroup

\subsection{Linear aggregation across subsystems}
\label{app:retrieval_aggregation}

\begin{corollary}
\label{cor:static_grouping}
Fix a set $J$ of subsystem indices $j=(\eta,p)$ from
Definition~\ref{def:mhssm}, a pair $(t,s)$ and all evaluated coefficients.
Suppose $u_s^j=V_s^j\tilde u_s$ for one shared
$\tilde u_s\in\R^{R'}$. Let $\hat B_s^j=B_s^jV_s^j$,
$\Phi^j=\Phi^j(t,s)\in\R^{N\times N}$ and
$C_t^j\in\R^{O\times N}$. Suppose the aggregation is linear in local
outputs, $\psi_t^\diamond(x_{t;\ell},(y^j)_j)=\sum_{j\in J}M_t^jy^j$,
with $M_t^j\in\R^{O_{\mathrm{out}}\times O}$ depending only on the
declared input window. Then its state-mediated operator factors as
\[
 \begin{gathered}
 W_{\mathrm{agg}}=\sum_{j\in J}M_t^jC_t^j\Phi^j\hat B_s^j
     =C_{\mathrm{agg}}\Phi^{\mathrm{stack}}\hat B_s^{\mathrm{stack}},\\
 \Phi^{\mathrm{stack}}=\bigoplus_{j\in J}\Phi^j,\qquad
 \hat B_s^{\mathrm{stack}}=(\hat B_s^j)_{j\in J}\in\R^{|J|N\times R'},\qquad
 C_{\mathrm{agg}}=[\,M_t^jC_t^j\,]_{j\in J}\in\R^{O_{\mathrm{out}}\times |J|N},
 \end{gathered}
\]
where stacking follows the index order of Definition~\ref{def:mhssm}.
Consequently, for arbitrary transfers,
\[
 \operatorname{rank}W_{\mathrm{agg}}
 \le\min\!\big(\operatorname{rank}C_{\mathrm{agg}},
                 \operatorname{rank}\hat B_s^{\mathrm{stack}}\big)
 \le\min(O_{\mathrm{out}},R').
\]
If all $\Phi^j$ are nonsingular, the maximum over unrestricted $C_{\mathrm{agg}}$
is $\min(O_{\mathrm{out}},\operatorname{rank}\hat B_s^{\mathrm{stack}})$.
\end{corollary}

\begin{proof}
Block multiplication gives the displayed factorization: each diagonal
transfer block pairs the corresponding write and read. The rank-product
inequality gives the bound, since the stacked write has $R'$ columns and
the aggregate read has $O_{\mathrm{out}}$ rows. No block must be invertible
for this step. If every block is nonsingular, so is
$\Phi^{\mathrm{stack}}$, and Theorem~\ref{thm:crosstalk} applies with state
width $|J|N$, effective read $C_{\mathrm{agg}}$ and output width
$O_{\mathrm{out}}$. The attainment is over unrestricted $C_{\mathrm{agg}}$; the constrained
read $[M_t^jC_t^j]_j$ need not realize a maximizing one.
\end{proof}

\subsection{Masked linear attention: proof and scope}
\label{app:theory_sec4_proofs}

\begin{proof}[Proof of Theorem~\ref{thm:mla_membership}]
For (i) apply the fixed-basis construction; for (ii) use the interior-order invariance of scalar masks.  Each case reads $W$
off the transfer~\eqref{eq:mla_transfer}.

(i) Under the fixed basis $Q$ of
Theorem~\ref{thm:simultaneous_diagonalization} the diagonal representative
$\Sigma^{Q}$ of~\eqref{eq:diagonal_equivalent_recurrence} has diagonal
transfer and, by Lemma~\ref{lem:similarity_invariance}, the same kernel $W$,
so
$\Phi(t,s)=Q\operatorname{diag}_n\!\big(\prod_{r=s+1}^{t}\lambda_n(x_r)\big)Q^{-1}$
and
$W[x](t,s)=\sum_n\big(\prod_{r=s+1}^{t}\lambda_n(x_r)\big)\,(C(x_t)Qe_n)(e_n^{\!\top}Q^{-1}B(x_s))$,
which is~\eqref{eq:mla_kernel} with the maps~\eqref{eq:mla_features},
token-local and shared across time because the original maps are and $Q$
is fixed (no invertibility of the transitions is used,
unlike the prefix frame of Lemma~\ref{lem:gauge_prefix});
unrolling~\eqref{eq:mla_operators} with $z_{-1}=Q^{-1}h_{-1}$ gives
\[
 z^{(n)}_t=\gamma^{(n)}(t,-1)(Q^{-1}h_{-1})_n
       +\sum_{s=0}^{t}\gamma^{(n)}(t,s)k^{(n)}(x_s)^{*}u_s.
\]
The initial-memory output is
$\operatorname{Re}_{\mathbb F}\sum_n
 q^{(n)}(x_t)\gamma^{(n)}(t,-1)(Q^{-1}h_{-1})_n
 =\operatorname{Re}_{\mathbb F}(C(x_t)\Phi(t,-1)h_{-1})$.
Together with the kernel action and unchanged feedthrough this reproduces
the complete layer output.
Conversely, given~\eqref{eq:mla_kernel} the recurrence~\eqref{eq:mla_recurrence}
has diagonal realized transitions
$\operatorname{diag}(\delta^{(1)}_t,\dots,\delta^{(S)}_t)$, so
Theorem~\ref{thm:simultaneous_diagonalization} with $Q=I$ places them in
$\mathcal D^{\mathbb F}(1,S)$ with time-shared maps, giving~\eqref{eq:mla_class_equality}.
Taking the diagonal realization's initial state to be the prescribed
$z_{-1}$ preserves its initialized output as well. The same calculation with step-indexed token-local maps and the same fixed $Q$ proves the
extension of part~(i) stated after the theorem.

(ii) In~\eqref{eq:mla_kernel} interior tokens enter only through the scalar
masks $\gamma^{(n)}(t,s)$, invariant under permutation of their factors, while the
query and key maps see only $x_t,x_s$; so every masked linear attention kernel and --- the constraint being
closed under limits --- every pointwise limit of them is
interior-transposition invariant.  A witnessed change therefore excludes $W$
from the class and its closure. If $W$ is the kernel of a realization
$\Sigma$, then $\Phi(t,s)$ must depend on the interior order (the endpoint
maps are fixed), which commuting transitions cannot do, so $\mathcal A_\Sigma$
is non-abelian.

\end{proof}

\begin{remark}
\label{rem:mla_positioning}
Theorem~\ref{thm:mla_membership}(ii) tests time-shared token-local
factors. Lemma~\ref{lem:gauge_prefix} gives every invertible transition
family a Gram form with prefix-dependent endpoints; the exclusion therefore
rests on time-shared token locality, not Gram factorization alone.
A nonlocal presentation of a kernel does not exclude it, since another
presentation may satisfy those hypotheses; exclusion needs a witness in the
declared domain. General softmax attention is not covered by~\eqref{eq:mla_kernel}:
its row normalization couples all summands over $s$ and need not admit
token-local query/key maps and scalar product masks. This does not exclude
special cases that admit such a representation.
\end{remark}

\par\begingroup
\phantomsection\label{app:mla_heads}
\paragraph{Head organization.}
Multiple heads are included in the same definition. Stack their supplied
values, and let $E_\eta$ and $F_\eta$ embed head $\eta$'s input and output
coordinates into the combined interfaces. Their concatenated kernel and
its modes are
\[
 W(t,s)=\sum_\eta F_\eta W_\eta(t,s)E_\eta^\top,\qquad
 q^{(\eta,n)}=F_\eta q_\eta^{(n)},\quad
 k^{(\eta,n)}=E_\eta k_\eta^{(n)},\quad
 \delta^{(\eta,n)}=\delta_\eta^{(n)}.
\]
This is a block-diagonal kernel before any output projection; a fixed
output projection is absorbed into the read directions. Removing the
block-support and parameter-sharing constraints gives the more general
combined kernel of Definition~\ref{def:linear_attention}.
This regrouping does not assert reduction to one conventional head
of~\eqref{eq:masked_attention_matrix}: that head applies one scalar
temporal score to every value coordinate, whereas distinct heads can
apply different scores. The combined mode representation retains those
differences through its endpoint supports and mode masks.

\par\endgroup

\subsection{Classification of existing architectures: details}
\label{app:classification_detail}

Shapes below follow Table~\ref{tab:specializations}: $N$ counts real
coordinates, with $N=2K$ for a fully paired $K$-complex representation.
Shapes use the exhibited linear interface reductions of Section~\ref{sec:def_general};
effective write rank and remaining aggregation are recorded separately.

We use the layer, mixer and SSM core vocabulary of
Section~\ref{sec:classification}; every window is measured on the core input.

\paragraph{Window and interface conventions.}
The table uses the smallest established window among the exhibited
presentations at fixed core input, state size and interfaces, without a
global minimality claim. All temporal preprocessing counts toward this
window. Two-endpoint forcing generally needs the preceding sample even
for constant $\lambda\ne1$; Euler or an identically zero previous write
removes that dependence. Evaluating an LTI recurrence over its full history
does not enlarge its window. Conversely, constant $A$ does not make
selective write/read maps or nonlinear preprocessing LTI.

Output gates belong to aggregation, and coefficient branches need not
factor through the supplied write features. If
$u_t^j=V_t^j\tilde u_t^j$ with $\tilde u_t^j\in\R^{R'}$, the effective write is
\[
 \hat B_t^j=B_t^jV_t^j,\qquad
 \operatorname{rank}\hat B_t^j\le\min(N,R,R').
\]
The Input-dep.\ column records selected discrete $A_t,B_t,C_t$, including
step and write-gain factors (endpoint coefficients for two-endpoint writes).
Driving features, feedthrough and aggregation are omitted from this column,
but their conditioning still counts toward the window. Moving factors
between these maps and $B,C$ can change the column without changing the
recurrence. Thus a dash does not imply a linear complete core: H3's
key--value product is a driving feature and its query contraction belongs
to aggregation, while GSS has gates around constant DSS coefficients.
DSS's horizon-dependent coefficient is input-independent at fixed horizon.

\subsubsection{Constant transitions}
\label{app:class_constant}

\paragraph{The S4 family.}
S4, S4D and DSS use one SISO LTI core per driving channel
\citep{gu2022efficiently,gu2022s4d,gupta2022diagonal}. S4 uses a
diagonal-plus-low-rank (DPLR) generator in its computational basis;
discretization gives a constant transition. The usual complex
parameterization retains conjugate symmetry to represent a real system.
The table does not identify this full DPLR system with a $K$-coordinate
complex-diagonal recurrence. Its class membership is conditional on the
discrete transition being diagonalizable.
At HiPPO-LegS initialization, the continuous generator $G$ is lower
triangular with distinct eigenvalues $-(n+1)$, and hence diagonalizable
\citep[Eq.~(2), Lemma~3.2]{gu2022efficiently}.
Learning the full DPLR generator preserves diagonalizability generically,
but does not guarantee it for every parameter setting.
For a positive step $\Delta$ with the inverse defined, bilinear
discretization gives $A=(I-\Delta G/2)^{-1}(I+\Delta G/2)$
\citep[Eq.~(3)]{gu2022efficiently}.
It preserves eigenvectors: $Gv=\lambda v$ implies
$Av=\frac{1+\Delta\lambda/2}{1-\Delta\lambda/2}v$.
S4D and DSS instead use diagonal complex generators and diagonal
transitions. DSS softmax fixes the kernel horizon $L$: its write
coefficient contains $(e^{L\lambda_i\Delta}-1)^{-1}$; DSS exponential
and the exponential variant used in GSS do not have this dependence.
S5~\citep{smith2023simplified} shares one such state across
all input channels, with matrix-valued $B,C$: its declared shape is
$(1,1;d,N,d)$, where $N=2K$ for $K$ complex coordinates.
Its channel mixing is in the write and read matrices, independently of
the diagonal transition.

\noindent Class: $\mathcal D^{\C}(1)$ for S4 (when diagonalizable), S4D, DSS and S5, because each is one constant, hence commuting, complex-diagonal transition per subsystem with token-local maps, so $\ell=1$.

\paragraph{H3 diagonal core and GSS.}
For H3~\citep{fu2023h3}, let $d_h$ be the key and value width per head.
Its shift core is described in Appendix~\ref{app:class_nondiagonal}.
After shifting, head $\eta$ forms
$\bar k_t^\eta(v_t^\eta)^{\!\top}\in\R^{d_h\times d_h}$.
The diagonal core filters each of these $d_h^2$ scalar streams, so its
declared shape is $(H,d_h^2;1,N,1)$; a query then
contracts the key axis through $\psi_t^\diamond$.
Here $k_{\mathrm{shift}}$ counts the taps of the finite shift filter, which equal the
shift core's state width.
The diagonal core's core-input window includes that filter and has bound
$k_{\mathrm{shift}}$, whereas the shift core itself has window one.
Only at $d_h=1$ do their subsystem
counts coincide. GSS~\citep{mehta2022gss} applies a DSS core to a
projected bottleneck of width $d_{\mathrm{ssm}}$. Surrounding products and gates in
both models do not change these cores' transition families.

\noindent Class: $\mathcal D^{\C}(k_{\mathrm{shift}})$ for H3's diagonal core and $\mathcal D^{\C}(1)$ for GSS, because both transitions are constant complex diagonal and the window is the shift filter's $k_{\mathrm{shift}}$ taps (GSS has none).

\paragraph{LRU.}
LRU~\citep{orvieto2023} uses one shared complex state,
$h_t=\Lambda h_{t-1}+\operatorname{diag}(\gamma)Bu_t$ and
$y_t=\operatorname{Re}(Ch_t)+Du_t$, with constant diagonal $\Lambda$.
Its shape is $(1,1;d,N,d)$, where $N=2K$ realifies the $K$ complex
coordinates; all maps are token-local.
Class: $\mathcal D^{\C}(1)$, because $\Lambda$ is constant complex diagonal and no preprocessing widens the window.

\paragraph{RetNet.}
RetNet~\citep{sun2023retnet} uses a headwise constant decay and per-coordinate frequencies,
$\alpha\operatorname{diag}_{k}(e^{i\omega_k})$ in complex coordinates with the cumulative rotation folded into its query and key maps and selective $B, C$, classified before the fold exactly as our own designs (Section~\ref{sec:hm_m1}).

Its retention normalization divides the state by $\sqrt{\kappa_t}$, where
$\kappa_t=\sum_{j\le t}\alpha^{t-j}=\alpha\kappa_{t-1}+1$ is the running sum of decay weights, so the
stored transition is $\alpha\sqrt{\kappa_{t-1}/\kappa_t}$~\citep[Section 2.2]{sun2023retnet}; GroupNorm
removes this scaling up to its $\varepsilon$, and the table lists the unscaled recurrence.
Class: $\mathcal D^{\C}(1)$, because the transitions are a constant complex diagonal, rescaled at
most by a positive scalar that depends on $t$ alone, and $B,C$ are token-local.

\subsubsection{Input-dependent real diagonal transitions}
\label{app:class_real}

\paragraph{Mamba 1, Mamba 2 and GLA.}
Mamba~1 derives coefficients from its processed value branch.
Mamba~2 projects values, $B,C$, steps and output gates separately.
The convolution and its sigmoid linear unit (SiLU) act on values and
$B,C$ only.
Mamba~1~\citep{gu2024mamba} makes $\Delta_t,B_t,C_t$ input-dependent while retaining one diagonal SISO core per channel. Its implementation uses the exponential-Euler write of~\eqref{eq:exp_euler}, whereas the paper states zero-order hold; Mamba~3 makes this distinction explicit~\citep[Section 3.1]{lahoti2026mamba3}. Its real transitions are the default; the complex variant~\citep[Section 3.6]{gu2024mamba}, suggested with S4D-Lin initialization for continuous modalities such as audio, instead belongs to $\mathcal D^{\C}(k_{\mathrm{conv}})$, each eigenvalue $e^{\Delta_ta_n}$ moving along a fixed spiral.
Mamba~2~\citep{dao2024transformers} restricts $A_t = \alpha_t I$, which exposes the state space duality of Section~\ref{sec:ssd}, equation~\eqref{eq:ssd_matrix}, and its chunkwise algorithm; the $P$ subsystems of a head share $\alpha_t$, which is what ``head'' denotes in Mamba~2.
GLA~\citep{yang2024gla} has a diagonal selective decay $A_t=\operatorname{diag}(\alpha_t)$ per head, per
coordinate like Mamba~1's, with key writes $B_t=k_t$, query reads $C_t=q_t^{\top}$ and values $u_t=v_t[p]$;
the reported model has no short convolution, so its window is one, and an optional
convolution enlarges it.
Class: $\mathcal D^{\R}(k_{\mathrm{conv}})$ for Mamba~1 and~2 (complex Mamba~1: $\mathcal D^{\C}(k_{\mathrm{conv}})$) and $\mathcal D^{\R}(1)$ for GLA, because the transitions are real diagonal, hence commuting, with the window set by the convolution width ($1$ for GLA without one).

\paragraph{RG-LRU and HGRN2.}
RG-LRU, the recurrent core of Griffin and Hawk~\citep{de2024griffin}, uses
$h_t=a_t\odot h_{t-1}+\sqrt{1-a_t^2}\odot(i_t\odot u_t)$.
Both gates depend on the convolved drive, not on $h_{t-1}$. The displayed scalar
factorization takes $A_t=a_t$, $B_t=\sqrt{1-a_t^2}\,i_t$ and $C_t=1$,
so its entry is $A,B$; its GeLU gate branch, like Mamba~1's, lies outside the core.
Taking each scalar state coordinate as a subsystem gives
$(d_{\mathrm{ssm}},1;1,1,1)$ and $\mathcal D^{\R}(k_{\mathrm{conv}})$,
with $k_{\mathrm{conv}}=4$ in the reported mixer. The gate projections'
block partition is not an additional recurrent-state axis.
HGRN2~\citep{qin2024hgrn2} has a real diagonal SSM core: per head, $Z_t=Z_{t-1}\operatorname{diag}(f_t)
+i_t(1-f_t)^{\!\top}$ and $y_t=Z_t o_t$. In our notation row $p$ of $Z_t$ is the state $h_t\in\R^N$ of
subsystem $p$, with $A_t=\operatorname{diag}(f_t)$, $B_t=1-f_t$, value $u_t=i_t[p]$ and read
$C_t=o_t^{\!\top}$.
Its $P$ rows are scalar-driven states of width $N$, giving
$(H,P;1,N,1)$; the reported square heads have $P=N$.
The layer-dependent lower bound in $f_t=\beta+(1-\beta)\odot g_t$
is learned and time-constant. All conditioning branches are token-local,
so the real diagonal selective core lies in $\mathcal D^{\R}(1)$;
an optional short convolution enlarges this bound to its width.
Class: $\mathcal D^{\R}(k_{\mathrm{conv}})$ for RG-LRU and $\mathcal D^{\R}(1)$ for HGRN2, because both are real diagonal (scalar for RG-LRU) with token-local gates, the window being the convolution width ($1$ for HGRN2 without one).

\paragraph{mLSTM, including its normalizer.}
The mLSTM of extended LSTM (xLSTM)~\citep[Eqs.~(19)--(26)]{beck2024xlstm}
has a matrix memory and a normalizer sharing the scalar gates:
\[
 Z_t=f_tZ_{t-1}+i_tv_tk_t^{\!\top},\qquad
 \mathsf n_t=f_t\mathsf n_{t-1}+i_tk_t.
\]
For each head, the $P=d_v$ memory rows and the normalizer are $P+1$
subsystems of state width $N=d_{qk}$: in our notation row $p$ of $Z_t$, and $\mathsf n_t$, is a state
$h_t\in\R^N$ with $A_t=f_tI_N$, $B_t=i_tk_t$, $C_t=q_t^{\!\top}$ and value $v_t[p]$, or $1$ for the
normalizer. Thus the table's shape is
$(H,P+1;1,N,1)$, with $H(P+1)N$ state coordinates.
The rows remain distinct subsystems because they are driven by different values.
The local reads are $Z_t[p,:]q_t$ and $\mathsf n_t^{\!\top}q_t$.
The stateless aggregation $\psi_t^\diamond$ divides the memory reads by
$\max(|\mathsf n_t^{\!\top}q_t|,1)$ and applies the output gate.
This is one affine recurrent system with a nonlinear aggregation;
the masked-linear-attention correspondence does not extend to that
aggregation. The input gate is $i_t=e^{\tilde i_t}$; the forget gate is
sigmoid or exponential.
The numerical path carries a per-head stabilizer
$m_t=\max(\log f_t+m_{t-1},\log i_t)$ and stores
$Z'_t=e^{-m_t}Z_t$, $\mathsf n'_t=e^{-m_t}\mathsf n_t$, an exact rescaling
in real arithmetic (the denominator floor becomes $e^{-m_t}$); the table
classifies the recurrence before stabilization.
In the released mixer, queries and keys pass through a width-$k_{\mathrm{conv}}$
convolution and SiLU, values bypass it, and both gate logits depend on
these query/key/value branches. Gates and SiLU act pointwise on input
features; only the convolution extends their core-input window to
$k_{\mathrm{conv}}$. No gate feeds back from the recurrent hidden state,
unlike sLSTM, so these nonlinear input maps preserve the affine core.
The xLSTM-7B variant~\citep[Sections~3.1--3.2]{beck2025xlstm7b} soft-caps the
gate preactivations, which changes only input maps, and removes the
convolution, giving window one.
Class: the mLSTM layer is in $\mathcal D^{\R}(k_{\mathrm{conv}})$, and the convolution-free mLSTM of xLSTM-7B, built from mLSTM blocks only, in $\mathcal D^{\R}(1)$, because the shared transition $f_tI_N$ is a real scalar, hence commuting and diagonal, the normalizer is one more subsystem, and the window is the query/key convolution width.

\subsubsection{Input-dependent rotations}
\label{app:class_rotations}

The models below realize their rotations in real arithmetic: in one fixed real
basis, every transition consists of real scalar rows and rotation--scaling pairs
$\rho\,\mathcal R(\theta)$. By Corollary~\ref{cor:abelian_closure}, these are exactly
the real transition families whose complexification is complex diagonal-equivalent,
which gives their $\mathcal D^{\C}$ entries in Table~\ref{tab:specializations}.

\paragraph{Mamba 3.}
Mamba~3~\citep{lahoti2026mamba3} has separate coefficient and value
branches, normalized $B,C$, token-dependent rotations and the two-endpoint
write of Section~\ref{sec:continuous}. Its MIMO variant lifts each scalar
value by a learned, data-independent $R_{\mathrm{M3}}$-vector. The internal
$(H,P;R_{\mathrm{M3}},N,R_{\mathrm{M3}})$ factorization therefore has input
rank one. Absorbing the lift and local gated contraction gives the table's
$(H,P;1,N,1)$ interface (Corollary~\ref{cor:m3_value_rank}). If branchwise
normalization across subsystems precedes contraction, retain
$(H,P;1,N,R_{\mathrm{M3}})$ and put the remaining operations in
$\psi_t^\diamond$. Token-local coefficient and feature branches together
with the previous-sample write give the unfolded core window two, hence
$\mathcal D^{\C}(2)$; earlier temporal preprocessing counts toward the
window relative to the declared core input.
Class: $\mathcal D^{\C}(2)$ (SISO and MIMO), because token-dependent rotation--scaling pairs share a fixed complex basis (commuting) and the two-endpoint write reaches back one token.

\paragraph{AUSSM and the declared core.}
AUSSM~\citep{karuvally2025aussm} assumes simultaneously diagonalizable
real skew-symmetric generators, with input-dependent generator and step
and constant continuous-time write/read matrices in its stated SSM.
After discretization, the write coefficient is $B_t=\Delta_t B$, while
$C_t=C$~\citep[Eq.~(4)]{karuvally2025aussm}; hence its entry is $A,B$.
The transition exponentials share a fixed complex eigenbasis. The table groups its channelwise cores
as one subsystem per driving channel; regrouping identically tied cores
into heads leaves diagonal equivalence unchanged. The released mixer's
optional width-$k_{\mathrm{conv}}$ convolution is included in its window bound.
As a mathematical extension, odd real widths can include an unpaired
scalar block, as allowed by Corollary~\ref{cor:abelian_closure}; the
released complex-coordinate implementation has even real width.
Class: $\mathcal D^{\C}(k_{\mathrm{conv}})$, because the unitary transitions share a fixed complex eigenbasis (commuting, complex diagonal) and the window is the optional convolution width.

\paragraph{Our designs.}
HybridMamba and SSD-MIMO are classified in Section~\ref{sec:hm_transition_laws} (Class and window):
$\mathcal D^{\R}(\ell)$ without complex rows and $\mathcal D^{\C}(\ell)$ with them, with $\ell=5$ by default.

\subsubsection{Inside $\mathcal S^{\mathbb F}(\ell)$, outside $\mathcal D^{\mathbb F}(\ell)$}
\label{app:class_nondiagonal}

\paragraph{H3 shift core.}
For H3~\citep{fu2023h3}, the shift core processes each of the $Hd_h$
key channels. Its nilpotent shift transition is defective for state width
$N>1$, so it is not diagonal-equivalent despite its
constant, hence commuting, transition. Its core-input window is one;
the diagonal core's post-shift window is given in Appendix~\ref{app:class_constant}.
Class: $\mathcal S^{\R}_{\mathrm{nd}}(1)$, because the constant real transition commutes but is defective, hence not diagonalizable even over $\C$, and the window is one.

\paragraph{Delta-rule transitions.}
The delta-rule family (DeltaNet~\citep{schlag2021fastweight,yang2024deltanet} and Gated DeltaNet~\citep{yang2025gateddelta}) is linear in the state, its transition $I - \beta_t k_t k_t^{\!\top}$ an input-dependent rank-one perturbation of the identity, and two such perturbations generically fail to commute,
\[
  \big[\,I - \beta_t k_t k_t^{\!\top},\; I - \beta_{t'} k_{t'} k_{t'}^{\!\top}\,\big]
  \;=\; \beta_t\beta_{t'}\,\big(k_t^{\!\top} k_{t'}\big)\,
        \big(k_t k_{t'}^{\!\top} - k_{t'} k_t^{\!\top}\big),
\]
which, for real keys, vanishes only for $k_t \perp k_{t'}$, $k_t \parallel k_{t'}$ or $\beta_t\beta_{t'} = 0$: the non-abelian claim requires an attained pair violating these conditions.
Class: $\mathcal S^{\R}_{\mathrm{nd}}(k_{\mathrm{conv}})$ for DeltaNet and Gated DeltaNet, because the real transitions are generically non-commuting (the commutator above) and their short convolutions on queries, keys and values give window $k_{\mathrm{conv}}$.

\paragraph{DeltaProduct and other noncommuting updates.}
DeltaProduct~\citep{siems2025deltaproduct} supplies $n_h$ keys, gates
and value vectors per token. Suppressing the head index and writing one
memory row as a column, let
\[
 A_{t,i}=I-\beta_{t,i}k_{t,i}k_{t,i}^{\!\top},\qquad
 h_{t,i}^p=A_{t,i}h_{t,i-1}^p+
          \beta_{t,i}k_{t,i}v_{t,i}[p],\qquad h_{t,0}^p=h_{t-1}^p.
\]
Collecting the $n_h$ updates gives
\[
 A_t=\rprod_{i=1}^{n_h}A_{t,i},\qquad
 (B_t)_{:,i}=\Big(\rprod_{q=i+1}^{n_h}A_{t,q}\Big)
             \beta_{t,i}k_{t,i},\qquad
 u_t^p=\operatorname{col}_{i=1}^{n_h}v_{t,i}[p],
\]
so $h_t^p=A_th_{t-1}^p+B_tu_t^p$ has write width $R=n_h$.
An $R=1$ description can instead absorb the independent values into
an input-dependent write column, but then $R$ no longer records the
supplied value interface. The table row is ungated DeltaProduct; Gated
DeltaProduct, $\widetilde A_t=\alpha_t A_t$, lies in the same generically
non-abelian $\mathcal S_{\mathrm{nd}}^{\R}(k_{\mathrm{conv}})$
class~\citep[Appendix B]{siems2025deltaproduct}.
Gated DeltaNet adds a scalar
decay to the one-update transition. RWKV-7~\citep{peng2025rwkv7} uses a
diagonal term plus an asymmetric rank-one correction (the sign may be
absorbed into a factor). For the column-state convention here its transition is
$\operatorname{diag}(w_t)-(a_t\odot\hat\kappa_t)\hat\kappa_t^{\!\top}$.
All its conditioning branches mix the current and preceding core-input
tokens; the first layer's value precursor, read by later layers, is part of
their declared core input. Group normalization, output gating and a
stateless bonus term on the current and preceding tokens act outside the
recurrent update.
Class: $\mathcal S^{\R}_{\mathrm{nd}}(k_{\mathrm{conv}})$ for (Gated) DeltaProduct and $\mathcal S^{\R}_{\mathrm{nd}}(2)$ for RWKV-7, because the real transitions are generically non-commuting (explicit pairs below) and the window is set by the short convolutions on queries, keys and values, resp.\ by RWKV-7's token shift over the current and preceding token.

For an explicit noncommuting pair, take $N\ge2$,
$k=e_1$ and $k'=(e_1+e_2)/\sqrt2$ with $\beta=\beta'=1/2$.
Then
\[
 [I-\tfrac12kk^{\!\top},I-\tfrac12k'k'^{\!\top}]
 =\tfrac18(e_1e_2^{\!\top}-e_2e_1^{\!\top})\ne0.
\]
Learned key projections can realize these two directions, and constant
sigmoid logits realize the gates. A short convolution can retain only
its current tap. Gated DeltaNet can use a constant scalar
decay law with $0<\alpha<1$; this only scales the commutator. For DeltaProduct, repeat each
key at all $n_h$ substeps with gate $1/2$: the product is
$I-(1-2^{-n_h})kk^{\!\top}$, giving the same obstruction.
For RWKV-7, take constant $w_t=c\mathbf1$ and $a_t=\tfrac12\mathbf1$;
its permitted decay range contains $c=e^{-e^{-1/2}/2}$, obtained with zero
decay logit. The same normalized removal keys give the displayed
commutator. By Lemma~\ref{lem:transition_boundary}(ii), one attained
noncommuting pair excludes these cores from $\mathcal D^{\C}(L)$, hence
from every $\mathcal D^{\mathbb F}(\ell)$, and the exclusion persists under
small perturbations of the weights; this is the sense of ``generic'' in
Table~\ref{tab:specializations}. An exposed change
(Section~\ref{sec:theory_mla}) further excludes a masked linear attention
kernel by Theorem~\ref{thm:mla_membership}(ii). These are realizable examples, not assertions about every
trained parameter setting. Scalar states, zero gates or restricted key
families can remain diagonal-equivalent.

\subsubsection{Outside the SSM class $\mathcal S^{\mathbb F}(\ell)$}
\label{app:class_outside}

The specified GRU, LSTM and sLSTM recurrences use nonlinear
hidden-state feedback in their gates. For example, sLSTM gate logits
include the preceding hidden state~\citep[Eqs.~(8)--(14)]{beck2024xlstm};
they are not coefficient maps of the core input alone as required by
Definition~\ref{def:field_ssm}. This contrasts with mLSTM's input-conditioned
gates and stateless nonlinear aggregation in Appendix~\ref{app:class_real}.

Full softmax attention stores an increasing list of keys and values.
That specified cache is not a recurrent state of fixed width $S$
independent of the horizon. Hyena~\citep{poli2023hyena} instead specifies
its long-convolution filter as a learned function of lag, rather than
through a fixed-width recurrent state. A convolution has a presentation as an SSM core
when its kernel admits $K_\tau=CA^\tau B$ with a state width
independent of the horizon; a finite-horizon shift register whose width
grows with the horizon does not give that presentation. By contrast, S4 and H3 generate their
kernels from a fixed-width state and therefore lie inside the class
(Appendix~\ref{app:class_constant}). Being outside the class is a statement about the architecture
as specified, not about which input--output maps it can compute.

\noindent Class: outside $\mathcal S$ for GRU, LSTM and sLSTM (nonlinear hidden-state feedback in the gates), softmax attention (cache growing with the horizon) and Hyena (kernel specified by lag, not a fixed-width state), because none is a fixed-width recurrence with core-input coefficient maps.
\par\endgroup

\section{Design details, state continuation and Mamba specializations}
\label{app:design_details}\label{app:implementation}

This appendix supplies the map constructions, continuation interface and
Mamba specializations behind Section~\ref{sec:hybridmamba}.
Numerical agreement is reported in Sections~\ref{app:numerics_exact}--\ref{app:numerics_redteam};
kernel algorithms and performance measurements are in
Appendix~\ref{app:kernel_details}. Experiment-specific settings are given
with the protocols in Appendix~\ref{app:exp_details}.

\par\begingroup
\subsection{Shared coefficients and endpoint maps}
\label{app:design_interface}
Here $\eta,p,n$ index head, channel within the head and state coordinate. The base
endpoint projections are shared across heads (in HybridMamba also across the $P$
subsystems); heads differ through static biases, steps and rates, and in SSD-MIMO the
biases also resolve the sibling.

For the per-coordinate scan, the log modulus must separate as
\begin{equation}
  \ell_t^{(\eta,p)}[n]
  = \sum_{m=1}^{2} c_m^{(\eta,p)}[n]\,s_{m,t}^{(\eta)}[n],
  \qquad \alpha_t^{(\eta,p)}[n]=\exp\!\big(\ell_t^{(\eta,p)}[n]\big),
  \label{eq:hm_separable}
\end{equation}
where a schedule may instead be head-resolved and hence omit $n$, but never
carries $p$: the $P$ subsystems of a head share the schedules, so the kernel evaluates
them once per head.

For static rows, the coefficient is $a_n^j=-\exp(A_{\log,n}^j)$.
Dynamic rows select the rate with static coefficient $-1$; potential rows
use the second schedule $\mu_t-\mu_{t-1}$ with the bounded gain
of~\eqref{eq:hm_bounded_mu}.

\paragraph{Write and read maps.}
\label{app:design_maps}

Let $\widehat B_t,\widehat C_t$ be the base maps, RMS-normalized
along the state axis when normalization is enabled, and $\phi_B,\phi_C$ the write and read gates
of~\eqref{eq:endpoint_gates}, $\phi_\bullet(s_t)=\phi(W^\bullet_gs_t)$ with gate activation $\phi$ (default
$\phi(v)=1+v$) and gate projections $W^B_g,W^C_g$. SSD-MIMO initializes these projections to zero;
HM uses uniform fan-in initialization when its optional gates are enabled,
with zero initialization available as an option. For HM the maps depend on the head only, $j=\eta$, and are
vectors; for SSD, $j=(\eta,r)$ and they have dimensions $N\times R$ and
$O\times N$. The default maps in both families are
\begin{equation}
\begin{aligned}
 \bar B_t^j&=\phi_B^j(s_t)\odot\widehat B_t+\beta^j,
 &\widetilde B_t^j&=\mathcal R(-\Theta_t^j)\bar B_t^j,\\
 \bar C_t^j&=\phi_C(s_t)\odot\widehat C_t+\gamma^j,
 &\widetilde C_t^j&=\bar C_t^j\mathcal R(\Theta_t^j+\nu_\eta).
\end{aligned}
\label{eq:ssd_write}
\end{equation}
The phase follows~\eqref{eq:hm_rotation_law}; the experiment protocols
specify the frequency sharing and initialization.
For SSD, $\phi_B^{\eta,r}(s_t)=\phi_B(s_t)[:,r]$ selects the sibling and scales every
input column; $\phi_C(s_t)$ resolves the output coordinate. Biases
$\beta^{\eta,r}\in\R^{N\times R}$ and
$\gamma^{\eta,r}\in\R^{O\times N}$ resolve the scan head; this per-scan-head bias
layout is required whenever a complex block is present. HM biases
$\beta^\eta,\gamma^\eta\in\R^{N_{\mathrm{hm}}}$ resolve the head,
where $N_{\mathrm{hm}}$ is the real width of the per-coordinate scan;
its gate features are head-shared. HM's optional post-rotation gates instead form
$\phi(W^B_gs_t+\beta_g)\odot(\widehat B_t+\beta^j)$, with the bias before the gate and an
optional gate bias $\beta_g$, and act after the rotation; they are pair-tied.

A gate projection emits one feature per gated complex pair and per gated
real row, for each sibling (write gate) or output coordinate (read gate) in
SSD-MIMO; each factor scales all columns of its sibling's map. Its factors
are expanded to the state axis as
\begin{equation}
 \phi_\bullet(s_t)|_{\{2k,2k+1\}}=\phi\big((W^\bullet_gs_t)[k]\big)I_2
 \quad\hbox{on a gated complex pair},\qquad
 \phi_\bullet(s_t)|_{\mathcal N_q}=1\quad\hbox{on an ungated block},
 \label{eq:hm_gates}
\end{equation}
for $\bullet\in\{B,C\}$. Here the first equality denotes the diagonal multiplication operator;
a pair-tied real factor is a complex scalar. Real blocks default to gated
and complex blocks to ungated. A gate projection exists only when enabled: in
SSD-MIMO by default for writes at $R>1$ and reads at $O>1$, in HybridMamba
never by default, and when enabled there, writes and reads are gated together. The experiment protocols state any
exceptions; complex-only SSD has no map gates by default even at $R,O>1$.

Pair-tied complex gates commute with rotation, so gating before or after the rotation
does not change the class or window of Section~\ref{sec:hm_transition_laws}. The default
gates before the rotation, as in~\eqref{eq:ssd_write}.

Static sibling weights compose with the output projection as
$W_{\mathrm{out}}(\zeta\otimes I_O)$; selected weights generally cannot.

\paragraph{Local and aggregate transport rank.}
\label{app:ssd_local_rank}
For the Euler write (\eqref{eq:ssdmimo_recurrence} with $\lambda\equiv1$), one sibling's
state-mediated temporal block can attain rank $\min(N,R,O)$ with nonsingular
transfer and free endpoint maps (Theorem~\ref{thm:crosstalk}); at $R=O=2$
and $N\ge2$, $B=[e_1\ e_2]$, $C=B^\top$, $\Phi=\alpha I$ with $\alpha\ne0$
give $C\Phi B=\alpha I_2$. The module attains this witness on a real block:
choose identity gates, zero biases, base maps proportional
to $[e_1\ e_2]$ and its transpose, and a nonzero scalar transfer.
RMS normalization and the step rescale the nonzero entries only, so the local rank remains two.
Mamba~3's implemented data-independent lift of a scalar
has effective input rank at most one at the corresponding local interface
(Corollary~\ref{cor:m3_value_rank}). These ranks concern supplied features at fixed
coefficient schedules; preprocessing may mix residual-token coordinates.

For one SSD-MIMO group, suppress $\eta,g$ and set
$B_s^{r,\mathrm{eff}}=\Delta_{r,s}\widetilde B_{r,s}\in\R^{N\times R}$ and stack
these effective writes over $r$ into $B_s^{\mathrm{eff,stack}}\in\R^{RN\times R}$. The Euler
state-mediated retrieval operator of the group is
\[
 W_{\mathrm{group}}(t,s)
 =\sum_{r=1}^{R}\zeta_{r,t}\widetilde C_{r,t}\Phi^r(t,s)B_s^{r,\mathrm{eff}}.
\]
This is the factorization of~\eqref{eq:aggregation_rank} with $M_t^r=\zeta_{r,t}I_O$, so
its rank is at most
$\min(O,\operatorname{rank}B_s^{\mathrm{eff,stack}})\le\min(O,R)$.
When $N\ge\min(R,O)$, one sibling can already attain this ceiling.
Different sibling transitions give different lag dependence, but
several scalar-input subsystems can also attain full aggregate rank, so the local-interface
distinction is not a whole-model expressivity ordering: comparisons require matched state
budgets and parameter sharing (Appendix~\ref{app:classification_detail}).

\subsection{State continuation and differentiability}
\label{app:state_validation}
\label{app:impl_mh_state}
Our native scans and mixers accept initial carries and return final ones: the recurrent state, the
accumulated phase, the previous write of the two-endpoint rule, the potential and the convolution
history, and the previous endpoint maps and drive where these are needed. Gradients flow through
both, so a long sequence can be processed as consecutive segments with the carries passed between
calls: keeping the carries in the graph backpropagates through all segments, detaching them gives
truncated backpropagation through time. Single-token steps use the same carries for generation.
With row blocks, each block carries its own state and the skip is applied once.

Continuation tests cover the native routes advertised as supporting full backpropagation through time and check that whole-sequence, token-by-token and split-sequence
evaluation agree in outputs, final carries and all gradients, including those of the initial
state, within arithmetic-derived tolerances. Splitting bounds the working memory of each call,
but backpropagating through all segments keeps each segment's saved tensors
(Section~\ref{sec:hybridmamba_complexity}), so total training memory still grows with the
backpropagated length; only detaching the carries bounds it. Without gradients nothing accumulates
across calls, so inference of arbitrary length runs in segments that fit in memory.

\subsection{Exact Mamba specializations and numerical evidence}
\label{app:design_writes}\label{app:mamba_mapping}
Table~\ref{tab:mamba_recovery} gives the defining settings. With matching
projections, normalization and initial states, the parameter correspondences
in Table~\ref{tab:m3_map} recover the same function in exact arithmetic. Copied-weight output
and gradient checks use declared floating-point tolerances; upstream
step/decode parity has not been tested.

\begin{table}[!ht]
\centering\footnotesize
\caption{Parameter correspondences for the SISO presets in
Table~\ref{tab:mamba_recovery}. Mamba~3 MIMO has no exact preset
(Corollary~\ref{cor:m3_value_rank}).}
\label{tab:m3_map}
\begin{tabular}{@{}lp{.79\linewidth}@{}}
\toprule
Target & Parameter correspondence \\
\midrule
Mamba 1 & $P=1$, step rank $\lceil d_{\mathrm{model}}/16\rceil$;
no B/C or output norm. Match input/output, convolution, B/C/step maps,
$A_{\log}$ and head $D$. \\
Mamba 2 & $R=O=1$, one B/C group, no B/C norm or biases.
Split packed projections and convolution; match step bias,
$A_{\log}$, head $D$ and output normalization; no static rate bias. \\
Mamba 3 SISO & $R=O=1$, no output norm; half of the state rows rotate
(in pairs) with $\omega=\pi$, $\nu=0$. $V\mapsto u$, $K,Q\mapsto B,C$; match the rate activation (heavy-tail by default),
$\log\alpha_t=\Delta_t a_t$ and the step $\Delta_t$ that scales the write,
B/C gains and biases, step bias, head $D$, output gate and angle/endpoint maps. \\
\bottomrule
\end{tabular}
\end{table}
Copied-weight checks for Mamba~3 match the softplus rate law on both sides;
they do not test our default heavy-tail configurations.

The matched M3exact and real-only R1O1-r (Mamba~3 settings) training runs are experimental
controls, not copied-weight parity tests; their downstream comparisons
are in Appendix~\ref{app:lm_final_breakdowns}.

\par\endgroup

\subsection{Numerical agreement with the official modules}
\label{app:numerics_exact}
\par\begingroup
\paragraph{Setup.}
We compare with the official Mamba modules, matching the softplus rate law
for Mamba~3, using PyTorch~2.14 and Triton~3.8. Here fp32 denotes 32-bit floating point and tf32
denotes TensorFloat-32 matrix-product operands. Official weights are copied into our preset
(Table~\ref{tab:m3_map}); both modules receive the same input and the same output cotangent. The reference is our
module in 64-bit floating point (fp64) through a pure-torch route (token-by-token recurrence, or the chunked torch route at the long
shapes; the two agree to $2.1\cdot10^{-15}$ at a small shape), which shares no code with either kernel. Errors are
normwise relative errors $\|a-b\|/\|b\|$ per tensor, with the pooled signed error $\langle a-b,b\rangle/\langle b,b\rangle$
as bias check. A tensor passes when the error is below $\min(\max(u\sqrt{n}\sqrt2,1.28u),12.8u)$ (bf16, $u=2^{-8}$) or
$\max(4u\sqrt n,10^{-6})$ (fp32, tf32 operands, $u=2^{-11}$) and the pooled signed error below a quarter of that bound;
$n=\mathsf B L$ for parameters and $n=64$ for $y$ and $\partial u$. This is the rule of the repository's kernel oracles;
an arm run with IEEE operands is held to the same bound. bf16 runs use autocast on both sides with bf16-exact inputs.
The comparison is end to end through the module (projections, scan, gate, output map), so bf16 rounding of the
projections enters both arms. Each cell reports the median (maximum) over five draws.
The full matrix runs on a Blackwell RTX PRO 5000; an A100-SXM4-80GB
cross-check repeats the main shape, long-shape Mamba~3 and selected stress cases.
The measured configuration adds the step bias in fp32 and saves HM boundary
checkpoints. These results do not certify untested execution configurations
or H100 numerical accuracy.

\paragraph{What each side computes in fp32.}
Official Mamba~3 casts $Q$, $K$, $V$, the trapezoid weight and the angles to bf16 before its scan for any input dtype
and keeps the step and the decay exponent in fp32 (\texttt{mamba3\_siso\_combined.py}); its backward multiplies in bf16
in the input-gradient kernel. Our fp32 route uses tf32 matrix-product operands by default and fp32 operands
(\texttt{precision="ieee"}) on request. {On the tested NVIDIA routes, Triton's ``IEEE'' mode uses
single-instruction, multiple-thread (SIMT) fp32 fused multiply-adds on CUDA
cores instead of tf32 Tensor Core products. Both modes support fp32
input/output and accumulation; multiplication precision and execution units
differ. ``IEEE'' means neither fp64 nor module-wide IEEE compliance.}
Official Mamba~2 uses Triton dots on fp32 operands (tf32 by default), and
official Mamba~1 is an fp32 scan. Hence fp32 rows compare each side in the arithmetic it performs, and only the bf16 rows
use the same operand precision on both sides.

\paragraph{Agreement in detail.}
Table~\ref{tab:numerics_exact} reports copied-weight agreement and each side's
error against the fp64 recurrence, which shares no code with either kernel.
At $\mathsf B=6$, $L=2048$, every tested tensor--seed comparison lies within
twice its normwise tolerance in both dtypes: $60/60$ for Mamba~1, $75/75$ for
Mamba~2 and $80/80$ for Mamba~3. The A100 cross-check also passes every comparison at this shape.
Median bf16 output differences between ours and official are
$3.5\cdot10^{-3}$, $1.4\cdot10^{-3}$ and $4.3\cdot10^{-3}$, respectively.
At $\mathsf B=2$, $L=16384$, the exception is Mamba~3's bf16 angle-projection
gradient: all five seeds exceed twice the tolerance, leaving $75/80$ passes.

For bf16 Mamba~3, ours and official have output errors against fp64 of
$5.0\cdot10^{-3}$ and $4.8\cdot10^{-3}$, and input-gradient errors of
$5.5\cdot10^{-3}$ on both sides. In fp32, official Mamba~1 agrees with ours
to about $3\cdot10^{-7}$ on the output. Mamba~2's ours--official difference
is about $6\cdot10^{-5}$, while our full-fp32 product mode is about
$10^{-6}$ from fp64. Official Mamba~3's internal bf16 casts give an output
error about $3.5\cdot10^{-3}$; ours is about $2.6\cdot10^{-4}$ with tf32
products and $10^{-6}$ with full-fp32 products. These are comparisons of the
arithmetic each implementation performs, not bitwise-equivalence claims.

\begin{table}[!htbp]
\centering\scriptsize
\caption{{Copied-weight agreement with official modules and relative error against fp64
($\mathsf B=6$, $L=2048$, $d_{\mathrm{inner}}=2048$; Blackwell).
Cells give median (maximum) over five seeds; ``param.\ grads'' takes the worst
tensor per seed. ``Ours'' uses default operands and adds the step bias in
fp32; IEEE uses fp32 matrix-product operands (SSD-MIMO, fp32 only).
bf16 runs use autocast on both sides. For fp32 input/output, official Mamba~1 uses
an fp32 scan, Mamba~2 tf32 dots, and Mamba~3 casts $Q,K,V$, trapezoid weight
and angles to bf16 inside its scan; our SSD recoveries default to tf32 operands.}}
\label{tab:numerics_exact}
\begin{adjustbox}{max width=\linewidth}
\begin{tabular}{@{}l l l c c c c@{}}
\toprule
Pair & prec. & quantity & ours vs.\ off. & off.\ vs.\ fp64 & ours vs.\ fp64 & ours (IEEE) vs.\ fp64 \\
\midrule
Mamba~1 (MH) & bf16 & $y$ & $3.5\,(3.6){\times}10^{-3}$ & $4.2\,(4.3){\times}10^{-3}$ & $4.5\,(4.6){\times}10^{-3}$ & -- \\
 &  & param.\ grads & $5.9\,(7.2){\times}10^{-3}$ & $8.0{\times}10^{-3}\,(1.1{\times}10^{-2})$ & $8.4{\times}10^{-3}\,(1.1{\times}10^{-2})$ & -- \\
 & fp32 & $y$ & $2.7\,(2.7){\times}10^{-7}$ & $8.9\,(9.0){\times}10^{-7}$ & $8.9\,(8.9){\times}10^{-7}$ & -- \\
 &  & param.\ grads & $2.2\,(3.3){\times}10^{-6}$ & $8.7{\times}10^{-6}\,(1.1{\times}10^{-5})$ & $8.6{\times}10^{-6}\,(1.1{\times}10^{-5})$ & -- \\
\addlinespace
Mamba~2 & bf16 & $y$ & $1.4\,(1.4){\times}10^{-3}$ & $4.5\,(4.5){\times}10^{-3}$ & $4.5\,(4.5){\times}10^{-3}$ & -- \\
 &  & param.\ grads & $5.9{\times}10^{-3}\,(1.1{\times}10^{-2})$ & $1.1\,(1.4){\times}10^{-2}$ & $1.1\,(1.5){\times}10^{-2}$ & -- \\
 & fp32 & $y$ & $6.5\,(8.2){\times}10^{-5}$ & $6.2\,(7.9){\times}10^{-5}$ & $1.7\,(1.8){\times}10^{-5}$ & $8.9\,(8.9){\times}10^{-7}$ \\
 &  & param.\ grads & $1.7\,(2.0){\times}10^{-3}$ & $1.5\,(1.7){\times}10^{-3}$ & $5.8\,(7.2){\times}10^{-4}$ & $1.6\,(2.4){\times}10^{-6}$ \\
\addlinespace
Mamba~3 SISO & bf16 & $y$ & $4.3\,(4.3){\times}10^{-3}$ & $4.8\,(4.8){\times}10^{-3}$ & $5.0\,(5.0){\times}10^{-3}$ & -- \\
 &  & param.\ grads & $4.7\,(5.1){\times}10^{-2}$ & $4.3\,(4.8){\times}10^{-2}$ & $2.0\,(2.1){\times}10^{-2}$ & -- \\
 & fp32 & $y$ & $3.6\,(3.6){\times}10^{-3}$ & $3.5\,(3.5){\times}10^{-3}$ & $2.6\,(2.7){\times}10^{-4}$ & $1.0\,(1.1){\times}10^{-6}$ \\
 &  & param.\ grads & $4.4\,(4.9){\times}10^{-2}$ & $4.4\,(4.9){\times}10^{-2}$ & $2.5\,(2.5){\times}10^{-3}$ & $2.7\,(2.7){\times}10^{-6}$ \\

\bottomrule
\end{tabular}
\end{adjustbox}
\end{table}

\subsection{Gradient accuracy of the rotation path against official Mamba~3}
\label{app:numerics_redteam}
\paragraph{What is tested.}
Official Mamba~3 and our Mamba~3 specialization (Table~\ref{tab:m3_map}) receive copied weights,
the same input and the same output cotangent, and both are scored against the fp64 evaluation of
Appendix~\ref{app:numerics_exact}. We report the gradients with respect to the angle projection
$W_q$ (with $q_t=W_qs_t$), written $\partial\theta$; the step projection, $\partial W_\Delta$; and the
input, $\partial u$. The previous-token case additionally reports the endpoint-weight projection gradient,
$\partial W_\lambda$.

\paragraph{Why the angle gradient is the sensitive one.}
For one scan head $(\eta,r)$, with its indices suppressed, the phase of~\eqref{eq:hm_rotation_law} is
a running sum, so with $G_{k,t}=\sum_{s\ge t}\partial L/\partial\Theta_{k,s}$ the reverse running sum
of phase gradients,
\begin{equation}\label{eq:dtheta_reverse}
 \frac{\partial L}{\partial q_{k,t}}=-\,\omega_k\operatorname{sech}^2(q_{k,t})\,\Delta_t\,G_{k,t},
 \qquad
 \frac{\partial L}{\partial\Delta_t}\ni-\sum_k\omega_k\tanh(q_{k,t})\,G_{k,t}.
\end{equation}
Three properties make $G_{k,t}$ numerically delicate. (i) It sums over all later tokens, so the
rounding of its terms accumulates with the context length. (ii) After a forgetting token
($\Delta_t|a_t|\gg1$), the exact contribution of earlier phases to later outputs has decayed, so the
exact $G_{k,t}$ is a small remainder of terms that each carry the operand rounding of the backward
matrix products, while the factor $\Delta_t$ in~\eqref{eq:dtheta_reverse} is large there; the
relative error of $\partial\theta$ is then large although every term is accurate to the operand
precision. (iii) Official Mamba~3 forms the in-chunk reverse sum as chunk total minus forward
prefix plus the current term, a difference of large quantities, from bf16 matrix products; ours
accumulates the reverse sum directly, with fp32 phases reduced modulo $2\pi$. This mechanism is
consistent with every measurement below; we did not isolate it by ablation.

\paragraph{Cases.}
Table~\ref{tab:numerics_redteam} includes ordinary inputs at increasing length,
strong rotation at $L=8192$, fast-forgetting heads, sparse emphasis spikes,
document boundaries and the previous-token limit of the two-endpoint write.
The stress matrix uses nine scenarios; its largest boundary strength is
$\Delta|a|\approx100$. All ratios are official over ours against fp64.
The full matrix is measured on Blackwell; the A100 cross-check is narrower.
Across the nine scenarios the median angle-gradient ratios range from
$1.4$ to $7.4$ in bf16 and from $14$ to $49$ in fp32, where official still
uses bf16 scan operands. At ordinary initialization, increasing $L$ from
$2048$ to $32768$ raises the bf16 angle-gradient error from
$4.4\cdot10^{-2}$ to $2.6\cdot10^{-1}$ for official and from
$1.7\cdot10^{-2}$ to $1.1\cdot10^{-1}$ for ours.

\begin{table}[!htbp]
\centering\scriptsize
\caption{{Gradient errors $\|a-b\|/\|b\|$ against fp64 for official Mamba~3 and ours,
with copied weights and identical input/cotangent: median (worst) over five
Blackwell draws. Ordinary inputs use $\mathsf B=1$, $d_{\mathrm{model}}=1024$
unless stated; stress patterns use $\mathsf B=2$, $L=2048$,
$d_{\mathrm{model}}=256$; strongly rotating context uses $\mathsf B=1$, $L=8192$.
$\partial W_\lambda$, $\partial b_\Delta$ and $\partial W_a$ denote endpoint-weight
projection, step-bias and rate-projection gradients.
Bold marks any draw exceeding the tolerance of Appendix~\ref{app:numerics_exact}
or a pooled signed error exceeding one quarter of it.
``Ours'' uses default operands (tf32 in fp32); IEEE uses fp32 operands.
Ratio is official/ours ($<1$: ours is less accurate). fp32 rows retain each
implementation's arithmetic, including official bf16 scan casts.}}
\label{tab:numerics_redteam}
\begin{adjustbox}{max width=\linewidth}
\begin{tabular}{@{}p{.27\linewidth} l l c c c r r@{}}
\toprule
Case & prec. & tensor & official & ours & ours (IEEE) & ratio & bound \\
\midrule
Ordinary inputs, $L{=}2048$ ($\mathsf B{=}1$) & bf16 & $\partial\theta$ & $4.4\,(4.7){\times}10^{-2}$ & $1.7\,(1.8){\times}10^{-2}$ & -- & $2.6$ & $0.05$ \\
 & fp32 & $\partial\theta$ & $4.4\,(4.9){\times}10^{-2}$ & $2.4\,(2.8){\times}10^{-3}$ & $2.4\,(2.8){\times}10^{-6}$ & $18.2$ & $0.088$ \\
Ordinary inputs, $L{=}8192$ ($\mathsf B{=}1$) & bf16 & $\partial\theta$ & $\mathbf{1.0\,(1.1){\times}10^{-1}}$ & $4.3\,(4.8){\times}10^{-2}$ & -- & $2.3$ & $0.05$ \\
 & fp32 & $\partial\theta$ & $9.5{\times}10^{-2}\,(1.1{\times}10^{-1})$ & $5.5\,(6.5){\times}10^{-3}$ & $5.4\,(5.6){\times}10^{-6}$ & $17.4$ & $0.18$ \\
Ordinary inputs, $L{=}32768$ ($\mathsf B{=}1$) & bf16 & $\partial\theta$ & $\mathbf{2.6\,(2.8){\times}10^{-1}}$ & $\mathbf{1.1\,(1.2){\times}10^{-1}}$ & -- & $2.2$ & $0.05$ \\
 & fp32 & $\partial\theta$ & $2.5\,(2.8){\times}10^{-1}$ & $1.5\,(1.6){\times}10^{-2}$ & $1.4\,(1.5){\times}10^{-5}$ & $16.8$ & $0.35$ \\
Ordinary inputs, $L{=}2048$, $d_{\mathrm{model}}{=}256$ & bf16 & $\partial\theta$ & $\mathbf{4.7\,(6.2){\times}10^{-2}}$ & $1.8\,(1.8){\times}10^{-2}$ & -- & $2.7$ & $0.05$ \\
Long context (phase $\lesssim 250$ rad, $L{=}8192$) & bf16 & $\partial\theta$ & $\mathbf{8.7\,(9.2){\times}10^{-2}}$ & $\mathbf{4.6\,(5.4){\times}10^{-2}}$ & -- & $1.9$ & $0.05$ \\
 & bf16 & $\partial W_\Delta$ & $\mathbf{7.3{\times}10^{-2}\,(1.1{\times}10^{-1})}$ & $\mathbf{3.7\,(5.1){\times}10^{-2}}$ & -- & $2.0$ & $0.05$ \\
 & fp32 & $\partial\theta$ & $7.3\,(9.1){\times}10^{-2}$ & $4.4\,(4.7){\times}10^{-3}$ & $6.0\,(6.2){\times}10^{-4}$ & $16.6$ & $0.18$ \\
Fast-forgetting heads (target $\Delta|a|\le 50$) & bf16 & $\partial\theta$ & $\mathbf{1.3\,(1.5){\times}10^{-1}}$ & $1.7\,(1.8){\times}10^{-2}$ & -- & $7.4$ & $0.05$ \\
 & fp32 & $\partial\theta$ & $\mathbf{1.4\,(1.6){\times}10^{-1}}$ & $2.8\,(3.1){\times}10^{-3}$ & $5.3\,(6.3){\times}10^{-6}$ & $49.2$ & $0.12$ \\
Emphasis spikes (10\% tokens, $\Delta|A|$ to $\approx 20$) & bf16 & $\partial\theta$ & $\mathbf{4.0\,(5.8){\times}10^{0}}$ & $\mathbf{1.5\,(1.6){\times}10^{0}}$ & -- & $2.7$ & $0.05$ \\
 & bf16 & $\partial u$ & $\mathbf{3.8\,(4.6){\times}10^{-2}}$ & $1.4\,(1.6){\times}10^{-2}$ & -- & $2.8$ & $0.044$ \\
 & fp32 & $\partial\theta$ & $\mathbf{4.7\,(5.7){\times}10^{0}}$ & $\mathbf{2.1\,(3.0){\times}10^{-1}}$ & $2.1\,(3.0){\times}10^{-4}$ & $22.5$ & $0.12$ \\
 & fp32 & $\partial u$ & $\mathbf{4.1\,(5.2){\times}10^{-2}}$ & $2.1\,(2.2){\times}10^{-3}$ & $2.1\,(2.6){\times}10^{-6}$ & $20.1$ & $0.016$ \\
Document boundary ($\Delta|A|\approx 100$) & fp32 & $\partial\theta$ & $\mathbf{3.2\,(3.6){\times}10^{2}}$ & $\mathbf{1.9\,(3.4){\times}10^{1}}$ & $2.3\,(3.1){\times}10^{-2}$ & $16.3$ & $0.12$ \\
 & fp32 & $\partial u$ & $\mathbf{2.6\,(3.6){\times}10^{-2}}$ & $1.6\,(2.6){\times}10^{-3}$ & $3.2\,(3.6){\times}10^{-6}$ & $16.0$ & $0.016$ \\
 & bf16 & $\partial\theta$ & $\mathbf{2.4\,(4.4){\times}10^{2}}$ & $\mathbf{1.7\,(2.1){\times}10^{2}}$ & -- & $1.4$ & $0.05$ \\
Previous-token write ($\lambda\approx 10^{-5}$) & fp32 & $\partial\theta$ & $\mathbf{6.9{\times}10^{-2}\,(4.8{\times}10^{-1})}$ & $4.0{\times}10^{-3}\,(2.5{\times}10^{-2})$ & $5.2{\times}10^{-6}\,(2.0{\times}10^{-5})$ & $17.1$ & $0.12$ \\
 & bf16 & $\partial\theta$ & $\mathbf{5.7{\times}10^{-2}\,(4.0{\times}10^{-1})}$ & $\mathbf{4.0{\times}10^{-2}\,(1.7{\times}10^{-1})}$ & -- & $1.4$ & $0.05$ \\

 & bf16 & $\partial W_\lambda$ & $\mathbf{2.2\,(6.5){\times}10^{-2}}$ & $\mathbf{2.0\,(6.6){\times}10^{-2}}$ & -- & $1.1$ & $0.05$ \\
\bottomrule
\end{tabular}
\end{adjustbox}
\end{table}

\paragraph{Why it could matter, and what we do not claim.}
Errors in $\partial\theta$ perturb the updates of the rotation parameters, most at long context, and
an fp32 run is only as exact as its scan. In our MQMKVAR and language-model runs, neither our models
nor the official Mamba~3 baselines showed a training instability attributable to these effects; the
measurements bound gradient accuracy on constructed or long inputs.

\paragraph{Remaining errors.}
With bf16, both implementations fail the angle-gradient tolerance at strong
boundaries, emphasis spikes and the previous-token write (by the worst draw).
In the previous-token case, both also fail the endpoint-weight projection-gradient tolerance.
In the strongly rotating $L=8192$ case, our angle and step-projection gradients
also fail on Blackwell: worst errors are $5.4\cdot10^{-2}$ and
$5.1\cdot10^{-2}$ against a $5\cdot10^{-2}$ bound. Neither fails in the
same-configuration A100 cross-check; official fails both on both devices.
On ordinary inputs at $L=32768$, ours also exceeds the bf16 angle-gradient
bound; official exceeds it already at $L=8192$.

With fp32 input/output and tf32 products, our angle gradient fails at strong
forgetting tokens. Full-fp32 products pass all nine stress scenarios. A separate
A100 probe with stronger boundaries ($\Delta|a|\approx300$), outside this matrix,
still gives an angle-gradient error $0.22$ against a $0.125$ bound even with
full-fp32 products. Strong forgetting suppresses the exact phase cotangent,
while the implementation forms a cancelling reverse sum of rounded terms;
the large step can amplify its error. This is a gradient-accuracy limitation,
not evidence of a training instability. Other gradients can also be less
accurate than official without exceeding tolerance; for example, the ordinary
bf16 step-bias error is $8.0\cdot10^{-3}$ versus $5.4\cdot10^{-3}$.

\paragraph{fp32 caveat.}
Official Mamba~3's fp32 rows measure bf16-operand arithmetic (Appendix~\ref{app:numerics_exact}); the
same-arithmetic comparison is the bf16 rows. The
cases use matched softplus rate laws, do not cover our default heavy-tail
configurations, and cover Blackwell with the stated A100 cross-checks.
\par\endgroup

\needspace{10\baselineskip}
\section{Chunk algorithms and computational cost}
\label{app:kernel_details}

\par\begingroup
\subsection{Routing, contractions and chunk factorization}
\label{app:gemm_details}

Let $z_t\in\R^d$ be the input and
$x_t^{(\eta,p)}=V_t^{(\eta,p)}z_t\in\R^R$ the routed feature of
subsystem $p$ in head $\eta$. Its write and read maps have shapes
$B_t^{(\eta,p)}\in\R^{N\times R}$ and
$C_t^{(\eta,p)}\in\R^{O\times N}$. The aggregation considered here sums
subsystem readouts. For a group $\mathcal I$ of heads sharing $B_t^{(p)}$
and $C_t^{(p)}$, stack the routed features as columns:
\begin{equation}
 X_{t,\mathcal I}^{(p)}=[x_t^{(\eta,p)}]_{\eta\in\mathcal I},\qquad
 U_{t,\mathcal I}^{(p)}=B_t^{(p)}X_{t,\mathcal I}^{(p)}.
\end{equation}
The write is an $[N,R][R,|\mathcal I|]$ GEMM even when the routed inputs
differ. Stacking the states vertically and the read maps horizontally gives
\begin{equation}
 Y_{t,\mathcal I}=
 [C_t^{(1)}\ \cdots\ C_t^{(P)}]
 \begin{bmatrix}h_{t,\mathcal I}^{(1)}\\ \vdots\\h_{t,\mathcal I}^{(P)}\end{bmatrix}.
\end{equation}
Identical subsystem inputs additionally permit stacking the write maps
into one GEMM; independent inputs require separate or batched products.
Selected aggregation weights can be absorbed into $C_t$ only if the
required sharing survives; fixed weights can compose with the output projection.

\paragraph{Chunk scores and routing.}
For scalar transitions shared across the heads in $\mathcal I$, let
$\Gamma_{t,s}^{(p)}=\prod_{u=s+1}^t\alpha_u^{(p)}$ for $s\le t$, zero otherwise.
Within a chunk of $T$ tokens, collect each read row and write column into
$C_{p,o},B_{p,r}\in\R^{T\times N}$. For zero incoming carry,
\begin{equation}
\begin{aligned}
 \mathrm{Score}^{(p,o,r)}&=\Gamma^{(p)}\odot(C_{p,o}B_{p,r}^{\top}),\\
 Y^{(o)}&=\sum_{p=1}^P\sum_{r=1}^R
             \mathrm{Score}^{(p,o,r)}X^{(p,r)},
 &X^{(p,r)}&\in\R^{T\times|\mathcal I|}.
\end{aligned}
\label{eq:block_ssd_small_apply}
\end{equation}
$X^{(p,r)}$ stacks routed coordinate $r$ across tokens and heads in $\mathcal I$.
The full output adds
$\sum_p C_t^{(p)}(\prod_{u=a}^t\alpha_u^{(p)})h_{a-1}^{(\eta,p)}$
for chunk start $a$. Each score construction is $[T,N][N,T]$;
each application is $[T,T][T,|\mathcal I|]$. Equivalently, these scores
form a block matrix of shape $TO\times TPR$; it need not be materialized.
If $X^{(p,r)}=X^{(r)}$ for all $p$, sum the scores over $p$ before application:
\begin{equation}
 Y^{(o)}=\sum_{r=1}^R
       \left(\sum_{p=1}^P\mathrm{Score}^{(p,o,r)}\right)X^{(r)}.
\end{equation}
This retains $PRO$ score products but reduces applications from $PRO$
to $RO$. A permutation of a shared local input can be absorbed into its
write map and does not enlarge this family.

The required sharing is of endpoint maps and history masks across GEMM
columns, not across tokens. In SSD-MIMO these columns are the $G$ groups
within one scan head $(\eta,r)$; biases, gates and phases generally
prevent reuse across scan heads.

\paragraph{Contraction orders and numerical decay evaluation.}
\label{app:design_execution}\label{sec:ssd_chunkwise}
Equation~\eqref{eq:block_ssd_small_apply} contracts endpoint maps first,
reusing scores across streams with shared maps. Alternatively, writing
$v_s=B_su_s$ first gives, in the matrix notation of Proposition~\ref{prop:ssd},
\begin{equation}\label{eq:ssd_matrix_v}
 Y^o-Y_{\mathrm{init}}^o=
    \bigl(\Gamma\odot(\mathsf C^o V^\top)\bigr)\mathbf1_L,
\end{equation}
where $V$ stacks $v_s^\top$. This uses one temporal decay application
per stream, but different routed inputs generally give different $V$.
SSD-MIMO uses the endpoint-first masked chunk contraction. Padding uses
identity transitions and zero writes; padded outputs are discarded.

Scalar decay can be absorbed into endpoint maps by the accumulator
change of frame in Corollary~\ref{cor:gauge_schedule}, but its exponential
intermediates can be large. The masked contraction instead evaluates
$a(D_i-D_j)$ difference-first for a per-head scalar $a$ and cumulative
nonnegative steps $D_i$ formed within each chunk (for selected rates the step carries the rate magnitude), avoiding subtraction of separately rounded $aD_i,aD_j$.
Per-coordinate decay admits a change of frame too, but loses the common
mask used here.

\paragraph{Cross-chunk factorization.}
\label{app:ssd_block_factors}
Write $c=T$ for the SSD chunk length (Section~\ref{sec:hybridmamba_complexity}) and partition the sequence into $m=\lceil L/c\rceil$ chunks
$I_k=\{kc,\ldots,e_k\}$, with $e_k=\min((k+1)c,L)-1$.
Let $U_k$ and $Y_k$ stack the inputs and outputs in chunk $k$, and define
its zero-entry write map $\mathsf W_k$, incoming-state read map
$\mathsf F_k$, and local input--output block $\mathsf D_k$ by
\begin{equation}
 \begin{aligned}
 \rho_k&=\Gamma_{e_k,kc-1},
 &\mathsf W_k&=[\,\Gamma_{e_k,s}B_s\,]_{s\in I_k},\\
 \mathsf F_k&=[\,\Gamma_{t,kc-1}C_t\,]_{t\in I_k}^{\mathrm{vertical}},
 &(\mathsf D_k)_{t,s}&=
 \begin{cases}\Gamma_{t,s}C_tB_s,&s\le t,\\0,&s>t.\end{cases}
 \end{aligned}
 \label{eq:ssd_chunk_factors}
\end{equation}
Thus $\mathsf W_k$ has shape $N\times |I_k|R$,
$\mathsf F_k$ has shape $|I_k|O\times N$, and $\mathsf D_k$ has shape
$|I_k|O\times |I_k|R$; $\mathsf D_k$ denotes a local temporal block,
not a feedthrough map. With $h_k^-=h_{kc-1}$, the chunk recurrence is
\begin{equation}
 h_{k+1}^-=\rho_k h_k^-+\mathsf W_kU_k,\qquad
 Y_k=\mathsf D_kU_k+\mathsf F_kh_k^-,\qquad h_0^-=h_{-1}.
 \label{eq:ssd_all_previous_chunks}
\end{equation}
For a short final chunk, $h_m^-=h_{L-1}$. Iteration includes every earlier
chunk and the supplied initial state.
Every off-diagonal block factors through the $N$-dimensional boundary
state: block $(k,\ell)$, $\ell<k$, equals
$\mathsf F_k\bigl(\prod_{j=\ell+1}^{k-1}\rho_j\bigr)\mathsf W_\ell$. The state recurrence evaluates their sum without materializing
those blocks. This is the same block factorization underlying the SSD
algorithm of Mamba~2~\citep{dao2024transformers}, expressed here for
write/read widths $(R,O)$.

\subsection{Two-endpoint evaluation and numerical formulation}
\label{app:two_endpoint_eval}
For scalar decay with paired rotation, set $\tilde h_t=\mathcal R(-\Theta_t)h_t$,
$\widetilde v_t=\widetilde B_tu_t$,
$n_t=\Delta_{t+1}(1-\lambda_{t+1})$ and $n_{L-1}=0$.
The shifted state $S_t=\tilde h_t+n_t\widetilde v_t$ gives the evaluated form
\begin{equation}
\begin{aligned}
 S_t&=\alpha_tS_{t-1}+w_t\widetilde v_t,
 &w_t&=\lambda_t\Delta_t+n_t,\\
 \widetilde C_t\tilde h_t&=\widetilde C_t\alpha_tS_{t-1}
                 +\lambda_t\Delta_t\,d_tu_t,
 &d_t&=\widetilde C_t\widetilde B_t.
\end{aligned}
\label{eq:hm_collapse}
\end{equation}
SSD-MIMO puts the current-token weight $\lambda_t\Delta_t$ directly on the
score diagonal. HM's coordinate scan instead evaluates the propagated
previous state and the current-token contribution separately; the same
shift applies with a diagonal transition in place of $\alpha_t$.
Both avoid a subtractive output correction. For continuation, the initial
shift is $S_{-1}=\tilde h_{-1}+\Delta_0(1-\lambda_0)\widetilde v_{-1}$,
using the carried previous write in the same folded frame.
At the last token $S_{L-1}=\tilde h_{L-1}$; the physical state is
$h_{L-1}=\mathcal R(\Theta_{L-1})S_{L-1}$. Physical and folded endpoint
maps must not be mixed.

\label{app:impl_execution}
Phases and state accumulators use fp32, with phases reduced modulo
$2\pi$. SSD decay exponents are evaluated difference-first without clamps;
its matrix-product phase derivatives use the corresponding rounded operands.
By default SSD-MIMO rounds matrix-product operands to the input/output
dtype, 16-bit for 16-bit input/output and TF32 for fp32 input/output,
with fp32 accumulation; HybridMamba's flat rows use IEEE fp32 at fp32
input/output. An IEEE fp32 operand mode is available for validation. The copied-weight
checks also evaluate each implementation against an independent fp64 recurrence
(Appendix~\ref{app:numerics_exact}). HM training saves boundary
checkpoints in the input/output dtype when it is 16-bit and in fp32
otherwise; without saved checkpoints the backward replays the states.
These precision choices and validation results are route-specific; they do
not certify every path or optional provider. Validation uses normwise relative error or,
near zero, a terms-scaled bound $|a-r|\le c u\sum_i|t_i|$, where $u$ is
roundoff unit, $c$ the tolerance multiplier and $t_i$ the contributing terms.

\subsection{Backward work and execution}
\label{app:backward_cost}
The adjoint recurrence is Lemma~\ref{lem:ssm_gradients}. SSD backward
uses saved or reconstructed chunk-boundary states and differentiates the score, application, write
aggregation and carry-read products; transposition preserves their matrix
sizes. Thus forward and backward have the same asymptotic bound
in~\eqref{eq:scan_work_bounds}, which counts score formation once per group
tile of width $G_t$ and covers repeated map preparation up to constant factors.
On routes that reconstruct boundary states, the backward may use a shorter
chunk than the forward without changing the recurrence. HM backward replays from its checkpoints and applies the elementwise
adjoint, including reduction of shared endpoint gradients across subsystems.
Head-shared feature adjoints accumulate into bounded fp32 buffers; prepared
per-head endpoint adjoints retain distinct heads while combining channel
partitions. Recomputed folded maps and backward scratch are released when no
longer needed. These reductions and shorter overlapping lifetimes reduce
peak workspace, without changing the saved inputs or checkpoints.
The memory terms of~\eqref{eq:scan_work_bounds} count, for SSD-MIMO with
$HR=d/G$ scan heads, the retained drive and pre-gate output, coefficient maps, rotation
phase and two-endpoint correction table, and for HM the inputs, its head-shared coefficient maps
and, when retained, one boundary checkpoint every $T_{\mathrm c}$ tokens. SSD's general
fused route reconstructs boundary states by default. {The
rotation-free SISO route, including its chunk-parallel schedule, also uses
the general backward with state replay by default. The specialized rotating
SISO backward retains the boundary states; the conditional
$\mathsf B dN\lceil L/T\rceil$ term covers this storage.} Rotating SISO
may also retain per-head folded maps, covered by the $\mathsf B LHRN$ term.
HM regenerates its head-expanded folded maps during backward instead of
retaining them across the forward--backward boundary; its fold nevertheless
saves a phase per rotating pair and head every $T_\phi$ tokens, giving the
separate phase-checkpoint term in~\eqref{eq:scan_work_bounds}. This fold
accepts head-resolved endpoint biases shared across $P$, and saves its phase
checkpoints independently of the recurrent-checkpoint switch.
These are conservative saved-memory bounds, excluding the rest of the model;
they do not substitute for peak-memory measurements.

\paragraph{Execution schedules.}
\label{app:kernel_hardware}\label{app:kernel_schedules}
HM per-coordinate rows use the coordinate scan; HM flat rows and SSD-MIMO
use the scalar-transition SSD scan, with $R=O=1$ for HM. Mamba~1/2/3
specializations use these respective restrictions.

The coordinate scan parallelizes associative recurrence evaluation within
each chunk. SSD supports a fused serial chunk loop and schedules that
compute zero-entry summaries, propagate their carries and evaluate chunks
or segments from their true incoming states. In particular, the rotation-free
$R=O=1$ route can select a chunk-parallel schedule using device and shape
calibration; other shapes use their supported fused or segmented schedules.
Chunk and tile choices affect execution but not the recurrence.

Optimized Mamba~2 and released Mamba~3 kernels also propagate boundary
states between chunks~\citep{dao2024transformers,lahoti2026mamba3}, so avoiding explicit off-diagonal
blocks gives no asymptotic advantage over them. Optional upstream providers
and native continuation routes require separate numerical validation.

\par\endgroup

\subsection{Per-shape speed and memory}
\label{app:speed_cells}
\par\begingroup
Section~\ref{sec:hybridmamba_complexity} gives the main SISO recovery
ratios and scan training costs. This appendix adds absolute recovery times,
inference and full-mixer costs, memory across shapes, and workload/precision
checks. All performance measurements use H100. Forward-only is abbreviated
fwd; f+b denotes forward and backward. Paired cells below always follow the
order specified in their caption.
\paragraph{Performance measurement scope.}
The performance tables use an H100 PCIe, PyTorch~2.14, Triton~3.8 and
\texttt{mamba\_ssm}~2.3.2.post1. Main timings and memory use bf16;
the separately labelled appendix ablation uses fp32 input/output.
The Mamba recoveries use copied official weights and pre-timing output checks;
the Mamba~3 recovery matches the installed baseline's softplus activation.
The exact kernel-scope boundaries and gradient leaves are specified per pair
in the accompanying harness method; mixer measurements cover the full module.
Table~\ref{tab:bench_cost} rows implement different functions and are not recovery speed ratios.
The rank rows use model head width $P=64$, with real kernel widths $G=32$
at R2 and $G=16$ at R4; these are not fixed-$G$ rank scaling experiments.

\paragraph{Selection and aggregation.}
Each comparison group is measured for three rounds with alternating arm order.
The reported time is the median of the three round values; a forward-and-backward
round is an event-timed mean of ten calls after eight warm-up calls. Selection
uses the first attempt passing the validity checks and a per-group canary spread
of at most $1.15$. Selection does not use the measured latency. Whole-run drift
is recorded rather than gated. The Table~\ref{tab:app_parity_summary} summaries use geometric means and
min--max ranges over four Large or six LM shapes; HM results beyond the
five main shapes retain the earlier buffer schedule, as labelled. Noise-only failures with no
accepted attempt remain explicitly n/a.

\paragraph{Memory.}
Extra saved is retained allocator growth between a gradient-enabled
forward and its backward, excluding parameters, caller-held inputs and the
output. The harness checks release after backward and freeing leaf gradients.
Peak is the maximum allocated memory during one forward and backward above
the resident baseline of parameters, caller inputs and output cotangent,
and includes outputs, temporary workspace and newly allocated gradient buffers.

Tables~\ref{tab:bench_parity} and~\ref{tab:bench_cost} additionally report
unique nonparameter backing storage held for backward, measured in separate
memory-only runs on the same implementation. This retained total includes
caller inputs and outputs when saved, deduplicates aliases, and includes
custom context tensors and inputs held for regeneration without unpacking
saved tensors. It excludes resident parameters. Allocation and post-backward
release checks validate the accounting. HM and paired Mamba~1 timings and
peaks use a complete five-shape campaign; retained totals match the earlier
memory-only measurements exactly. SSD-MIMO and Mamba~2/3 retain their accepted
measurements; the memory-only runs reproduce their matched-pair peaks within $0.5\%$.
The remaining appendix tables retain the extra-saved metric. Retained storage
and peak allocation overlap and must not be added; neither is total-model
training memory.

\paragraph{Rank configurations and coverage.}
The real R2O2 and R4O4 configurations use forward/backward chunks $32/32$
and $16/16$, respectively. Chunk selection uses production-mixer workloads
at $N=128$, over 18 shared shapes for R2 and six LM shapes for R4.
Launch configurations are selected using five rounds on two representative
workloads per rank at $N=64$, then measured at the paper geometry $N=128$.
This is a targeted configuration screen, not an exhaustive search for an
optimum on every measured shape. Tuning and compilation precede timing.
Complex R1/R2 retain their measured configurations. Complex R4 uses training
chunks $16/16$ at kernel width $16$. Its launch configurations use the generic
light search for this chunk size during warmup, outside the reported timings.
The real R2/R4 and complex R4 training configurations have kernel and eager-mixer
forward-and-backward measurements on all five paper shapes, and saved/peak
memory at the LM shape; other training entries are n/a. Complex R4's
forward-only results retain the original measurements because its inference
configuration is unchanged.

\paragraph{Complex R4 speed and memory.}
A bounded configuration screen compared chunks $16$ and $32$; a separate run
supplies the reported table entries. The same-weight fp64 output/input-gradient
check and gradient-presence check passed. At the LM shape, chunk $16$ reduces
kernel forward-and-backward time from $70.89$ to $63.71$\,ms, while kernel
peak memory rises from $2549$ to $3125$\,MiB and mixer peak memory from
$2818$ to $3394$\,MiB. Extra-saved memory is unchanged ($974$ and
$1359$\,MiB). This is a speed/peak-memory tradeoff, not a uniform improvement.

\paragraph{Limits of the evidence.}
The Mamba~3 recovery's LM kernel forward-and-backward cell varied by about
$10\%$ between earlier processes (approximately $2.90$ versus $3.19$\,ms), while
the official arm was stable; the cause remains unresolved. HM measurement records
include launch configurations but some calibration-source fields are null, so
complete calibration-source provenance is not claimed for them. The real rank rows additionally record calibration file hashes, active groups and cached
launch configurations. Their short numerical check compares outputs and input
gradients against the same-weight fp64 Torch reference; it is not an exhaustive
numerical-validation campaign.

\paragraph{HM temporary-buffer schedule.}
The optimized backward schedule reduces redundant gradient-buffer slots and
releases scratch before subsequent adjoint operations, reducing overlapping
lifetimes. It does not reduce the activations or checkpoints retained for
backward. Table~\ref{tab:hm_workspace_tradeoff} compares both schedules on all
five main shapes using absolute measurements from each. The campaign includes
forward-only, eager and CUDA-graph training times, kernel/mixer peak allocation,
and retained storage. This buffer-schedule comparison is separate from the copied-weight fp64
recovery studies of Appendix~\ref{app:numerics_exact}.
The additional five LM shapes and the small/medium HM workloads were not
remeasured; their earlier-schedule results remain explicitly labelled.

\begin{table}[!htbp]
\centering\scriptsize
\caption{Benchmark shapes: batch $\mathsf B$, drive width $d_{\mathrm{inner}}$, length $L$, elements
$\mathsf BLd_{\mathrm{inner}}$ of one drive-sized tensor and its size in bf16. The four large shapes share
$2^{27}$ elements, $256$\,MiB per tensor. Table~\ref{tab:bench_cost} uses the large shapes and the first
language-model shape (LM); Table~\ref{tab:app_parity_summary} uses all ten, with earlier-schedule HM
measurements identified in its caption.}
\label{tab:bench_shapes}
\begin{tabular}{@{}l r r r r r@{}}
\toprule
Shape & $\mathsf B$ & $d_{\mathrm{inner}}$ & $L$ & elements & MiB (bf16) \\
\midrule
Balanced & 16 & 4096 & 2048 & $128\cdot2^{20}$ & 256 \\
Long & 2 & 2048 & 32768 & $128\cdot2^{20}$ & 256 \\
Wide & 4 & 8192 & 4096 & $128\cdot2^{20}$ & 256 \\
Batch & 128 & 1024 & 1024 & $128\cdot2^{20}$ & 256 \\
\midrule
LM & 6 & 2048 & 2048 & $24\cdot2^{20}$ & 48 \\
LM & 32 & 2048 & 2048 & $128\cdot2^{20}$ & 256 \\
LM & 2 & 2048 & 16384 & $64\cdot2^{20}$ & 128 \\
LM & 6 & 4096 & 2048 & $48\cdot2^{20}$ & 96 \\
LM & 1 & 4096 & 32768 & $128\cdot2^{20}$ & 256 \\
LM & 8 & 512 & 8192 & $32\cdot2^{20}$ & 64 \\
\bottomrule
\end{tabular}
\end{table}

\begin{table}[!htbp]
\centering\footnotesize
\caption{Same-function summaries: ours/official ratios, geometric mean (min--max) over four Large or six LM shapes. Extra-saved and peak memory refer to the full mixer. Table~\ref{tab:bench_parity} instead summarizes the five shapes shared with Table~\ref{tab:bench_cost}. H100, bf16.}
\label{tab:app_parity_summary}
\adjustbox{max width=\linewidth}{\begin{tabular}{@{}lrrrrrr@{}}
\toprule
Baseline / set & Kernel fwd & Kernel f+b & Mixer eager f+b & Mixer graph f+b & Extra saved & Peak \\
\midrule
Mamba 1 / Large & 0.74 (0.73--0.75) & 0.72 (0.68--0.79) & 0.88 (0.82--0.94) & 0.88 (0.84--0.91) & 1.06 (1.05--1.06) & 0.86 (0.86--0.86) \\
Mamba 1 / LM\textsuperscript{a} & 0.76 (0.71--0.88) & 0.79 (0.69--0.86) & 0.91 (0.82--0.95) & 0.89 (0.83--0.94) & 1.06 (1.05--1.06) & 0.86 (0.86--0.87) \\
Mamba 2 / Large & 0.81 (0.71--0.91) & 0.90 (0.85--0.95) & 1.04 (1.00--1.07) & 1.06 (1.01--1.10) & 1.05 (1.02--1.11) & 1.26 (1.23--1.28) \\
Mamba 2 / LM & 1.06 (0.77--2.11) & 1.01 (0.85--1.52) & 1.15 (1.04--1.48) & 1.08 (1.03--1.11) & 1.07 (1.03--1.19) & 1.24 (1.12--1.27) \\
Mamba 3 SISO / Large & 0.69 (0.65--0.75) & 0.83 (0.77--0.89) & 0.84 (0.76--0.89) & 0.85 (0.76--0.92) & 1.08 (1.07--1.08) & 1.01 (1.01--1.03) \\
Mamba 3 SISO / LM & 0.74 (0.65--1.13) & 0.85 (0.77--1.17) & 0.83 (0.76--0.95) & 0.82 (0.76--0.88) & 1.08 (1.07--1.08) & 1.01 (1.01--1.03) \\
\bottomrule
\end{tabular}
}
\par\smallskip\noindent\begin{minipage}{\linewidth}\footnotesize
\textsuperscript{a}The six-LM HM summary retains the earlier backward-buffer schedule in full;
only its first LM shape was remeasured, so the aggregate is not mixed with
the updated five-shape campaign.
\end{minipage}

\end{table}

\begin{table}[!htbp]
\centering\footnotesize
\caption{{HybridMamba / Mamba~1 recovery ($N=16$): ours/official ratio followed by
(ours/official) time in ms. H100, bf16; shapes: Table~\ref{tab:bench_shapes}. The first five rows use the optimized backward-buffer schedule.}}
\label{tab:app_parity_m1}
\adjustbox{max width=\linewidth}{\begin{tabular}{@{}lrrrr@{}}
\toprule
Shape & Kernel fwd & Kernel f+b & Mixer eager f+b & Mixer graph f+b \\
\midrule
\texttt{large\_balanced} & 0.74 (2.132/2.867) & 0.68 (9.958/14.614) & 0.87 (26.581/30.707) & 0.86 (27.067/31.482) \\
\texttt{large\_long} & 0.75 (2.618/3.512) & 0.79 (12.482/15.826) & 0.90 (23.051/25.744) & 0.90 (23.458/25.962) \\
\texttt{large\_wide} & 0.75 (2.148/2.870) & 0.72 (10.411/14.551) & 0.94 (40.201/42.576) & 0.91 (39.831/43.717) \\
\texttt{large\_batch} & 0.73 (2.095/2.850) & 0.72 (10.101/14.115) & 0.82 (17.933/21.848) & 0.84 (18.304/21.810) \\
\texttt{lm\_b6\_d2048\_l2048} & 0.76 (0.418/0.549) & 0.83 (2.304/2.773) & 0.94 (4.432/4.712) & 0.91 (4.389/4.835) \\
\texttt{lm\_b32\_d2048\_l2048}\textsuperscript{a} & 0.71 (2.082/2.934) & 0.69 (10.257/14.898) & 0.82 (20.727/25.255) & 0.83 (21.349/25.608) \\
\texttt{lm\_b2\_d2048\_l16384}\textsuperscript{a} & 0.74 (1.288/1.737) & 0.80 (6.370/7.949) & 0.91 (11.783/12.900) & 0.90 (12.100/13.446) \\
\texttt{lm\_b6\_d4096\_l2048}\textsuperscript{a} & 0.75 (0.812/1.089) & 0.79 (4.379/5.526) & 0.92 (10.706/11.681) & 0.89 (10.901/12.205) \\
\texttt{lm\_b1\_d4096\_l32768}\textsuperscript{a} & 0.74 (2.667/3.626) & 0.77 (12.394/16.081) & 0.90 (28.934/32.083) & 0.90 (29.345/32.527) \\
\texttt{lm\_b8\_d512\_l8192}\textsuperscript{a} & 0.88 (0.651/0.744) & 0.86 (3.290/3.807) & 0.95 (5.241/5.514) & 0.94 (5.211/5.561) \\
\bottomrule
\end{tabular}
}
\par\smallskip\noindent\begin{minipage}{\linewidth}\footnotesize
\textsuperscript{a}Additional LM shapes retain the earlier backward-buffer schedule and are
excluded from the updated five-shape mean.
\end{minipage}

\end{table}

\begin{table}[!htbp]
\centering\footnotesize
\caption{{SSD-MIMO / Mamba~2 recovery ($N=128$): ours/official ratio followed by
(ours/official) time in ms. H100, bf16; shapes: Table~\ref{tab:bench_shapes}.}}
\label{tab:app_parity_1b_m2}
\adjustbox{max width=\linewidth}{\begin{tabular}{l r r r r }
\toprule
shape & kernel fwd & kernel f+b & mixer eager f+b & mixer graph f+b \\
\midrule
\texttt{large\_balanced} & 0.84 (1.598/1.907) & 0.91 (8.717/9.559) & 1.07 (24.798/23.078) & 1.10 (25.403/23.144) \\
\texttt{large\_long} & 0.71 (1.551/2.197) & 0.87 (9.287/10.644) & 1.05 (19.615/18.678) & 1.08 (20.142/18.653) \\
\texttt{large\_wide} & 0.91 (1.524/1.671) & 0.95 (8.322/8.720) & 1.02 (36.799/36.249) & 1.05 (37.989/36.244) \\
\texttt{large\_batch} & 0.78 (1.855/2.369) & 0.85 (9.878/11.579) & 1.00 (17.766/17.679) & 1.01 (18.353/18.179) \\
\texttt{lm\_b6\_d2048\_l2048} & 2.11 (0.687/0.325) & 1.52 (3.069/2.019) & 1.48 (5.199/3.524) & 1.10 (4.007/3.628) \\
\texttt{lm\_b32\_d2048\_l2048} & 0.77 (1.678/2.193) & 0.85 (9.021/10.617) & 1.04 (19.378/18.614) & 1.07 (19.985/18.645) \\
\texttt{lm\_b2\_d2048\_l16384} & 0.83 (0.817/0.989) & 0.93 (4.712/5.043) & 1.12 (10.091/9.043) & 1.08 (10.282/9.514) \\
\texttt{lm\_b6\_d4096\_l2048} & 1.14 (0.737/0.649) & 1.00 (3.500/3.490) & 1.11 (9.480/8.564) & 1.08 (9.816/9.120) \\
\texttt{lm\_b1\_d4096\_l32768} & 0.79 (1.527/1.931) & 0.94 (9.107/9.656) & 1.09 (25.270/23.255) & 1.11 (25.928/23.380) \\
\texttt{lm\_b8\_d512\_l8192} & 1.19 (0.736/0.619) & 0.95 (3.178/3.361) & 1.13 (5.469/4.834) & 1.03 (4.980/4.830) \\
\bottomrule
\end{tabular}
}

\end{table}

\begin{table}[!htbp]
\centering\footnotesize
\caption{{SSD-MIMO / Mamba~3 SISO recovery ($N=128$): ours/official ratio followed by
(ours/official) time in ms. H100, bf16; shapes: Table~\ref{tab:bench_shapes}.}}
\label{tab:app_parity_1c_m3}
\adjustbox{max width=\linewidth}{\begin{tabular}{l r r r r }
\toprule
shape & kernel fwd & kernel f+b & mixer eager f+b & mixer graph f+b \\
\midrule
\texttt{large\_balanced} & 0.69 (2.670/3.879) & 0.84 (9.168/10.979) & 0.87 (23.451/26.940) & 0.87 (24.007/27.606) \\
\texttt{large\_long} & 0.65 (4.906/7.539) & 0.77 (13.430/17.516) & 0.76 (21.421/28.347) & 0.76 (21.947/28.868) \\
\texttt{large\_wide} & 0.66 (2.756/4.180) & 0.83 (9.735/11.734) & 0.89 (35.569/39.946) & 0.92 (37.279/40.734) \\
\texttt{large\_batch} & 0.75 (2.936/3.935) & 0.89 (9.758/11.025) & 0.86 (16.017/18.690) & 0.86 (16.600/19.271) \\
\texttt{lm\_b6\_d2048\_l2048} & 1.13 (0.946/0.838) & 1.17 (2.901/2.481) & 0.95 (4.276/4.483) & 0.84 (3.844/4.564) \\
\texttt{lm\_b32\_d2048\_l2048} & 0.69 (2.738/3.940) & 0.84 (9.242/10.989) & 0.84 (17.876/21.277) & 0.83 (18.387/22.064) \\
\texttt{lm\_b2\_d2048\_l16384} & 0.66 (2.532/3.836) & 0.77 (6.953/9.061) & 0.76 (10.971/14.472) & 0.76 (11.212/14.708) \\
\texttt{lm\_b6\_d4096\_l2048} & 0.68 (1.041/1.541) & 0.83 (3.720/4.501) & 0.87 (9.215/10.559) & 0.88 (9.600/10.942) \\
\texttt{lm\_b1\_d4096\_l32768} & 0.65 (4.877/7.507) & 0.77 (13.422/17.541) & 0.81 (27.501/33.950) & 0.82 (28.207/34.548) \\
\texttt{lm\_b8\_d512\_l8192} & 0.70 (1.402/1.993) & 0.82 (3.891/4.769) & 0.77 (5.283/6.892) & 0.77 (5.303/6.922) \\
\bottomrule
\end{tabular}
}

\end{table}

\begin{table}[!htbp]
\centering\footnotesize
\caption{Inference times (ms): each cell gives scan kernel / eager full mixer. Shapes and configurations match Table~\ref{tab:bench_cost}. Forward-only complex R4 uses its unchanged inference configuration.}
\label{tab:app_scope_fwd}\label{tab:app_mixer_fwd}
\adjustbox{max width=\linewidth}{\begin{tabular}{@{}lrrrrr@{}}
\toprule
Model & Balanced & Long & Wide & Batch & LM \\
\midrule
Mamba 1 (official) & 21.49 / 26.21 & 23.66 / 27.40 & 21.56 / 30.34 & 23.78 / 26.40 & 3.72 / 4.39 \\
Mamba 2 (official) & 1.93 / 6.67 & 2.21 / 5.22 & 1.68 / 10.52 & 2.36 / 4.65 & 0.33 / 0.96 \\
Mamba 3 SISO (official) & 3.89 / 9.30 & 7.54 / 11.62 & 4.19 / 13.21 & 3.94 / 6.66 & 0.84 / 1.56 \\
Mamba 3 MIMO R=4 (official) & 15.82\textsuperscript{a} / n/a\textsuperscript{a} & 47.50\textsuperscript{a} / n/a\textsuperscript{a} & 18.18\textsuperscript{a} / n/a\textsuperscript{a} & 15.07\textsuperscript{a} / n/a\textsuperscript{a} & 3.42\textsuperscript{a} / n/a\textsuperscript{a} \\
HM m1exact (P=1) & 14.79 / 20.83 & 18.28 / 23.76 & 14.77 / 28.02 & 15.20 / 18.48 & 2.84 / 3.89 \\
SSD m2exact (R=O=1) & 1.63 / n/a\textsuperscript{b} & 1.50 / 4.82 & 1.55 / 10.67 & 1.79 / n/a\textsuperscript{b} & 0.69 / 1.37 \\
SSD m3exact (R=O=1) & 2.68 / 7.33 & 4.91 / 7.51 & 2.76 / 11.14 & 2.95 / 5.30 & 0.94 / 1.38 \\
HM-0, P=64 & 21.07 / 25.58 & 27.74 / 31.06 & 21.87 / 30.39 & 21.19 / 23.42 & 4.25 / 4.81 \\
HM-75, P=64 & 22.82 / 27.52 & 30.37 / 33.33 & 22.68 / 31.94 & 24.42 / 26.59 & 5.20 / 5.84 \\
HM-100, P=64 & 22.64 / 27.57 & 30.54 / 33.71 & 22.58 / 31.80 & 24.09 / 26.33 & 5.26 / 5.93 \\
HM-0, P=32 & 21.16 / 25.60 & 27.90 / 31.42 & 21.89 / 30.79 & 21.58 / 23.58 & 4.26 / 4.86 \\
HM-75, P=32 & 24.52 / 29.90 & 32.15 / 34.97 & 24.72 / 34.21 & 26.37 / 28.52 & 5.54 / 6.17 \\
HM-100, P=32 & 24.86 / 30.08 & 32.67 / 35.63 & 24.84 / 34.64 & 26.41 / 29.02 & 5.65 / 6.23 \\
SSD R1O1 real & 3.15 / 8.25 & 3.08 / 6.37 & 3.08 / n/a\textsuperscript{b} & 3.38 / 5.86 & 0.92 / 1.58 \\
SSD R1O1 complex & 3.29 / 8.15 & 5.24 / 8.36 & 3.39 / n/a\textsuperscript{b} & 3.40 / 6.21 & 1.03 / 1.64 \\
SSD R2O2 real & n/a\textsuperscript{c} & n/a\textsuperscript{c} & n/a\textsuperscript{c} & n/a\textsuperscript{c} & n/a\textsuperscript{c} \\
SSD R2O2 complex & 11.14 / 18.43 & 16.07 / 21.11 & 10.57 / 22.00 & 13.04 / 17.33 & 2.18 / 3.23 \\
SSD R4O4 real & n/a\textsuperscript{c} & n/a\textsuperscript{c} & n/a\textsuperscript{c} & n/a\textsuperscript{c} & n/a\textsuperscript{c} \\
SSD R4O4 complex & 53.25 / 61.09 & 79.75 / 85.99 & 52.00 / 64.87 & 56.12 / 61.78 & 11.46 / 12.66 \\
\bottomrule
\end{tabular}
}
\par\smallskip\noindent\begin{minipage}{\linewidth}\footnotesize
\textsuperscript{a}Profiler GPU time at kernel scope; eager mixer unavailable because of wrapper overhead.
\textsuperscript{b}No attempt passed the comparison-group canary gate.
\textsuperscript{c}Not remeasured for the accepted training configuration.
\end{minipage}
\end{table}

\begin{table}[!htbp]
\centering\footnotesize
\caption{Full-mixer training times (ms): each cell gives eager / CUDA-graph f+b. Shapes match Table~\ref{tab:bench_cost}. Mamba~3 MIMO has no eager comparison because of persistent wrapper overhead; its graph measurement is retained.}
\label{tab:app_mixer_eager}\label{tab:app_mixer_graph}
\adjustbox{max width=\linewidth}{\begin{tabular}{@{}lrrrrr@{}}
\toprule
Model & Balanced & Long & Wide & Batch & LM \\
\midrule
Mamba 1 (official) & 137.0 / 137.4 & 131.4 / 132.5 & 145.8 / 147.2 & 149.0 / 151.3 & 22.68 / 22.91 \\
Mamba 2 (official) & 22.69 / 23.91 & 18.48 / 18.83 & 36.74 / 35.98 & 17.45 / 18.25 & 3.55 / 3.67 \\
Mamba 3 SISO (official) & 26.95 / 27.72 & 28.30 / 28.87 & 40.11 / 39.92 & 18.80 / 19.43 & 4.48 / 4.55 \\
Mamba 3 MIMO R=4 (official) & n/a\textsuperscript{a} / 88.87 & n/a\textsuperscript{a} / 176.7 & n/a\textsuperscript{a} / 111.3 & n/a\textsuperscript{a} / 80.91 & n/a\textsuperscript{a} / 17.83 \\
HM m1exact (P=1) & 85.71 / 86.63 & 94.93 / 95.40 & 104.3 / 105.4 & 77.08 / 77.58 & 17.21 / 17.15 \\
SSD m2exact (R=O=1) & 24.80 / 25.40 & 19.62 / 20.14 & 36.80 / 37.99 & 17.77 / 18.35 & 5.20 / 4.01 \\
SSD m3exact (R=O=1) & 23.45 / 24.01 & 21.42 / 21.95 & 35.57 / 37.28 & 16.02 / 16.60 & 4.28 / 3.84 \\
HM-0, P=64 & 110.9 / 111.8 & 150.2 / 149.3 & 126.0 / 126.7 & 103.5 / 103.9 & 21.96 / 21.80 \\
HM-75, P=64 & 123.3 / 123.8 & 166.8 / 165.4 & 140.1 / 140.0 & 118.9 / 119.0 & 25.11 / 24.80 \\
HM-100, P=64 & 122.9 / 123.4 & 170.8 / 172.6 & 137.6 / 138.7 & 119.0 / 119.4 & 24.60 / 24.35 \\
HM-0, P=32 & 111.2 / 111.8 & 150.7 / 151.5 & 127.5 / 128.2 & 104.5 / 104.9 & 21.85 / 21.80 \\
HM-75, P=32 & 132.6 / 132.8 & 180.8 / 180.1 & 150.0 / 150.5 & 125.1 / 125.5 & 27.33 / 27.07 \\
HM-100, P=32 & 131.7 / 132.4 & 187.0 / 187.4 & 147.2 / 148.4 & 125.7 / 125.9 & 27.12 / 26.83 \\
SSD R1O1 real & 28.98 / 29.53 & 24.01 / 24.39 & 40.09 / 41.59 & 22.33 / 22.78 & 5.34 / 4.93 \\
SSD R1O1 complex & 27.27 / 28.19 & 25.31 / 25.78 & 40.13 / 40.09 & 20.36 / 20.71 & 5.06 / 4.73 \\
SSD R2O2 real & 64.37 / n/a\textsuperscript{c} & 65.99 / n/a\textsuperscript{c} & 77.52 / n/a\textsuperscript{c} & 60.37 / n/a\textsuperscript{c} & 11.70 / n/a\textsuperscript{c} \\
SSD R2O2 complex & 89.98 / 90.84 & 96.67 / 97.04 & 102.9 / 103.8 & 88.96 / 89.37 & 17.04 / 16.86 \\
SSD R4O4 real & 207.8 / n/a\textsuperscript{c} & 218.7 / n/a\textsuperscript{c} & 216.0 / n/a\textsuperscript{c} & 218.3 / n/a\textsuperscript{c} & 38.73 / n/a\textsuperscript{c} \\
SSD R4O4 complex & 355.9 / n/a\textsuperscript{c} & 383.4 / n/a\textsuperscript{c} & 364.7 / n/a\textsuperscript{c} & 376.9 / n/a\textsuperscript{c} & 67.67 / n/a\textsuperscript{c} \\
\bottomrule
\end{tabular}
}
\par\smallskip\noindent\begin{minipage}{\linewidth}\footnotesize
\textsuperscript{a}Profiler GPU time at kernel scope; eager mixer unavailable because of wrapper overhead.
\textsuperscript{c}Not remeasured for the accepted training configuration.
\end{minipage}
\end{table}

\begin{table}[!htbp]
\centering\footnotesize
\caption{Scan-core memory (MiB): each cell gives extra-saved / peak allocation, with caller-held inputs and outputs excluded from extra-saved storage (Appendix~\ref{app:speed_cells}). Shapes match Table~\ref{tab:bench_cost}; n/a marks unavailable measurements, not zero memory.}
\label{tab:app_held_scope}\label{tab:app_peak_scope}
\adjustbox{max width=\linewidth}{\begin{tabular}{@{}lrrrrr@{}}
\toprule
Model & Balanced & Long & Wide & Batch & LM \\
\midrule
Mamba 1 (official) & 68 / 886 & 66 / 931 & 72 / 868 & 129 / 1090 & 14 / 178 \\
Mamba 2 (official) & 8 / 1140 & 8 / 1158 & 8 / 1100 & 8 / 1464 & 2 / 239 \\
Mamba 3 SISO (official) & 2144 / 4016 & 2161 / 4049 & 2136 / 4000 & 2193 / 4113 & 405 / 759 \\
Mamba 3 MIMO R=4 (official) & 637 / 8438 & 702 / 8502 & 604 / 8405 & 832 / 8634 & 132 / 1618 \\
HM m1exact (P=1) & 544 / 1448 & 546 / 1574 & 528 / 1376 & 608 / 1699 & 106 / 284 \\
SSD m2exact (R=O=1) & 16 / 2920 & 24 / 2930 & 12 / 2884 & 40 / 3072 & 4 / 555 \\
SSD m3exact (R=O=1) & 2128 / 3832 & 2144 / 3850 & 2120 / 3792 & 2176 / 4008 & 402 / 728 \\
HM-0, P=64 & 1123 / 2269 & 1147 / 2326 & 1096 / 2196 & 1257 / 2346 & 220 / 453 \\
HM-75, P=64 & 1967 / 5375 & 2028 / 5470 & 1915 / 5300 & 2253 / 5519 & 390 / 1043 \\
HM-100, P=64 & 1158 / 6352 & 1215 / 6394 & 1116 / 6288 & 1389 / 6446 & 234 / 1223 \\
HM-0, P=32 & 1163 / 2333 & 1187 / 2539 & 1136 / 2260 & 1297 / 2370 & 228 / 464 \\
HM-75, P=32 & 2027 / 8531 & 2085 / 8623 & 1973 / 8455 & 2311 / 8633 & 398 / 1632 \\
HM-100, P=32 & 1202 / 10516 & 1259 / 10558 & 1160 / 10452 & 1433 / 10570 & 241 / 2001 \\
SSD R1O1 real & 316 / 3252 & 332 / 3270 & 308 / 3212 & 364 / 3428 & 62 / 619 \\
SSD R1O1 complex & 2388 / 4092 & 2404 / 4110 & 2380 / 4052 & 2436 / 4268 & 451 / 776 \\
SSD R2O2 real & n/a\textsuperscript{c} & n/a\textsuperscript{c} & n/a\textsuperscript{c} & n/a\textsuperscript{c} & 245 / 1050 \\
SSD R2O2 complex & 2264 / 7498 & 2344 / 7639 & 2225 / 7404 & 2504 / 8051 & 440 / 1442 \\
SSD R4O4 real & n/a\textsuperscript{c} & n/a\textsuperscript{c} & n/a\textsuperscript{c} & n/a\textsuperscript{c} & 584 / 2334 \\
SSD R4O4 complex & n/a\textsuperscript{c} & n/a\textsuperscript{c} & n/a\textsuperscript{c} & n/a\textsuperscript{c} & 974 / 3125 \\
\bottomrule
\end{tabular}
}
\par\smallskip\noindent\begin{minipage}{\linewidth}\footnotesize
\textsuperscript{c}Not remeasured for the accepted training configuration.
\end{minipage}
\end{table}

\begin{table}[!htbp]
\centering\footnotesize
\caption{Full-mixer memory (MiB): each cell gives extra-saved / peak allocation, using the exclusions defined above. Shapes match Table~\ref{tab:bench_cost}; n/a marks unavailable measurements, not zero memory.}
\label{tab:app_held_mixer}\label{tab:app_peak_mixer}
\adjustbox{max width=\linewidth}{\begin{tabular}{@{}lrrrrr@{}}
\toprule
Model & Balanced & Long & Wide & Batch & LM \\
\midrule
Mamba 1 (official) & 1056 / 3299 & 1048 / 3321 & 1196 / 3468 & 1165 / 3533 & 209 / 639 \\
Mamba 2 (official) & 965 / 3793 & 945 / 3823 & 1103 / 3875 & 968 / 4068 & 187 / 733 \\
Mamba 3 SISO (official) & 3120 / 5648 & 3117 / 5649 & 3251 / 5819 & 3176 / 5737 & 595 / 1073 \\
HM m1exact (P=1) & 1518 / 3334 & 1498 / 3329 & 1648 / 3456 & 1580 / 3568 & 297 / 647 \\
SSD m2exact (R=O=1) & 997 / 4814 & 1001 / 4807 & 1123 / 4955 & 1072 / 5001 & 198 / 920 \\
SSD m3exact (R=O=1) & 3360 / 5720 & 3357 / 5707 & 3490 / 5866 & 3415 / 5888 & 640 / 1089 \\
HM-0, P=64 & 2103 / 3905 & 2106 / 3929 & 2220 / 4024 & 2246 / 3976 & 413 / 771 \\
HM-75, P=64 & 3206 / 7014 & 3249 / 7079 & 3296 / 7129 & 3509 / 7160 & 629 / 1361 \\
HM-100, P=64 & 2398 / 7992 & 2439 / 8006 & 2500 / 8120 & 2650 / 8092 & 476 / 1577 \\
HM-0, P=32 & 2151 / 3977 & 2154 / 4002 & 2276 / 4104 & 2293 / 4007 & 421 / 781 \\
HM-75, P=32 & 3275 / 10179 & 3315 / 10242 & 3371 / 10301 & 3576 / 10282 & 642 / 1952 \\
HM-100, P=32 & 2452 / 12166 & 2491 / 12178 & 2560 / 12300 & 2702 / 12224 & 485 / 2323 \\
SSD R1O1 real & 1567 / 4903 & 1577 / 4904 & 1692 / 5044 & 1664 / 5113 & 307 / 939 \\
SSD R1O1 complex & 3897 / 6001 & 3909 / 6003 & 4021 / 6141 & 4000 / 6217 & 744 / 1145 \\
SSD R2O2 real & n/a\textsuperscript{c} & n/a\textsuperscript{c} & n/a\textsuperscript{c} & n/a\textsuperscript{c} & 600 / 1289 \\
SSD R2O2 complex & 4072 / 8682 & 4179 / 8838 & 4152 / 8755 & 4460 / 9368 & 795 / 1682 \\
SSD R4O4 real & n/a\textsuperscript{c} & n/a\textsuperscript{c} & n/a\textsuperscript{c} & n/a\textsuperscript{c} & 968 / 2602 \\
SSD R4O4 complex & n/a\textsuperscript{c} & n/a\textsuperscript{c} & n/a\textsuperscript{c} & n/a\textsuperscript{c} & 1359 / 3394 \\
\bottomrule
\end{tabular}
}
\par\smallskip\noindent\begin{minipage}{\linewidth}\footnotesize
\textsuperscript{c}Not remeasured for the accepted training configuration.
\end{minipage}
\end{table}

\begin{table}[!htbp]
\centering\footnotesize
\caption{{HM backward-buffer ablation: geometric-mean optimized/unreduced-buffer
ratios over Table~\ref{tab:bench_cost}'s five shapes ($<1$: lower time or peak).
Times include forward and backward; Mixer is the eager full module.
Both schedules use identical models and retained-for-backward storage.
At $P=32$, HM-75/100 mixer time rises $3.9\%$/$7.8\%$ on geometric mean,
up to $18\%$ on one shape; m1exact is essentially unchanged.
H100, bf16, $N=128$.}}
\label{tab:hm_workspace_tradeoff}
\begin{tabular}{@{}lrrrr@{}}
\toprule
Model & Kernel time & Mixer time & Kernel peak & Mixer peak \\
\midrule
HM-0, P=64 & 0.960 & 0.958 & 0.516 & 0.648 \\
HM-75, P=64 & 0.981 & 0.993 & 0.486 & 0.554 \\
HM-100, P=64 & 0.992 & 1.038 & 0.503 & 0.567 \\
HM-0, P=32 & 0.963 & 0.962 & 0.525 & 0.652 \\
HM-75, P=32 & 1.010 & 1.039 & 0.762 & 0.793 \\
HM-100, P=32 & 1.025 & 1.078 & 0.830 & 0.850 \\
\bottomrule
\end{tabular}

\end{table}

\paragraph{Smaller workloads and compilation.}
Small and medium workloads expose launch overhead and include cases where
our implementations are slower. Table~\ref{tab:app_small_summary} retains
these comparisons under the same protocol; its HM row retains the earlier
backward-buffer schedule, which was not remeasured on these shapes.
Extra-saved/peak memory refers to the full mixer. Compiled Mamba~2 mixer f+b ratios have geometric means
$1.08$ on Large and $1.25$ on LM (Table~\ref{tab:app_compile_m2}). The tested
official Mamba~3 compilation path fails, so it has no comparable column.
Mamba~2 held-memory regeneration is disabled under compilation.

\begin{table}[!htbp]
\centering\footnotesize
\caption{Small and medium workload recovery ratios, ours/official: geometric mean (min--max). The baselines use the same state sizes as the main recovery comparison. H100, bf16.}
\label{tab:app_small_summary}
\adjustbox{max width=\linewidth}{\begin{tabular}{@{}lrrrrrr@{}}
\toprule
Baseline / set & Kernel fwd & Kernel f+b & Mixer eager f+b & Mixer graph f+b & Extra saved & Peak \\
\midrule
Mamba 1 / Small+medium\textsuperscript{a} & 1.22 (0.73--2.34) & 1.22 (0.70--2.38) & 1.44 (0.90--2.28) & 0.99 (0.88--1.16) & 1.04 (0.99--1.06) & 1.07 (0.86--1.50) \\
Mamba 2 / Small+medium & 2.08 (1.17--3.06) & 1.35 (0.94--1.71) & 1.35 (1.02--1.73) & 1.08 (0.94--1.21) & 1.26 (1.01--1.91) & 1.07 (0.81--1.28) \\
Mamba 3 SISO / Small+medium & 0.95 (0.44--1.89) & 0.96 (0.39--1.59) & 0.95 (0.43--1.72) & 0.70 (0.27--0.98) & 0.93 (0.59--1.08) & 1.01 (0.83--1.08) \\
\bottomrule
\end{tabular}
}
\par\smallskip\noindent\begin{minipage}{\linewidth}\footnotesize
\textsuperscript{a}HM uses the earlier backward-buffer schedule.
\end{minipage}

\end{table}

\begin{table}[!htbp]
\centering\footnotesize
\caption{Compiled Mamba~2 mixer f+b: ours/official ratio and (ours/official) absolute time in ms. H100, bf16.}
\label{tab:app_compile_m2}\label{tab:app_compile_m2_detail}
\adjustbox{max width=\linewidth}{\begin{tabular}{l r }
\toprule
shape & mixer compile f+b \\
\midrule
\texttt{large\_balanced} & 1.13 (25.689/22.788) \\
\texttt{large\_long} & 1.08 (20.362/18.813) \\
\texttt{large\_wide} & 1.05 (37.356/35.626) \\
\texttt{large\_batch} & 1.06 (18.783/17.671) \\
\texttt{lm\_b6\_d2048\_l2048} & 1.63 (5.612/3.438) \\
\texttt{lm\_b32\_d2048\_l2048} & 1.10 (20.316/18.418) \\
\texttt{lm\_b2\_d2048\_l16384} & 1.19 (10.517/8.875) \\
\texttt{lm\_b6\_d4096\_l2048} & 1.15 (9.730/8.466) \\
\texttt{lm\_b1\_d4096\_l32768} & 1.17 (26.387/22.544) \\
\texttt{lm\_b8\_d512\_l8192} & 1.33 (6.279/4.725) \\
\bottomrule
\end{tabular}
}

\end{table}

\begin{table}[!htbp]
\centering\footnotesize
\caption{Why the Mamba~3 MIMO scan metric differs: elapsed time (wall) versus summed profiler GPU-kernel time (prof), in ms, with prof/wall ratio. The profiler excludes persistent wrapper overhead; the two timing metrics are not interchangeable.}
\label{tab:app_mimo_profiler}
\adjustbox{max width=\linewidth}{\begin{tabular}{l r r r r r r}
\toprule
shape & MIMO wall & MIMO prof & MIMO prof/wall & SISO wall & SISO prof & SISO prof/wall \\
\midrule
Bal.\ 16/4096/2048 & 710.1 & 65.76 & 0.093 & 10.91 & 10.74 & 0.985 \\
Long 2/2048/32768 & 721.7 & 148.9 & 0.206 & 17.54 & 17.13 & 0.977 \\
Wide 4/8192/4096 & 729.2 & 73.92 & 0.101 & 11.65 & 11.52 & 0.989 \\
Batch 128/1024/1024 & 721.9 & 65.82 & 0.091 & 10.99 & 10.83 & 0.986 \\
LM 6/2048/2048 & 719.5 & 14.27 & 0.020 & 2.49 & 2.36 & 0.949 \\
\bottomrule
\end{tabular}
}

\end{table}

\subsection{Baseline tracking with fp32 input and output}
\label{app:fp32_baseline_tracking}
Table~\ref{tab:fp32_baseline_tracking} compares the three baseline recoveries
on H100 PCIe with fp32 input and output. It uses copied weights, output checks
before timing, three-round medians and the first comparison-group attempt
passing the validity and noise gates. Kernel timings use the aligned scope;
mixer timings cover an eager forward and backward of the full module.
The SSD recoveries here use $R=O=1$; fp32 memory is not measured.
HM timings retain the earlier backward-buffer schedule; the buffer
optimization was measured only in bf16.

\begin{table}[!htbp]
\centering\scriptsize
\caption{Baseline tracking with fp32 input/output: time ratios ours/official
($<1$ means ours is faster), geometric mean with min--max underneath over four
Large or six LM shapes. Mamba~1 uses IEEE arithmetic, Mamba~2 uses tf32,
and our Mamba~3 recovery uses its opt-in bf16-operand mode to match the
official implementation's internal operand casts. The Mamba~3 row is not
a comparison of full-fp32 scan arithmetic. PyTorch~2.14, Triton~3.8.}
\label{tab:fp32_baseline_tracking}
\adjustbox{max width=\linewidth}{\begin{tabular}{@{}l ccc ccc@{}}
\toprule
 & \multicolumn{3}{c}{Large} & \multicolumn{3}{c}{LM} \\
\cmidrule(lr){2-4}\cmidrule(lr){5-7}
Pair (our operands) & kernel fwd & kernel f+b & mixer f+b & kernel fwd & kernel f+b & mixer f+b \\
\midrule
Mamba 1 (IEEE) & \shortstack{0.54\\{\tiny (0.51--0.65)}} & \shortstack{0.53\\{\tiny (0.45--0.65)}} & \shortstack{0.97\\{\tiny (0.87--1.02)}} & \shortstack{0.58\\{\tiny (0.51--0.67)}} & \shortstack{0.61\\{\tiny (0.50--0.68)}} & \shortstack{0.99\\{\tiny (0.95--1.03)}} \\
Mamba 2 (tf32) & \shortstack{1.36\\{\tiny (1.16--1.58)}} & \shortstack{1.00\\{\tiny (0.97--1.08)}} & \shortstack{1.15\\{\tiny (1.11--1.19)}} & \shortstack{1.54\\{\tiny (1.15--1.70)}} & \shortstack{1.02\\{\tiny (0.92--1.09)}} & \shortstack{1.18\\{\tiny (1.15--1.22)}} \\
Mamba 3 (bf16 operands) & \shortstack{0.88\\{\tiny (0.76--0.98)}} & \shortstack{0.94\\{\tiny (0.85--1.02)}} & \shortstack{0.95\\{\tiny (0.91--0.98)}} & \shortstack{0.88\\{\tiny (0.76--1.11)}} & \shortstack{0.94\\{\tiny (0.85--1.12)}} & \shortstack{0.94\\{\tiny (0.85--1.02)}} \\
\bottomrule
\end{tabular}
}
\end{table}

With our default tf32 operands instead, the Mamba~3 kernel forward-and-backward
ratio is $4.03$ on Large and $3.57$ on LM, and the eager mixer ratio is $2.00$
and $2.06$, respectively. These ratios describe the measured operand configurations; no subsequent
fp32 tuning is included.

\par\endgroup

\section{Experimental tasks and protocols}
\label{app:exp_details}

\subsection{MQMKVAR: generator and supervision}
\label{app:mqmkvar_protocol}
\par\begingroup

MQMKVAR combines MQAR's context-first, multi-query layout and
power-law query placement adapted from
Zoology~\citep{arora2023zoologymeasuringimprovingrecall}
with multi-token keys and variable-length values as in MAD's fuzzy
in-context recall~\citep{poli2024mad}. Keys and values use disjoint vocabularies.

For medium and hard, key lengths remain variable at evaluation. The generator
controls prefix/substring overlap, and masks every query answer slot in
the input with padding. The complete value list must be produced from
state without seeing earlier answers. Figure~\ref{fig:recall_tasks} in the
introduction shows an actual proper-prefix example and the label alignment.
{The first value is scored at the last key token unless the key is a
proper prefix of another stored key. In that case, the last key token is
loss-masked and targets start at the following padding token, which confirms
key completion. Subsequent values are scored at successive padding positions.
This convention makes answer alignment depend on context prefix relations;
predictions at masked positions remain unconstrained.} Multi-token binding, variable-length values and the
unambiguous last-key-token convention are inherited task features.
The generator also offers an explicit completion marker (\emph{fire}
mode), which the reported runs do not use. {The task fixes the target
values and their positions, not how a model routes them through its state.}

The overlap parameter selects parent/subkey pairs: for $K$ stored keys
and rate $\rho$, the generator requests $\operatorname{round}(K\rho/2)$
pairs, with a proper contiguous subkey per parent when a distinct one
can be sampled. Thus $\rho$ specifies pair participation, not the
fraction of keys that are themselves proper prefixes; substring overlap
need not create an ambiguous endpoint. Table~\ref{tab:mq_protocol} gives
$\rho=0.4$ and $0.8$ for medium and hard. At unit lengths the binding
rule reduces to single-token MQAR, but the generators are not equivalent:
medium/hard permit repeated queries, use loss-masked padding between them,
evaluate on a held-out layout and draw query gaps from a power law with
exponent parameter $0.01$; easy uses Zoology's native slot sampling.

\subsection{MQMKVAR: training, selection and results}
\label{app:mqmkvar_analysis}

The optimizer is AdamW with $\beta=(0.9,0.95)$, no weight decay,
bf16 mixed precision with fp32 master weights, and no gradient clipping.
The base recipe uses peak learning rate $10^{-3}$, with two warmup epochs
and per-epoch cosine decay to $10^{-4}$. Easy and medium use this recipe without per-model learning-rate tuning;
hard additionally uses the sweep below.

\begin{table}[H]
\centering\small
\caption{Base training recipe by tier (peak LR $10^{-3}$); the hard
sweep changes its peak LR. Each run uses one of seeds $101,202,303$. The metric is final held-out token accuracy: correct supervised
value tokens divided by all supervised value tokens.}
\label{tab:mq_protocol}
\begin{tabular}{@{}l r c c c r r r@{}}
\toprule Tier & Length & Key length & Value length & Interleave & Batch & Epochs & Updates \\
\midrule
Easy & 1024 & 1 & 1 & 0 & 64 & 32 & 50,016 \\
Medium & 1024 & 2--4 & 2--6 & 0.4 & 64 & 32 & 50,016 \\
Hard & 2048 & 5--12 & 5--12 & 0.8 & 32 & 16 & 50,000 \\
\bottomrule
\end{tabular}

\end{table}

\label{app:mqmkvar_ledger}

There are 100,000 training, 10,000 validation
and 10,000 test examples, vocabulary size 8192 and 32 stored entries;
the data seed is 12345 and final-test seed is 99999. The native easy
layout queries each key once, uses unique values, excludes random
non-query tokens and samples query slots without replacement under the
power law of the medium and hard tiers. The two warmup epochs use an
initial learning-rate factor $10^{-8}$. All prescribed completed seeds,
including low-accuracy runs, enter the summaries, all under the
token-weighted estimand.

\paragraph{Configuration descriptions.}
Suffixes -r and -c mark real and complex state rows. All listed SSMs use $N=128$ and two layers. Residual and inner
widths and non-embedding counts are in Table~\ref{tab:mqmkvar_main}.
M1 uses expansion two and a separate step per channel. M2 and M3 retain
head width 64 and one endpoint-sharing group. M3 is the upstream
heavy-tail-rate configuration. The attention reference has two layers,
one attention head, learned absolute positions, pre-layer normalization,
a tied output head and dropout $0.1$.

SSD uses $P=64$: R2O2-r/c have seven/eight heads, R4O4-r/c six,
and R1O1-r/c nine. HM uses $P=16$. Table~\ref{tab:mq_configurations}
specifies every row allocation, rank, decay law and gate mask;
counts refer to real coordinates, with two per complex coordinate.
Complex gates, when enabled, are pair-tied as in~\eqref{eq:hm_gates}.
Every arm is width-matched to the parameter band
(Table~\ref{tab:mqmkvar_main}); gate-removal arms change only the residual width.

\paragraph{Endpoint and rotation specifications.}
M1/M2 use unnormalized endpoint maps without additive biases
($\beta=\gamma=0$ in~\eqref{eq:endpoint_gates}): M1 selects them from
the processed drive and uses a low-rank step projection; M2 has one
sharing group and joint drive/B/C convolution. M3 uses convolution-free,
normalized maps with per-head biases. SSD uses~\eqref{eq:ssd_write},
with coefficients read from the mixer input ($s_t=x_t$) and the gate masks in
Table~\ref{tab:mq_configurations}. Biases initialize to ones at
$R=1$, otherwise to the sibling spread; gate features initialize to zero.
HM maps read the processed drive ($s_t=u_t$), are RMS-normalized and have
\emph{no additive map biases} in these experiments; all HM rows are
ungated. Under this parameter-matched recipe, map gates cost HM model width;
in a study on an earlier HM revision this trade did not improve hard-tier recall,
so HM defaults to ungated maps. Potential ($\mu$) rows, tested on the same earlier
revision, reached lower and less stable hard-tier recall than dynamic rows, and no
reported configuration uses them. Both own-model families use a width-four drive convolution and
SiLU, no B/C convolution, no output normalization, and the two-endpoint
write with selected $\lambda_t$. All initial recurrent states are zero.

Default learned rotation means head-shared angle features $q_{k,t}$,
per-pair frequencies $\omega_k=\pi\tanh\varphi_k$ initialized at
$\varphi_k=3.8$, and a learned per-head/pair read offset $\nu$
initialized to zero, as in~\eqref{eq:hm_rotation_law}. The angle projection
uses its default fan-in draw. The head-frequency variant instead uses
one $q_{\eta,t}$ and $\omega_\eta$ shared across all pairs of a head,
with the angle projection draw scaled by $0.1\sqrt3$.
Dynamic rates use the heavy-tail activation with magnitude floored at
$10^{-4}$; SSD selects one rate per head, and HM selects per state row
with complex pairs tied. Static HM rates vary per channel and state row,
again tied within each complex pair. The selected step varies per head
in HM and per sibling in SSD. Initial steps are log-uniform in
$[10^{-3},10^{-1}]$, floored at $10^{-4}$; endpoint-weight projections
have zero bias, giving $\lambda=1/2$ at zero selection. Static HM
rate magnitudes start at $1,2,\ldots$ within each row block, counting
a complex pair once.

\begingroup\footnotesize
\begin{longtable}{@{}p{.26\linewidth}c p{.29\linewidth}p{.31\linewidth}@{}}
\caption{Exact MQ row configurations. Counts denote real coordinates, so 64 complex rows mean 32 complex coordinates. All SSM rows use $N=128$; the text defines the endpoint recipes and rotation.}\label{tab:mq_configurations}\\
\toprule Configuration & $R/O$ & Row partition and law & Gates, rotation and write \\
\midrule\endfirsthead
\toprule Configuration & $R/O$ & Row partition and law & Gates, rotation and write \\
\midrule\endhead
\bottomrule\endfoot
M1 & $1/1$ & 128 real, coordinate-static & No map gates/rotation; Euler; M1 endpoints. \\
M2 & $1/1$ & 128 real, head-static & No map gates/rotation; Euler; M2 endpoints. \\
M3 & $1/1$ & 64 complex + 64 real, head-dynamic & No map gates; fixed frequency $\pi$, no read offset; two endpoints; M3 maps. \\
Attention & --- & No recurrent state & One attention head; learned absolute positions. \\
\midrule
R1O1-r & $1/1$ & 128 real, head-dynamic & Ungated; no rotation; two endpoints; SSD maps. \\
R1O1-c & $1/1$ & 64 complex + 64 real, head-dynamic & Ungated; default learned rotation; two endpoints; SSD maps. \\
R2O2-r & $2/2$ & 128 real, head-dynamic & Real rows gated; no rotation; two endpoints; SSD maps. \\
R2O2-r, gates off & $2/2$ & 128 real, head-dynamic & Ungated; no rotation; two endpoints; SSD maps. \\
R2O2-c & $2/2$ & 64 complex + 64 real, head-dynamic & Only real rows gated; default learned rotation; two endpoints; SSD maps. \\
R2O2-c, gates off & $2/2$ & 64 complex + 64 real, head-dynamic & Ungated; default learned rotation; two endpoints; SSD maps. \\
R2O2-c, head frequency & $2/2$ & 64 complex + 64 real, head-dynamic & Only real rows gated; one frequency/angle feature per head, scaled angle initialization; two endpoints; SSD maps. \\
R4O4-r & $4/4$ & 128 real, head-dynamic & Real rows gated; no rotation; two endpoints; SSD maps. \\
R4O4-c & $4/4$ & 64 complex + 64 real, head-dynamic & Only real rows gated; default learned rotation; two endpoints; SSD maps. \\
\midrule
HM $A$+dynamic & $1/1$ & 64 real static + 64 real dynamic & Ungated; no rotation; two endpoints; HM maps. \\
HM complex-$A$ mix & $1/1$ & 48 real static + 48 real dynamic + 32 complex static & Ungated; default learned rotation; two endpoints; HM maps. \\
HM complex-dynamic mix & $1/1$ & 48 real static + 48 real dynamic + 32 complex dynamic & Ungated; default learned rotation; two endpoints; HM maps. \\
HM all-$A$ real & $1/1$ & 128 real static & Ungated; no rotation; two endpoints; HM maps. \\
HM all-$A$ complex & $1/1$ & 128 complex static & Ungated; default learned rotation; two endpoints; HM maps. \\
\end{longtable}
\endgroup

All 54 fixed-recipe cells contain three completed seeds. The reported
head-frequency arm uses ungated complex and gated real rows;
superseded configurations and diagnostic reruns are excluded.

\paragraph{Learning-rate sweep and selection.}
\label{app:mq_lr_sweep}

All 18 configurations use the common hard-tier LR grid
$\{5\times10^{-4},10^{-3},2\times10^{-3}\}$ with seeds $101,202,303$;
the $10^{-3}$ runs are reused from the shared recipe. Geometry and other
hyperparameters are fixed within each arm. For each configuration, selection maximizes the three-seed mean final
held-out test token accuracy among rates with three completed seeds; exact ties prefer $10^{-3}$,
then the smaller rate. Selecting by the logged validation means instead of the test means picks the same rate
for every arm. For runs that predate the metric fix, these validation logs average token accuracy
per batch rather than globally and were not re-scored; all test endpoints were. Because the selected rate's test
mean is also the reported value, the hard column is a best-of-three over rates per arm and is
slightly optimistic; the validation check limits but does not remove this.

The extra M3/attention $10^{-4}$ cells are reported but excluded from
selection to preserve a common grid. Their LR is constant after warmup
because the cosine floor equals the peak. Table~\ref{tab:mq_lr_sweep}
contains every scheduled seed and identifies the only incomplete cell.

\paragraph{Attention position ablation.}
Replacing the reference's learned positions by rotary embeddings ($\theta=10^4$, full head width),
with everything else unchanged, lowers held-out token accuracy at peak LR $10^{-3}$ from $22.2\%$ to
$0.7\%$ on medium ($0.74$, $0.73$, $0.72$ for seeds $101,202,303$) and from $1.1\%$ to $0.3\%$ on hard
($0.31$, $0.29$, $0.29$). Its code contains the metric fix, so these are global token accuracies.

\paragraph{Divergence and execution differences.}
The M1 seed that diverged at $2\times10^{-3}$ (Table~\ref{tab:mq_lr_sweep})
ended in a final-RMSNorm overflow; diagnostic reruns are excluded.
Non-finite/zero-gradient guards, GPU architectures and code snapshots
differed across cells, including A100/H100 M3 sweep runs versus RTX PRO 5000 (Blackwell)
runs in its shared-rate cell, so repeated seed IDs do not define paired
comparisons and LR is not the only execution difference.

\begingroup\footnotesize\setlength{\tabcolsep}{3pt}
\begin{longtable}{@{}p{.235\linewidth}*{5}{>{\centering\arraybackslash}p{.14\linewidth}}@{}}
\caption{Complete MQMKVAR results: held-out token accuracy (percent). Each cell shows the completed-seed mean above seeds 101, 202 and 303; all cells complete 3/3 attempts except $\dagger$ (2/3). Underlining marks the best eligible hard-tier mean over the three common rates; ties prefer $10^{-3}$, then the smaller rate. D denotes the non-escaping, diverged M1 seed described in Appendix~\ref{app:mq_lr_sweep}.}\label{tab:mq_complete}\label{tab:mqmkvar_final_seeds}\label{tab:mq_lr_sweep}\\
\toprule
Configuration & Easy & Medium & \multicolumn{3}{c}{Hard} \\
\cmidrule(lr){4-6}
Peak LR & $10^{-3}$ & $10^{-3}$ & $5\times10^{-4}$ & $10^{-3}$ & $2\times10^{-3}$ \\
\midrule
\endfirsthead
\toprule
Configuration & Easy & Medium & \multicolumn{3}{c}{Hard} \\
\cmidrule(lr){4-6}
Peak LR & $10^{-3}$ & $10^{-3}$ & $5\times10^{-4}$ & $10^{-3}$ & $2\times10^{-3}$ \\
\midrule
\endhead
\bottomrule\endfoot
M1\footnote{M1 at $2\times10^{-3}$: $48.9$ (diverged seed counted as $0$) / $73.3$ (completed seeds, $2/3$). The rate is ineligible.} & \shortstack{99.9\\{\scriptsize 99.9, 100.0, 100.0}} & \shortstack{89.3\\{\scriptsize 88.8, 89.7, 89.4}} & \shortstack{67.1\\{\scriptsize 69.2, 61.5, 70.6}} & \shortstack{\underline{69.6}\\{\scriptsize 72.8, 68.8, 67.3}} & \shortstack{73.3$^{\dagger}$\\{\scriptsize 74.1, 72.6, D}} \\[3pt]
M2 & \shortstack{100.0\\{\scriptsize 100.0, 100.0, 100.0}} & \shortstack{91.7\\{\scriptsize 88.7, 89.1, 97.3}} & \shortstack{\underline{42.0}\\{\scriptsize 66.0, 30.3, 29.6}} & \shortstack{36.4\\{\scriptsize 52.0, 29.5, 27.8}} & \shortstack{39.9\\{\scriptsize 46.9, 43.5, 29.4}} \\[3pt]
M3 & \shortstack{99.9\\{\scriptsize 99.6, 100.0, 100.0}} & \shortstack{76.6\\{\scriptsize 84.7, 58.1, 87.1}} & \shortstack{\underline{27.8}\\{\scriptsize 48.2, 14.3, 21.0}} & \shortstack{12.1\\{\scriptsize 2.8, 2.2, 31.2}} & \shortstack{1.1\\{\scriptsize 2.9, 0.1, 0.2}} \\[3pt]
Attention & \shortstack{100.0\\{\scriptsize 100.0, 100.0, 100.0}} & \shortstack{22.2\\{\scriptsize 22.2, 22.2, 22.2}} & \shortstack{0.5\\{\scriptsize 0.4, 0.3, 0.7}} & \shortstack{\underline{1.1}\\{\scriptsize 1.9, 0.9, 0.4}} & \shortstack{0.0\\{\scriptsize 0.0, 0.0, 0.0}} \\[3pt]
\midrule
R1O1-r & \shortstack{100.0\\{\scriptsize 100.0, 100.0, 100.0}} & \shortstack{89.6\\{\scriptsize 89.7, 89.7, 89.4}} & \shortstack{\underline{67.0}\\{\scriptsize 63.9, 71.7, 65.4}} & \shortstack{62.3\\{\scriptsize 71.6, 67.4, 47.9}} & \shortstack{35.4\\{\scriptsize 15.0, 41.1, 50.1}} \\[3pt]
R1O1-c & \shortstack{99.9\\{\scriptsize 100.0, 100.0, 99.9}} & \shortstack{89.8\\{\scriptsize 89.6, 90.3, 89.5}} & \shortstack{\underline{69.4}\\{\scriptsize 67.6, 61.2, 79.3}} & \shortstack{48.9\\{\scriptsize 6.4, 67.8, 72.5}} & \shortstack{57.5\\{\scriptsize 67.9, 52.4, 52.1}} \\[3pt]
R2O2-r & \shortstack{100.0\\{\scriptsize 100.0, 100.0, 100.0}} & \shortstack{91.3\\{\scriptsize 91.0, 91.5, 91.3}} & \shortstack{79.5\\{\scriptsize 80.1, 78.3, 79.9}} & \shortstack{\underline{82.0}\\{\scriptsize 81.7, 81.1, 83.3}} & \shortstack{77.5\\{\scriptsize 77.9, 79.0, 75.6}} \\[3pt]
R2O2-r, gates off & \shortstack{100.0\\{\scriptsize 100.0, 100.0, 100.0}} & \shortstack{89.4\\{\scriptsize 89.6, 89.3, 89.4}} & \shortstack{\underline{60.3}\\{\scriptsize 53.1, 51.0, 76.9}} & \shortstack{59.9\\{\scriptsize 68.7, 50.4, 60.5}} & \shortstack{2.6\\{\scriptsize 0.2, 3.8, 3.7}} \\[3pt]
R2O2-c & \shortstack{100.0\\{\scriptsize 100.0, 100.0, 100.0}} & \shortstack{91.1\\{\scriptsize 90.9, 91.5, 90.8}} & \shortstack{82.2\\{\scriptsize 80.0, 83.0, 83.6}} & \shortstack{\underline{82.6}\\{\scriptsize 83.5, 82.1, 82.3}} & \shortstack{61.1\\{\scriptsize 81.7, 18.5, 83.2}} \\[3pt]
R2O2-c, gates off & \shortstack{100.0\\{\scriptsize 100.0, 100.0, 100.0}} & \shortstack{90.4\\{\scriptsize 90.5, 90.8, 89.9}} & \shortstack{79.1\\{\scriptsize 82.0, 75.7, 79.5}} & \shortstack{\underline{82.2}\\{\scriptsize 82.6, 82.6, 81.4}} & \shortstack{0.3\\{\scriptsize 0.4, 0.4, 0.1}} \\[3pt]
R2O2-c, head frequency & \shortstack{100.0\\{\scriptsize 100.0, 100.0, 100.0}} & \shortstack{91.0\\{\scriptsize 90.8, 90.5, 91.7}} & \shortstack{\underline{82.0}\\{\scriptsize 80.9, 81.7, 83.2}} & \shortstack{80.8\\{\scriptsize 82.0, 78.7, 81.9}} & \shortstack{80.7\\{\scriptsize 81.7, 79.4, 81.0}} \\[3pt]
R4O4-r & \shortstack{100.0\\{\scriptsize 100.0, 100.0, 100.0}} & \shortstack{91.5\\{\scriptsize 92.0, 91.3, 91.1}} & \shortstack{78.2\\{\scriptsize 78.4, 76.5, 79.7}} & \shortstack{\underline{82.0}\\{\scriptsize 83.4, 80.9, 81.8}} & \shortstack{53.6\\{\scriptsize 3.2, 75.7, 81.9}} \\[3pt]
R4O4-c & \shortstack{100.0\\{\scriptsize 100.0, 100.0, 100.0}} & \shortstack{90.5\\{\scriptsize 91.6, 90.4, 89.6}} & \shortstack{\underline{82.2}\\{\scriptsize 81.9, 82.7, 82.0}} & \shortstack{81.9\\{\scriptsize 82.5, 80.4, 82.7}} & \shortstack{79.9\\{\scriptsize 79.4, 80.6, 79.7}} \\[3pt]
\midrule
HM $A$+dynamic & \shortstack{100.0\\{\scriptsize 100.0, 100.0, 100.0}} & \shortstack{93.5\\{\scriptsize 96.0, 93.2, 91.2}} & \shortstack{79.6\\{\scriptsize 76.8, 81.0, 81.1}} & \shortstack{\underline{80.6}\\{\scriptsize 80.6, 79.8, 81.4}} & \shortstack{59.7\\{\scriptsize 27.9, 79.5, 71.7}} \\[3pt]
HM complex-$A$ mix & \shortstack{100.0\\{\scriptsize 100.0, 100.0, 100.0}} & \shortstack{95.1\\{\scriptsize 90.9, 96.4, 98.1}} & \shortstack{62.7\\{\scriptsize 26.5, 80.9, 80.6}} & \shortstack{\underline{79.6}\\{\scriptsize 79.7, 79.1, 80.0}} & \shortstack{60.1\\{\scriptsize 27.3, 81.6, 71.5}} \\[3pt]
HM complex-dynamic mix & \shortstack{100.0\\{\scriptsize 100.0, 100.0, 100.0}} & \shortstack{96.9\\{\scriptsize 96.5, 96.9, 97.2}} & \shortstack{\underline{80.7}\\{\scriptsize 81.5, 79.8, 80.8}} & \shortstack{79.2\\{\scriptsize 78.8, 80.3, 78.5}} & \shortstack{64.8\\{\scriptsize 80.8, 74.3, 39.4}} \\[3pt]
HM all-$A$ real & \shortstack{100.0\\{\scriptsize 100.0, 100.0, 100.0}} & \shortstack{91.2\\{\scriptsize 90.3, 92.9, 90.3}} & \shortstack{\underline{79.2}\\{\scriptsize 78.2, 80.1, 79.4}} & \shortstack{73.9\\{\scriptsize 62.1, 80.0, 79.5}} & \shortstack{43.2\\{\scriptsize 74.8, 26.9, 27.8}} \\[3pt]
HM all-$A$ complex & \shortstack{100.0\\{\scriptsize 100.0, 100.0, 100.0}} & \shortstack{92.3\\{\scriptsize 89.5, 96.9, 90.5}} & \shortstack{66.0\\{\scriptsize 79.6, 74.6, 43.8}} & \shortstack{\underline{69.0}\\{\scriptsize 61.4, 74.6, 71.0}} & \shortstack{47.8\\{\scriptsize 71.0, 20.8, 51.7}} \\[3pt]
\midrule
\multicolumn{6}{@{}l}{Extra hard-tier rate $10^{-4}$: outside the common selection grid} \\
M3 & \multicolumn{5}{c}{\shortstack{29.7\\{\scriptsize 11.2, 3.4, 74.4}}} \\[3pt]
Attention & \multicolumn{5}{c}{\shortstack{2.9\\{\scriptsize 0.5, 7.7, 0.5}}} \\[3pt]
\end{longtable}
\endgroup

\par\endgroup

\subsection{FineWeb-Edu pretraining}
\label{app:fineweb_protocol}
\par\begingroup

FineWeb-Edu's 100-billion-token sample~\citep{penedo2024fineweb} is tokenized
with the Llama-3.1 tokenizer (vocabulary $128{,}256$, a beginning-of-sequence
token prepended and no end-of-sequence token). Shards 1--138 of 140
supply training, shard 139 supplies validation and shard 140 supplies the
held-out test set. Context length is 2048 without crossing shard boundaries;
training uses a seed-0 permutation per epoch. The backbone has tied
embeddings and pre-normalized mixer and Swish-gated linear unit (SwiGLU) branches, with RMSNorm
$\epsilon=10^{-5}$, residual width $d_{\mathrm{model}}=1024$, driving width
$d=2048$ and SSM state width $N=128$. Embedding initialization has standard
deviation $0.02$; residual output weights are rescaled by
$1/\sqrt{2n_{\mathrm{layers}}}$, where $n_{\mathrm{layers}}$ is the number
of backbone layers. The parameter match adjusts depth first and trims
multilayer perceptron (MLP) width by at most $5\%$ only when necessary; attention, whose mixer has fewer parameters,
instead widens its MLP to $2861$.

Training uses fused AdamW with $\beta=(0.9,0.95)$, $\epsilon=10^{-8}$,
gradient clipping at 1 and weight decay $0.1$ on linear, embedding and
convolution weights. Biases, gains, static state parameters and sibling
mixing weights are exempt; rate-producing linear weights are decayed.
The global batch is $491{,}520$ tokens: eight GPUs, six sequences per
micro-batch and five accumulation steps. Mixed precision uses bf16 with
fp32 master weights and optimizer state. The learning rate warms up over
$0.5$B tokens to $1.5\times10^{-3}$ and stays constant through $2$B.
The $1$B/$2$B milestones are preserved along the same trajectory, without
resetting optimizer or data position, at updates 2035 and 4070.
Mamba~3 SISO/MIMO, SSD-MIMO R2O2-c/r and HM-75 continue from the $2$B checkpoint
to update 20346 (approximately $10$B tokens), with cosine decay to
$10^{-5}$. These later checkpoints therefore use an annealed schedule,
unlike the constant-LR $2$B comparison.

Validation uses $50\times6\times2048$ tokens approximately every $100$M
training tokens. Milestone test loss uses $200\times6\times2048$ tokens
from the separate test shard. All reported arms use seed 0 on eight GH200 GPUs;
the Mamba and attention references run without compilation or CUDA graphs for the $1$B/$2$B trunk;
the upstream determinism option was enabled for them, but bitwise determinism was
not verified, and our arms are not bit-deterministic (their backward uses atomic reductions). No speed comparison is made.

\paragraph{Reported mixer configurations.}
Each configuration's identifier C$x$ is given where it is specified below and in Table~\ref{tab:lm_final_geometry}; the unused identifiers C6b, C8 and C10 belong to configurations that were not run.
Table~\ref{tab:lm_final_geometry} gives the depth and MLP-width match.
These are the configurations used in the reported runs. All SSMs have $N=128$; headed SSMs have 32 heads
of driving width 64. Unless noted, our models use two-endpoint writes,
RMS-normalized write/read maps, a SiLU output gate and no output RMSNorm.
Initial steps follow Appendix~\ref{app:mqmkvar_analysis};
linear maps use uniform fan-in initialization and the output projections
use the shared residual rescaling.

\begin{table}[!htbp]
\centering\small
\caption{Language-model geometry and total parameter counts. Mamba~3 SISO is the anchor; configuration details are given in the text.}
\label{tab:lm_final_geometry}
\setlength{\tabcolsep}{3pt}
\adjustbox{max width=\linewidth}{%
\begin{tabular}{@{}lrrr@{}}
\toprule
Model & Layers & MLP width & Parameters (M) \\
\midrule
Mamba~1 (C1) & 28 & 1983 & 507.88 \\
Mamba~2 (C2) & 29 & 2048 & 505.25 \\
Mamba~3 SISO (C3) & 29 & 2048 & 507.95 \\
Mamba~3 MIMO (C4) & 27 & 2048 & 504.54 \\
Attention (C5) & 29 & 2861 & 507.91 \\
\addlinespace[2pt]
SSD-MIMO R1O1-r (C11) & 29 & 1958 & 507.89 \\
SSD-MIMO R2O2-r (C7) & 27 & 2048 & 506.03 \\
SSD-MIMO R2O2-c (C6) & 28 & 1983 & 507.93 \\
SSD-MIMO R4O4-r (C9) & 25 & 2016 & 507.87 \\
SSD-MIMO M3exact (C12) & 29 & 2048 & 507.95 \\
SSD-MIMO R1O1-r (M3 settings) (C14) & 29 & 2048 & 507.00 \\
HM-0 (C13) & 28 & 2024 & 507.87 \\
HM-25 (C15) & 28 & 2024 & 507.88 \\
HM-75 (C16) & 28 & 2024 & 507.91 \\
HM-100 (C17) & 28 & 2024 & 507.92 \\
\bottomrule
\end{tabular}}
\end{table}

\paragraph{Mamba references and attention.}
Mamba~1 (C1) uses a rank-64 step projection and width-four drive
convolution; its coefficients read the processed drive. Mamba~2 (C2)
uses a width-four joint drive/B/C convolution, one endpoint-sharing group,
static scalar rates per head and gated output RMSNorm. Mamba~3 SISO (C3)
is convolution-free, with one endpoint-sharing group, normalized and biased
maps, two-endpoint writes and no output RMSNorm. Half its coordinate pairs
rotate with fixed frequencies $\pi$; the selected rate uses the heavy-tail
activation, floored at $10^{-4}$. Mamba~3 MIMO (C4) adds its authors'
rank-four lift. These references retain their upstream initializations.
Attention (C5) has 16 heads of width 64, causal scaled dot-product attention,
RoPE base $5\times10^5$, bias-free projections, no grouped-query sharing
and no dropout.

\paragraph{SSD-MIMO configurations.}
R2O2-c (C6) has $R=O=2$ and 32 value channels per head/write sibling,
with half real and half complex rows. Coefficients read the mixer input ($s_t=x_t$);
only the drive passes through a width-four convolution and SiLU.
A heavy-tail rate is selected per head, floored at $10^{-4}$.
Only real rows have coordinate gates. The endpoint maps have per-head/sibling
biases, initialized to ones for the first sibling and discrete-cosine
spread for the others; gate features initialize to zero. Static and selected
sibling weights combine the readouts, starting at $1/R$ with zero selected
projection. Complex rows use the default learned rotation of
Appendix~\ref{app:mqmkvar_analysis}. The skip starts at one.
R2O2-r (C7) makes all rows real and gated. R4O4-r (C9) uses $R=O=4$
and 16 value channels per sibling. R1O1-r (C11) uses $R=O=1$ and 64
value channels per head, retaining the real-row gates and static/selected
sibling scale, with all-ones endpoint biases.

M3exact (C12) uses $R=O=1$ and Mamba~3 SISO's coefficient configuration:
no convolution, coordinate gates or sibling mixing, fixed frequencies
$\pi$, no read offset or rate/endpoint projection bias, per-head skip
and no output normalization. Endpoint RMSNorm uses $\epsilon=10^{-5}$;
its gains, biases and skips start at one. The rate law, step initialization
and weight-decay groups match Mamba~3 SISO; random draws, scan arithmetic and
projection grouping can differ. R1O1-r with Mamba~3 settings (C14) removes rotation and the angle
projection from M3exact, keeping the other settings; it differs from R1O1-r in using Mamba~3's coefficient settings
instead of ours (no convolution, gates or sibling weighting) and, through the parameter match, in MLP
width ($2048$ versus $1958$). At $1$B tokens the two real rank-one runs reach about the same loss ($3.2462$
versus $3.2473$ for R1O1-r, one seed each), while rotation (M3exact) lowers it by $0.013$ nats.
Against R1O1-r with Mamba~3 settings, M3exact's paired differences are $+0.7\,[+0.0,+1.5]$ ZS ($p=0.049$),
$+1.0\,[+0.5,+1.6]$ Recall-6 and $-3.5\,[-5.2,-1.8]$ NIAH points (one comparison, so
unadjusted), all three significant, zero-shot only marginally: rotation helps cloze recall but not NIAH at $1$B.

\paragraph{HybridMamba configurations.}
HM-0 (C13) uses 64 real static and 64 real dynamic rows,
without coordinate gates or rotation. Coefficients read the drive after its width-four convolution and SiLU ($s_t=u_t$); normalized endpoints have per-head
biases, and the skip is per head. Static rates start at $a_n=-n$ within
the block; dynamic rows use the heavy-tail activation without a floor.
Steps use a full-rank projection. Endpoint biases start at zero, and
normalization gains and skips at one. HM-25 (C15) replaces the allocation by
$25\%$ complex dynamic, $37.5\%$ real static and $37.5\%$ real dynamic
rows; the complex rows add the learned frequencies and read offset
specified for R2O2-c. Other settings follow HM-0.
HM-75 (C16) uses $75\%$ complex dynamic rows (48 pairs), $12.5\%$ real static and $12.5\%$ real
dynamic rows (16 each; $507.91$M parameters). HM-100 (C17) makes all 128 rows complex dynamic (64 pairs), with no static or real rows
($507.92$M parameters). Their complex rows, rotation law and all other settings are as for HM-25; the four HM arms
differ in parameter count by $0.02\%$ or less. HM-75 was trained on the earlier code up to $2$B tokens and, for the
$2$--$10$B continuation, on the later code of the merged kernels; on a fixed batch the two
codes differ in loss by $8\times10^{-5}$ nats. HM-100 was trained on the later code, so its comparison with HM-75 at $2$B carries
the same offset, negligible against their $0.006$ loss difference.

\subsection{Language-model evaluation breakdowns}
\label{app:lm_final_breakdowns}

Each arm has one training seed. Absolute means are item-level macro means
rounded half-up to one decimal; paired differences use the unrounded
item-level scores and can differ from subtracting the displayed means.
Unreported milestones are marked
with a dash, rather than filled from other checkpoints.

Tables~\ref{tab:lm_milestone_1b}--\ref{tab:lm_milestone_10b} give, per
milestone, loss/perplexity and zero-shot scores (for configurations not in
Table~\ref{tab:lm_final_main}), all cloze tasks and the twelve NIAH cells.

\paragraph{Zero-shot evaluation.}
The Language Model Evaluation Harness~\citep{gao2024harness} scores
LAMBADA~\citep{paperno2016lambada}, HellaSwag~\citep{zellers2019hellaswag},
PIQA~\citep{bisk2019piqa}, ARC-Easy/Challenge~\mbox{\citep{clark2018arc}},
WinoGrande~\citep{sakaguchi2019winogrande} and
OpenBookQA~\mbox{\citep{mihaylov2018openbook}}.
HellaSwag and ARC-Challenge use normalized accuracy; the other tasks use
plain accuracy. ZS weights the seven task scores equally.

\paragraph{Cloze recall.}
Recall-6 is the equally weighted mean of SWDE, FDA, SQuAD, TriviaQA,
NQ-1024 and DROP. All six use Hazy Research's cloze versions, as in
Mamba~3 Table~4~\citep{lahoti2026mamba3}, rather than the original
question-answering evaluation formats. This follows the recall-focused
protocols of Based and Just Read Twice~\citep{arora2024simple,arora2024jrt};
SWDE, SQuAD and FDA originate in web extraction, reading comprehension
and EVAPORATE's extraction collection
\citep{hao2011swde,rajpurkar2016squad,arora2023evaporate}.
Scores depend slightly on the evaluation batch size through bf16 rounding, but no item's score depends on the content of the other items in its batch;
truncating to 2K tokens loses the answer on $2.0\%$ of NQ items and on no
TriviaQA or DROP item.

\paragraph{NIAH diagnostic.}
NIAH (RULER-style, our implementation)~\citep{hsieh2024ruler} uses
\texttt{single\_1}, \texttt{single\_2}, \texttt{single\_3} and
\texttt{multikey\_1} at 512, 1024 and 2048 tokens, with 100 items per cell.
The diagnostic mean weights all twelve cells equally. Generation uses at most 64 tokens;
an answer is scored after skipping a leading colon or whitespace prefix, up to the first newline,
and counts as correct if it contains the needle's value.
Without the prefix skip, outputs of the form ``:'' + newline + answer, which occur in every arm
($7.4\%$ of items, mostly \texttt{single\_1}), would score zero. We regenerated these items with each
checkpoint's training code; the old rule applied to the regenerated outputs reproduces $2{,}319$ of
the $2{,}320$ stored scores, and all non-NIAH scores are unchanged.

\begin{table}[p]
\centering\scriptsize
\setlength{\tabcolsep}{3pt}
\caption{All reported configurations at $1$B training tokens, one seed per arm. Test loss is in nats; PPL is perplexity; task scores are percentages. HellaSwag and ARC-C use normalized accuracy. MK and S1--S3 denote multikey\_1 and single\_1--3; NIAH averages twelve cells. Protocols: Appendix~\ref{app:lm_final_breakdowns}.}
\label{tab:lm_milestone_1b}
\textit{(a) Test loss / perplexity and zero-shot accuracy}\par\smallskip
\adjustbox{max width=\linewidth}{%
\begin{tabular}{@{}lrrrrrrrrr@{}}
\toprule
Model & Loss / PPL & LAMB. & Hella. & PIQA & ARC-E & ARC-C & Wino. & OBQA & Avg. \\
\midrule
Mamba~1 & 3.2962 / 27.01 & 22.5 & 31.3 & 63.2 & 48.4 & 23.9 & 52.8 & 19.0 & 37.3 \\
Mamba~2 & 3.2715 / 26.35 & 19.6 & 31.6 & 63.0 & 50.1 & 24.4 & 50.0 & 17.4 & 36.6 \\
Mamba~3 SISO & 3.2367 / 25.45 & 23.4 & 31.4 & 62.7 & 51.3 & 24.9 & 51.5 & 19.6 & 37.8 \\
Mamba~3 MIMO & 3.2317 / 25.32 & 21.2 & 31.8 & 61.9 & 51.4 & 25.3 & 49.4 & 20.4 & 37.3 \\
Attention & 3.3457 / 28.38 & 20.1 & 29.5 & 60.7 & 46.3 & 23.4 & 49.1 & 17.4 & 35.2 \\
\addlinespace[2pt]
SSD-MIMO R1O1-r & 3.2473 / 25.72 & 22.5 & 31.4 & 61.9 & 50.8 & 25.2 & 49.7 & 19.8 & 37.3 \\
SSD-MIMO R2O2-r & 3.2569 / 25.97 & 23.6 & 30.8 & 62.5 & 51.0 & 24.1 & 49.2 & 19.0 & 37.2 \\
SSD-MIMO R2O2-c & 3.2524 / 25.85 & 22.3 & 31.4 & 61.0 & 51.4 & 26.8 & 51.9 & 20.4 & 37.9 \\
SSD-MIMO R4O4-r & 3.2777 / 26.51 & 20.6 & 30.7 & 61.5 & 50.5 & 23.9 & 50.7 & 18.4 & 36.6 \\
SSD-MIMO M3exact & 3.2335 / 25.37 & 22.7 & 31.5 & 63.2 & 51.3 & 23.7 & 52.4 & 18.6 & 37.6 \\
SSD-MIMO R1O1-r (M3 settings) & 3.2462 / 25.69 & 22.8 & 31.5 & 62.1 & 50.3 & 24.1 & 48.8 & 18.8 & 36.9 \\
HM-0 & 3.2504 / 25.80 & 22.7 & 31.7 & 61.6 & 51.3 & 25.5 & 50.5 & 19.4 & 37.5 \\
HM-25 & 3.2485 / 25.75 & 22.2 & 31.9 & 61.6 & 52.1 & 23.9 & 51.5 & 18.2 & 37.3 \\
HM-75 & 3.2418 / 25.58 & 23.3 & 31.6 & 62.5 & 51.3 & 23.8 & 51.5 & 16.6 & 37.2 \\
HM-100 & 3.2511 / 25.82 & 25.5 & 31.6 & 62.1 & 50.9 & 24.7 & 51.4 & 19.8 & 38.0 \\
\bottomrule
\end{tabular}}

\medskip
\textit{(b) Cloze recall}\par\smallskip
\adjustbox{max width=\linewidth}{%
\begin{tabular}{@{}lrrrrrrr@{}}
\toprule
Model & SWDE & FDA & SQuAD & TriviaQA & NQ-1024 & DROP & Avg. \\
\midrule
Mamba~1 & 4.7 & 0.6 & 12.2 & 28.6 & 5.3 & 10.2 & 10.3 \\
Mamba~2 & 4.3 & 1.3 & 13.7 & 31.6 & 6.1 & 10.1 & 11.2 \\
Mamba~3 SISO & 6.8 & 1.4 & 17.7 & 34.5 & 7.5 & 15.5 & 13.9 \\
Mamba~3 MIMO & 7.8 & 2.1 & 17.1 & 35.4 & 8.5 & 14.2 & 14.2 \\
Attention & 21.5 & 10.0 & 20.6 & 33.0 & 8.8 & 14.3 & 18.0 \\
\addlinespace[2pt]
SSD-MIMO R1O1-r & 6.5 & 1.9 & 18.6 & 33.2 & 7.7 & 14.0 & 13.7 \\
SSD-MIMO R2O2-r & 9.2 & 1.6 & 19.9 & 35.5 & 9.8 & 15.3 & 15.2 \\
SSD-MIMO R2O2-c & 8.1 & 2.5 & 17.2 & 35.7 & 8.5 & 12.7 & 14.1 \\
SSD-MIMO R4O4-r & 8.0 & 1.5 & 19.8 & 36.2 & 9.5 & 17.0 & 15.3 \\
SSD-MIMO M3exact & 6.8 & 1.8 & 17.5 & 36.0 & 8.7 & 15.5 & 14.4 \\
SSD-MIMO R1O1-r (M3 settings) & 6.6 & 1.5 & 18.6 & 33.7 & 7.9 & 12.1 & 13.4 \\
HM-0 & 6.2 & 0.8 & 16.4 & 32.2 & 6.4 & 15.2 & 12.9 \\
HM-25 & 6.4 & 2.0 & 18.3 & 35.4 & 6.9 & 13.6 & 13.8 \\
HM-75 & 6.7 & 2.0 & 17.9 & 35.3 & 7.0 & 15.5 & 14.1 \\
HM-100 & 6.9 & 1.6 & 18.9 & 36.2 & 6.8 & 14.2 & 14.1 \\
\bottomrule
\end{tabular}}

\medskip
\textit{(c) NIAH (RULER-style, our implementation)}\par\smallskip
\adjustbox{max width=\linewidth}{%
\begin{tabular}{@{}lrrrrrrrrrrrrr@{}}
\toprule
 & \multicolumn{4}{c}{512 tokens} & \multicolumn{4}{c}{1024 tokens} & \multicolumn{4}{c}{2048 tokens} & \\
Model & MK & S1 & S2 & S3 & MK & S1 & S2 & S3 & MK & S1 & S2 & S3 & Avg. \\
\midrule
Mamba~1 & 5.0 & 5.0 & 0.0 & 0.0 & 1.0 & 0.0 & 0.0 & 0.0 & 0.0 & 0.0 & 0.0 & 0.0 & 0.9 \\
Mamba~2 & 4.0 & 95.0 & 21.0 & 0.0 & 7.0 & 96.0 & 7.0 & 0.0 & 4.0 & 61.0 & 2.0 & 0.0 & 24.8 \\
Mamba~3 SISO & 12.0 & 98.0 & 49.0 & 1.0 & 7.0 & 82.0 & 10.0 & 0.0 & 3.0 & 17.0 & 7.0 & 0.0 & 23.8 \\
Mamba~3 MIMO & 10.0 & 99.0 & 60.0 & 1.0 & 7.0 & 90.0 & 17.0 & 0.0 & 3.0 & 41.0 & 5.0 & 0.0 & 27.8 \\
Attention & 20.0 & 98.0 & 91.0 & 53.0 & 16.0 & 97.0 & 67.0 & 33.0 & 17.0 & 89.0 & 36.0 & 11.0 & 52.3 \\
\addlinespace[2pt]
SSD-MIMO R1O1-r & 13.0 & 100.0 & 65.0 & 0.0 & 9.0 & 100.0 & 23.0 & 1.0 & 7.0 & 77.0 & 8.0 & 0.0 & 33.6 \\
SSD-MIMO R2O2-r & 15.0 & 100.0 & 55.0 & 1.0 & 10.0 & 100.0 & 10.0 & 0.0 & 8.0 & 84.0 & 8.0 & 0.0 & 32.6 \\
SSD-MIMO R2O2-c & 14.0 & 99.0 & 70.0 & 0.0 & 9.0 & 98.0 & 16.0 & 0.0 & 8.0 & 91.0 & 8.0 & 0.0 & 34.4 \\
SSD-MIMO R4O4-r & 14.0 & 99.0 & 79.0 & 5.0 & 12.0 & 99.0 & 21.0 & 0.0 & 7.0 & 93.0 & 8.0 & 0.0 & 36.4 \\
SSD-MIMO M3exact & 14.0 & 98.0 & 58.0 & 0.0 & 10.0 & 90.0 & 12.0 & 0.0 & 6.0 & 39.0 & 7.0 & 0.0 & 27.8 \\
SSD-MIMO R1O1-r (M3 settings) & 18.0 & 100.0 & 55.0 & 0.0 & 9.0 & 99.0 & 13.0 & 0.0 & 5.0 & 73.0 & 4.0 & 0.0 & 31.3 \\
HM-0 & 9.0 & 64.0 & 46.0 & 1.0 & 7.0 & 23.0 & 15.0 & 0.0 & 5.0 & 8.0 & 7.0 & 0.0 & 15.4 \\
HM-25 & 12.0 & 53.0 & 49.0 & 3.0 & 6.0 & 27.0 & 10.0 & 3.0 & 5.0 & 7.0 & 5.0 & 0.0 & 15.0 \\
HM-75 & 14.0 & 98.0 & 52.0 & 2.0 & 11.0 & 63.0 & 5.0 & 1.0 & 5.0 & 21.0 & 5.0 & 0.0 & 23.1 \\
HM-100 & 11.0 & 15.0 & 9.0 & 1.0 & 4.0 & 5.0 & 3.0 & 0.0 & 1.0 & 1.0 & 0.0 & 0.0 & 4.2 \\
\bottomrule
\end{tabular}}

\end{table}

\begin{table}[p]
\centering\scriptsize
\setlength{\tabcolsep}{3pt}
\caption{As Table~\ref{tab:lm_milestone_1b}, at $2$B training tokens.}
\label{tab:lm_milestone_2b}
\textit{(a) Test loss / perplexity and zero-shot accuracy, configurations not in Table~\ref{tab:lm_final_main}}\par\smallskip
\adjustbox{max width=\linewidth}{%
\begin{tabular}{@{}lrrrrrrrrr@{}}
\toprule
Model & Loss / PPL & LAMB. & Hella. & PIQA & ARC-E & ARC-C & Wino. & OBQA & Avg. \\
\midrule
SSD-MIMO R4O4-r & 3.0913 / 22.01 & 23.2 & 34.1 & 62.9 & 53.5 & 25.5 & 50.5 & 20.2 & 38.6 \\
SSD-MIMO R1O1-r (M3 settings) & 3.0681 / 21.50 & 24.7 & 35.0 & 64.2 & 54.7 & 25.9 & 51.8 & 20.0 & 39.5 \\
\bottomrule
\end{tabular}}

\medskip
\textit{(b) Cloze recall}\par\smallskip
\adjustbox{max width=\linewidth}{%
\begin{tabular}{@{}lrrrrrrr@{}}
\toprule
Model & SWDE & FDA & SQuAD & TriviaQA & NQ-1024 & DROP & Avg. \\
\midrule
Mamba~1 & 9.0 & 1.5 & 18.1 & 36.0 & 6.8 & 17.2 & 14.8 \\
Mamba~2 & 8.3 & 2.4 & 21.1 & 37.6 & 8.6 & 17.2 & 15.8 \\
Mamba~3 SISO & 12.6 & 2.5 & 23.3 & 36.3 & 8.5 & 17.4 & 16.8 \\
Mamba~3 MIMO & 13.5 & 2.3 & 21.8 & 37.3 & 10.7 & 15.3 & 16.8 \\
Attention & 37.2 & 20.9 & 21.6 & 39.5 & 13.5 & 17.3 & 25.0 \\
\addlinespace[2pt]
SSD-MIMO R2O2-r & 16.8 & 3.9 & 22.7 & 38.5 & 10.7 & 17.3 & 18.3 \\
SSD-MIMO R2O2-c & 11.7 & 1.9 & 23.7 & 38.4 & 10.1 & 16.0 & 17.0 \\
SSD-MIMO R4O4-r & 11.8 & 3.4 & 23.5 & 39.0 & 10.6 & 17.8 & 17.7 \\
SSD-MIMO R1O1-r (M3 settings) & 10.6 & 2.2 & 23.2 & 37.5 & 8.1 & 15.0 & 16.1 \\
HM-0 & 8.9 & 2.0 & 23.9 & 37.7 & 7.5 & 17.1 & 16.2 \\
HM-25 & 9.6 & 3.1 & 23.4 & 39.1 & 8.2 & 18.5 & 17.0 \\
HM-75 & 11.7 & 3.8 & 22.2 & 38.3 & 8.3 & 17.1 & 16.9 \\
HM-100 & 10.5 & 3.5 & 22.4 & 39.8 & 7.6 & 17.7 & 16.9 \\
\bottomrule
\end{tabular}}

\medskip
\textit{(c) NIAH (RULER-style, our implementation)}\par\smallskip
\adjustbox{max width=\linewidth}{%
\begin{tabular}{@{}lrrrrrrrrrrrrr@{}}
\toprule
 & \multicolumn{4}{c}{512 tokens} & \multicolumn{4}{c}{1024 tokens} & \multicolumn{4}{c}{2048 tokens} & \\
Model & MK & S1 & S2 & S3 & MK & S1 & S2 & S3 & MK & S1 & S2 & S3 & Avg. \\
\midrule
Mamba~1 & 11.0 & 85.0 & 45.0 & 0.0 & 12.0 & 42.0 & 13.0 & 0.0 & 8.0 & 10.0 & 7.0 & 0.0 & 19.4 \\
Mamba~2 & 20.0 & 100.0 & 71.0 & 4.0 & 9.0 & 100.0 & 28.0 & 1.0 & 7.0 & 78.0 & 11.0 & 0.0 & 35.8 \\
Mamba~3 SISO & 17.0 & 97.0 & 32.0 & 5.0 & 9.0 & 90.0 & 6.0 & 3.0 & 4.0 & 39.0 & 3.0 & 2.0 & 25.6 \\
Mamba~3 MIMO & 18.0 & 95.0 & 56.0 & 1.0 & 8.0 & 80.0 & 18.0 & 0.0 & 6.0 & 45.0 & 2.0 & 0.0 & 27.4 \\
Attention & 37.0 & 98.0 & 85.0 & 62.0 & 24.0 & 88.0 & 65.0 & 45.0 & 23.0 & 97.0 & 41.0 & 16.0 & 56.8 \\
\addlinespace[2pt]
SSD-MIMO R2O2-r & 18.0 & 100.0 & 79.0 & 15.0 & 16.0 & 100.0 & 19.0 & 6.0 & 9.0 & 97.0 & 8.0 & 1.0 & 39.0 \\
SSD-MIMO R2O2-c & 22.0 & 95.0 & 81.0 & 18.0 & 12.0 & 90.0 & 23.0 & 6.0 & 9.0 & 73.0 & 11.0 & 2.0 & 36.8 \\
SSD-MIMO R4O4-r & 20.0 & 100.0 & 71.0 & 38.0 & 7.0 & 100.0 & 12.0 & 6.0 & 4.0 & 61.0 & 6.0 & 3.0 & 35.7 \\
SSD-MIMO R1O1-r (M3 settings) & 25.0 & 100.0 & 78.0 & 3.0 & 16.0 & 99.0 & 29.0 & 0.0 & 7.0 & 70.0 & 12.0 & 0.0 & 36.6 \\
HM-0 & 20.0 & 84.0 & 65.0 & 5.0 & 12.0 & 31.0 & 26.0 & 2.0 & 11.0 & 6.0 & 7.0 & 2.0 & 22.6 \\
HM-25 & 25.0 & 37.0 & 55.0 & 3.0 & 14.0 & 16.0 & 16.0 & 3.0 & 8.0 & 7.0 & 9.0 & 3.0 & 16.3 \\
HM-75 & 25.0 & 100.0 & 84.0 & 19.0 & 13.0 & 79.0 & 31.0 & 7.0 & 9.0 & 25.0 & 10.0 & 2.0 & 33.7 \\
HM-100 & 16.0 & 82.0 & 39.0 & 7.0 & 10.0 & 36.0 & 9.0 & 3.0 & 9.0 & 7.0 & 4.0 & 0.0 & 18.5 \\
\bottomrule
\end{tabular}}

\end{table}

\begin{table}[p]
\centering\scriptsize
\setlength{\tabcolsep}{3pt}
\caption{As Table~\ref{tab:lm_milestone_1b}, at $10$B training tokens. Loss and zero-shot scores are in Table~\ref{tab:lm_final_main}.}
\label{tab:lm_milestone_10b}
\textit{(a) Cloze recall}\par\smallskip
\adjustbox{max width=\linewidth}{%
\begin{tabular}{@{}lrrrrrrr@{}}
\toprule
Model & SWDE & FDA & SQuAD & TriviaQA & NQ-1024 & DROP & Avg. \\
\midrule
Mamba~3 SISO & 24.4 & 15.6 & 33.4 & 53.1 & 14.9 & 22.7 & 27.3 \\
Mamba~3 MIMO & 26.6 & 12.2 & 35.7 & 52.0 & 16.6 & 24.4 & 27.9 \\
\addlinespace[2pt]
SSD-MIMO R2O2-r & 26.6 & 11.4 & 33.7 & 52.3 & 16.8 & 23.2 & 27.3 \\
SSD-MIMO R2O2-c & 29.9 & 11.6 & 33.7 & 52.3 & 16.5 & 24.3 & 28.0 \\
HM-75 & 27.9 & 6.0 & 36.2 & 52.9 & 16.5 & 25.9 & 27.6 \\
\bottomrule
\end{tabular}}

\medskip
\textit{(b) NIAH (RULER-style, our implementation)}\par\smallskip
\adjustbox{max width=\linewidth}{%
\begin{tabular}{@{}lrrrrrrrrrrrrr@{}}
\toprule
 & \multicolumn{4}{c}{512 tokens} & \multicolumn{4}{c}{1024 tokens} & \multicolumn{4}{c}{2048 tokens} & \\
Model & MK & S1 & S2 & S3 & MK & S1 & S2 & S3 & MK & S1 & S2 & S3 & Avg. \\
\midrule
Mamba~3 SISO & 20.0 & 100.0 & 66.0 & 28.0 & 9.0 & 99.0 & 35.0 & 14.0 & 6.0 & 73.0 & 10.0 & 5.0 & 38.8 \\
Mamba~3 MIMO & 33.0 & 100.0 & 82.0 & 44.0 & 18.0 & 100.0 & 43.0 & 19.0 & 8.0 & 100.0 & 12.0 & 3.0 & 46.8 \\
\addlinespace[2pt]
SSD-MIMO R2O2-r & 9.0 & 100.0 & 48.0 & 21.0 & 15.0 & 100.0 & 20.0 & 11.0 & 8.0 & 97.0 & 8.0 & 1.0 & 36.5 \\
SSD-MIMO R2O2-c & 13.0 & 100.0 & 71.0 & 34.0 & 14.0 & 100.0 & 41.0 & 10.0 & 9.0 & 100.0 & 18.0 & 3.0 & 42.8 \\
HM-75 & 30.0 & 100.0 & 80.0 & 50.0 & 20.0 & 94.0 & 19.0 & 15.0 & 7.0 & 29.0 & 8.0 & 1.0 & 37.8 \\
\bottomrule
\end{tabular}}

\end{table}

\FloatBarrier

\paragraph{HM complex fraction.}
Table~\ref{tab:lm_complex_fraction} varies only the fraction of complex dynamic rows in the HM arm. $75\%$ has the
lowest loss and the highest NIAH at both milestones (one seed each); zero-shot and recall means
differ by at most $1.2$ points, significantly only for HM-75's $1$B Recall-6 lead over HM-0. All-complex (HM-100), which removes the static and the real dynamic rows together,
has a lower NIAH than HM-75.
\begin{table}[!htbp]
\centering\footnotesize
\caption{HybridMamba arms HM-0 to HM-100 that differ only in the fraction of state rows that are complex dynamic rows (all arms: $507.9$M parameters, 28 layers, MLP width 2024; the remaining rows are static-$A$ and real dynamic rows, except at $100\%$). Test loss (nats) and mean scores (percent), one seed per arm, followed by paired differences in the three means. $^{*}$: Holm-adjusted $p<0.05$ within the original source family and measure (seven comparisons in the HM-75 sources, three in the HM-100 sources); intervals cover evaluation items, not training seeds. }
\label{tab:lm_complex_fraction}
\setlength{\tabcolsep}{3pt}
\adjustbox{max width=\linewidth}{%
\begin{tabular}{@{}lrrrrrrrr@{}}
\toprule
 & \multicolumn{4}{c}{1B tokens} & \multicolumn{4}{c}{2B tokens} \\
\cmidrule(lr){2-5}\cmidrule(l){6-9}
Model & Loss & ZS & Recall-6 & NIAH & Loss & ZS & Recall-6 & NIAH \\
\midrule
HM-0 (0\% complex) & 3.2504 & 37.5 & 12.9 & 15.4 & 3.0745 & 39.9 & 16.2 & 22.6 \\
HM-25 (25\% complex) & 3.2485 & 37.3 & 13.8 & 15.0 & 3.0701 & 39.8 & 17.0 & 16.3 \\
HM-75 (75\% complex) & 3.2418 & 37.2 & 14.1 & 23.1 & 3.0639 & 39.7 & 16.9 & 33.7 \\
HM-100 (100\% complex) & 3.2511 & 38.0 & 14.1 & 4.2 & 3.0698 & 39.6 & 16.9 & 18.5 \\
\midrule
\multicolumn{9}{@{}l}{\emph{Paired differences (percentage points, $95\%$ item-bootstrap interval)}} \\
HM-75 $-$ HM-0 & --- & $-0.3\,[-1.1,+0.5]$ & $+1.2\,[+0.6,+1.7]^{*}$ & $+7.7\,[+5.8,+9.5]^{*}$ & --- & $-0.2\,[-0.9,+0.5]$ & $+0.7\,[+0.1,+1.3]$ & $+11.1\,[+9.0,+13.2]^{*}$ \\
HM-75 $-$ HM-25 & --- & $-0.1\,[-0.9,+0.6]$ & $+0.3\,[-0.3,+0.9]$ & $+8.1\,[+6.2,+9.9]^{*}$ & --- & $-0.1\,[-0.9,+0.6]$ & $-0.1\,[-0.7,+0.5]$ & $+17.3\,[+15.2,+19.5]^{*}$ \\
HM-100 $-$ HM-75 & --- & $+0.8\,[+0.0,+1.5]$ & $+0.1\,[-0.5,+0.6]$ & $-18.9\,[-20.7,-17.1]^{*}$ & --- & $-0.1\,[-0.8,+0.7]$ & $+0.0\,[-0.6,+0.7]$ & $-15.2\,[-17.3,-13.1]^{*}$ \\
HM-75 $-$ M3 SISO & --- & $-0.6\,[-1.3,+0.1]$ & $+0.2\,[-0.4,+0.7]$ & $-0.7\,[-2.4,+0.9]$ & --- & $-0.2\,[-0.9,+0.6]$ & $+0.1\,[-0.5,+0.7]$ & $+8.1\,[+6.1,+10.2]^{*}$ \\
HM-100 $-$ M3 SISO & --- & $+0.2\,[-0.6,+0.9]$ & $+0.2\,[-0.3,+0.8]$ & $-19.7\,[-21.4,-18.0]^{*}$ & --- & $-0.3\,[-1.0,+0.5]$ & $+0.2\,[-0.5,+0.8]$ & $-7.1\,[-9.1,-5.1]^{*}$ \\
\bottomrule
\end{tabular}}
\end{table}

\FloatBarrier

\paragraph{Paired comparisons.}
The table reports selected paired differences and $95\%$ item-bootstrap intervals,
using 20,000 bootstrap replicates and resampling seed 11. Holm adjustment
is applied across all comparisons for each measure within its source
table: ten at $1$B/$2$B, five at $10$B, two for the separate R1O1-r/M3exact
comparison, one for M3exact against R1O1-r with Mamba~3 settings, and four for the attention comparisons at each
of $1$B and $2$B, seven in each HM-75 source at $1$B/$2$B, four in the HM-75 source at $10$B and three in each HM-100 source. Grouping the contrasts by milestone does not change those families.
Only comparisons discussed in Sections~\ref{sec:hybridmamba}
and~\ref{sec:experiments} appear below. The intervals are not simultaneous
Holm intervals, and no significance test is attached to a test-loss
difference.
\begin{table}[!htbp]
\centering\footnotesize
\caption{Paired comparisons discussed in the text: differences in percentage points with $95\%$ item-bootstrap intervals. $^{*}$: Holm-adjusted $p<0.05$ within the original source family and measure, including comparisons omitted here. Intervals cover evaluation items, not training seeds. All recall contrasts use Recall-6. M1/M2: Mamba~1/2; M3 SISO/MIMO: Mamba~3 SISO/MIMO; R2O2-c/r: SSD-MIMO R2O2-c/r; M3exact: SSD-MIMO\'s Mamba~3 SISO preset; R1O1-r (M3): M3exact without rotation; HM-$x$: HybridMamba with $x\%$ complex dynamic rows. Bootstrap families: Appendix~\ref{app:lm_final_breakdowns}.}
\label{tab:lm_comparisons_selected}
\setlength{\tabcolsep}{3pt}
\adjustbox{max width=\linewidth}{%
\begin{tabular}{@{}clrrr@{}}
\toprule
Tokens & Comparison & ZS & Recall & NIAH \\
\midrule
$1$B & M3exact $-$ M3 SISO & $-0.2\,[-0.9,+0.5]$ & $+0.5\,[-0.1,+1.0]$ & $+4.0\,[+2.5,+5.5]^{*}$ \\
\addlinespace
$2$B & R2O2-r $-$ M3 SISO & $-1.1\,[-1.9,-0.4]^{*}$ & $+1.6\,[+0.9,+2.2]^{*}$ & $+13.4\,[+11.6,+15.3]^{*}$ \\
$2$B & HM-0 $-$ M1 & $+0.2\,[-0.5,+1.0]$ & $+1.4\,[+0.8,+2.0]^{*}$ & $+3.2\,[+1.3,+5.0]^{*}$ \\
$2$B & HM-0 $-$ M2 & $+0.7\,[+0.0,+1.5]$ & $+0.3\,[-0.3,+1.0]$ & $-13.2\,[-15.2,-11.2]^{*}$ \\
$2$B & HM-25 $-$ M1 & $+0.2\,[-0.6,+0.9]$ & $+2.2\,[+1.6,+2.8]^{*}$ & $-3.1\,[-5.0,-1.2]^{*}$ \\
$2$B & HM-25 $-$ M2 & $+0.7\,[-0.1,+1.4]$ & $+1.1\,[+0.5,+1.7]^{*}$ & $-19.4\,[-21.5,-17.3]^{*}$ \\
$2$B & R1O1-r (M3) $-$ M3 SISO & $-0.4\,[-1.2,+0.3]$ & $-0.6\,[-1.3,-0.0]$ & $+11.0\,[+9.1,+12.9]^{*}$ \\
\addlinespace
$2$B & Attention $-$ M3 SISO & $-1.6\,[-2.4,-0.8]^{*}$ & $+8.2\,[+7.4,+9.1]^{*}$ & $+31.2\,[+28.6,+33.7]^{*}$ \\
\addlinespace
$2$B & HM-75 $-$ HM-0 & $-0.2\,[-0.9,+0.5]$ & $+0.7\,[+0.1,+1.3]$ & $+11.1\,[+9.0,+13.2]^{*}$ \\
$2$B & HM-75 $-$ HM-25 & $-0.1\,[-0.9,+0.6]$ & $-0.1\,[-0.7,+0.5]$ & $+17.3\,[+15.2,+19.5]^{*}$ \\
$2$B & HM-75 $-$ M3 SISO & $-0.2\,[-0.9,+0.6]$ & $+0.1\,[-0.5,+0.7]$ & $+8.1\,[+6.1,+10.2]^{*}$ \\
$2$B & HM-75 $-$ M1 & $+0.0\,[-0.7,+0.8]$ & $+2.1\,[+1.5,+2.8]^{*}$ & $+14.2\,[+12.2,+16.3]^{*}$ \\
$2$B & HM-75 $-$ M2 & $+0.5\,[-0.2,+1.3]$ & $+1.0\,[+0.4,+1.7]^{*}$ & $-2.1\,[-4.1,-0.1]^{*}$ \\
\addlinespace
$2$B & HM-100 $-$ HM-75 & $-0.1\,[-0.8,+0.7]$ & $+0.0\,[-0.6,+0.7]$ & $-15.2\,[-17.3,-13.1]^{*}$ \\
$2$B & HM-100 $-$ M3 SISO & $-0.3\,[-1.0,+0.5]$ & $+0.2\,[-0.5,+0.8]$ & $-7.1\,[-9.1,-5.1]^{*}$ \\
\addlinespace
$10$B & R2O2-c $-$ M3 SISO & $+0.2\,[-0.6,+0.9]$ & $+0.7\,[-0.1,+1.5]$ & $+4.0\,[+1.9,+6.1]^{*}$ \\
$10$B & R2O2-c $-$ M3 MIMO & $-0.3\,[-1.1,+0.4]$ & $+0.1\,[-0.7,+0.9]$ & $-4.1\,[-6.1,-2.1]^{*}$ \\
$10$B & R2O2-r $-$ M3 SISO & $-0.5\,[-1.2,+0.3]$ & $+0.0\,[-0.8,+0.8]$ & $-2.3\,[-4.3,-0.2]^{*}$ \\
$10$B & M3 MIMO $-$ M3 SISO & $+0.5\,[-0.2,+1.2]$ & $+0.6\,[-0.2,+1.4]$ & $+8.1\,[+6.1,+10.1]^{*}$ \\
\addlinespace
$10$B & HM-75 $-$ M3 SISO & $+0.1\,[-0.7,+0.8]$ & $+0.2\,[-0.6,+1.0]$ & $-1.0\,[-3.2,+1.3]$ \\
$10$B & HM-75 $-$ M3 MIMO & $-0.4\,[-1.2,+0.4]$ & $-0.4\,[-1.1,+0.4]$ & $-9.1\,[-11.2,-6.9]^{*}$ \\
\addlinespace
\bottomrule
\end{tabular}}
\end{table}

\FloatBarrier
\par\endgroup

\end{document}